\documentclass[11pt,a4paper,twoside,openright]{report}
\usepackage[export]{adjustbox}
\usepackage{algorithm}
\usepackage[algo2e,boxruled,linesnumbered]{algorithm2e} 
\usepackage{algpseudocode}
\usepackage{amsmath}
\usepackage{amssymb}
\usepackage[backend=biber,sorting=none]{biblatex}
\usepackage[inline]{enumitem}
\usepackage{fancyhdr}
\usepackage{geometry}
\usepackage{graphicx}
\usepackage[final]{hyperref}
\usepackage{cleveref}
\usepackage[utf8]{inputenc}
\usepackage{lscape}
\usepackage{mathtools}
\usepackage{pifont}
\usepackage{subcaption}
\usepackage{titlesec}
\usepackage{xcolor}
\crefname{chapter}{Chapter}{Chapters}
\crefname{section}{Section}{Sections}
\crefname{subsection}{Section}{Sections}
\crefname{subsubsection}{Section}{Sections}
\crefname{appendix}{Appendix}{Appendices}
\crefname{figure}{Figure}{Figures}
\crefname{subfigure}{Figure}{Figures}
\crefname{table}{Table}{Tables}
\crefname{equation}{Equation}{Equations}
\crefname{subequations}{Equation}{Equations}
\crefname{algorithm}{Algorithm}{Algorithms}

\newcommand{\cmark}{\ding{51}}

\SetCommentSty{mycommfont}
\titleformat{\paragraph}
{\normalfont\normalsize\bfseries}{\theparagraph}{1em}{}
\titlespacing*{\paragraph}
{0pt}{3.25ex plus 1ex minus .2ex}{1.5ex plus .2ex}

\graphicspath{ {./pictures/} }

\begin{document}
    \pagenumbering{gobble}
    \thispagestyle{plain}
\begin{center}
  \large
    POLITECNICO DI MILANO\\
    School of Industrial and Information Engineering\\
  \normalsize
    Master of Science Degree in Computer Science and Engineering\\
    Department of Electronics, Information and Bioengineering\\
    \vspace*{1cm}
    \begin{figure}[htbp]
      \begin{center}
        \includegraphics[width=5cm]{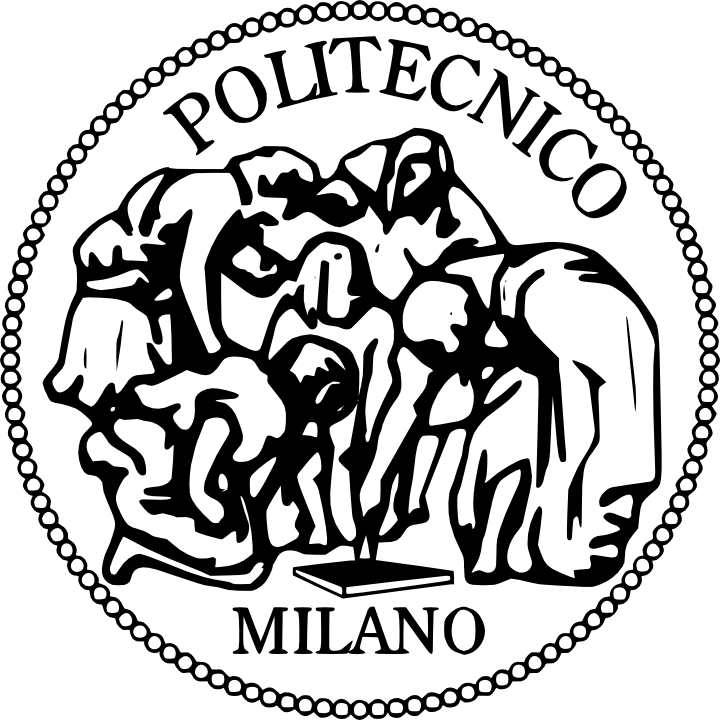}
      \end{center}
    \end{figure}
  \vspace*{1.5cm}
  \LARGE
    \textbf{Non-planar Object Detection and Identification by Features Matching and Triangulation Growth}\\
\end{center}
\vspace*{2cm}
\large
    \begin{flushleft}
        Supervisor:    Giacomo Boracchi, Ph.D. \\
        Co-supervisor: Simone Gasparini, Ph.D.
    \end{flushleft}
    \begin{flushright}
        Author:
        \\
        Filippo Leveni, 875208
    \end{flushright}
    \vspace*{1.5cm}
    \begin{center}
        Academic Year 2017-2018
    \end{center}
\clearpage
    \thispagestyle{empty}
    \normalfont
    \cleardoublepage
    \thispagestyle{plain}
\begin{flushright}
    \itshape{A mio fratello, la parte migliore di me...}
\end{flushright}

    \thispagestyle{empty}
    \cleardoublepage
    \pagenumbering{arabic}
    \newpage
\chapter*{Summary}
\addcontentsline{toc}{chapter}{Summary}

Object detection and identification is surely a fundamental topic in the computer vision field; it plays a crucial role in many applications such as object tracking, industrial robots control, image retrieval, etc.\\
We propose a feature-based approach for detecting and identifying distorted occurrences of a given template in a scene image by incremental grouping of feature matches between the image and the template.
For this purpose, we consider the Delaunay triangulation of template features as an useful tool through which to be guided in this iterative approach. The triangulation is treated as a graph and, starting from a single triangle, neighboring nodes are considered and the corresponding features are identified; then matches related to them are evaluated to determine if they are worthy to be grouped. This evaluation is based on local consistency criteria derived from geometric and photometric properties of local features.\\
Our solution allows the identification of the object in situations where geometric models (e.g. homography) does not hold, thus enable the detection of objects such that the template is non planar or when it is planar but appears distorted in the image.\\
We show that our approach performs just as well or better than application of homography-based RANSAC in scenarios in which distortion is nearly absent, while when the deformation becomes relevant our method shows better description performance.
    \thispagestyle{empty}
    \cleardoublepage
    \chapter*{Ringraziamenti}
\addcontentsline{toc}{chapter}{Ringraziamenti}
Vorrei ringraziare tutte le persone che mi hanno accompagnato ed aiutato a raggiungere questo traguardo.
\\\\
Innanzitutto ringrazio Giacomo per il suo costante supporto tecnico e, non meno importante, morale. Grazie anche a Simone che, nonostante la distanza, è riuscito a darmi preziosi consigli.
\\\\
Grazie ai miei genitori per i grandi sacrifici che hanno fatto per permettermi di arrivare fino a qui.
Grazie a mia mamma, che mi ha insegnato ad essere me stesso, ed a mio papà, che mi ha insegnato ad essere indipendente. Grazie a mio fratello, che con la sua sola presenza mi fa sentire a casa. Grazie a mia nonna Giuseppina, o meglio conosciuta come "Pinuccia", che mi ha insegnato a pretendere il massimo da me stesso.
\\\\
Grazie ai miei amici di sempre che non se ne sono mai andati, Mizal e Andrea, probabilmente senza di voi non sarei riuscito ad arrivare fin qua.\\
Grazie a Greta che, nonostante tutto, continua a starmi vicino.\\
Grazie anche a Valentina, la prima compagna conosciuta in università, che è diventata subito un'amica speciale, anche se rimane una brutta persona.
\begin{flushright}
    \emph{Filippo}
\end{flushright}

    \thispagestyle{empty}
    \normalfont
    \cleardoublepage
    \tableofcontents
    \newpage
     
    \chapter{Introduction}
\label{ch:chapter_1}

\begin{quotation}
    {
     \footnotesize
     \noindent
     \emph{"Men love to wonder, \\
            and that is the seed of science"}
    \begin{flushright}
         Ralph Waldo Emerson
     \end{flushright}
    }
\end{quotation}
\vspace{0.5cm}
\noindent
In recent years, computer vision has become a fundamental field in engineering. This is mainly due to the increasing pervasiveness of technology in everyday tasks, which necessarily requires a greater understanding of the surrounding environment.\\
Computer Vision, according to the British Machine Vision Association\cite{ref:reference_43} is defined as:
\begin{quotation}
    \emph{"Computer vision is concerned with the automatic extraction, analysis and understanding of useful information from a single image or a sequence of images. It involves the development of a theoretical and algorithmic basis to achieve automatic visual understanding."}
\end{quotation}
Thus computer vision seeks to automate tasks that are usually fulfilled by the human visual system.
\begin{figure}[h]
  \begin{center}
    \includegraphics[width=0.8\textwidth]{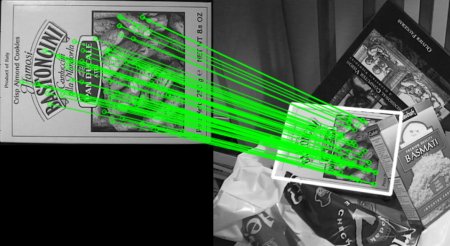}
    \caption{Example of feature-based object detection and identification. Figure from \cite{ref:reference_8}.}
  \end{center}
\end{figure}
\par
Object detection and identification is a particular subfield of computer vision whose goal is to identify instances of certain objects in digital images and videos. Frequent applications of this technology are face detection and recognition, object tracking, industrial robots control and image retrieval\cite{ref:reference_47}\cite{ref:reference_48}\cite{ref:reference_49}\cite{ref:reference_50}\cite{ref:reference_51}\cite{ref:reference_52}.\\
Most of object detection algorithms and methods operate under the assumption that the target is known and possibly occur multiple times in the scene, and are composed of 3 major steps:
\begin{enumerate*}[label=(\roman*)]
    \item features (or regions of interest) are detected in every image independently by algorithms such as SIFT\cite{ref:reference_5}, BRISK\cite{ref:reference_22} and FREAK\cite{ref:reference_24}. Features can be any points or regions but they should be repeatably detectable, i.e. it should be possible to detect them under changing conditions such as changing illumination, rotation, scale, etc.
    \item Features extracted from the scene and the template are matched in order to establish correspondences. This step usually makes use of nearest neighbor search\cite{ref:reference_17} based on Euclidean distance between feature descriptors which can result in a set of matches corrupted by a large number of outliers.
    \item Verify the matches in order to filter out outliers. The verification is often done by enforcing the homography or the epipolar geometry constraint with the largest amount of inliers and discarding all outliers.
\end{enumerate*}
\\
This approach presents two main limitations. First, accepting the hypothesis with the largest amount of inliers enables only the detection of the main instance of the object present in the scene. Second, using the homography hypothesis allows to correctly identify only templates that are planar and not distorted.\\
\\
\emph{The purpose of this thesis is to investigate and develop a feature-based method that can solve the detection and identification problem in presence of multiple occurrences of the object of interest in a depicted scenario. In particular we aim to deal with:
\begin{itemize}
    \item[--] Deformable objects depicted in a planar configuration in the template.
    \item[--] Rigid non-planar objects.
\end{itemize}}
\noindent
\\
Several feature-based detection algorithms and variants of the above described approach have been investigated over the past few years, due to the huge amount of applications where they can be applied. For instance \cite{ref:reference_41} proposes a pairwise dissimilarity measure formulating object-detection as an unsupervised clustering problem. In \cite{ref:reference_44} the authors set up an integer quadratic programming problem, where the cost function has terms based on similarity of corresponding point descriptors as well as the geometric distortion between pairs of corresponding feature points. While \cite{ref:reference_45} formulates the matching task as an energy minimization problem by defining a complex objective function of the appearance and the spatial arrangement of the features.\\
Differently from these algorithms, our idea was born from the identification of the Delaunay triangulation as a useful tool for encapsulating information about the mutual position of template features. Starting from this consideration, we will use the triangulation to perform a guided and iterative grouping of feature matches.
\begin{figure}
  \centering
  \begin{subfigure}[!ht]{0.28\linewidth}
    \includegraphics[width=\linewidth]{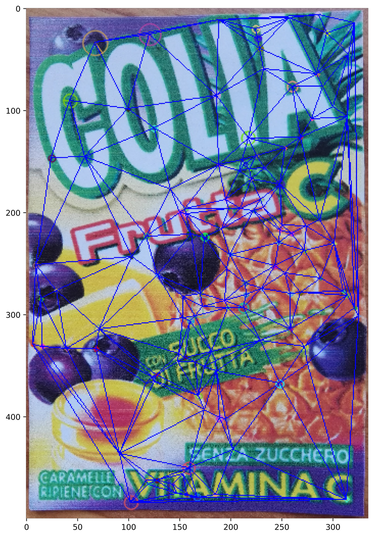}
    \caption{Template Delaunay triangulation built from template features coordinates.}
    \label{fig:figure_89}
  \end{subfigure}
  \begin{subfigure}[!ht]{0.53\linewidth}
    \includegraphics[width=\linewidth]{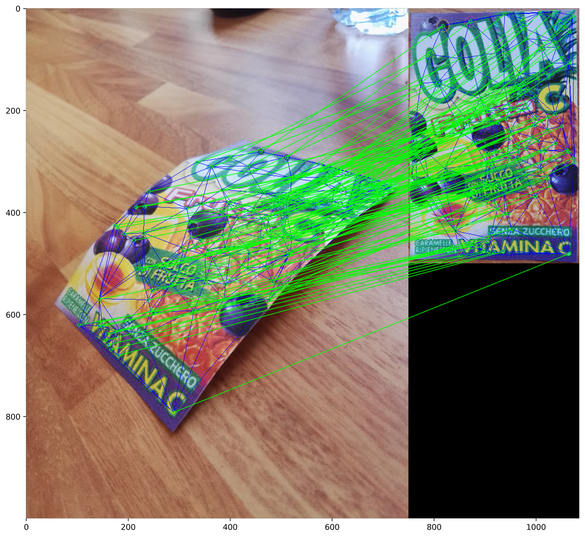}
    \caption{In this image we can see correspondences between positions of template and scene features. These matches has been grouped in an iterative way guided by the triangulation in \cref{fig:figure_89}.}
  \end{subfigure}
\end{figure}
\par
The contribution of this work is the development of an algorithm for object detection and identification based on the iterative growth of a set containing correct correspondences between the template and one of its occurrence in the scene. The growth is guided by a Delaunay triangulation of the features position in the template. This is done in order to enforce locality constraints, to preserve the mutual position of features in the template and in each of the object occurrence in the scene. At each iteration, correspondences that refer to features nearby to those whose correspondences are already part of the set are evaluated to determine if they are worthy to be grouped. This evaluation is based on local consistency criteria derived from geometric and photometric properties of local features.\\
Extensive tests, to assess the performance of the proposed solution in challenging scenarios where multiple instances of the template appears distorted in the scene, have demonstrated the effectiveness of this approach. We compared our method with a baseline detection algorithm where RANSAC is used to enforce an homography to perform geometric verification of feature matches. Experimental results indicate that, although our algorithm is more time consuming, it is able to rectify distorted objects and to align them with the template in order to perform image-by-image comparison. Therefore we believe that further research on triangulation-based approaches to drive the object detection and identification task can lead to significant results.\\
\\
The thesis is structured as follows:
\begin{itemize}
    \item In \cref{ch:chapter_2} we introduce background concepts necessary to understand this research, then we review literature on feature-based object detection and identification.
    \item In \cref{ch:chapter_3} we first formally define the problem we address, then we present the proposed approach for non-planar object detection and identification by features matching and triangulation growth.
    \item In \cref{ch:chapter_4} we present details that are necessary to implement our ideas in an efficient and effective algorithm.
    \item In \cref{ch:chapter_5} we assess the performance of the proposed algorithm, introducing dataset, evaluation procedure and results.
    \item In \cref{ch:chapter_6} we summarize the contributions of this work, present our conclusions and point out possible improvements of the approach.
    \item In \cref{appendix_A} we collect an extensive version of the evaluation results.
\end{itemize}
    \thispagestyle{empty}
    \chapter{State of the art}
\label{ch:chapter_2}

\noindent
In \cref{sec:section_2.1} we present the necessary instruments to understand our work. In \cref{sec:section_2.2} the state of the art in the feature-based object detection field are presented.
\section{Background}
    \label{sec:section_2.1}
    \subsection{Homography and Affine transformations}
        \label{sec:section_2.1.1}
        In 2D spaces an $homography$ $transformation$ is a projective transformation from one coordinates system to another. The mapping is done according to:
        \begin{equation}
            \begin{bmatrix}
                x' \\
                y' \\
                z'
            \end{bmatrix}
            =
            M
            \begin{bmatrix}
                x \\
                y \\
                z
            \end{bmatrix}
            \label{eq:equation_36}
        \end{equation}
        Where for the $homography$ $transformation$
        \begin{equation}
            M = M_H =
            \begin{bmatrix}
                m_{1} & m_{2} & m_{3} \\
                m_{4} & m_{5} & m_{6} \\
                m_{7} & m_{8} & m_{9}
            \end{bmatrix}
        \end{equation}
        is defined only up to scale, thus the total number of degrees of freedom is 8.
        $Affine$ $transformations$ are special cases of the former, in which:
        \begin{equation}
            M = M_A =
            \begin{bmatrix}
                m_{1} & m_{2} & m_{3} \\
                m_{4} & m_{5} & m_{6} \\
                0     & 0     & 1
            \end{bmatrix}
        \end{equation}
        \cref{eq:equation_36} refers to $homogeneous$ $coordinates$, system of coordinates used in projective geometry.
    \subsection{Triangulation of a set of points}
        \label{sec:section_2.1.2}
\begin{figure}[!h]
            \centering
            \includegraphics[width=0.4\linewidth]{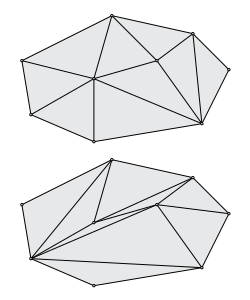}
            \caption{Two different triangulations of the same set of points in the plane. Figure from \cite{ref:reference_54}.}
            \label{fig:figure_81}
        \end{figure}

        A $triangulation$ of a set of points $P$ in the plane is defined as a maximal set of non-crossing edges between points of $P$ (examples of triangulation in \cref{fig:figure_81}).
        \subsubsection{Delaunay triangulation}
\begin{figure}[!h]
                \centering
                \includegraphics[width=0.4\linewidth]{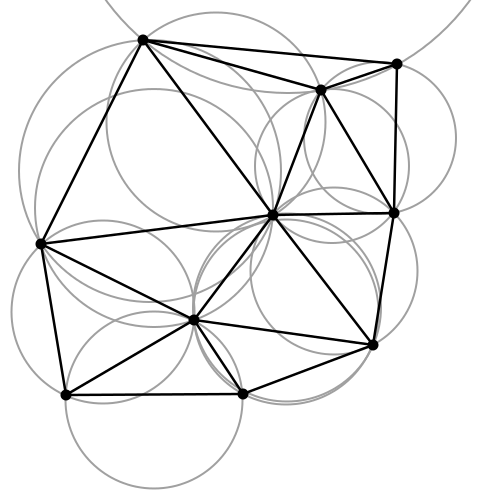}
                \caption{Planar Delaunay triangulation with circumcircles shown. Figure from \cite{ref:reference_55}.}
                \label{fig:figure_82}
            \end{figure}

            A special case of triangulation is the $Delaunay$ $triangulation$.  The $Delaunay$ $triangulation$ of a set of 2D points $P$ is a particular triangulation for which no point $p \in P$ is inside the circumcircle of any triangle of the triangulation (\cref{fig:figure_82}).
            \\
            An important property is that, in the plane, the $Delaunay$ $triangulation$ maximizes the global minimum angle. This is useful in our work because avoids skinny triangles.
        \subsubsection{Constrained Delaunay triangulation}
            A $constrained$ $Delaunay$ $triangulation$ is a generalization of the Delaunay triangulation that forces certain segments to be present in the triangulation. Since a Delaunay triangulation is almost always unique, often a $constrained$ $Delaunay$ $triangulation$ contains edges that do not satisfy the Delaunay condition (the requirement that the circumcircles of all triangles have no internal points). Thus a $constrained$ $Delaunay$ $triangulation$ often is not a Delaunay triangulation itself. 
    \subsection{Features matching}
        \label{sec:section_2.1.3}
        \subsubsection{Local features}
            \label{subsubsec:subsubsec_2.1.3.1}
\begin{figure}[!h]
                \centering
                \includegraphics[width=0.4\linewidth]{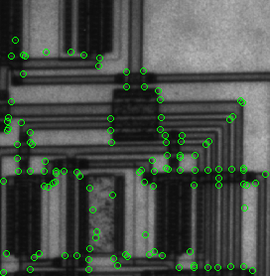}
                \caption{Example of positions where features are detected. For the detection FAST\cite{ref:reference_46} method has been used. Figure from \cite{ref:reference_56}.}
                \label{fig:figure_83}
            \end{figure}

            There is no salient definition of $local feature$, but in general it could be defined as an "interesting" part of the image (i.e. an image pattern which differs from its immediate neighborhood). It is usually associated with a change of an image property or several properties simultaneously (features example in \cref{fig:figure_83}). These considered properties are color intensities and texture.
            \\
            The main goal of $local feature$ representation is to distinctively represent the image based on some salient regions while remaining invariant to viewpoint and illumination changes. Thus, the image is represented based on its local structures by a set of local feature descriptors extracted from a set of image regions called interest regions (regions around the position where the feature is located).
            \\
            These descriptors allow to effectively match local structures between images.
        \subsubsection{Detector}
            \label{subsubsec:subsubsec_2.1.3.2}
\begin{figure}[!h]
                    \centering
                    \includegraphics[width=0.8\linewidth]{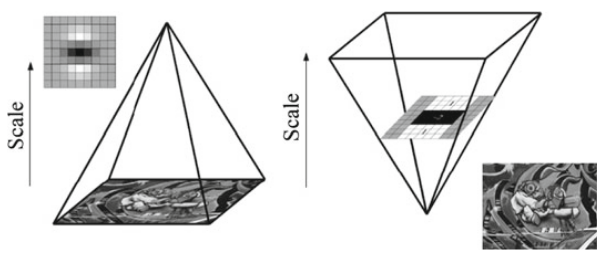}
                    \caption{Scale-space structure construction. Figure from \cite{ref:reference_57}.}
                    \label{fig:figure_84}
                \end{figure}

\begin{figure}[!h]
                    \centering
                    \includegraphics[width=\linewidth]{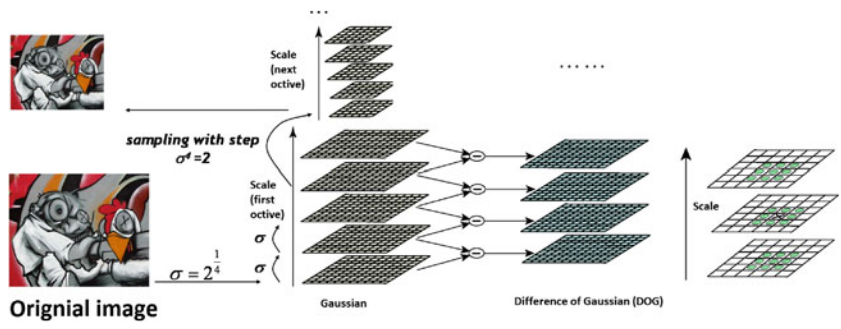}
                    \caption{3D scale-space extrema search. Figure from \cite{ref:reference_57}.}
                    \label{fig:figure_85}
                \end{figure}

            Feature $detector$ is the algorithm responsible to locate features in a given image.
            Since information around features position is used as the starting point and main primitive for subsequent algorithms, these algorithms will often only be as good as its feature detector.
            \\
             We want to obtain a sparse set of local measurements that capture the essence of the underlying input images and encode their interesting structures. To meet this goal, the feature detectors must have certain properties.
             \begin{itemize}
                 \item \textbf{Robustness}, the feature detection algorithm should be able to detect the same feature locations independent of scaling, rotation, shifting, photometric deformations, compression artifacts, and noise.
                 \item \textbf{Accuracy}, the feature detection algorithm should accurately localize the image features (same pixel locations), especially for image matching tasks, where precise correspondences are needed to estimate the epipolar geometry.
                 \item \textbf{Generality}, the feature detection algorithm should be able to detect features that can be used in different applications.
                 \item \textbf{Efficiency}, the feature detection algorithm should be able to detect features in new images quickly to support real-time applications.
                 \item \textbf{Quantity}, the feature detection algorithm should be able to detect all or most of the features in the image. Where, the density of detected features should reflect the information content of the image for providing a compact image representation.
                 \item \textbf{Repeatability}, the feature detection algorithm should be able to detect the same features of the same scene or object repeatedly under variety of viewing conditions. It directly depends on the other properties like robustness, accuracy and quantity.
             \end{itemize}
             There are different types of feature detectors. For this work it is interesting to introduce a specific multi-scale detector: $Difference$ $of$ $Gaussian$ $(DoG)$, proposed by Lowe\cite{ref:reference_5}.
             We first explain the concept of scale-space pyramid, then $DoG$ is introduced.
             \paragraph{Scale-space pyramid}
                In order to address scale changes, multi-scale detectors assume that the change in scale is the same in all directions (i.e. uniform) and they search for stable features across all possible scales, using a kernel function of scale, known as scale space. In particular, the image size is varied and a filter (e.g. Gaussian filter) is repeatedly applied to subsequent layers, or the original image is left unchanged and the filter size is varied, as shown in \cref{fig:figure_84}.
             \paragraph{Difference of Gaussian (DoG)}
                \label{par:paragraph_DoG}
                $DoG$ is an efficient algorithm based on local 3D extrema in the scale-space pyramid built with Gaussian filters. $DoG$ is used to efficiently detect stable features from scale-space extrema. The $DoG$ function $D(x, y, \sigma)$ can be computed subctracting adjacent scale levels of a Gaussian pyramid separated by a factor k.
                \begin{equation}
                    \begin{aligned}
                        D(x, y, \sigma) &= (G(x, y, k\sigma) - G(x, y, \sigma)) * I(x, y) \\
                        &= L(x, y, k\sigma) - L(x, y, \sigma)
                    \end{aligned}
                \end{equation}
                Thus a detector based on $DoG$ searches for 3D scale space extrema of the $DoG$ function, as shown in \cref{fig:figure_85}.
                After that an extrema (feature) is recognized, the corresponding interest point (i.e. key point) is identified with that location $p(x, y)$ and scale $s$.
        \subsubsection{Descriptor}
            \label{subsubsec:subsubsec_2.1.3.3}
\begin{figure}[!h]
                    \centering
                    \includegraphics[width=0.8\linewidth]{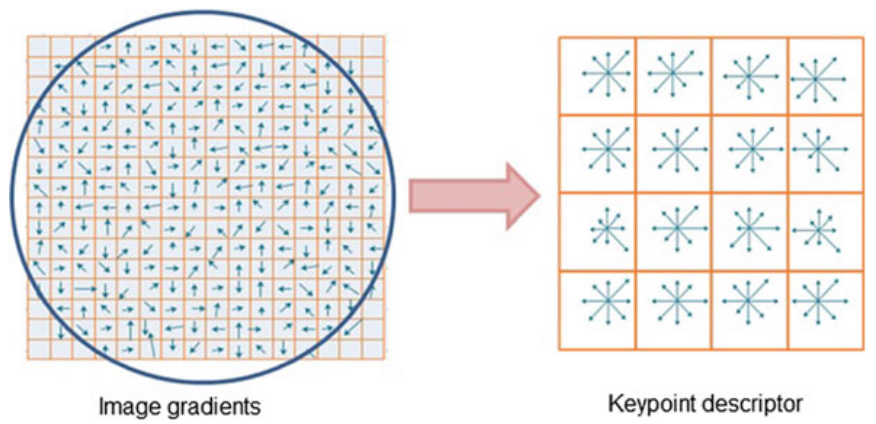}
                    \caption{SIFT descriptor for a 16x16 pixels patch and a 4x4 descriptor array. Figure from \cite{ref:reference_57}.}
                    \label{fig:figure_86}
                \end{figure}

            Once a key point has been detected from an image at a location $p(x, y)$ and scale $s$, the image structure in a neighborhood of $p$ needs to be encoded in a suitable descriptor for matching.
            \paragraph{Scale Invariant Feature Transform (SIFT)}
                In SIFT\cite{ref:reference_5} algorithm a number of key points are detected in the image using the $DoG$ operator introduced in \cref{par:paragraph_DoG}. The points are selected as local extrema of the $DoG$ function, and for each key point a descriptor vector is extracted. Over a neighborhood around the point of interest, the local orientation of the image is estimated (let us call it $\theta$) using the local image properties to provide invariance against rotation. Next, a descriptor is computed for each detected point, based on local image information at the charachteristic scale. The SIFT descriptor builds an histogram of gradient orientations of points in a region around the key point position, finds the highest orientation value and any other values that are within 80\% of it, and uses these orientations for building the descriptor. The description stage of the SIFT algorithm starts by sampling the image gradient magnitudes and orientations in a 16x16 region around each key point position using its scale to select the level of Gaussian blur for the image. Then, a set of orientation histograms is created where each histogram contains samples from a 4x4 subregion of the original neighborhood and having eight orientations bins in each. A Gaussian weighting function with $\sigma$ equal to half the region size is used to assign weight to the magnitude of each sample point and gives higher weights to gradients closer to the center of the region, which are less affected by positional changes. The descriptor is then formed from a vector containing the values of all the orientation histograms entries. Since there are 4x4 histograms each with 8 bins, the vector has 4x4x8 = 128 elements for each key point. Finally the descriptor vector is normalized to unit length to gain invariance to affine changes in illumination. \cref{fig:figure_86} illustrates a schematic representation of the SIFT descriptor, where the gradient orientations and magnitudes are computed at each pixel and then weighted by a Gaussian function (indicated by the circle). A weighted gradient orientation histogram is then computed for each subregion.
                Morel and Yu\cite{ref:reference_18} proved that SIFT is fully invariant with respect to only zoom, rotation and translation. Therefore, they introduced affine-SIFT (A-SIFT), which simulates all image views obtainable by varying the camera axis orientation parameters (latitude and longitude angles).
        \subsubsection{Features matching}
            \label{subsubsec:subsubsec_2.1.3.4}
\begin{figure}[!h]
                \centering
                \includegraphics[width=0.8\linewidth]{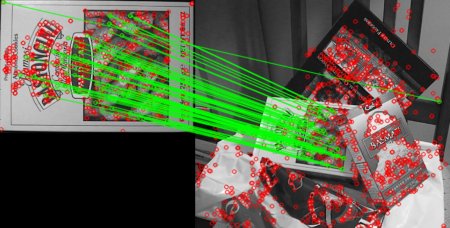}
                \caption{Matching image regions based on their local descriptors. Figure from \cite{ref:reference_7}.}
                \label{fig:figure_87}
            \end{figure}

            Features matching is the task of establishing correspondences between two images of the same scene/object. A common approach to features matching consists of detecting a set of key points each associated with its own descriptor, from image data. Once the points and their descriptors have been extracted from two or more images, the next step is to establish correspondences between these images as illustrated in \cref{fig:figure_87}.
            \\
            Suppose to have two images $I^1$, $I^2$ and for each image we have computed descriptors for key points:
            \begin{equation}
                \begin{aligned}
                    &D^1 = \{d^1_1, d^1_2, \dots, d^1_n\},
                    \\
                    &D^2 = \{d^2_1, d^2_2, \dots, d^2_m\}.
                \end{aligned}
            \end{equation}
            The aim is to find, for each $d^1_i \in D^1$, the best correspondence in $D^2$. To this end, the distance between each couple of points descriptors $\|d^1_i - d^2_j\|$ is computed. Then, for each $d^1_i \in D^1$ the nearest neighbour descriptor in $D^2$ is considered, and a match is created between the two key points of the descriptors. Furthermore, to suppress matching candidates for which the correspondence may be regarded as ambiguous, the ratio between the distances to the nearest and the next nearest descriptor is required to be less than some threshold (ratio test, introduced in \cite{ref:reference_5}).
    \subsection{Classic template matching with homography validation}
        \label{sec:section_2.1.4}
        The more common approach of template matching with homography validation is driven by the RANdom SAmple Consensus\cite{ref:reference_4} (RANSAC) algorithm. Let us first introduce RANSAC, then the matching approach is described.
        \subsubsection{RANSAC}
            \label{subsubsec:subsubsection_2.1.4.1}
\begin{figure}[!h]
                \centering
                \begin{subfigure}[!h]{0.4\linewidth}
                    \includegraphics[width=\linewidth]{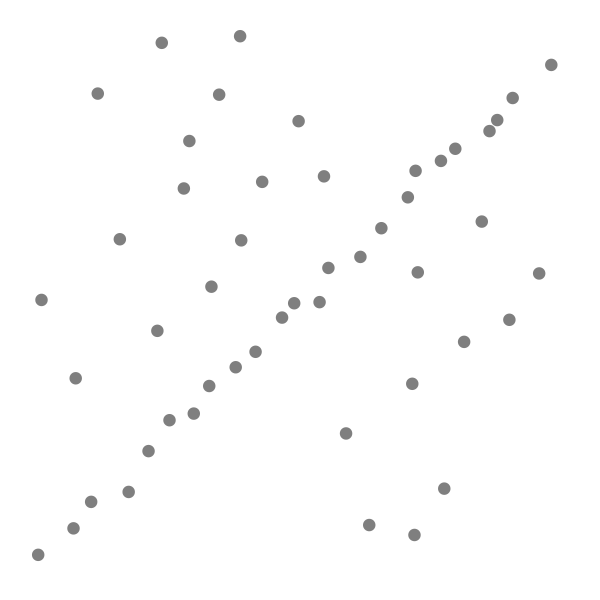}
                    \caption{Data set with many outliers.}
                    \label{fig:figure_88a}
                \end{subfigure}
                \begin{subfigure}[!h]{0.4\linewidth}
                    \includegraphics[width=\linewidth]{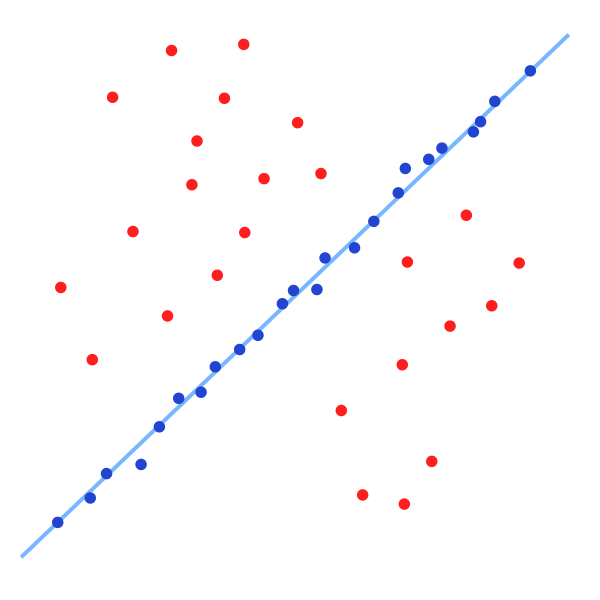}
                    \caption{Fitted line with RANSAC. Outliers have no influence on the result.}
                    \label{fig:figure_88b}
                \end{subfigure}
                \caption{Example of application of RANSAC. Figure from \cite{ref:reference_58}.}
                \label{fig:figure_88}
            \end{figure}

            Random sample consensus (RANSAC) is an iterative method to estimate parameters of a mathematical model from a set of observed data that contains outliers. It is a non-deterministic algorithm in the sense that it produces a reasonable result only with a certain probability, with this probability increasing as more iterations are allowed. A basic assumption is that the data consists of "inliers", i.e., data whose distribution can be explained by some set of model parameters, though may be subject to noise, and "outliers" which are data that do not fit the model. The outliers can come, for example, from extreme values of the noise or from erroneous measurements or incorrect hypotheses about the interpretation of data.
            \\
            The RANSAC algorithm is essentially composed of three steps that are iteratively repeated:
            \begin{enumerate}
                \item In the first step, a sample subset containing minimal data items is randomly selected from the input dataset. A fitting model and the corresponding model parameters are computed using only the elements of this sample subset. The cardinality of the sample subset is the smallest sufficient to determine the model parameters.
                \item In the second step, the algorithm checks which elements of the entire dataset are consistent with the model instantiated by the estimated model parameters obtained from the first step. A data element will be considered as an outlier if it does not fit the model within some error threshold that defines the maximum deviation attributable to the effect of noise. The points that fit the estimated model are considered as part of the consensus set.
                \item Afterwards, the model may be improved by reestimating it using all members of the consensus set.
            \end{enumerate}
            This procedure is repeated a fixed number of times, each time producing either a model which is rejected because too few points are part of the consensus set, or a refined model together with a corresponding consensus set size. Example of application of RANSAC in \cref{fig:figure_88}.
        \\
        \par
\begin{figure}[!h]
                    \centering
                    \includegraphics[width=0.5\linewidth]{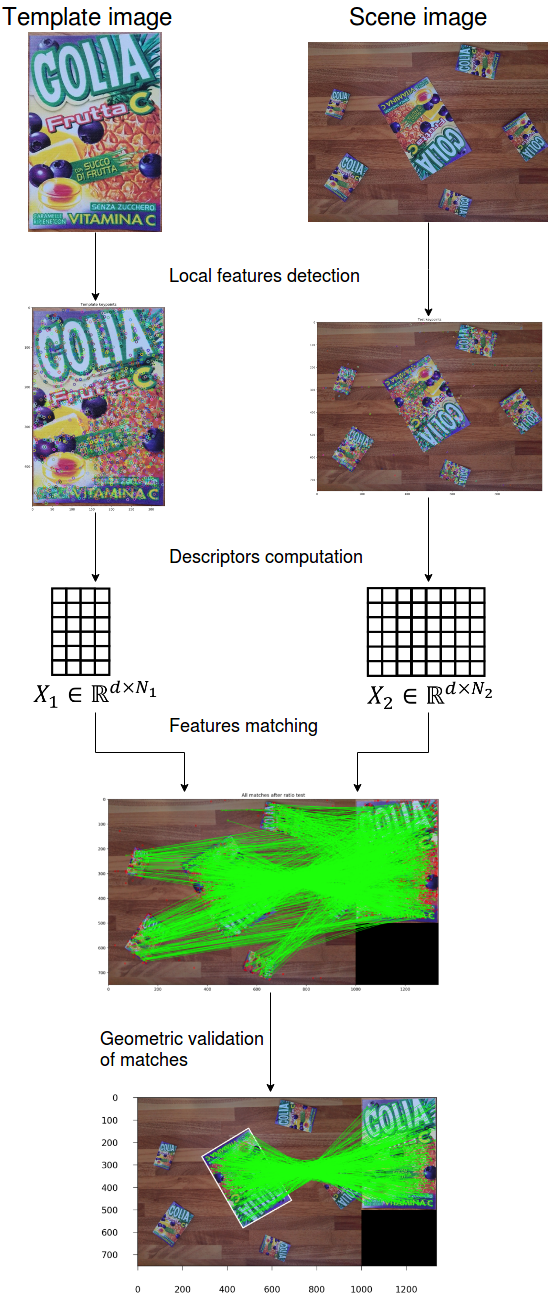}
                    \caption{Classic template matching pipeline.}
                    \label{fig:figure_92}
        \end{figure}

        Template matching with homography validation driven by RANSAC algorithm follows the following steps:
        \begin{enumerate}
            \item Identification of features in template and scene image and extraction of their key points and descriptors (as explained in \cref{subsubsec:subsubsec_2.1.3.1}, \ref{subsubsec:subsubsec_2.1.3.2}, \ref{subsubsec:subsubsec_2.1.3.3}).
            \item Features matching (introduced in \cref{subsubsec:subsubsec_2.1.3.4}).
            \item Validate an homography model through RANSAC algorithm (\cref{subsubsec:subsubsection_2.1.4.1}), using matches as observed data.
        \end{enumerate}
        \cref{fig:figure_92} shows the classic template matching pipeline.
\section{Related works}
    \label{sec:section_2.2}
    Most of the existing feature-based object detection approaches, follow a similar scheme. First, given a pair of images, descriptors such as SIFT\cite{ref:reference_5}, SURF\cite{ref:reference_19}, LBP\cite{ref:reference_20}, ORB\cite{ref:reference_21}, BRISK\cite{ref:reference_22}, MSER\cite{ref:reference_23} and FREAK\cite{ref:reference_24} are extracted from them. Then the nearest neighbours (in the descriptor space) to each descriptor in the first image are found in the second image. Unfortunately many matches are outliers. Lowe, in its influential paper \cite{ref:reference_5}, has proposed a simple method to discard ambiguous matches, called "ratio test". This test consists in computing the ratio between the distance to the nearest neighbour over the distance to the second nearest neighbour. Then descriptors whose ratio is over a certain threshold (Lowe proposed 0.8 in \cite{ref:reference_5}) are eliminated. However, matches can be wrong even after this test, thus they have to be verified geometrically.
    \\
    Many works goal is to enrich tentative matches (matches not already geometrically verified). Fragoso and Turk \cite{ref:reference_33} introduce an alternative score to Lowe ratio. Interesting is ASIFT\cite{ref:reference_18}, a technique that generates multiple synthetic affine transformation of an input image, and detects features on them. \cite{ref:reference_37} formulates image matching as a joint optimization problem, in which adaptive descriptor selection (among five ones: SIFT\cite{ref:reference_5}, LIOP\cite{ref:reference_38}, DAISY\cite{ref:reference_39}, GB\cite{ref:reference_40} and RI, the last one extracts the grey-level pixel intensities of the feature regions in a raster scan order) and progressive correspondence enrichment are alternately conducted. \cite{ref:reference_34}\cite{ref:reference_35} attempts to validate tentative matches preserving local neighborhood structures and eliminating the ones inconsistent with it. \cite{ref:reference_36} proposes an efficient method for estimating the number of correct matches without explicitly computing them, but analyzing the spatial order of the features position, as projected to the x-axis of the image.
    Since our work exploits properties in common to most of the key point descriptors, these approaches could be applied to improve our results.
    \\
    Many geometric verifications are done through RANSAC\cite{ref:reference_4} algorithm. The fundamental idea (as introduced in \cref{subsubsec:subsubsection_2.1.4.1}) is the estimation of an homography (or epipolar geometry for 3D scenes) from random samples, then the hypothesis with the largest support is kept.
    \\
    Other works related to RANSAC generate hypothesis from one match. Since each key point has location and shape (oriented disc in SIFT or ellipse in MSER), it is possible to compute a transformation normalizing the image patch to a unit circle in the origin (a similarity for SIFT and an affine transformation for MSER). Thus, having a matching pair of key points, it is possible to combine transformations of both key points to retrieve a transformation from one image to the other. This approach is interesting because, being the minimal sample size equal to one, and therefore the set of all possible samples very small, it allows the computation of all possible samples, reducing randomness. In particular Philbin $et$ $al.$ \cite{ref:reference_25} compute an hypothesis from a single match, find its inliers, refine it into an affine transformation using the inliers, and again find inliers. They do this for every match and keep the largest set of inliers. Vedaldi and Zisserman \cite{ref:reference_26} follow a similar approach but refine the transformation into a full homography.
    \\
    An alternative to RANSAC are Hough transform\cite{ref:reference_27}\cite{ref:reference_28} based approaches. They use all matches to vote for transformations that are consistent with them and finally select the one with the largest support. Although this allows handling higher outlier contamination, it has high memory requirements since all possible transformations have to be represented (done by discretizing space of parameters). Lowe in \cite{ref:reference_5} has utilized the Hough transform to verify SIFT matches.
    \\
    An interesting case is represented by \cite{ref:reference_1}, in which it is presented a method for object recognition and localization taking into account the geometric consistency of matched key points concurrently with their descriptor vector similarity. This approach merges geometric with photometric information in a straightforward manner and we found it inspirational for our work.
    \\
    Previous considered works extract the hypothesis with the largest support. However the presence of multiple models in a scene is possible. \cite{ref:reference_29}\cite{ref:reference_30} use RANSAC to greedily extract multiple hypotheses (i.e. find the best hypothesis, remove its inliers from the set of all matches and repeat until there is enough matches), while \cite{ref:reference_31} formulate the MultiRANSAC algorithm which estimates multiple transformations in every round. \cite{ref:reference_32} claims that exctracting the hypothesis with the most inliers in not the best strategy and design an energy-based approach called PEARL.
    \\
    RANSAC based object detection algorithms reach amazing results in terms of accuracy and execution time. Unfortunately homography is not powerful enough to describe non-planar deformations. In order to account for these situations some works has been proposed. \cite{ref:reference_41} formulates object-based image matching as an unsupervised multi-class clustering problem on a set of candidate key point matches, proposing a pairwise dissimilarity measure that exploits both photometric similarity and pairwise geometric consistency. In order to perform template matching in unconstrained environments \cite{ref:reference_42} proposes Best-Buddies Similarity, a parameter-free similarity measure between two sets of points, based on counting the number of pairs of points in source and target sets, where each point is the nearest neighbour of the other.
    \\
    In this our work we present a method able to deal with non-planar smooth deformations exploiting geometric relationship between template key points through constrained Delaunay triangulation (introduced in \cref{sec:section_2.1.2}).
    \thispagestyle{empty}
    \chapter{Proposed solution}
\label{ch:chapter_3}

\section{Problem formulation}
    \label{sec:section_4.0}
    Our goal is to detect and identify multiple instances of a template object $T$ in a scene image $S$. To this purpose we will pursue a feature-based approach. To reach our goal we are given:
    \begin{itemize}
        \item Template image $T$, depicts the object to be detected.
        \item Scene image $S$ in which $p$ occurrences of the template are present, let's call the occurrences $S_1$, $S_2$, \dots, $S_p$ ($p$ is unknown).
    \end{itemize}
    In particular we focus on the following non-trivial situation:
    \begin{quotation}
        \emph{Object depicted in image $T$ is planar and object occurrences in $S$ can be smoothly distorted and seen from different points of view.}
    \end{quotation}
    We are able now to define our goal:
    \begin{equation}
        \forall S_i \in S, \text{ identify it as an occurrence of } T.
        \label{eq:equation_2}
    \end{equation}

\section{Key points extraction}
    The first step in any features-based object detection algorithm is to extract key points and descriptors from template and scene image. In particular, key points and descriptors, are extracted considering the positions of the image features, identified by a specific detector. In this work we use SIFT\cite{ref:reference_5} algorithm for features detection, key points and descriptors extraction (described in \cref{subsubsec:subsubsec_2.1.3.3}).\\
    After the execution of SIFT algorithm, we have key points and descriptors for both template and scene image. In particular it returns:
    \begin{subequations}
        \begin{equation}
            K^T = \{k^T_1, k^T_2, \dots, k^T_n\},
            \label{eq:equation_3a}
        \end{equation}
        \begin{equation}
            K^S = \{k^S_1, k^S_2, \dots, k^S_m\}
            \label{eq:equation_3b}
        \end{equation}
    \end{subequations}
    sets of template and scene image key points respectively, and:
    \begin{subequations}
        \begin{equation}
            D^T = \{d^T_1, d^T_2, \dots, d^T_n\},
            \label{eq:equation_38a}
        \end{equation}
        \begin{equation}
            D^S = \{d^S_1, d^S_2, \dots, d^S_m\}
            \label{eq:equation_38b}
        \end{equation}
    \end{subequations}
    sets of template and scene key points descriptors.
    Where:
    \begin{equation}
        k^\ast_i = (x^\ast_i, y^\ast_i, s^\ast_i, \theta^\ast_i) \in \mathbb{R}^4
    \end{equation}
    contains the position coordinates ($x^\ast_i$, $y^\ast_i$) and scale $s^\ast_i$ where the feature i\textsuperscript{th} has been detected, together with the orientation $\theta^\ast_i$ assigned to that feature by its descriptor. While:
    \begin{equation}
        d^\ast_i \in \mathbb{R}^{128}
    \end{equation}
    is the SIFT descriptor of the i\textsuperscript{th} feature (illustrated in \cref{fig:figure_86}).
    \par
    After that, for each element in $K^S$ the corresponding key point in $K^T$ is established through the matching procedure introduced in \cref{subsubsec:subsubsec_2.1.3.4}. This step produces the following set of matches, one for each element in $K^S$:
    \begin{equation}
        M = \{m_1, m_2, \dots, m_m\}.
        \label{eq:equation_4}
    \end{equation}
    Assuming that the set $M$ is composed of matches that passed the Lowe's ratio test\cite{ref:reference_5} while the sets $K^S$ and $K^T$ of key points involved in at least one match.

\section{Delaunay triangulation as a model of the template configuration}
    \label{sec:section_4.2}
\begin{figure}[htbp]
      \begin{center}
        \includegraphics[width=0.8\textwidth]{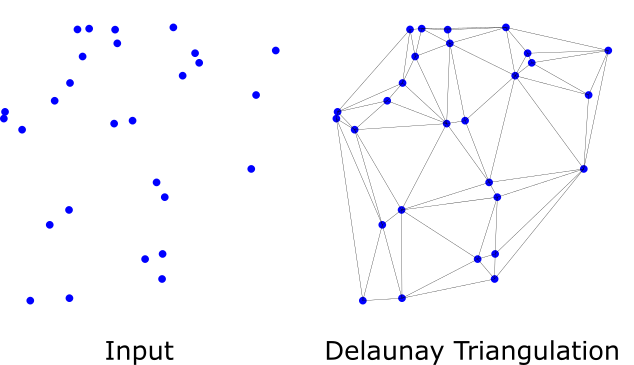}
        \caption{Example of Delaunay triangulation applied to a set of points. Figure from \cite{ref:reference_59}.}
        \label{fig:figure_1}
      \end{center}
    \end{figure}

    In order to describe non-planar deformation of the template we avoid the direct estimation of a global transformation. In particular we approach the problem through an iterative procedure enforcing local regularity constraints.
    \\
    We have identified in the Delaunay triangulation a very useful and intuitive tool to describe the mutual geometric relationship between nearby key points. \cref{fig:figure_1} shows an example of Delaunay triangulation.
    \subsection{Ideal mesh deformation}
        \label{subsec:subsection_4.2.1}
\begin{figure}
          \centering
          \begin{subfigure}[!ht]{0.28\linewidth}
            \includegraphics[width=\linewidth]{pictures/exposition_images/figure_2}
            \caption{Template Delaunay triangulation built from key points coordinates.}
            \label{fig:figure_2}
          \end{subfigure}
          \begin{subfigure}[!ht]{0.31\linewidth}
            \includegraphics[width=\linewidth]{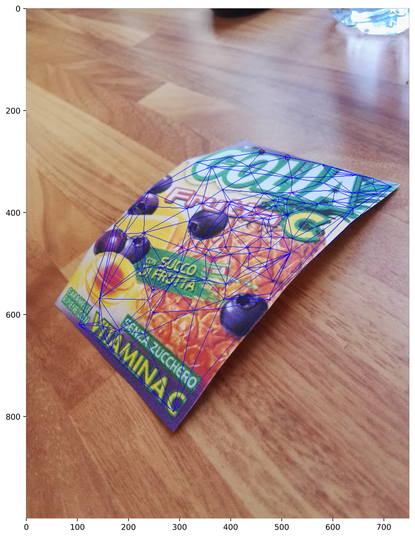}
            \caption{Scene constrained Delaunay triangulation. It is built thanks to key point matches showed in \cref{fig:figure_4}, forcing the triangulation edges (in \cref{fig:figure_2}) to be present for corresponding key points. In this way it is possible to replicate the template mesh in the scene.}
            \label{fig:figure_3}
          \end{subfigure}
          \begin{subfigure}[!ht]{0.53\linewidth}
            \includegraphics[width=\linewidth]{pictures/exposition_images/figure_4}
            \caption{In this image we can see correspondences between positions of template and scene key points, together with the triangulation in the template and its replication in scene image.}
            \label{fig:figure_4}
          \end{subfigure}
        \caption{Example of ideal situation in which it is possible to build an exact correspondence between template and scene triangulation (situation described in \cref{subsec:subsection_4.2.1}).}
        \end{figure}

        The following is a brief and practical explanation of why we believe the Delaunay triangulation is useful for our purpose.\\
        Let suppose to construct the Delaunay triangulation from coordinates (x, y) of the key points contained in $K^T$. That triangulation is a mesh lying on a plane, the template image one (\cref{fig:figure_2}). Imagine now that the object, together with its key point positions, is smoothly distorted in 3D and after that it is acquired in a new image (i.e. the scene image), producing the $K^S$ set.\\
        Suppose that an affine robust descriptor is computed at the extracted key points position and scale, we are able to match template object key points to their counterpart in the scene, generating the set $M$. Thus a constrained Delaunay triangulation can be computed from coordinates of the key points contained in $K^S$, forcing the previous triangulation edges to be present for corresponding key points (\cref{fig:figure_2}, \ref{fig:figure_3}, \ref{fig:figure_4}).\\
        In this way we are able to model local deformations of the template object as affine transformations between corresponding triangles in the previous triangulations.
        We can now re-define the \cref{eq:equation_3a}, (\ref{eq:equation_3b}) and (\ref{eq:equation_4}) for this specific and ideal case:
        \begin{subequations}
            \begin{equation}
                K^T = \{k^T_1, k^T_2, \dots, k^T_n\},
                \label{eq:equation_5a}
            \end{equation}
            \begin{equation}
                K^S = \{k^S_1, k^S_2, \dots, k^S_n\}.
                \label{eq:equation_5b}
            \end{equation}
            \label{eq:equation_5}
        \end{subequations}
        Note that we set $m = n$ underlining the fact that, in this ideal case, all object key points are found in the scene; and:
        \begin{equation}
            M = \{m_1, m_2, \dots, m_n\},
            \label{eq:equation_6}
        \end{equation}
        where
        \begin{equation}
            m_i = (k^T_i, k^S_i).
            \label{eq:equation_7}
        \end{equation}
        In which, for the sake of simplicity, we have assumed that corresponding key points have the same index in $K^T$ and $K^S$.\\
        Now we should have confidence in the usefulness of the Delaunay triangulation for our goal. However we considered an ideal case, and the reality is not so nice.
    \subsection{Challenges in the real world}
        \label{subsec:subsection_4.2.2}
        Although the previous approach is simple and straightforward, in the reality there are some issues that make it no more trivial to apply.
        \begin{itemize}
            \item \textbf{Lack of key points}. The absence of some object key point in the scene does not allow to replicate correctly the template mesh in the scene. This absence of key points can be formalized as:
            \begin{subequations}
                \begin{equation}
                    K^T = \{k^T_1, k^T_2, \dots, k^T_n\},
                    \label{eq:equation_8a}
                \end{equation}
                \begin{equation}
                    K^S = \{k^S_1, k^S_2, \dots, k^S_m\},
                    \label{eq:equation_8b}
                \end{equation}
                \label{eq:equation_8}
            \end{subequations}
            \begin{equation}
                M = \{m_1, m_2, \dots, m_m\}.
                \label{eq:equation_9}
            \end{equation}
            With $m < n$.
            \item \textbf{Wrong matches}. The presence of wrong matches in $M$ does not permit to replicate correctly the template mesh in the scene or does not permit to replicate it at all (in case of multiple matches for a single template key point). In formulas:
            \begin{subequations}
                \begin{equation}
                    K^T = \{k^T_1, k^T_2, \dots, k^T_n\}
                    \label{eq:equation_10a}
                \end{equation}
                \begin{equation}
                    K^S = \{k^S_1, k^S_2, \dots, k^S_m\}
                    \label{eq:equation_10b}
                \end{equation}
                \label{eq:equation_10}
            \end{subequations}
            \begin{equation}
                M = \{m_1, m_2, \dots, m_m\}
                \label{eq:equation_11}
            \end{equation}
            Where $m \geq n$. Supposing as before that corresponding key points have the same index, in this case it happens that
            \begin{equation}
                \exists m_i \in M \mid m_i = (k^T_j, k^S_i) \land j \neq i
                \label{eq:equation_12}
            \end{equation}
            \item \textbf{Multiple template object occurrences}. The presence of multiple object occurrences in scene image produces many true matches for each template key point. This does not permit to replicate the template mesh in the scene because the key point correspondences are not univocal. In other words:
            \begin{subequations}
                \begin{equation}
                    K^T = \{k^T_1, k^T_2, \dots, k^T_n\}
                    \label{eq:equation_13a}
                \end{equation}
                \begin{equation}
                    K^S = \{k^S_1, k^S_2, \dots, k^S_{p*m}\}
                    \label{eq:equation_13b}
                \end{equation}
                \label{eq:equation_13}
            \end{subequations}
            \begin{equation}
                M = \{m_1, m_2, \dots, m_{p*m}\}
                \label{eq:equation_14}
            \end{equation}
            Where $p$ is the number of object occurrences in the scene (as specified in \cref{sec:section_4.0}). This implies that
            \begin{equation}
                \forall l \in \{1, \dots, p\} \text{ } m_{l*i} = (k^T_i, k^S_{l*i})
                \label{eq:equation_15}
            \end{equation}
        \end{itemize}
        In the reality all these problems can be potentially present, often simultaneously.

\section{Incremental correspondences growth}
    \label{sec:section_4.3}
    In this section we present the approach and the ingredients used to reach our goal in presence of problems presented in \cref{subsec:subsection_4.2.2}.
    \subsection{The idea}
        Our approach is based on the iterative growth of the set $N$, which contains the correspondences between key points in $K^T$ and $K^S$. The growth is guided by a triangulation of the position of the key points in $T$ (\cref{fig:figure_2}); in particular we iteratively consider new triangles in order to group in a set the correspondences related to their key points. Having a set of correspondences (\cref{fig:figure_4}), it is possible to replicate the triangulation in $S$ (replicate edges between corresponding key points, \cref{fig:figure_3}). Therefore, we can say that a collateral aim is to replicate, over each occurrence of the template object in $S$, the triangulation mesh of $T$, enforcing that key points satisfy locality constraints to preserve the mutual position of features. Thus the gain is double: object localization by correspondences and deformation description by meshes.\\
        The set of grouped correspondences between key points in $K^T$ and $K^S$ is formalized in the concept of \emph{seed}.
        \subsubsection{\emph{Seed} formalization}
            \label{subsec:subsection_4.3.2}
            Given:
            \begin{subequations}
                \begin{equation}
                    B^T = \{b^T_1, b^T_2, \dots, b^T_m\} \subseteq K^T
                    \label{eq:equation_16a}
                \end{equation}
                \begin{equation}
                    B^S = \{b^S_1, b^S_2, \dots, b^S_m\} \subseteq K^S
                    \label{eq:equation_16b}
                \end{equation}
                \label{eq:equation_37}
            \end{subequations}
            A \emph{seed} is:
            \begin{subequations}
                \begin{equation}
                    N = \{n_1, n_2, \dots, n_m\} \subseteq M
                    \label{eq:equation_16c}
                \end{equation}
                \label{eq:equation_16}
            \end{subequations}
            Where
            \begin{equation}
                n_i = (b^T_i, b^S_i) \text{, is a match.}
                \label{eq:equation_17}
            \end{equation}
            Since the situation in which $m = 3$ is the most recurrent in the following sections, we appropriate to give a specific name to the \emph{seed} in that case: \emph{matching triangle}.\\
        \\
        In light of the concept of \emph{seed}, we can now explain our approach. The method can be divided in two fundamental parts:
        \begin{enumerate}
            \item Initial \emph{seed} selection
            \item \emph{Seed} expansion
        \end{enumerate}
        \subsubsection{Initial \emph{seed} selection}
            \label{subsubsec:subsubsection_4.3.1.1}
            The incremental approach suggests starting from the minimal component of a triangulation: the triangle. Thus the starting point is to identify a triangle of key points in $S$ that corresponds to a triangle of key points in $T$.  The explicit correspondences between these two triplets of key points is called \emph{matching triangle} and it is formalized in \cref{subsec:subsection_4.3.2}.
            To do so consider for a moment the ideal case in which no one of the problems in \cref{subsec:subsection_4.2.2} is present (i.e. assume the conditions described in \cref{subsec:subsection_4.2.1}). In this case it is sufficient to choose whatever triangle of the triangulation in $T$ and there exists a corresponding triangle in $S$ that we could easily find by selecting the corresponding key points.
            Unfortunately, the problems highlighted in \cref{subsec:subsection_4.2.2} are present and we have to deal with them, thus we designed specific solutions:
            \begin{itemize}
                \item \textbf{Redundant triangulation}. This solution addresses the problem \emph{lack of key points}. In particular, instead of search for triangles in $S$ starting from triangles of the plain Delaunay triangulation in $T$; we consider some expansions of each triangle in $T$, taking into account neighbours of its vertices (we consider the triangulation as a graph) and composing triplets of them (examples of expanded triangles in \cref{fig:figure_5}). Starting from these triplets we then look for the corresponding triangles in $S$.  In this way we can deal with the absence of some key points in $K^S$.
                \begin{figure}
                  \centering
                  \begin{subfigure}[!ht]{0.4\linewidth}
                    \includegraphics[width=\linewidth]{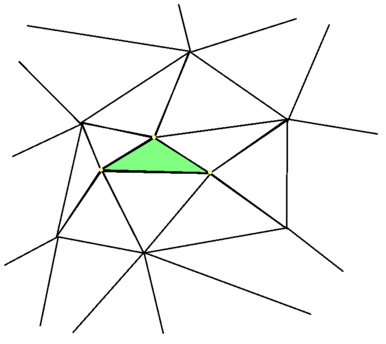}
                    \caption{Triangle of triangulation in $T$.}
                    \label{fig:figure_5a}
                  \end{subfigure}
                  \\
                  \begin{subfigure}[!ht]{0.32\linewidth}
                    \includegraphics[width=\linewidth]{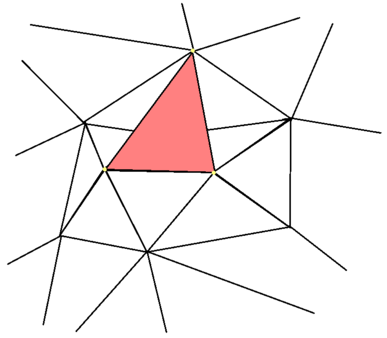}
                  \end{subfigure}
                  \begin{subfigure}[!ht]{0.32\linewidth}
                    \includegraphics[width=\linewidth]{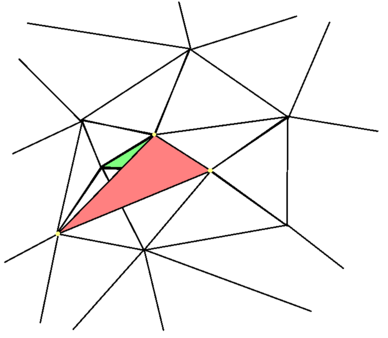}
                  \end{subfigure}
                  \begin{subfigure}[!ht]{0.32\linewidth}
                    \includegraphics[width=\linewidth]{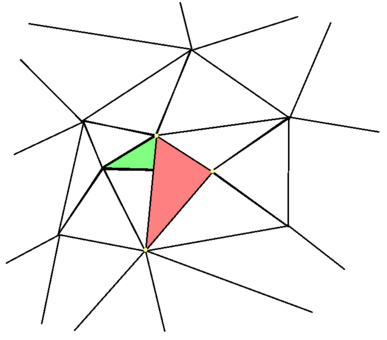}
                  \end{subfigure}
                  \begin{subfigure}[!ht]{0.32\linewidth}
                    \includegraphics[width=\linewidth]{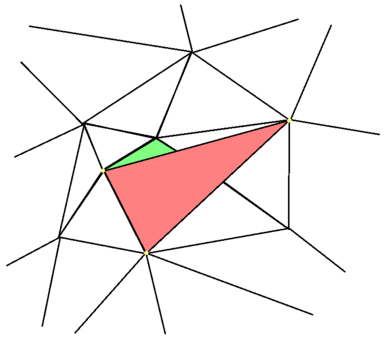}
                  \end{subfigure}
                  \begin{subfigure}[!ht]{0.32\linewidth}
                    \includegraphics[width=\linewidth]{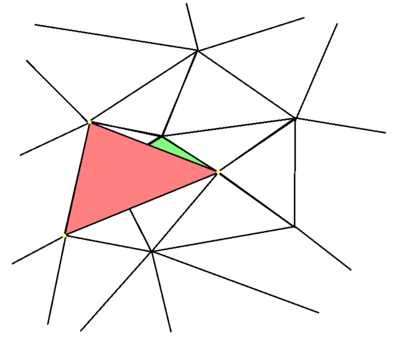}
                  \end{subfigure}
                  \begin{subfigure}[!ht]{0.32\linewidth}
                    \includegraphics[width=\linewidth]{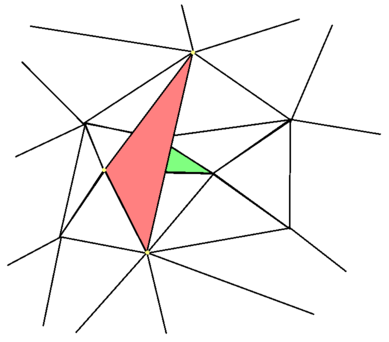}
                  \end{subfigure}
                  \begin{subfigure}[!ht]{0.32\linewidth}
                    \includegraphics[width=\linewidth]{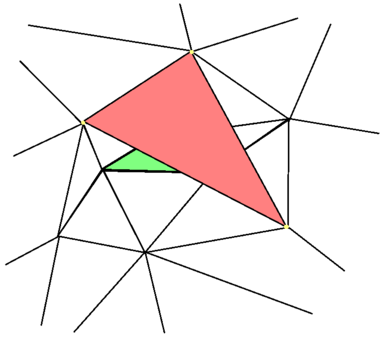}
                  \end{subfigure}
                  \begin{subfigure}[!ht]{0.32\linewidth}
                    \includegraphics[width=\linewidth]{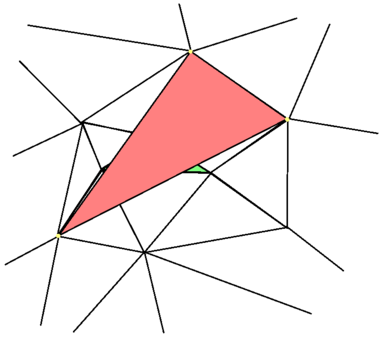}
                  \end{subfigure}
                  \caption{In order to be robust to the lack of key points in $K^S$, some expansions of the triangle in \cref{fig:figure_5a} are extracted (the considered triangle belongs to the triangulation in $T$). This is done taking into account neighbouring vertices (we consider the triangulation in $T$ as a graph) and building triplets of them. Triangulation in \textcolor{black}{$black$}, one triangle of the triangulation in \textcolor{green}{$green$} and expanded triangles in \textcolor{red}{$red$}.}
                  \label{fig:figure_5}
                \end{figure}
                \item \textbf{Consistency scores}. This solution addresses the problems \emph{wrong matches} and \emph{multiple object occurrences}. Both these problems results in the possibility that a triangle in $T$ corresponds to more triangles in $S$, many of them wrong. To solve this issue we can exploit the information embedded in key points and their descriptors. In \cref{subsec:subsection_4.3.5} we introduce consistency scores that are useful for this goal. They assign an higher score to \emph{matching triangles} whose key points preserve a certain consistency between the images.
            \end{itemize}
            We can now decompose the procedure of initial \emph{seed} selection in the following steps:
            \begin{enumerate}
                \item \emph{Replace the Delaunay triangulation of key points in $T$ with a redundant coverage}. As explained in the \emph{redundant triangulation} point, this step consists in the creation of a redundant triangulation of key points in $T$.  In particular triangles are no more disjointed but overlapped (\cref{fig:figure_5}).
                \item \emph{For each triangle in $T$ the corresponding ones in $S$ are identified, and the matching triangles are composed}. Since it is possible that for every key point in $K^T$ there are many matching key points in $K^S$, this could lead to the fact that there are many triangles in $S$ for each triangle in $T$. Therefore at this step are created all the \emph{matching triangles} that can be inferred from each triangle in $T$. 
                \item \emph{Evaluation of consistency score for each matching triangle}. As explained in the \emph{consistency scores} point, at this step it is evaluated the goodness of \emph{matching triangles} (through scores introduced in \cref{subsec:subsection_4.3.5}).
                \item \emph{Selection of the best matching triangle}. The \emph{matching triangle} with the highest score is selected as the initial \emph{seed}.
            \end{enumerate}
            After this phase we have our initial \emph{seed} that is composed by a triplet of key point matches (as formalized in \cref{subsec:subsection_4.3.2}).
        \subsubsection{\emph{Seed} expansion}
            \label{subsubsec:subsubsection_4.3.1.2}
\begin{figure}[!ht]
                \centering
                \includegraphics[width=0.32\linewidth]{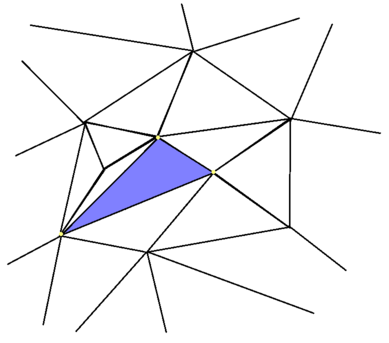}
                \caption{Constrained Delaunay triangulation computed on a new set of key points, forcing boundary sides of the triangle associated to the \emph{seed} to be present. In particular the key points whose position intersect with that triangle (except the triangle vertices) has been eliminated before computing the triangulation (for comparison see \cref{fig:figure_5a}).}
                \label{fig:figure_6a}
            \end{figure}

\begin{figure}[!ht]
                \centering
                \includegraphics[width=0.32\linewidth]{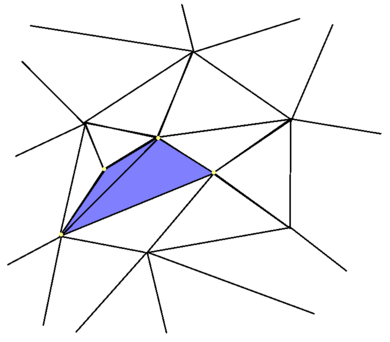}
                \caption{Constrained Delaunay triangulation computed after one iteration of \emph{seed} expansion. The boundary sides of the convex hull associated to the \emph{seed} has been forced to be present in the triangulation.}
                \label{fig:figure_7}
            \end{figure}

            The purpose of the \emph{seed} expansion phase is to add to the \emph{seed} additional correspondences between key points in $K^T$ and $K^S$. The growth of the \emph{seed} is guided by the triangulation of key points position in $T$.\\
            In the first expansion iteration the initial \emph{seed} can refer to an expanded triangle (this can be observed in \cref{fig:figure_5}), it can happen that no triangle of the triangulation in $T$ is associated to our \emph{seed}. To solve this issue we can simply eliminate the key points in $K^T$ whose position intersect with the triangle in $T$ associated to our \emph{seed} (except the triangle vertices) and compute a constrained Delaunay triangulation forcing the boundary sides of that triangle to be present (\cref{fig:figure_6a}).\\
            Analogously, in the subsequent expansion iterations, it can happen that the boundary of the convex hull generated from key points in $B^T$ (introduced in \cref{subsec:subsection_4.3.2} in \cref{eq:equation_16a}) does not correspond to edges of the triangulation generated from key points in $K^T$. Thus the key points in $K^T$ that intersect with the convex hull are eliminated (except the ones in $B^T$), then a constrained Delaunay triangulation is computed forcing the convex hull boundary sides to be present (\cref{fig:figure_7}).
            \par
            After the computation of the constrained Delaunay triangulation, let us consider again for a moment the ideal case in which no one of the problems in \cref{subsec:subsection_4.2.2} is present (i.e. assume the conditions described in \cref{subsec:subsection_4.2.1}). Here we can focus on the triangulation in $T$, in particular on the convex hull generated from key points in $B^T$. We can identify in the adjacent triangles (those with a side in common with it; for example in \cref{fig:figure_6a} we can see three triangles with a side in common with the highlighted one) the ones from which to retrieve the correspondences to be added to our \emph{seed}.
            \par
            Unfortunately we have to deal with the previous problems, therefore we present specific solutions:
            \begin{itemize}
                \item \textbf{Redundant adjacent triangles}. Our aim is to expand the \emph{seed}, and in the ideal case we saw that it is possible to directly group correspondences from triangles adjacent to the convex hull generated from key points in $B^T$. However, the problem \emph{lack of key points} is present and we address it. Therefore we search for expanded versions of the adjacent triangles. In particular, key points in common between the convex hull and an adjacent triangle are fixed. The third vertex, on the other hand, is chosen among the neighbours of the last vertex of the adjacent triangle (examples of adjacent expanded triangles in \cref{fig:figure_6}). Starting from these triplets we then look for the corresponding triangles in $S$. In this way we can deal with the absence of some key points in $K^S$.
                \begin{figure}
                  \centering
                  \begin{subfigure}[!ht]{0.4\linewidth}
                    \includegraphics[width=\linewidth]{pictures/exposition_images/figure_6a}
                    \caption{Convex hull generated in $T$ from key points of $B^T$.}
                    \label{fig:figure_6aa}
                  \end{subfigure}
                  \\
                  \begin{subfigure}[!ht]{0.32\linewidth}
                    \includegraphics[width=\linewidth]{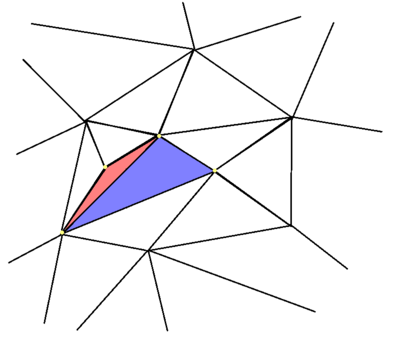}
                  \end{subfigure}
                  \begin{subfigure}[!ht]{0.32\linewidth}
                    \includegraphics[width=\linewidth]{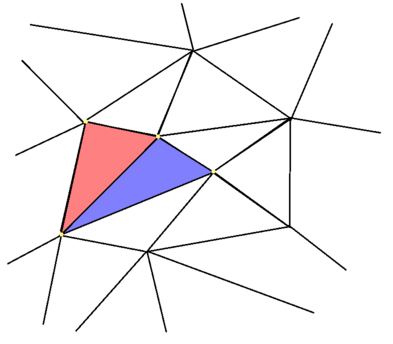}
                  \end{subfigure}
                  \begin{subfigure}[!ht]{0.32\linewidth}
                    \includegraphics[width=\linewidth]{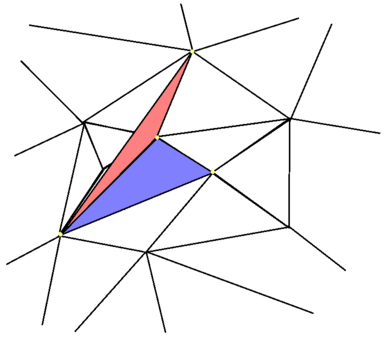}
                  \end{subfigure}
                  \begin{subfigure}[!ht]{0.32\linewidth}
                    \includegraphics[width=\linewidth]{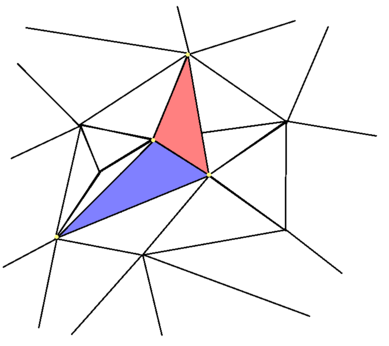}
                  \end{subfigure}
                  \begin{subfigure}[!ht]{0.32\linewidth}
                    \includegraphics[width=\linewidth]{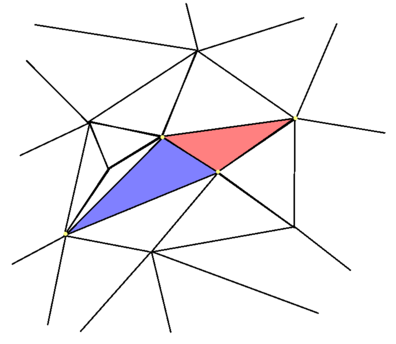}
                  \end{subfigure}
                  \begin{subfigure}[!ht]{0.32\linewidth}
                    \includegraphics[width=\linewidth]{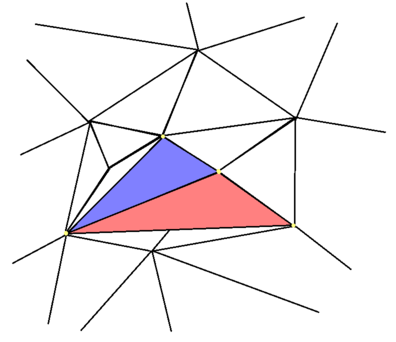}
                  \end{subfigure}
                  \begin{subfigure}[!ht]{0.32\linewidth}
                    \includegraphics[width=\linewidth]{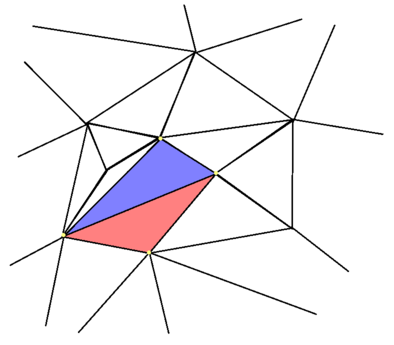}
                  \end{subfigure}
                  \caption{In order to be robust to the lack of key points in $K^S$, some expanded triangles adjacent to \cref{fig:figure_6aa} are extracted. This is done by fixing the key points in common between the convex hull and an adjacent triangle. Then the third vertex is chosen among the neighbours of the last vertex of the adjacent triangle. The procedure is done many times for each side of the convex hull, generating the triangles shown in the figure. Triangulation in \textcolor{black}{$black$}, convex hull in \textcolor{blue}{$blue$}, expanded triangles in \textcolor{red}{$red$}.}
                  \label{fig:figure_6}
                \end{figure}
                \item \textbf{Consistency scores and geometric checks}. This solution addresses the problems \emph{wrong matches} and \emph{multiple object occurrences}. Both these problems results in the possibility that a triangle in $T$ corresponds to more triangles in $S$, many of them wrong. To solve this issue we can exploit the information embedded in key points and their descriptors. In \cref{subsec:subsection_4.3.3} and \ref{subsec:subsection_4.3.4} we introduce geometric checks that discard \emph{matching triangles} geometrically incoherent with the \emph{seed}; while in \cref{subsec:subsection_4.3.5} we introduce consistency scores that assign an higher value to \emph{matching triangles} whose key points preserve a certain consistency between the images.
            \end{itemize}
            The expansion phase is iterated many times, until no growth happens. After each expansion iteration we have our \emph{seed} composed by an additional correspondence between key points in $K^T$ and $K^S$. In other words, the cardinality of the set $N$ (introduced in \cref{subsec:subsection_4.3.2} in \cref{eq:equation_16c}) increases by one.\\
            We decompose the procedure of \emph{seed} expansion in the following steps:
            \begin{enumerate}
                \item \emph{Update of the constrained Delaunay triangulation in $T$}. Key points whose position intersect with the convex hull in $T$ generated from points associated to our \emph{seed} are eliminated. Then a constrained Delaunay triangulation is computed forcing the boundary sides of the convex hull to be present.
                \item \emph{Extraction of expanded triangles from those adjacent to the convex hull in $T$ associated to the \emph{seed}}. As explained in the \emph{redundant adjacent triangles} point, this step consists in the creation of redundant triangles of key points in $T$. Triangles overlap with the triangulation (\cref{fig:figure_6}).
                \item \emph{For each triangle in $T$ the corresponding ones in $S$ are identified, and the matching triangles are composed}.
                \item \emph{Evaluation of geometric checks for each matching triangle}. As explained in the \emph{consistency scores and geometric checks} point, at this step it is evaluated the geometric coherence of \emph{matching triangles} (through scores introduced in \cref{subsec:subsection_4.3.3} and \ref{subsec:subsection_4.3.4}.
                \item \emph{Evaluation of consistency score for each matching triangle}.
                \item \emph{Selection of the best matching triangle}. The \emph{matching triangle} with the highest score is kept.
                \item \emph{Seed updating}. The correspondences of the best \emph{matching triangle} are added to the \emph{seed} (the ones that are not already part of it).
                \item Go to step 1.
            \end{enumerate}
        These are the principles of our approach. In \cref{subsec:subsection_4.3.3} and \ref{subsec:subsection_4.3.4} we introduce the two geometric checks, while the consistency score is presented in \cref{subsec:subsection_4.3.5}. The formal algorithm is then described in \cref{subsec:subsection_4.3.6}.
    \subsection{Geometric checks}
        \label{subsec:subsection_4.3.3}
        Every time that a \emph{matching triangle} is composed in the \emph{seed expansion} phase, its geometric coherence with the \emph{seed} has to be checked.\\
        We divide the \emph{geometric checks} in two steps:
        \begin{enumerate}
            \item Non-intersection check
            \item Local coherence check
        \end{enumerate}
        \subsubsection{Non-intersection check}
            \label{subsubsec_subsubsection_4.3.3.1}
\begin{figure}[!ht]
                \centering
                \includegraphics[width=0.5\linewidth]{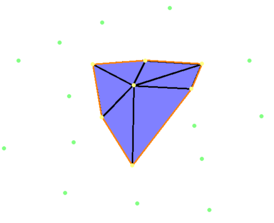}
                \caption{Key points related to the \emph{seed} in \textcolor{yellow}{$yellow$}, key points not related to the \emph{seed} in \textcolor{green}{$green$}, convex hull sides in \textcolor{orange}{$orange$}.}
                \label{fig:figure_8}
            \end{figure}

\begin{figure}
              \centering
              \begin{subfigure}[!ht]{0.32\linewidth}
                \includegraphics[width=\linewidth]{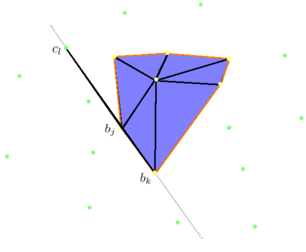}
                \caption{$c_l$ is on the line connecting $b^T_j$ and $b^T_k$.}
                \label{fig:figure_9a}
              \end{subfigure}
              \begin{subfigure}[!ht]{0.32\linewidth}
                \includegraphics[width=\linewidth]{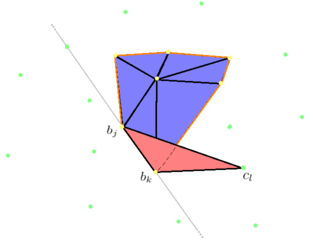}
                \caption{$c_l$ is in the half-plane containing the convex hull.}
                \label{fig:figure_9b}
              \end{subfigure}
              \begin{subfigure}[!ht]{0.32\linewidth}
                \includegraphics[width=\linewidth]{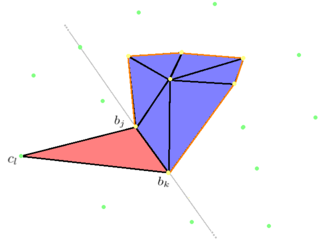}
                \caption{$c_l$ is in the half-plane that does not contain the convex hull.}
                \label{fig:figure_9c}
              \end{subfigure}
            \caption{Example of the three situations}
            \label{fig:figure_9}
            \end{figure}

            Both the \emph{seed} and a \emph{matching triangle} can be projected in $T$ and $S$, in the sense that it is possible to extract the key points from $K^T$ and $K^S$ which are related from correspondences of the \emph{seed} or the \emph{matching triangle}. Then we consider the \emph{projection polygon} as the convex hull generated from the key points extracted in this way.\\
            Non-intersection check consists in the exclusion of \emph{matching triangles} whose \emph{projection polygon} in $T$ or in $S$ intersects with the \emph{projection polygon} of the \emph{seed} in the same image.\\
            We focus only on $T$, the concept is identical for $S$. Let introduce formal ingredients through which explain the concept:
            \begin{subequations}
                \begin{equation}
                    B^T \subseteq K^T
                    \label{eq:equation_18a}
                \end{equation}
                set of key points related to a \emph{seed} match, as defined in \cref{eq:equation_16a};
                \begin{equation}
                    S = \{s_1, s_2, \dots, s_o\}
                    \label{eq:equation_18c}
                \end{equation}
                where
                \begin{equation}
                    s_i = (b^T_i, b^T_j)
                \end{equation}
                set of sides of the convex hull generated from the position of key points in $B^T$;
                \begin{equation}
                    C = \{c_1, c_2, \dots, c_n\} = K^T \setminus B^T 
                    \label{eq:equation_18b}
                \end{equation}
                set of key points in $K^T$ that are not related to any match of the \emph{seed}.
                \label{eq:equation_18}
            \end{subequations}
            \\
            Notice that any possible \emph{matching triangle} is related to two points of $B^T$ and one point of $C$. The \cref{fig:figure_8} illustrates this situation.
            Focusing on a certain convex hull side $s_i \in S$ we can identify triangles composed by the two key points of the side $b^T_j$, $b^T_k \in B^T$, and an external one, $c_l \in C$.\\
            We recognize three different situations, depending on the position of $c_l$ with respect to the line connecting $b^T_j$ and $b^T_k$:
            \begin{enumerate}
                \item $c_l$ is on the line connecting $b^T_j$ and $b^T_k$ (\cref{fig:figure_9a}).
                \item $c_l$ is in the half-plane containing the convex hull (\cref{fig:figure_9b}).
                \item $c_l$ is in the half-plane that does not contain the convex hull (\cref{fig:figure_9c}).
            \end{enumerate}
            In \cref{fig:figure_9a} the triangle is degenerate, thus the \emph{matching triangle} is discarded. In \cref{fig:figure_9b} the triangle intersects with the convex hull and the \emph{matching triangle} is discarded. These eliminations are motivated by the fact that it is always possible to choose another side $s_j \in S$ through which the triangle generated with $c_l$ is not degenerate and does not intersect the convex hull. \cref{fig:figure_9c} represents the only case accepted.
    \subsubsection{Local coherence check}
        \label{subsec:subsection_4.3.4}
\begin{figure}[!ht]
            \centering
            \includegraphics[width=0.5\linewidth]{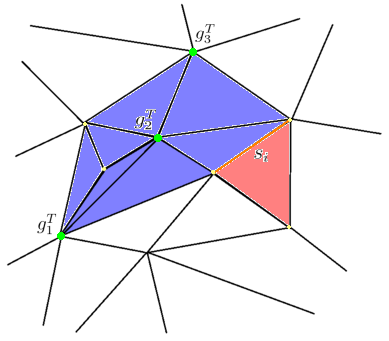}
            \caption{The local coherence of a \emph{matching triangle} is evaluated taking into account, in $T$, the key points that are close to the common side between the convex hull of the \emph{seed} and the boundary of the \emph{matching triangle}. Triangulation in \textcolor{black}{$black$}, triangles of the \emph{seed} in \textcolor{blue}{$blue$}, triangle of the \emph{matching triangle} in \textcolor{red}{$red$}, common side in \textcolor{orange}{$orange$}, close key points (elements of $G^T$) in \textcolor{green}{$green$}.}
            \label{fig:figure_90}
        \end{figure}

        The purpose of the \emph{local coherence check} is to test the coherence, in a local context, of a \emph{matching triangle} candidate for the \emph{seed} expansion (expansion procedure explained in \cref{subsubsec:subsubsection_4.3.1.2}).\\
        In \cref{subsubsec_subsubsection_4.3.3.1} we said that both the \emph{seed} and a \emph{matching triangle} can be projected in $T$ and $S$. In particular, given:
        \begin{equation}
            N \text{, the \emph{seed}}
        \end{equation}
        and
        \begin{equation}
            N' \text{, the \emph{matching triangle}}
        \end{equation}
        we can identify:
        \begin{subequations}
            \begin{equation}
                B^T \text{, the projection of the \emph{seed} in $T$;}
            \end{equation}
            \begin{equation}
                B^S \text{, the projection of the \emph{seed} in $S$;}
            \end{equation}
            \begin{equation}
                B^{T\prime} \text{, the projection of the \emph{matching triangle} in $T$;}
            \end{equation}
            \begin{equation}
                B^{S\prime} \text{, the projection of the \emph{matching triangle} in $S$.}
            \end{equation}
        \end{subequations}
        Given the constrained Delaunay triangulation computed from key points in $K^T$, it is important to remember that (as explained in \cref{subsubsec:subsubsection_4.3.1.2}) the \emph{matching triangle} $N'$ has been built from the key points of a triangle adjacent to the convex hull in $T$ associated to the \emph{seed}. Thus we are sure that exists an edge (or side, as specified in \cref{eq:equation_18c}) of the triangulation that belongs to both convex hull of the \emph{seed} and boundary of the \emph{matching triangle}. We name this specific edge as $s_i$.\\
        We test the local coherence of the \emph{matching triangle} $N'$ by evaluating to what extent key points related to some correspondences of the \emph{seed} $N$ are geometrically compliant with the affine transformation that maps key points from $B^{T\prime}$ to $B^{S\prime}$. The key points of $B^T$ and $B^S$ of which we want to check compliance are those geometrically close to the key points of $B^{T\prime}$ and $B^{S\prime}$. We define the set of key points to check as:
        \begin{subequations}
            \begin{equation}
                G^T = \{g^T_1, g^T_2, \dots, g^T_n\} \subset B^T
            \end{equation}
            and
            \begin{equation}
                G^S = \{g^S_1, g^S_2, \dots, g^S_n\} \subset B^S.
            \end{equation}
        \end{subequations}
        In particular, consider the constrained Delaunay triangulation computed from $K^T$ as a graph; $G^T$ is the set of key points of $B^T$ at distance 1 from key points of $B^{T\prime}$ (concept shown in \cref{fig:figure_90}). From $G^T$ are excluded key points lying on $s_i$. $G^S$, on the other hand, is the set of key points in $K^S$ related to those of $G^T$ by the correspondences of the \emph{seed}, in fact:
        \begin{equation}
            \forall g^S_i \in G^S \quad \exists n_i \in N \mid n_i = (g^T_i, g^S_i).
        \end{equation}
        The affine transformation $\mathcal{A}$, through which check the compliance of the key points in $G^T$ and $G^S$, is the one such that:
        \begin{equation}
            \forall b^{T\prime}_i \in B^{T\prime} \quad b^{S\prime}_i = \mathcal{A}*b^{T\prime} \text{.}
        \end{equation}
        The compliance is checked through the normalized median back-projection error of the mapping of the key points of $G^T$ in their correspondents in $G^S$ through $\mathcal{A}$ (correspondences are in $N$). In particular, given:
        \begin{equation}
            e_i = \|{g^S_i - \mathcal{A}*g^T_i}\| \text{, the single key point back-projection error}
            \label{eq:equation_20}
        \end{equation}
        the median back-projection error is:
        \begin{equation}
            \tilde{e} = Med(e_1, e_2, \dots, e_n),
            \label{eq:equation_19}
        \end{equation}
        where $n$ is the cardinality of both $G^T$ and $G^S$.
        $\tilde{e}$ is then normalized using a sigmoid:
        \begin{equation}
            \hat{\tilde{e}} = sigmoid(\tilde{e}) = \frac{1}{1 + e^{(\tilde{e}-t)*s}}
            \label{eq:equation_21}
        \end{equation}
        with $t = 10$ and $s = 0.5$.\\
        Whether $\hat{\tilde{e}}$ does not reach a value of $0.7$, the \emph{matching triangle} considered is discarded. The numerical values of $t$ and $s$ are chosen in order to produce good experimental results.\\
        \par
        We introduced the \emph{geometric checks} (respectively in \cref{subsec:subsection_4.3.3} and \ref{subsec:subsection_4.3.4}) that we apply to candidate \emph{matching triangles} in order to filter out those ones that are geometrically incoherent (or poorly coherent) with the \emph{seed}.\\
        In \cref{subsec:subsection_4.3.5} we introduce consistency scores used to identify the best candidate \emph{matching triangle} according to information embedded in key points of $B^{T\prime}$ and $B^{S\prime}$.
    \subsection{Consistency scores}
        \label{subsec:subsection_4.3.5}
        \emph{Consistency scores} take into account both geometric and photometric properties of key points to evaluate the goodness of a \emph{matching triangle}.\\
        We use \emph{consistency scores} proposed by \cite{ref:reference_1}, plus an additional one that is meant to evaluate the consistency of the scale to which the key points are extracted. 
        It is important to remember that:
        \begin{equation}
            N' = {n^{\prime}_1, n^{\prime}_2, n^{\prime}_3}
        \end{equation}
        is the \emph{matching triangle},
        \begin{subequations}
            \begin{equation}
                B^{T\prime} = {b^{T\prime}_1, b^{T\prime}_2, b^{T\prime}_3}
            \end{equation}
            \begin{equation}
                B^{S\prime} = {b^{S\prime}_1, b^{S\prime}_2, b^{S\prime}_3}
            \end{equation}
        \end{subequations}
        are the projections of the \emph{matching triangle} in $T$ and $S$ respectively,
        \begin{equation}
            b^{\ast\prime}_i = (x^{\ast\prime}_i, y^{\ast\prime}_i, s^{\ast\prime}_i, \theta^{\ast\prime}_i) \in \mathbb{R}^4
        \end{equation}
        is a key point of the set $B^{T\prime}$ or $B^{S\prime}$ and
        \begin{equation}
            d^{\ast\prime}_i \in \mathbb{R}^{128}
        \end{equation}
        is the SIFT local descriptor of the key point $b^{\ast\prime}_i$.\\
        The scores presented in this section exploit the knowledge of key points position ($x^{\ast\prime}_i, y^{\ast\prime}_i$), orientation $\theta^{\ast\prime}_i$, scale $s^{\ast\prime}_i$ and descriptor $d^{\ast\prime}_i$ in order to prefer \emph{matching triangles} whose projections $B^{T\prime}$ and $B^{S\prime}$ preserve some properties between $T$ and $S$.\\
        We highlight the fact that not all the \emph{consistency scores} are used both in \emph{initial seed selection} and \emph{seed expansion} phase. In particular a reduced set of them is used in the \emph{seed expansion} phase, while the whole set is used in the \emph{initial seed selection} phase. The motivation behind the inclusion or not of a certain score is given directly in the considered score section.\\
        The \emph{consistency scores} are:
        \begin{itemize}
            \item[--] Description vector consistency score,
            \item[--] Position consistency score,
            \item[--] Orientation consistency score,
            \item[--] Scale-ratio consistency score,
        \end{itemize}
        and they will be presented in \cref{subsubsec:subsubsection_4.3.5.1}, \ref{subsubsec:subsubsection_4.3.5.2}, \ref{subsubsec:subsubsection_4.3.5.3}, \ref{subsubsec:subsubsection_4.3.5.4} respectively.
        \subsubsection{Description vector consistency score}
            \label{subsubsec:subsubsection_4.3.5.1}
            \begin{figure}[!ht]
                \centering
                \includegraphics[width=\linewidth]{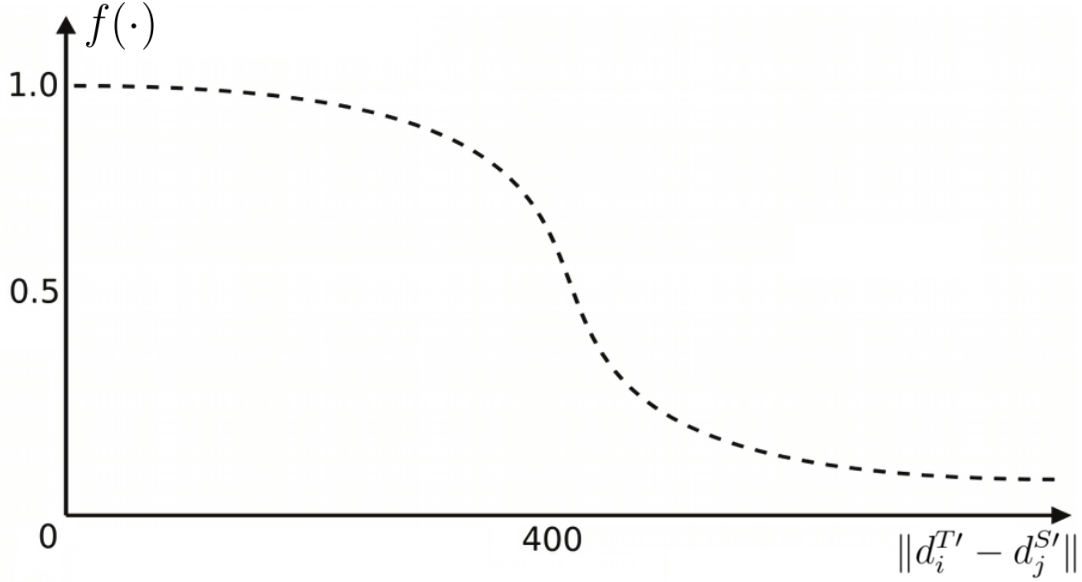}
                \caption{The function $f(\cdot)$ computes the probability $p_{ij}$ that a pair of SIFT key points $(b^{T\prime}_i, b^{S\prime}_j)$ represents a correct match. The output of $f(\cdot)$ varies with the Euclidean distance $\| d^{T\prime}_i - d^{S\prime}_j \|$ between description vectors of the key points. Figure from \cite{ref:reference_1}, and adapted to our notation.}
                \label{fig:figure_10}
            \end{figure}
            The explanation of description vector consistency score was taken largely from \cite{ref:reference_1} and adapted to our notation.\\
            Each SIFT key point is related to a 128 dimensional description vector, which is a scale and rotation invariant description of the pixel neighbourhood around the interest point. Additionally, this description is robust to lighting changes as the pixel intensity histogram, which forms the description vector, is normalized. This description vector is the information through which SIFT key points are matched between images.\\
            If two key points, one is $T$ and one in $S$, have similar description vectors, they are likely to be centered on the same point of the object.\\
            Similarity of SIFT key points is normally determined by the Euclidean distance between their description vectors, for example given two key points $b^{T\prime}_i$ and $b^{S\prime}_j$ the description vector distance is given by:
            \begin{equation}
                \| d^{T\prime}_i - d^{S\prime}_j \| = \sqrt{\sum_{i=1}^{128} (d^{T\prime}_i - d^{S\prime}_j)^2}.
            \end{equation}
            where $d^{T\prime}_i, d^{S\prime}_j \in \mathbb{R}^{128}$.
            In general, the lower the descriptor distance between two key points, the more likely they are to match. To determine the SIFT description vector consistency we need to quantify this relationship. In \cite{ref:reference_1} it is suggested to define a probability density function that describes the likelihood that a match between two key points (e.g. $b^{T\prime}_i$ and $b^{S\prime}_j$) is correct, in function of the distance $\| d^{T\prime}_i - d^{S\prime}_j \|$ between their description vectors. To do this the authors of \cite{ref:reference_1} recorded the descriptor distances between key points with similar orientation on the same point in different images of the same scene, under varying lighting conditions; for non-matching key points many images of different random scenes were taken and the descriptor distances between all pairs of key points recorded. This data resulted in a function
            \begin{equation}
                f(\| d^{T\prime}_i - d^{S\prime}_j \|) = \frac{1}{1 + e^{(\| d^{T\prime}_i - d^{S\prime}_j \|-\mu)*\sigma}} = p_{ij},
            \end{equation}
            with $\mu = 400$ and $\sigma = 0.015$; which given the Euclidean distance $\| d^{T\prime}_i - d^{S\prime}_j \|$ between key point description vectors returns the probability $p_{ij}$ that their match is correct. This function has a sigmoid shape and it is shown in \cref{fig:figure_10}. The numerical value of $\mu$ and $\sigma$ comes from \cite{ref:reference_1} in which it was determined empirically to give good results.\\
            The description vector consistency score $\mathcal{DV}$ of a \emph{matching triangle} $N'$ is equal to the average $p_{ij}$ value of corresponding key points:
            \begin{equation}
                \mathcal{DV}(N') = \frac{p_{11} + p_{22} + p_{33}}{3}
            \end{equation}
        \subsubsection{Position consistency score}
            \label{subsubsec:subsubsection_4.3.5.2}
            \begin{figure}[!ht]
                \centering
                \includegraphics[width=\linewidth]{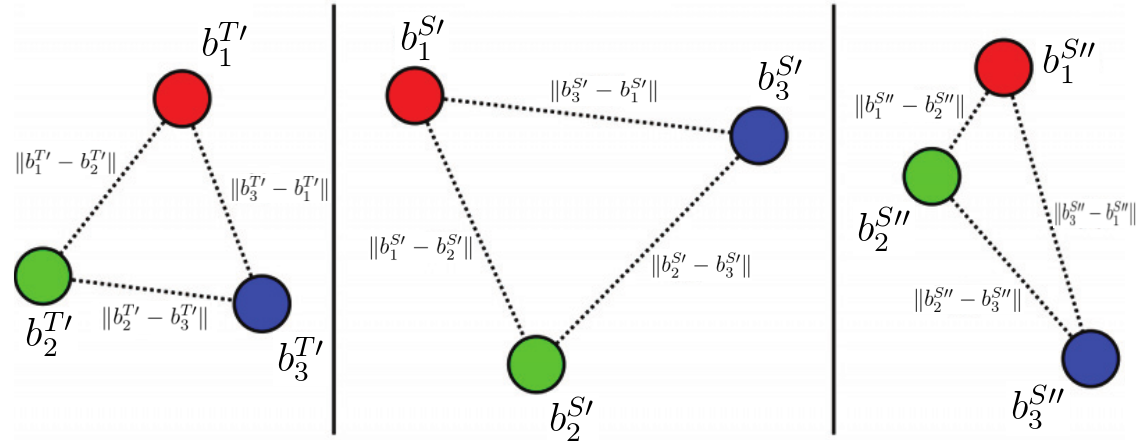}
                \caption{The triplet of matching key point pairs $(b^{T\prime}_1, b^{S\prime}_1)$, $(b^{T\prime}_2, b^{S\prime}_2)$, $(b^{T\prime}_3, b^{S\prime}_3)$ shows position consistency, whereas $(b^{T\prime}_1, b^{S\prime\prime}_1)$, $(b^{T\prime}_2, b^{S\prime\prime}_2)$, $(b^{T\prime}_3, b^{S\prime\prime}_3)$ not. Figure from \cite{ref:reference_1}, and adapted to our notation.}
                \label{fig:figure_11}
            \end{figure}
            The explanation of position consistency score was taken largely from \cite{ref:reference_1} and adapted to our notation.\\
            Position consistency refers to the fact that key points in $T$ should have the same positions relative to one another as the corresponding key points in $S$. This is a form of homographic consistency. Position consistency is defined for triplets of matching key point pairs. A matching key point pair is a tuple $(b^{T\prime}_i, b^{S\prime}_i)$.\\
            Let the triplet of matching key point pairs be $(b^{T\prime}_1, b^{S\prime}_1)$, $(b^{T\prime}_2, b^{S\prime}_2)$, and $(b^{T\prime}_3, b^{S\prime}_3)$; the authors of \cite{ref:reference_1} determine the consistency of this triplet by first finding the perimeters of the triangles formed by the $b^{T\prime}_1, b^{T\prime}_2, b^{T\prime}_3$ and $b^{S\prime}_1, b^{S\prime}_2, b^{S\prime}_3$: $p^{T\prime}$ and $p^{S\prime}$ respectively. Then the normalized sides length of the two triangles is computed by taking the length of the sides and scaling it to the perimeter of the respective triangle, e.g. indicating\\
            \begin{equation}
                \|b^{T\prime}_1 - b^{T\prime}_2\| = \left\|\begin{bmatrix}
                                                               x^{T\prime}_1 \\
                                                               y^{T\prime}_1
                                                           \end{bmatrix} - \begin{bmatrix}
                                                                               x^{T\prime}_2 \\
                                                                               y^{T\prime}_2
                                                                           \end{bmatrix}\right\|
            \end{equation}
            the normalized side length is
            \begin{equation}
                \widehat{\|b^{T\prime}_1 - b^{T\prime}_2\|} = \frac{\|b^{T\prime}_1 - b^{T\prime}_2\|}{p^{T\prime}}.
            \end{equation}
            For a position consistent \emph{matching triangle} the difference between corresponding normalized sides length of $T$ and $S$ will be small. The purpose of normalizing the sides length is to account for varying scales and image resolution. \cref{fig:figure_11} shows cases of position consistent and position inconsistent triplet of matching key point pairs; for example we have that:
            \begin{equation}
                \widehat{\|b^{T\prime}_1 - b^{T\prime}_2\|} \sim \widehat{\|b^{S\prime}_1 - b^{S\prime}_2\|} \text{ holds},
            \end{equation}
            whereas
            \begin{equation}
                \widehat{\|b^{T\prime}_1 - b^{T\prime}_2\|} \sim \widehat{\|b^{S\prime\prime}_1 - b^{S\prime\prime}_2\|} \text{ does not hold}.
            \end{equation}
            It is important to note that due to perspective effects the position consistency as defined above is not guaranteed to hold between two views of an object. This is apparently in contrast to our goal to deal with distorted occurrences of $T$ in $S$ because distortions correspond to different perspective effects for different parts of the object. For this reason the position consistency score is used only in the \emph{initial seed selection} phase, in order to increase the probability to select a good \emph{matching triangle} in the initial phase.\\
            The position consistency score $\mathcal{P}$ of a \emph{matching triangle} $N'$ is the normalized sum of the absolute differences between the corresponding normalized sides length of the triangles formed by key points of $B^{T\prime}$ and $B^{S\prime}$. Taking \cref{fig:figure_11} as an example:
            \begin{equation}
                \mathcal{P}(N') = e^{-\frac{1}{2} (\frac{x - \mu}{\sigma})^2}
            \end{equation}
            where
            \begin{equation}
                \begin{aligned}
                    x = &\left\| \widehat{\|b^{T\prime}_1 - b^{T\prime}_2\|} - \widehat{\|b^{S\prime}_1 - b^{S\prime}_2\|} \right\| + \left\| \widehat{\|b^{T\prime}_2 - b^{T\prime}_3\|} - \widehat{\|b^{S\prime}_2 - b^{S\prime}_3\|} \right\| + 
                    \\
                    &+ \left\| \widehat{\|b^{T\prime}_3 - b^{T\prime}_1\|} - \widehat{\|b^{S\prime}_3 - b^{S\prime}_1\|} \right\|,
                \end{aligned}
            \end{equation}
            $\mu = 0$ and $\sigma = 0.2$. The numerical value of $\mu$ and $\sigma$ comes from \cite{ref:reference_1} in which it was determined empirically to give good results.
        \subsubsection{Orientation consistency score}
            \label{subsubsec:subsubsection_4.3.5.3}
            \begin{figure}[!ht]
                \centering
                \includegraphics[width=\linewidth]{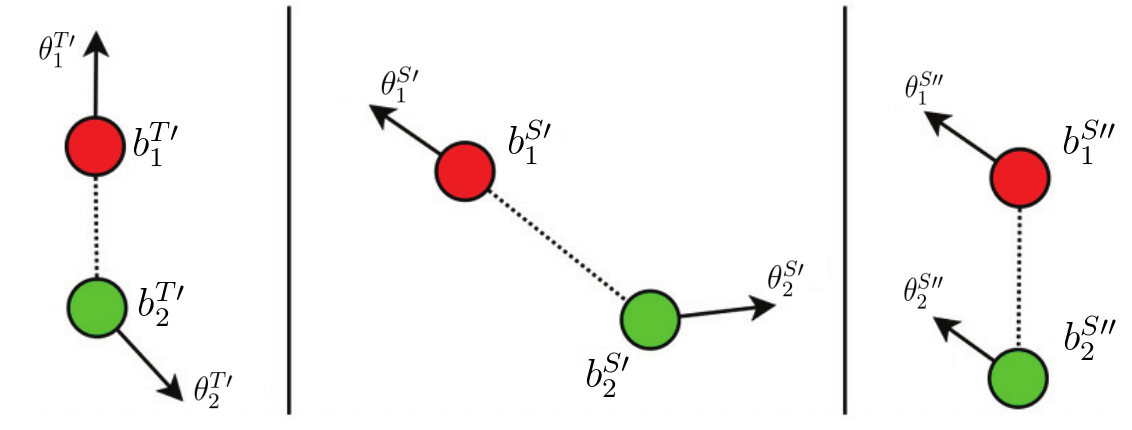}
                \caption{The couple of matching key point pairs $(b^{T\prime}_1, b^{S\prime}_1)$, $(b^{T\prime}_2, b^{S\prime}_2)$ shows orientation consistency, whereas $(b^{T\prime}_1, b^{S\prime\prime}_1)$, $(b^{T\prime}_2, b^{S\prime\prime}_2)$ not. Figure from \cite{ref:reference_1}, and adapted to our notation.}
                \label{fig:figure_12}
            \end{figure}
            The explanation of orientation consistency score was taken largely from \cite{ref:reference_1} and adapted to our notation.\\
            Each SIFT key point has an orientation angle $\theta^{\ast\prime}_i$ which refers to the direction of the image gradient at the key point position. If an object is rotated, the orientation of a key point on that object will also rotate, but will remain constant relative to other key points on the same object (since those key points are also rotated).\\
            A \emph{matching triangle} exhibits orientation consistency if the relative orientations between key points in $B^{T\prime}$ are equal to the corresponding relative orientations between key points in $B^{S\prime}$. Furthermore, the orientation of a key point $b^{T\prime}_i \in B^{T\prime}$ relative to a line between it and any other key point $b^{T\prime}_j \in B^{T\prime}$ should be equal to the orientation of the corresponding key point $b^{S\prime}_i \in B^{S\prime}$ relative to the line between it and the key point $b^{S\prime}_j \in B^{S\prime}$. \cref{fig:figure_12} shows cases of orientation consistent and orientation inconsistent corresponding key points; for example we have that:
            \begin{equation}
                \theta^{T\prime}_1 \cdot \theta^{T\prime}_2 \sim \theta^{S\prime}_1 \cdot \theta^{S\prime}_2 \text{ holds},
            \end{equation}
            whereas
            \begin{equation}
                \theta^{T\prime}_1 \cdot \theta^{T\prime}_2 \sim \theta^{S\prime\prime}_1 \cdot \theta^{S\prime\prime}_2 \text{ does not hold}.
            \end{equation}
            Furthermore it can be seen that:
            \begin{equation}
                \theta^{T\prime}_1 \cdot \begin{bmatrix}
                                             x^{T\prime}_1 - x^{T\prime}_2 \\
                                             y^{T\prime}_1 - y^{T\prime}_2
                                         \end{bmatrix} \sim \theta^{S\prime}_1 \cdot \begin{bmatrix}
                                                                                         x^{S\prime}_1 - x^{S\prime}_2 \\
                                                                                         y^{S\prime}_1 - y^{S\prime}_2
                                                                                     \end{bmatrix} \text{ holds},
            \end{equation}
            whereas
            \begin{equation}
                \theta^{T\prime}_1 \cdot \begin{bmatrix}
                                             x^{T\prime}_1 - x^{T\prime}_2 \\
                                             y^{T\prime}_1 - y^{T\prime}_2
                                         \end{bmatrix} \sim \theta^{S\prime\prime}_1 \cdot \begin{bmatrix}
                                                                                         x^{S\prime\prime}_1 - x^{S\prime\prime}_2 \\
                                                                                         y^{S\prime\prime}_1 - y^{S\prime\prime}_2
                                                                                     \end{bmatrix} \text{ does not hold}.
            \end{equation}
            Also in this case it is important to note that due to perspective effects the orientation consistency as defined above is not guaranteed to hold between two views of an object. Thus this score is used only in the \emph{initial seed selection} phase, in order to increase the probability to select a good \emph{matching triangle} in the initial phase.\\
            To determine the orientation consistency score of a \emph{matching triangle} first consider the orientation consistency score of two matching key point pairs. Denote the two matching key point pairs as $(b^{T\prime}_1, b^{S\prime}_1)$ and $(b^{T\prime}_2, b^{S\prime}_2)$, where $b^{T\prime}_1, b^{T\prime}_2 \in B^{T\prime}$ while $b^{S\prime}_1, b^{S\prime}_2 \in B^{S\prime}$. For $b^{T\prime}_1$ and $b^{T\prime}_2$ calculate three values:
            \begin{itemize}
                \item[--] $a^{T\prime}_1 = $ angle difference between $\theta^{T\prime}_1$ and $\theta^{T\prime}_2$.
                \item[--] $a^{T\prime}_2 = $ angle difference between $\theta^{T\prime}_1$ and the vector $\begin{bmatrix}
                                                                                             x^{T\prime}_1 - x^{T\prime}_2 \\
                                                                                             y^{T\prime}_1 - y^{T\prime}_2
                                                                                        \end{bmatrix}$.
                \item[--] $a^{T\prime}_3 = $ angle difference between $\theta^{T\prime}_2$ and the vector $\begin{bmatrix}
                                                                                             x^{T\prime}_2 - x^{T\prime}_1 \\
                                                                                             y^{T\prime}_2 - y^{T\prime}_1
                                                                                        \end{bmatrix}$.
            \end{itemize}
            Similarly the values $a^{S\prime}_1, a^{S\prime}_2, a^{S\prime}_3$ are calculated for the two key points $b^{S\prime}_1$ and $b^{S\prime}_2$. The sum of the absolute difference between corresponding values is the orientation consistency score of the two matching key point pairs:
            \begin{equation}
                x_{12} = \left\| a^{T\prime}_1 - a^{S\prime}_1 \right\| + \left\| a^{T\prime}_2 - a^{S\prime}_2 \right\| + \left\| a^{T\prime}_3 - a^{S\prime}_3 \right\|,
            \end{equation}
            The orientation consistency score $\mathcal{O}$ of a \emph{matching triangle} $N'$ is defined as the normalized average orientation consistency score across the three possible pairs of matching key point:
            \begin{equation}
                \mathcal{O}(N') = e^{-\frac{1}{2} (\frac{\bar{x} - \mu}{\sigma})^2}
            \end{equation}
            where
            \begin{equation}
                \bar{x} = \frac{x_{12} + x_{23} + x_{31}}{3},
            \end{equation}
            $\mu = 0$ and $\sigma = 1.75$. The numerical value of $\mu$ comes from \cite{ref:reference_1} while the value of $\sigma$ were determined empirically to give us good results.
        \subsubsection{Scale-ratio consistency score}
            \label{subsubsec:subsubsection_4.3.5.4}
            \begin{figure}[!ht]
                \centering
                \includegraphics[width=\linewidth]{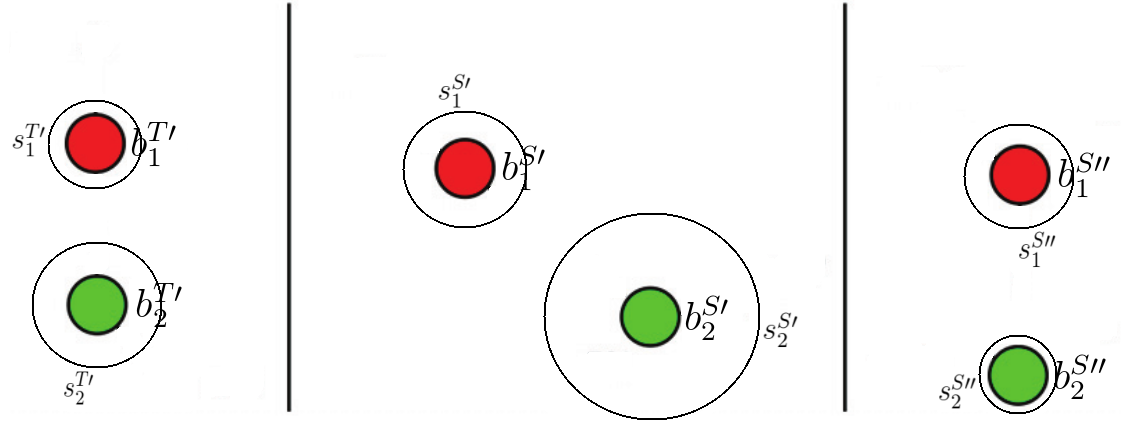}
                \caption{The scale $s^{\ast\prime}_i$ of a key point $b^{\ast\prime}_i$ is represented as a circle around the key point. In particular, the scale is higher (higher level of the scale-space) if the circle is wider. The couple of matching key point pairs $(b^{T\prime}_1, b^{S\prime}_1)$, $(b^{T\prime}_2, b^{S\prime}_2)$ shows scale-ratio consistency, whereas $(b^{T\prime}_1, b^{S\prime\prime}_1)$, $(b^{T\prime}_2, b^{S\prime\prime}_2)$ not. Figure from \cite{ref:reference_1}, and adapted to our notation.}
                \label{fig:figure_13}
            \end{figure}
            This is the only score that does not come from \cite{ref:reference_1}. We introduce a new score because we want exploit all the information contained in the SIFT key point. In fact using only the other three scores, the information about key point scale is lost. Each key point has a scale $s^{\ast\prime}_i$ that refers to the level of the scale-space in which the correspondent feature was detected (feature detection introduced in \cref{subsubsec:subsubsec_2.1.3.2}). A \emph{matching triangle} exhibits scale-ratio consistency if the mutual scale ratios between key points in $B^{T\prime}$ are equal to the mutual scale ratios between corresponding key points in $B^{S\prime}$. \cref{fig:figure_13} shows cases of scale-ratio consistent and scale-ratio inconsistent corresponding key points; for example we have that:
            \begin{equation}
                \frac{s^{T\prime}_1}{s^{T\prime}_2} \sim \frac{s^{S\prime}_1}{s^{S\prime}_2} \text{ holds},
            \end{equation}
            while
            \begin{equation}
                \frac{s^{T\prime}_1}{s^{T\prime}_2} \sim \frac{s^{S\prime\prime}_1}{s^{S\prime\prime}_2} \text{ does not hold}.
            \end{equation}
            Again it is important to note that due to perspective effects the scale-ratio consistency is not guaranteed to hold between two views of an object. However, differently from \emph{position consistency} and \emph{orientation consistency}, in presence of perspective effects the scale-ratio consistency is more likely to hold when the key points involved are close together. Since we are considering \emph{matching triangles} through which expand our \emph{seed}, this means that, thanks to this consistency, we prefer \emph{matching triangles} whose projections $B^T$ and $B^S$ produce smaller triangles in presence of perspectives. Smaller triangles are preferred (they lead to more dense triangulation meshes, so to higher deformation description capability), thus this score is used not only in the \emph{initial seed selection} phase, but also in the \emph{seed expansion} phase.
            \par
            To determine the scale-ratio consistency score of a \emph{matching triangle} first consider the scale-ratio score of two matching key point pairs. Denote the two matching key point pairs as $(b^{T\prime}_1, b^{S\prime}_1)$ and $(b^{T\prime}_2, b^{S\prime}_2)$, where $b^{T\prime}_1, b^{T\prime}_2 \in B^{T\prime}$ while $b^{S\prime}_1, b^{S\prime}_2 \in B^{S\prime}$. For $b^{T\prime}_1$ and $b^{T\prime}_2$ calculate two values:
            \begin{itemize}
                \item[--] $r^{T\prime}_1 = \frac{s^{T\prime}_1}{s^{T\prime}_2}$.
                \item[--] $r^{T\prime}_2 = \frac{s^{T\prime}_2}{s^{T\prime}_1}$.
            \end{itemize}
            Similarly the values $r^{S\prime}_1, r^{S\prime}_2$ are calculated for the two key points $b^{S\prime}_1$ and $b^{S\prime}_2$. The sum of the absolute difference between corresponding values is the scale-ratio consistency score of the two feature match pairs:
            \begin{equation}
                x_{12} = \left\| r^{T\prime}_1 - r^{S\prime}_1 \right\| + \left\| r^{T\prime}_2 - r^{S\prime}_2 \right\|,
            \end{equation}
            The scale-ratio consistency score $\mathcal{SR}$ of a \emph{matching triangle} $N'$ is defined as the normalized average scale-ratio consistency score across the three possible pairs of matching key point:
            \begin{equation}
                \mathcal{SR}(N') = e^{-\frac{1}{2} (\frac{\bar{x} - \mu}{\sigma})^2}
            \end{equation}
            where
            \begin{equation}
                \bar{x} = \frac{x_{12} + x_{23} + x_{31}}{3},
            \end{equation}
            $\mu = 0$ and $\sigma = 10$. The numerical values of $\mu$ and $\sigma$ were determined empirically to give us good results.
        \subsubsection{Total consistency score}
            \label{subsubsec:subsubsection_4.3.5.5}
            Given a \emph{matching triangle} $N'$, we compute a total consistency score that sums up all the previous scores. Since the \emph{initial seed selection} phase and \emph{seed expansion} phase depend from different sets of consistency scores, we define two different total consistency scores:
            \begin{equation}
                \begin{aligned}
                    \mathcal{CCS}(N') = e^{-\frac{1}{2} (\frac{\left\|
                    \tiny
                    \begin{bmatrix}
                        \mathcal{DV}(N') \\ \mathcal{P}(N') \\ \mathcal{O}(N') \\ \mathcal{SR}(N')
                    \end{bmatrix} \right\|_{- \mu}}{\sigma})^2}
                \end{aligned}
                \label{eq:equation_22}
            \end{equation}
            \begin{equation}
                \mathcal{RCS}(N') = e^{-\frac{1}{2} (\frac{\left\|
                \tiny
                \begin{bmatrix}
                    \mathcal{DV}(N') \\ \mathcal{SR}(N')
                \end{bmatrix} \right\|_{- \mu}}{\sigma})^2}
                \label{eq:equation_23}
            \end{equation}
            where $\mu = 2$ and $\sigma = 0.2$. The value of $\mu$ is determined to preserve the order between vectors norm. The standard deviation $\sigma$ were determined empirically to give us good results.\\
            $\mathcal{CCS}$, used in the \emph{initial seed selection} phase, extracts the total consistency score from the complete set of consistency scores; while $\mathcal{RCS}$, used in the \emph{seed expansion} phase, from the reduced set of consistency scores.\\
            Both in the \emph{initial seed selection} and \emph{seed expansion} phases, \emph{matching triangles} that do not reach a total consistency score of $0.6$ are discarded. Also this threshold value was determined empirically.
    \subsection{The algorithm}
        \label{subsec:subsection_4.3.6}
        Since our approach is feature-based, we first introduce the pseudocode of a typical feature-based object detection and identification algorithm: \cref{alg:algorithm_1}.
        \begin{algorithm2e}[H]
            \caption{Feature-based object detection algorithm}
            \label{alg:algorithm_1}
            \SetAlgoLined
            \DontPrintSemicolon
            \SetKwInOut{Input}{Input}
            \SetKwInOut{Output}{Output}
            
            \Input{$T$ $\gets$ template image\\
                   $S$ $\gets$ scene image}
            \Output{$O$ $\gets$ object occurrences}
            
            \SetKwFunction{FMain}{feature\_based\_object\_detection}
            \SetKwProg{Fn}{function}{:}{}
            \Fn{\FMain{$T, S$}}{
                $K^T, D^T$ $\gets$ features\_detection($T$) \tcp*{$K^T, D^T$ defined respectively in \hspace*{\fill} \cref{eq:equation_3a} and (\ref{eq:equation_38a})}
                $K^S, D^S$ $\gets$ features\_detection($S$) \tcp*{$K^S, D^S$ defined respectively in \hspace*{\fill} \cref{eq:equation_3b} and (\ref{eq:equation_38b})}
                $M$ $\gets$ features\_matching($D^T, D^S$) \tcp*{$M$ defined in \cref{eq:equation_4}}
                $O$ $\gets$ object\_detection($T, S, K^T, K^S, D^T, D^S, M$)\;
                \textbf{return} $O$\;
            }
            \textbf{end}
        \end{algorithm2e}
        In this work the $features\_detection$ is done through SIFT\cite{ref:reference_5} algorithm (presented in \cref{subsubsec:subsubsec_2.1.3.3}), while $features\_matching$ (described in \cref{subsubsec:subsubsec_2.1.3.4}) through the usual nearest neighbour match (using FLANN\cite{ref:reference_17}, a library for performing fast approximate nearest neighbor searches in high dimensional spaces). What really differ between various feature-based object detection and identification algorithms is the $object\_detection$ phase. In the following we expand the $object\_detection$ phase of our algorithm.\\
        \\
        The "classical" situations, in which the template $T$ contains a planar object and the object occurrences in $S$ are not distorted, can be solved through the iterative application of RANSAC\cite{ref:reference_4} algorithm over the set of key point matches $M$, in order to estimate homographies that relates $T$ and object occurrences $S_1$, $S_2$, \dots, $S_p$. However, the situations we aim to manage include distortions of these occurrences, thus we developed an approach based on the selection of initial \emph{seeds} (formalized in \cref{subsec:subsection_4.3.2}) and their subsequent expansion. Both \emph{initial seed selection} and \emph{seed expansion}, described respectively in \cref{subsubsec:subsubsection_4.3.1.1} and \ref{subsubsec:subsubsection_4.3.1.2}, exploit the Delaunay triangulation of the position of key points in $K^T$.\\
        \\
        Analogously to before we can see in \cref{alg:algorithm_2} the pseudocode of the $object\_detection$ routine.
        
        
        \begin{algorithm2e}[H]
            \caption{Object detection}
            \label{alg:algorithm_2}
            \footnotesize
            \SetAlgoLined
            \DontPrintSemicolon
            \SetKwInOut{Input}{Input}
            \SetKwInOut{Output}{Output}
            \SetKwRepeat{Do}{do}{while}
            
            \Input{$T$ $\gets$ template image\\
                   $S$ $\gets$ scene image\\
                   $K^T$ $\gets$ template key points\\
                   $K^S$ $\gets$ scene key points\\
                   $D^T$ $\gets$ descriptors of template key points\\
                   $D^S$ $\gets$ descriptors of scene key points\\
                   $M$ $\gets$ features matches}
            \Output{$O$ $\gets$ object occurrences}
            
            \SetKwFunction{FMain}{object\_detection}
            \SetKwProg{Fn}{function}{:}{}
            \Fn{\FMain{$T, S, K^T, K^S, D^T, D^S, M$}}{
                $N_{expanded}$ $\gets$ \{\} \tcp*{$N_{expanded}$ is the set of expanded \emph{seeds}}
                $terminate$ $\gets$ false\;
                \Do{$\neg terminate$}{
                    $K^T, K^S, D^T, D^S, M$ $\gets$ keypoints\_and\_matches\_filtering($K^T, K^S, D^T, D^S, M, N_{expanded}$)\;
                    $N$ $\gets$ initial\_seed\_selection($K^T, K^S, D^T, D^S, M$) \tcp*{$N$ is a \emph{seed}, as described in \hspace*{\fill} \cref{subsec:subsection_4.3.2}}
                    $N$ $\gets$ seed\_expansion($K^T, K^S, D^T, D^S, M, N$)\;
                    \uIf{$N$ was properly expanded}{
                        $N_{expanded} \gets N_{expanded} \cup N$\;
                    }
                    \uElse{
                        $terminate$ $\gets$ true\;
                    }
                }
                $O$ $\gets$ photometric\_filtering($T, S, N_{expanded}$)\;
                \textbf{return} $O$\;
            }
            \textbf{end}
        \end{algorithm2e}
        
        In \cref{alg:algorithm_2} we can see some subfunctions, in particular:
        \begin{itemize}
            \item $keypoints\_and\_matches\_filtering$ eliminates matches from $M$ that belong to some \emph{seed} in $N_{expanded}$. Furthermore key points in $K^T$ and $K^S$ that are no more associated to a match in $M$ are eliminated. Formally:
            \begin{subequations}
                \begin{equation}
                    N_i \text{ is a \emph{seed}}
                \end{equation}
                and
                \begin{equation}
                    N_{expanded} = \{N_1, N_2, \dots, N_n\}
                \end{equation}
            \end{subequations}
            is the set of expanded \emph{seeds}. Thus $keypoints\_and\_matches\_filtering$ performs the following:
            \begin{equation}
                \begin{aligned}
                    &\forall m_i \in M \text{ if } \exists N_j \in N_{expanded} \mid m_i \in N_j \text{ then } m_i
                    \\
                    &\text{ is removed from } M;
                \end{aligned}
            \end{equation}
            after that
            \begin{equation}
                \forall k^T_i \in K^T \text{ if } \nexists m_j \in M \mid m_j = (k^T_i, \ast) \text{ then } k^T_i \text{ is removed from } K^T
            \end{equation}
            and
            \begin{equation}
                \forall k^S_i \in K^S \text{ if } \nexists m_j \in M \mid m_j = (\ast, k^S_i) \text{ then } k^S_i \text{ is removed from } K^S.
            \end{equation}
            \item $initial\_seed\_selection$ create a Delaunay triangulation from the position of key points in $K^T$, builds a set of \emph{matching triangles} (concept introduced in \cref{subsec:subsection_4.3.2}) and selects the \emph{matching triangle} $N$ that is considered the best by the total consistency score $\mathcal{CCS}$ (see \cref{subsec:subsection_4.3.5} for consistency scores). The \emph{initial seed selection} procedure is described in \cref{subsubsec:subsubsection_4.3.1.1}, while this function is better presented in \cref{subsubsec:subsubsection_4.3.6.1}.
            \item $seed\_expansion$ expands the previous extracted \emph{seed} $N$ (as explained in \cref{subsubsec:subsubsection_4.3.1.2}). This is done exploiting a constrained Delaunay triangulation of key points in $K^T$ and searching for \emph{matching triangles} through which expand the \emph{seed} $N$. Then the geometric coherence of these \emph{matching triangles} is evaluated (by checks presented in \cref{subsec:subsection_4.3.3} and \ref{subsec:subsection_4.3.4}), finally the best \emph{matching triangle} $N'$ is identified by the total consistency score $\mathcal{RCS}$. This procedure is explained better in \cref{subsubsec:subsubsection_4.3.6.2}.
            \item $photometric\_filtering$ first performs reconstruction, for each object identified by a \emph{seed} in $N_{expanded}$, through Thin Plate Splines\cite{ref:reference_2} (technique for data interpolation and smoothing widely used as non-rigid transformation model in image alignment and shape matching); then photometric differences between $T$ and reconstructed object instances are extracted. \emph{Seeds} in $N_{expanded}$ that refer to the same object are identified and, based on the previously computed differences, only the best one is kept. Procedure better explained in \cref{sec:section_5.1}.
        \end{itemize}
        
        \subsubsection{Initial \emph{seed} selection}
            \label{subsubsec:subsubsection_4.3.6.1}
            \begin{algorithm2e}[H]
                \caption{Initial \emph{seed} selection}
                \label{alg:algorithm_3}
                \SetAlgoLined
                \DontPrintSemicolon
                \SetKwInOut{Input}{Input}
                \SetKwInOut{Output}{Output}
                
                \Input{$K^T$ $\gets$ template key points\\
                       $K^S$ $\gets$ scene key points\\
                       $D^T$ $\gets$ descriptors of template key points\\
                       $D^S$ $\gets$ descriptors of scene key points\\
                       $M$ $\gets$ features matches}
                \Output{$N$ $\gets$ \emph{seed}}
                
                \SetKwFunction{FMain}{initial seed selection}
                \SetKwProg{Fn}{function}{:}{}
                \Fn{\FMain{$K^T, K^S, D^T, D^S, M$}}{
                    $R^T$ $\gets$ redundant\_triangulation\_computation($K^T$) \tcp*{$R^T$ defined in \hspace*{\fill} \cref{eq:equation_24}}
                    $Q$ $\gets$ matching\_triangles\_set\_composition($M, R^T$) \tcp*{$Q$ defined in \hspace*{\fill} \cref{eq:equation_25}}
                    $U$ $\gets$ consistency\_scores\_evaluation($Q$) \tcp*{$U$ defined in \cref{eq:equation_28}}
                    $N$ $\gets$ best\_matching\_triangle\_selection($Q, U$)\;
                    \textbf{return} $N$\;
                }
                \textbf{end}
            \end{algorithm2e}
            \cref{alg:algorithm_3} is the formalization of the concepts described in \cref{subsubsec:subsubsection_4.3.1.1}.
            \begin{itemize}
                \item $redundant\_triangulation\_computation$ computes a set of key point triplets from the Delaunay triangulation of key points in $K^T$ (examples in \cref{fig:figure_5}). Its output has the following form:
                \begin{equation}
                    R^T = \{r^T_1, r^T_2, \dots, r^T_n\},
                    \label{eq:equation_24}
                \end{equation}
                where
                \begin{equation}
                    r^T_i = \{k^T_{i1}, k^T_{i2}, k^T_{i3}\} \subseteq K^T
                \end{equation}
                is a triplet of template key points.
                \item $matching\_triangles\_set\_composition$ creates many \emph{matching triangles} of those that can be generated from every element of $R^T$. In particular:
                \begin{equation}
                    Q = \{N_1, N_2, \dots, N_m\},
                    \label{eq:equation_25}
                \end{equation}
                where
                \begin{equation}
                    N_i = \{n_{i1}, n_{i2}, n_{i3}\} \subseteq M
                \end{equation}
                is a \emph{matching triangle}.\\
                It is important to note that:
                \begin{equation}
                    \forall N_i \in Q \quad \exists r^T_j \in R^T \mid \forall n_{ik} \in N_i \quad n_{ik} = (k^T_{jk}, \ast)
                \end{equation}
                \item $consistency\_scores\_evaluation$ computes the function $\mathcal{CCS}$ (defined in \cref{eq:equation_22}) for each \emph{matching triangle} in $Q$. Formally:
                \begin{equation}
                    U = \{u_1, u_2, \dots, u_m\},
                    \label{eq:equation_28}
                \end{equation}
                where
                \begin{equation}
                    \forall N_i \in Q \quad \exists! u_i = \mathcal{CCS}(N_i) \in U.
                \end{equation}
                \item $best\_matching\_triangle\_selection$ first ranks \emph{matching triangles} according to their score. Then the best one is selected as the new selected \emph{seed}.
            \end{itemize}
            
        \subsubsection{\emph{Seed} expansion}
            \label{subsubsec:subsubsection_4.3.6.2}
            
            \begin{algorithm2e}[H]
                \caption{\emph{Seed} expansion}
                \label{alg:algorithm_4}
                \SetAlgoLined
                \DontPrintSemicolon
                \SetKwInOut{Input}{Input}
                \SetKwInOut{Output}{Output}
                \SetKwRepeat{Do}{do}{while}
                
                \Input{$K^T$ $\gets$ template image features\\
                       $K^S$ $\gets$ scene image features\\
                       $D^T$ $\gets$ descriptors of template key points\\
                       $D^S$ $\gets$ descriptors of scene key points\\
                       $M$ $\gets$ features matches\\
                       $N$ $\gets$ \emph{seed}}
                \Output{$N^{\prime}$ $\gets$ expanded \emph{seed}}
                
                \SetKwFunction{FMain}{seed\_expansion}
                \SetKwProg{Fn}{function}{:}{}
                \Fn{\FMain{$K^T, K^S, D^T, D^S, M, N$}}{
                    $N^{\prime} \gets N$\;
                    $terminate$ $\gets$ false\;
                    \Do{$\neg terminate$}{
                        $N^{\prime}$ $\gets$ seed\_expansion\_step($K^T, K^S, D^T, D^S, M, N^{\prime}$)\;
                        \uIf{$\neg (N^{\prime}$ was expanded$)$}{
                            $terminate$ $\gets$ true\;
                        }
                    }
                    \textbf{return} $N^{\prime}$\;
                }
                \textbf{end}
            \end{algorithm2e}

            \begin{itemize}
                \item $seed\_expansion\_step$ executes a single expansion step of the \emph{seed} $N'$ (as illustrated in \cref{subsubsec:subsubsection_4.3.1.2}). \cref{alg:algorithm_5} presents better the expansion step.
            \end{itemize}
            
            
            \begin{algorithm2e}[H]
                \caption{Seed expansion step}
                \label{alg:algorithm_5}
                \SetAlgoLined
                \DontPrintSemicolon
                \SetKwInOut{Input}{Input}
                \SetKwInOut{Output}{Output}
                
                \Input{$K^T$ $\gets$ template image features\\
                       $K^S$ $\gets$ scene image features\\
                       $D^T$ $\gets$ descriptors of template key points\\
                       $D^S$ $\gets$ descriptors of scene key points\\
                       $M$ $\gets$ features matches\\
                       $N^{\prime}$ $\gets$ \emph{seed}}
                \Output{$N^{\prime\prime}$ $\gets$ expanded \emph{seed}}
                
                \SetKwFunction{FMain}{seed\_expansion\_step}
                \SetKwProg{Fn}{function}{:}{}
                \Fn{\FMain{$K^T, K^S, D^T, D^S, M, N^{\prime}$}}{
                    $W$ $\gets$ constrained\_triangulation($K^T, N^{\prime}$) \tcp*{$W$ defined in \cref{eq:equation_29}}
                    $R^T$ $\gets$ expanded\_adjacent\_triangles\_extraction($W, N^{\prime}$) \tcp*{$R^T$ defined in \hspace*{\fill} \cref{eq:equation_24}}
                    $Q$ $\gets$ matching\_triangles\_set\_composition($M, R^T$) \tcp*{$Q$ defined in \hspace*{\fill} \cref{eq:equation_25}}
                    $Q$ $\gets$ geometric\_checks($Q$)\;
                    $U$ $\gets$ consistency\_scores\_evaluation($Q$) \tcp*{$U$ defined in \cref{eq:equation_28}}
                    $N$ $\gets$ best\_matching\_triangle\_selection($Q, U$)\;
                    $N^{\prime\prime}$ $\gets$ $N^{\prime} \cup N$\;
                    \textbf{return} $N^{\prime\prime}$\;
                }
                \textbf{end}
            \end{algorithm2e}
            \cref{alg:algorithm_5} formalizes the procedure introduced in \cref{subsubsec:subsubsection_4.3.1.2}.
            \begin{itemize}
                \item $constrained\_triangulation$ (as described in \cref{subsubsec:subsubsection_4.3.1.2}) computes the constrained Delaunay triangulation of the key points in $K^T$ forcing segments of the convex hull generated from the projection $B^T$ of the \emph{seed} to be present (illustrated in \cref{fig:figure_6a}). Furthermore, key points that are internal to the convex hull but are not in $B^T$ are not taken into consideration for the triangulation. The output $W$ has a graph form:
                \begin{equation}
                    W = (V, E),
                \end{equation}
                where
                \begin{equation}
                    V = \{k^T_1, k^T_2, \dots, k^T_n\} \subseteq K^T
                \end{equation}
                with
                \begin{equation}
                    \forall k^T_i \in V \quad k^T_i \in B^T \lor k^T_i \text{ external to the convex hull generated from } B^T,
                \end{equation}
                and
                \begin{equation}
                    E = \{e_1, e_2, \dots, e_m\}
                    \label{eq:equation_29}
                \end{equation}
                with
                \begin{equation}
                    e_i = (j, l),
                \end{equation}
                and $j, l$ indices of elements in $V$.
                \item $expanded\_adjacent\_triangles\_extraction$ computes a set of key point triplets from the triangles adjacent to the convex hull generated from $B^T$ (illustrated in \cref{fig:figure_6}).
                \item $matching\_triangles\_set\_composition$ extracts, in the same way as \cref{alg:algorithm_3}, many \emph{matching triangles} of those that can be generated from every element of $R^T$.
                \item $geometric\_checks$ performs both geometric checks to all the \emph{matching triangles} in $Q$. In particular, first the \emph{non-intersection check} is computed; then the \emph{local coherence check} is performed on the remaining \emph{matching triangles}. These checks are widely explained in \cref{subsec:subsection_4.3.3}.
                \item $consistency\_scores\_evaluation$ computes the function $\mathcal{RCS}$ (described in \cref{eq:equation_23}) for each \emph{matching triangle} that in $Q$. Formally:
                \begin{equation}
                    U = \{u_1, u_2, \dots, u_m\},
                    \label{eq:equation_30}
                \end{equation}
                where
                \begin{equation}
                    \forall N_i \in Q \quad \exists! u_i = \mathcal{RCS}(N_i) \in U.
                \end{equation}
                \item $best\_matching\_triangle\_selection$ first ranks \emph{matching triangles} according to their score. Then the best \emph{matching triangle} is selected as the one to be added to the \emph{seed}.
            \end{itemize}
    \thispagestyle{empty}
    \chapter{Implementation details}
\label{ch:chapter_4}

\section{Photometric filtering}
    \label{sec:section_5.1}
    This function is presented here because it is not a central concept in our work. In particular it is a way to eliminate \emph{seeds} that refer to the same object occurrence, keeping only the best one.
    
    \begin{algorithm2e}[H]
        \caption{Photometric filtering}
        \label{alg:algorithm_6}
        \SetAlgoLined
        \DontPrintSemicolon
        \SetKwInOut{Input}{Input}
        \SetKwInOut{Output}{Output}
        
        \Input{$T$ $\gets$ template image\\
               $S$ $\gets$ scene image\\
               $N_{expanded}$ $\gets$ expanded \emph{seeds}}
        \Output{$O$ $\gets$ object occurrences}
        
        \SetKwFunction{FMain}{photometric\_filtering}
        \SetKwProg{Fn}{function}{:}{}
        \Fn{\FMain{$T, S, N_{expanded}$}}{
            $Z$ $\gets$ objects\_occurrences\_rectification($S, N_{expanded}$) \tcp*{$Z$ defined in \hspace*{\fill} \cref{eq:equation_32}}
            $J$ $\gets$ photometric\_differences\_extraction($T, Z$) \tcp*{$J$ defined in \hspace*{\fill} \cref{eq:equation_33}}
            $O$ $\gets$ intersecting\_seeds\_filtering($N_{expanded}, J$)\;
            \textbf{return} $O$\;
        }
        \textbf{end}
    \end{algorithm2e}
    \begin{itemize}
        \item $objects\_occurrences\_rectification$ first computes Thin Plate Splines\cite{ref:reference_2} transformations from all the \emph{seeds} in $N_{expanded}$. More specifically, $\forall N_i \in N_{expanded}$ computes the TPS transformation that maps $B^S_i$ to $B^T_i$.\\
        Then these transformations are separately applied to the scene image. Thus
        \begin{equation}
            Z = \{z_1, z_2, \dots, z_n\}
            \label{eq:equation_32}
        \end{equation}
        is the set of rectified object occurrences.
        \item $photometric\_differences\_extraction$ extracts photometric differences between $T$ and elements of $Z$. First pixelwise RGB differences are computed. Second these differences are normed over the color dimension. Finally the median of the histograms of the normed images are extracted as representatives of the reconstruction goodness. The output of this function:
        \begin{equation}
            J = \{j_1, j_2, \dots, j_n\}
            \label{eq:equation_33}
        \end{equation}
        is the set of these medians, one for each \emph{seed} in $N_{expanded}$.
        \item $intersecting\_seeds\_filtering$ identifies the \emph{seeds} that refer to the same object occurrence in $S$ and keeps only the best ones. In particular \emph{seeds} whose convex hulls in $S$ intersect are considered to be referred to the same object occurrence, and they are grouped together. Then for each group the \emph{seed} with lowest value of $j_i \in J$ is selected as representative for the specific object occurrence in $S$.
    \end{itemize}

\section{Multiple \emph{seeds} approach}
    \label{sec:section_5.2}
\begin{figure}
        \centering
        \begin{subfigure}[!ht]{0.45\linewidth}
            \includegraphics[width=\linewidth]{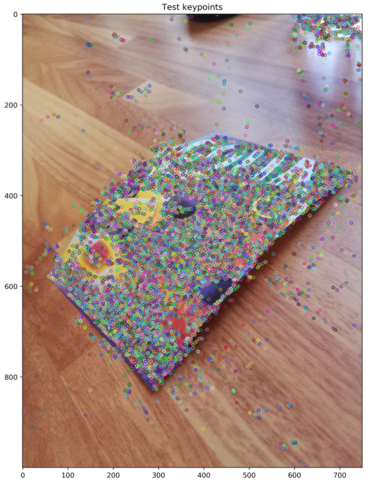}
            \caption{Scene image key points.}
            \label{fig:figure_20}
        \end{subfigure}
        \begin{subfigure}[!ht]{0.40\linewidth}
            \includegraphics[width=\linewidth]{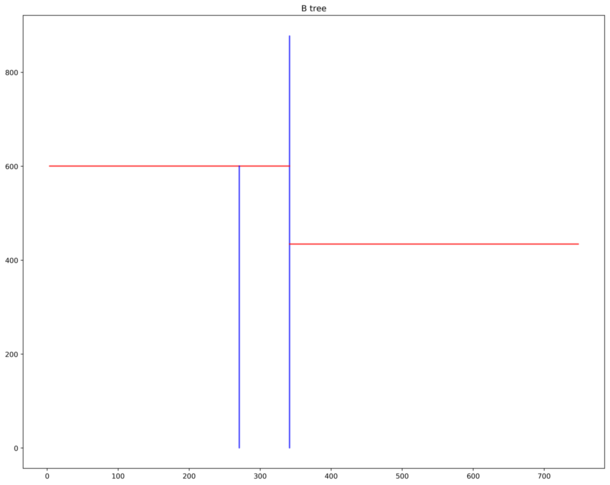}
            \caption{Kd-tree built on key points of \cref{fig:figure_20}. We have imposed 5 leaves to be present.}
            \label{fig:figure_21}
        \end{subfigure}
        \caption{Example of scene image key points and kd-tree built from them.}
    \end{figure}

\begin{figure}
        \centering
        \begin{subfigure}[!ht]{0.32\linewidth}
            \includegraphics[width=\linewidth]{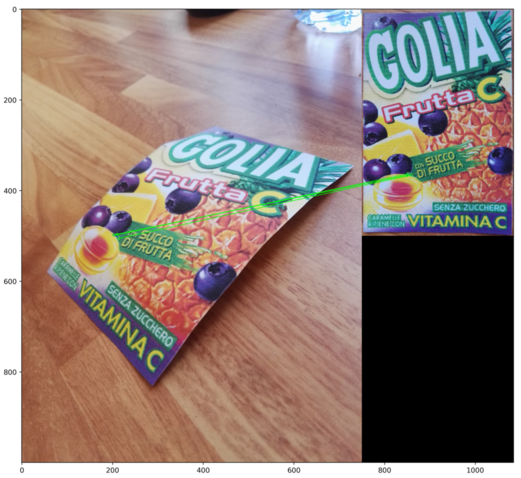}
        \end{subfigure}
        \begin{subfigure}[!ht]{0.32\linewidth}
            \includegraphics[width=\linewidth]{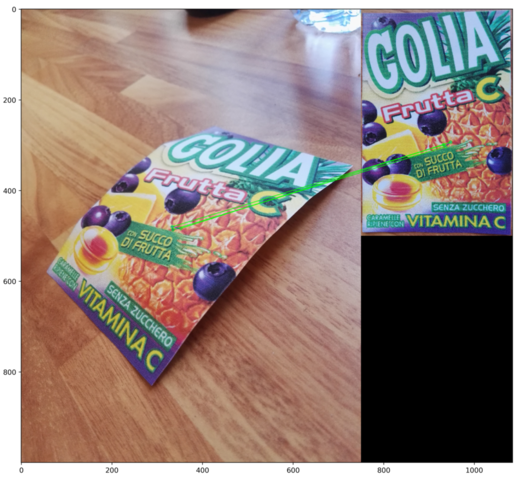}
        \end{subfigure}
        \begin{subfigure}[!ht]{0.32\linewidth}
            \includegraphics[width=\linewidth]{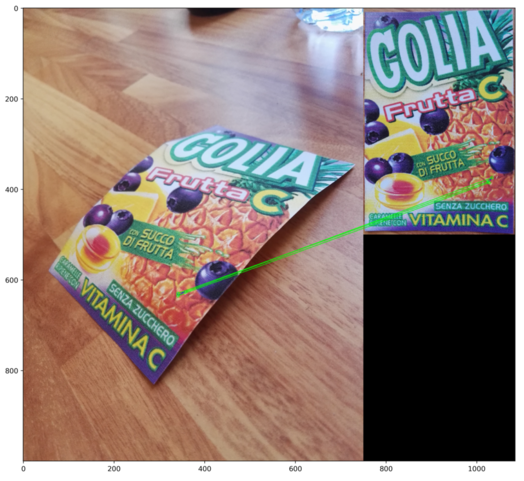}
        \end{subfigure}
        \begin{subfigure}[!ht]{0.32\linewidth}
            \includegraphics[width=\linewidth]{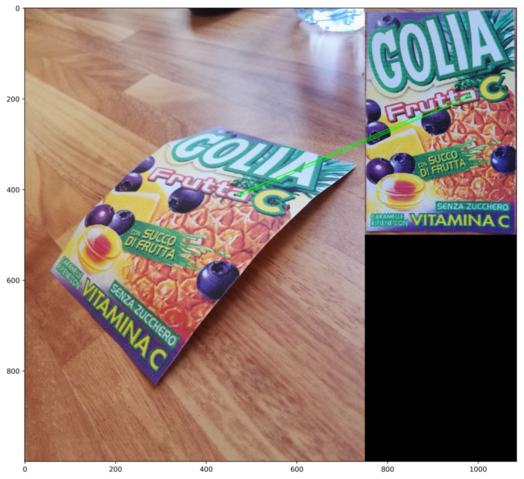}
        \end{subfigure}
        \begin{subfigure}[!ht]{0.32\linewidth}
            \includegraphics[width=\linewidth]{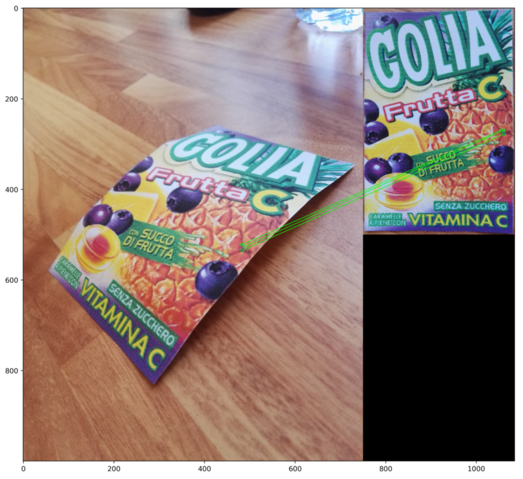}
        \end{subfigure}
        \caption{Multiple seeds extracted exploiting the kd-tree shown in \cref{fig:figure_21}.}
        \label{fig:figure_22}
    \end{figure}

    We considered the situation in which only one initial \emph{seed} is extracted at a time (described in \cref{subsubsec:subsubsection_4.3.1.1}). However our method can be easily generalized to the case in which multiple initial \emph{seeds} are extracted.\\
    This approach has many benefits:
    \begin{itemize}
        \item \emph{Multiple objects detection in a complete expansion iteration}. One \emph{seed} allows to group the correspondences between key points in $K^T$ and key points in $K^S$ referred to a single object occurrence. If multiple \emph{seeds} are generated, there is a probability that some of them refer to different object occurrences. This means that more object occurrences can be identified in a single complete expansion iteration.
        \item \emph{Seeds merging}. Whether more \emph{seeds} refer to the same object occurrence, they can be merged together. Thus the correspondences grouping is accomplished faster for the considered object occurrence.
        \item \emph{Parallelization}. Each \emph{seed} can be treated independently of the others. This means that their expansions can be naturally parallelized.
        \item \emph{Uniform scene coverage}. Each initial selected \emph{seed} can be drawn in a way that the spatial distribution of the key points in $S$ is exploited. We construct a kd-tree among key points in $K^S$, imposing a certain number of leaves to be present. These leaves represent 2-dimensional intervals in $S$. Thus the best \emph{seed} for each leaf can be selected. We consider that a \emph{seed} $N$ is contained in a leaf whether its projection $B^S$ has a key point in the leaf interval. In \cref{fig:figure_21} we can see an example of such kd-tree(with 5 leaves imposed), built from key points of \cref{fig:figure_20}. Then in \cref{fig:figure_22} we can see the extracted \emph{seeds}.
    \end{itemize}
    In order to apply this method we modified some parts of the previous presented algorithm.\\
    In particular:
    \begin{itemize}
        \item In \cref{alg:algorithm_3} the $best\_matching\_triangle\_selection$ function is improved with the kd-tree. Thus the extraction of multiple \emph{seeds} is achieved.
        \item In \cref{alg:algorithm_2} the $seed\_expansion$ function is applied to every initial selected \emph{seed} in a parallelized way. Furthermore the possibility to merge \emph{seeds} is exploited here.
    \end{itemize}
    In our work we use the multiple \emph{seeds} approach presented in \cref{sec:section_5.2}, exploiting its benefits, except for the parallelization one.
    \thispagestyle{empty}
    \chapter{Experiments}
\label{ch:chapter_5}

\noindent
In this chapter we present the experiments performed in order to evaluate our approach.
\\
We introduce the dataset, representative of the situations we want to manage, in \cref{sec:section_6.1}. Our figures of merit in \cref{subsec:subsection_6.2.1}, comparison technique in \cref{subsec:subsection_6.2.2} and used tools in \cref{sec:section_6.3}. Finally we present results in \cref{sec:section_6.4}.
\section{Dataset}
    \label{sec:section_6.1}
    All the images have been acquired by the 13 megapixel camera of a Huawei P8. At first they had a resolution of 4160x3120 pixels, but for the experiments they have been resized to 1000x750 pixels.
    \subsection{Template}
\begin{figure}
            \centering
            \begin{minipage}[!h]{0.4\textwidth}
                \includegraphics[width=0.8\linewidth]{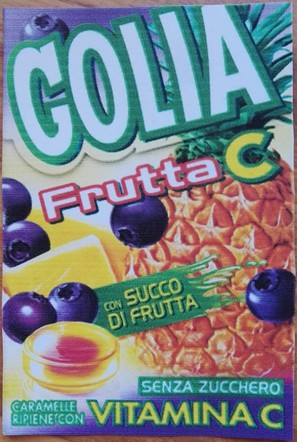}
                \caption{Template image.}
                \label{fig:figure_24a}
            \end{minipage}
            \begin{minipage}[!h]{0.4\textwidth}
                \includegraphics[width=0.8\linewidth]{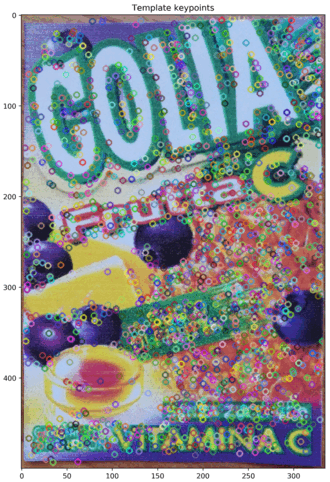}
                \caption{Key points extracted from \cref{fig:figure_24a}.}
                \label{fig:figure_24b}
            \end{minipage}
        \end{figure}

        In our approach only one template image is needed, as anticipated in \cref{sec:section_4.0}. We can see in \cref{fig:figure_24a} the image used as template, while in \cref{fig:figure_24b} the key points extracted from it through SIFT\cite{ref:reference_5}. It is important to notice that \cref{fig:figure_24b} shows a fairly uniform distribution of key points. In our method, this is an important feature for a template because it allows to better describe deformations.
    \subsection{Scenes}
        \label{subsec:subsection_6.1.2}
\begin{figure}[!h]
            \centering
            \includegraphics[width=0.5\linewidth]{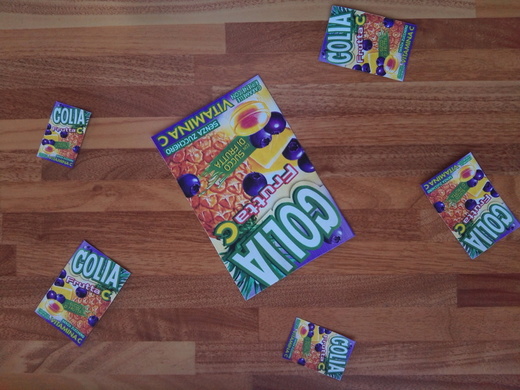}
            \caption{Scene with many undistorted object occurrences.}
            \label{fig:figure_25}
        \end{figure}

\begin{figure}
            \centering
            \begin{subfigure}[!h]{0.32\linewidth}
                \includegraphics[width=\linewidth]{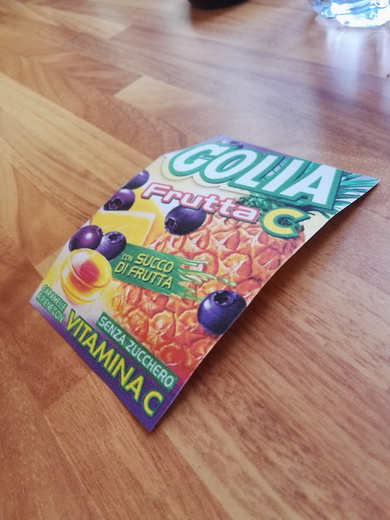}
            \end{subfigure}
            \begin{subfigure}[!h]{0.32\linewidth}
                \includegraphics[width=\linewidth]{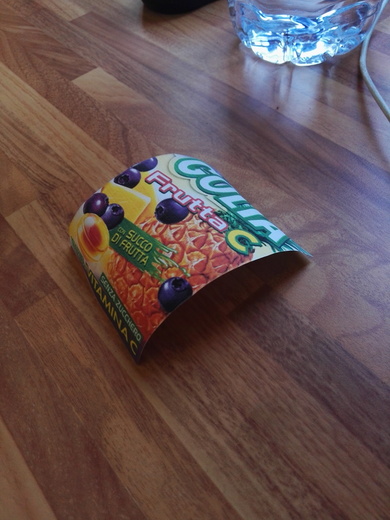}
            \end{subfigure}
            \begin{subfigure}[!h]{0.32\linewidth}
                \includegraphics[width=\linewidth]{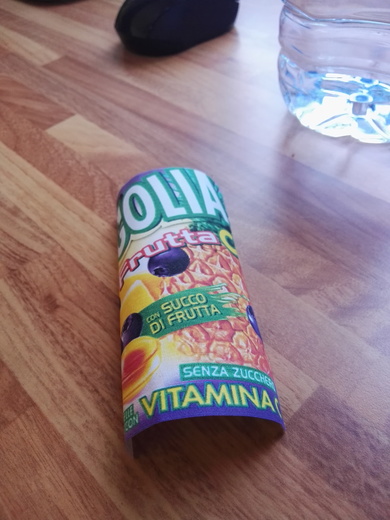}
            \end{subfigure}
            \begin{subfigure}[!h]{0.32\linewidth}
                \includegraphics[width=\linewidth]{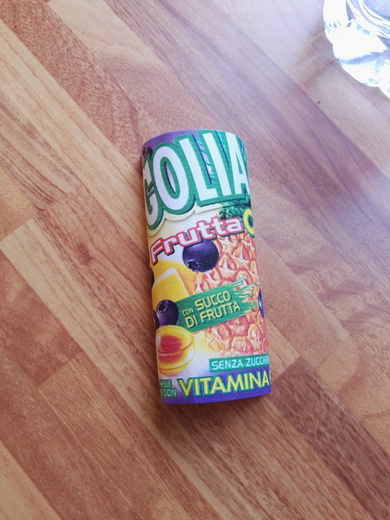}
            \end{subfigure}
            \begin{subfigure}[!h]{0.32\linewidth}
                \includegraphics[width=\linewidth]{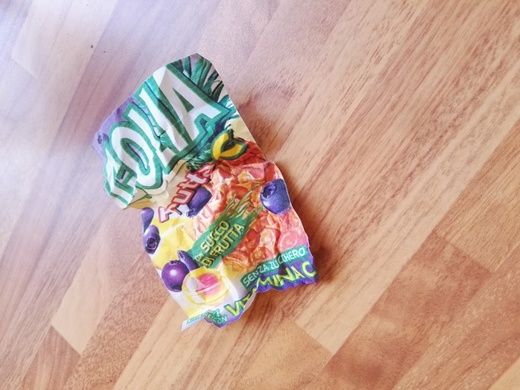}
            \end{subfigure}
            \begin{subfigure}[!h]{0.32\linewidth}
                \includegraphics[width=\linewidth]{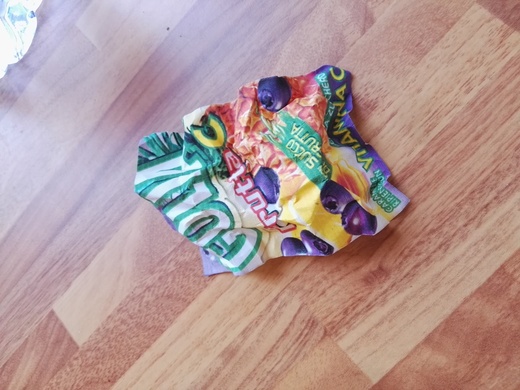}
            \end{subfigure}
            \caption{Scenes with one distorted object occurrence each.}
            \label{fig:figure_26}
        \end{figure}

\begin{figure}[!h]
            \centering
            \includegraphics[width=0.5\linewidth]{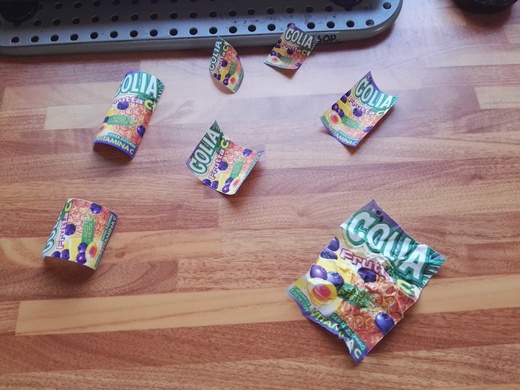}
            \caption{Scene with many distorted object occurrences.}
            \label{fig:figure_27}
        \end{figure}

        We test our method in three types of scene situations, with incremental difficulty.
        \begin{itemize}
            \item Many undistorted object occurrences (\cref{fig:figure_25}).
            \item One distorted object occurrence (\cref{fig:figure_26}).
            \item Many distorted object occurrences (\cref{fig:figure_27}).
        \end{itemize}
\section{Evaluation procedure}
    \label{sec:section_6.2}
    \subsection{Figures of merit}
        \label{subsec:subsection_6.2.1}
\begin{figure}[!h]
            \centering
            \includegraphics[width=0.5\textwidth]{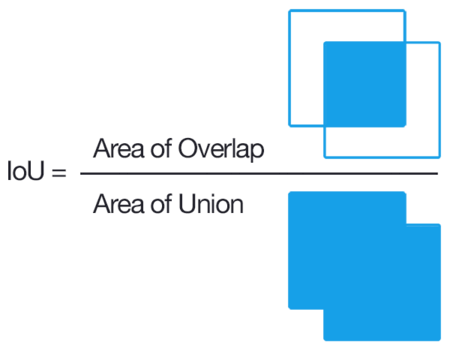}
            \caption{$IoU$ equation.}
            \label{fig:figure_28}
        \end{figure}

        The performance of our method is evaluated the by three values:
        \begin{itemize}
            \item Number of correctly detected objects in each scene image, with respect to all the ones present. This value represents the capacity of our method to identify all objects position.
            \item $IoU$ (Intersection over Union) measure (\cref{fig:figure_28}). In particular a ground truth mask is given for every object occurrence in each scene in order to compute this value. This measure shows the ability to group matches in order to cover as more object area as possible.
            \begin{equation}
                IoU \in [0, 1] \text{, with 0 the worse and 1 the best.}
                \label{eq:equation_34}
            \end{equation}
            \item Median of the difference image. This value is introduced in \cref{sec:section_5.1}, in \cref{eq:equation_33}. This last value represents goodness of the object reconstruction.
            \begin{equation}
                Median (\equiv j) \in [0, 255] \text{, with 255 the worse and 0 the best.}
                \label{eq:equation_35}
            \end{equation}
        \end{itemize}
    \subsection{Comparison technique}
        \label{subsec:subsection_6.2.2}
        We compare the performance of our approach with those of a feature-based object detection algorithm driven by RANSAC\cite{ref:reference_4}, in which we search for an homography transformation between template object image and object instances in the scene image.
\section{Tools and libraries}
    \label{sec:section_6.3}
    Both our algorithm and the RANSAC\cite{ref:reference_4} driven one are implemented in python3.6.4\cite{ref:reference_9}. The second one comes from official online tutorials (\cite{ref:reference_6},\cite{ref:reference_7},\cite{ref:reference_8}), and has been adapted to be able to deal with many object occurrences. To support the implementation of our approach, we used the following tools and libraries:
    \begin{itemize}
        \item \textbf{NumPy}. Numpy\cite{ref:reference_10} is a fundamental package intended for scientific computing in Python. It provides multidimensional array objects and many routines to operate on them. NumPy array is the fundamental structure through which images are represented in our algorithm.
        \item \textbf{OpenCV}. OpenCV (Open Computer Vision library)\cite{ref:reference_12} is an open source computer vision library. It provides a common infrastructure for computer vision applications and allows to accelerate the implementation of automated perception. Furthermore, this library has python wrappers and it is fully compatible with NumPy arrays. We use this library to manipulate images.
        \item \textbf{Matplotlib}. Matplotlib\cite{ref:reference_11} is a Python 2D plotting library which produces publication quality figures in a variety of hardcopy formats and interactive environments across platforms. We use this to show images of algorithm phases.
        \item \textbf{NetworkX}. NetworkX\cite{ref:reference_13} is a Python package for the creation, manipulation, and study of the structure, dynamics, and functions of complex networks. We use NetworkX to treat triangulation as a graph when searching for triangles.
        \item \textbf{Shapely}. Shapely\cite{ref:reference_14} is a Python package for set-theoretic analysis and manipulation of planar features. We use this library to perform geometric checks.
        \item \textbf{Python Triangle}. Python Triangle\cite{ref:reference_15} is a python wrapper around Jonathan Richard Shewchuk’s two-dimensional quality mesh generator and delaunay triangulator library\cite{ref:reference_16}. We use Python Triangle to generate triangulations and convex hulls.
    \end{itemize}
\section{Results}
    \label{sec:section_6.4}
    As anticipated in \cref{subsec:subsection_6.1.2}, in this section we present the results obtained in different scenarios. In particular, in \cref{subsec:subsection_6.4.1} we consider the case in which many object instances are present in the scene and the homography hypothesis holds. In \cref{subsec:subsection_6.4.2} we show results of different situations in which a single distorted object occurrence is present in the scene. Finally, in \cref{subsec:subsection_6.4.3} we generalize to the case in which many distorted objects are present in a single scene.
    \subsection{Many undistorted object occurrences}
        \label{subsec:subsection_6.4.1}
        \begin{figure}[!h]
            \centering
            \includegraphics[width=0.5\linewidth]{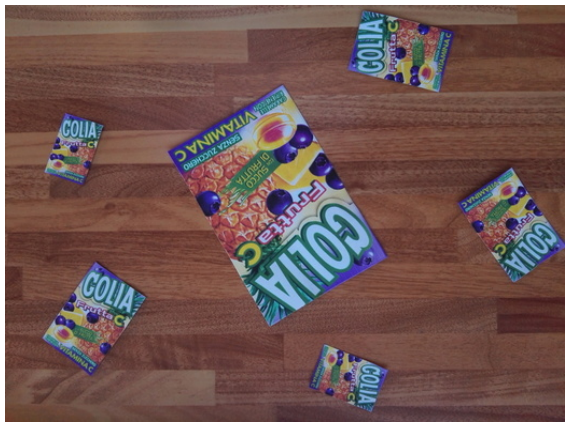}
            \caption{Many undistorted object occurrences.}
            \label{fig:figure_25repeated}
        \end{figure}
        \begin{figure}[!h]
            \centering
            \begin{subfigure}[!h]{0.32\linewidth}
                \includegraphics[width=\linewidth]{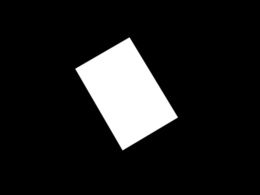}
            \end{subfigure}
            \begin{subfigure}[!h]{0.32\linewidth}
                \includegraphics[width=\linewidth]{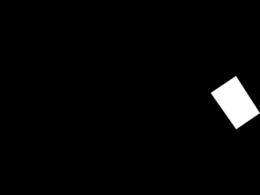}
            \end{subfigure}
            \begin{subfigure}[!h]{0.32\linewidth}
                \includegraphics[width=\linewidth]{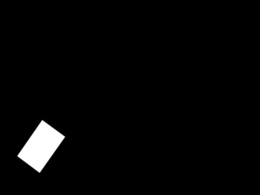}
            \end{subfigure}
            \begin{subfigure}[!h]{0.32\linewidth}
                \includegraphics[width=\linewidth]{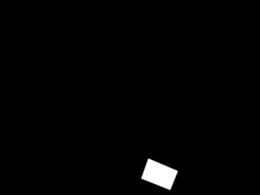}
            \end{subfigure}
            \begin{subfigure}[!h]{0.32\linewidth}
                \includegraphics[width=\linewidth]{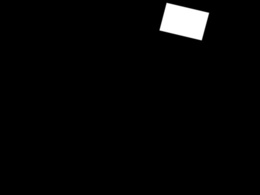}
            \end{subfigure}
            \begin{subfigure}[!h]{0.32\linewidth}
                \includegraphics[width=\linewidth]{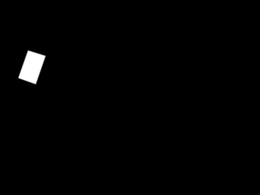}
            \end{subfigure}
            \caption{Ground truth mask for each object occurrence in \cref{fig:figure_25repeated}.}
            \label{fig:figure_29}
        \end{figure}
        First of all we can see in \cref{fig:figure_25repeated} (the same as \cref{fig:figure_25}) the scene image in which we want to search for template object occurrences, while in \cref{fig:figure_29} the ground truth masks used to compute the $IoU$ measure for each object found.
        \\
        \par
        Now we can show results of the execution of the object detection algorithm driven by RANSAC, with these input images. Subsequently results of our algorithm are shown.
        \paragraph{Feature-based object detection algorithm driven by RANSAC}
            We refer to the first object found by the algorithm (\cref{fig:figure_30}) to describe the different images illustrating the detection (images showed in \cref{appendix_A.1.1}), then for the others it is analogous.
            Let us list meanings of the different sub-images composing \cref{fig:figure_30}.
            \begin{itemize}
                \item \cref{fig:figure_30a} shows key point matches that agree with the found homography, together with the template image rectangle projected in the scene through this transformation.
                \item \cref{fig:figure_30b} shows the object occurrence rectified through the homography, together with inlier matches.
                \item \cref{fig:figure_30c} shows rectified object occurrence and template figure. Furthermore, the histogram of the rectified object image has been matched, in each color channel, to that of the template. \cref{fig:figure_30d}, \ref{fig:figure_30h}, \ref{fig:figure_30i}, \ref{fig:figure_30j} represent operations done to these images.
                \item \cref{fig:figure_30d} shows the difference, in each color channel, between images in \cref{fig:figure_30c}. The difference is shown with corresponding color intensity.
                \item \cref{fig:figure_30h} shows the histogram of \cref{fig:figure_30d}, shown with the respective colors.
                \item \cref{fig:figure_30i} shows the image in \cref{fig:figure_30d} to which it was computed the norm over the color dimension. Furthermore inlier key points are displayed.
                \item \cref{fig:figure_30j} shows the histogram of \cref{fig:figure_30i}, together with its median. The median has been introduced in \cref{sec:section_5.1} (in \cref{eq:equation_33}) and it is used as evaluation criteria, as explained in \cref{subsec:subsection_6.2.1}.
            \end{itemize}
        \paragraph{Our algorithm}
            Analogously to before, we describe the different images illustrating the detection (from \cref{appendix_A.1.2}).
            The caption of \cref{fig:figure_36} indicates whether the found object comes from the expansion of a single \emph{seed}, or from the union of many \emph{seeds}.
            \begin{itemize}
                \item \cref{fig:figure_36a} shows the key point matches belonging to the expanded \emph{seed}, together with the triangulation meshes.
                \item \cref{fig:figure_36b} shows the object occurrence rectified through Thin Plate Splines\cite{ref:reference_2}, together with \emph{seed} matches.
                \item \cref{fig:figure_36c} shows the portions of template and rectified object occurrence contained in the convex hull generated from \emph{seed} key points. Furthermore the histogram of the rectified object image has been matched, in each color channel, to the template image one. \cref{fig:figure_36d}, \ref{fig:figure_36h}, \ref{fig:figure_36i}, \ref{fig:figure_36j} represent operations done to these images.
                \item \cref{fig:figure_36d} shows the difference, in each color channel, between images in \cref{fig:figure_36c}. The difference is shown with corresponding color intensity.
                \item \cref{fig:figure_36h} shows the histogram of \cref{fig:figure_36d}, shown with the respective colors.
                \item \cref{fig:figure_36i} shows the image in \cref{fig:figure_36d} to which it was computed the norm over the color dimension. Furthermore inlier key points are displayed.
                \item \cref{fig:figure_36j} shows the histogram of \cref{fig:figure_36i}, together with its median.
            \end{itemize}
        \par
        \begin{figure}[!h]
            \centering
            \begin{subfigure}[!h]{0.48\linewidth}
                \includegraphics[width=\linewidth]{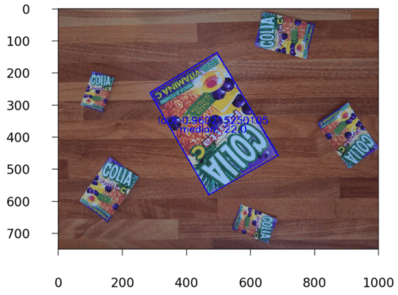}
            \end{subfigure}
            \begin{subfigure}[!h]{0.48\linewidth}
                \includegraphics[width=\linewidth]{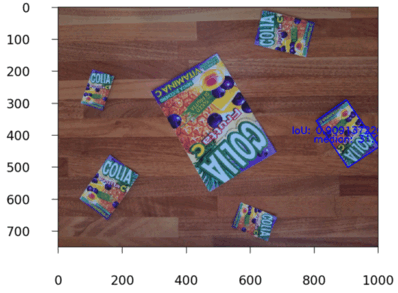}
            \end{subfigure}
            \begin{subfigure}[!h]{0.48\linewidth}
                \includegraphics[width=\linewidth]{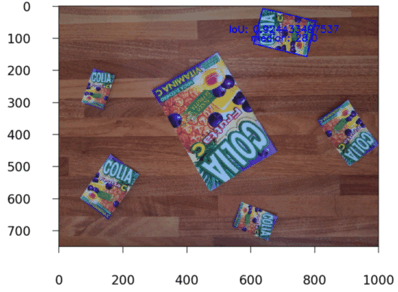}
            \end{subfigure}
            \begin{subfigure}[!h]{0.48\linewidth}
                \includegraphics[width=\linewidth]{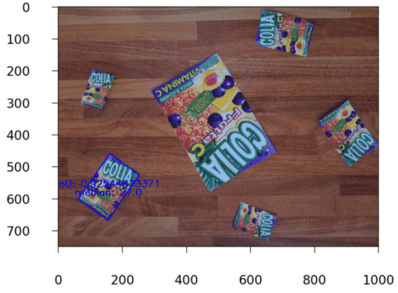}
            \end{subfigure}
            \begin{subfigure}[!h]{0.48\linewidth}
                \includegraphics[width=\linewidth]{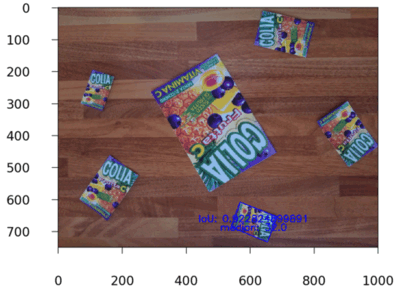}
            \end{subfigure}
            \begin{subfigure}[!h]{0.48\linewidth}
                \includegraphics[width=\linewidth]{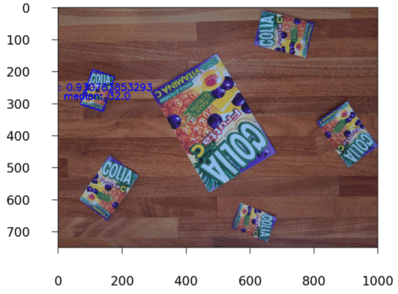}
            \end{subfigure}
            \caption{Object occurrences found by homography-based RANSAC algorithm.}
            \label{fig:figure_75}
        \end{figure}
        \begin{figure}[!h]
            \centering
            \begin{subfigure}[!h]{0.48\linewidth}
                \includegraphics[width=\linewidth]{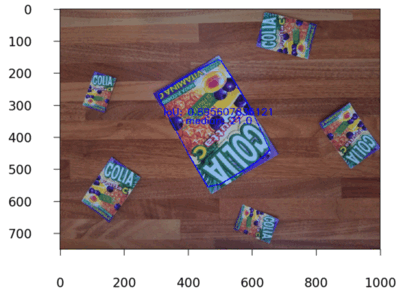}
            \end{subfigure}
            \begin{subfigure}[!h]{0.48\linewidth}
                \includegraphics[width=\linewidth]{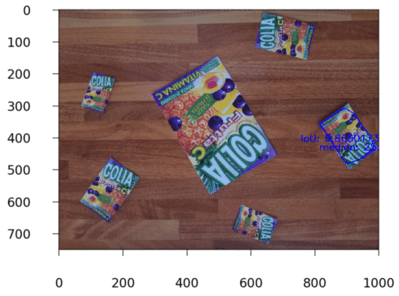}
            \end{subfigure}
            \begin{subfigure}[!h]{0.48\linewidth}
                \includegraphics[width=\linewidth]{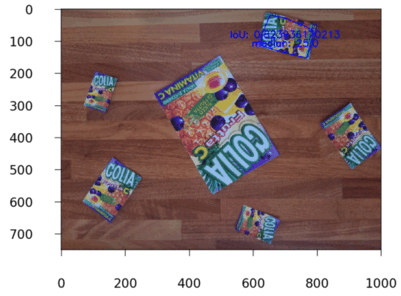}
            \end{subfigure}
            \begin{subfigure}[!h]{0.48\linewidth}
                \includegraphics[width=\linewidth]{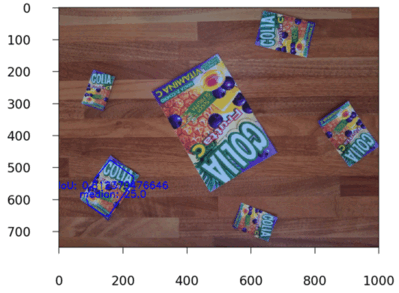}
            \end{subfigure}
            \begin{subfigure}[!h]{0.48\linewidth}
                \includegraphics[width=\linewidth]{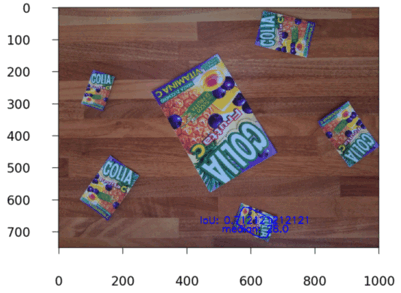}
            \end{subfigure}
            \begin{subfigure}[!h]{0.48\linewidth}
                \includegraphics[width=\linewidth]{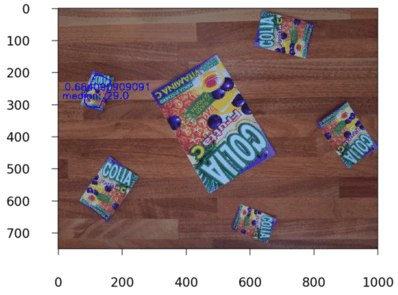}
            \end{subfigure}
            \caption{Object occurrences found by our algorithm.}
            \label{fig:figure_76}
        \end{figure}
        \begin{table}[!ht]
            \centering
            \begin{subtable}{\textwidth}
                \centering
                \begin{tabular}{c|cccccc|}
                    \cline{2-7}
                                                 & \multicolumn{6}{c|}{Object occurrences}                                                                                        \\ \cline{2-7} 
                                                 & \multicolumn{1}{c|}{1\textsuperscript{st}} & \multicolumn{1}{c|}{2\textsuperscript{nd}} & \multicolumn{1}{c|}{3\textsuperscript{rd}} & \multicolumn{1}{c|}{4\textsuperscript{th}} & \multicolumn{1}{c|}{5\textsuperscript{th}} & 6\textsuperscript{th} \\ \hline
                    \multicolumn{1}{|c|}{Identified}  & \cmark                 & \cmark                 & \cmark                 & \cmark                 & \cmark                 & \cmark  \\ \cline{1-1}
                    \multicolumn{1}{|c|}{$IoU$}  & 0.9602                 & 0.9091                 & 0.9244                 & 0.9254                 & 0.9228                 & 0.9308  \\ \cline{1-1}
                    \multicolumn{1}{|c|}{Median} & 22.0                   & 31.0                   & 28.0                   & 27.0                   & 32.0                   & 32.0  \\ \hline
                \end{tabular}
                \caption{Homography-based RANSAC}
                \label{tab:table_1a}
            \end{subtable}
            \\\vspace{5mm}
            \begin{subtable}{\textwidth}
                \centering
                \begin{tabular}{c|cccccc|}
                    \cline{2-7}
                                                 & \multicolumn{6}{c|}{Object occurrences}                                                                                        \\ \cline{2-7} 
                                                 & \multicolumn{1}{c|}{1\textsuperscript{st}} & \multicolumn{1}{c|}{2\textsuperscript{nd}} & \multicolumn{1}{c|}{3\textsuperscript{rd}} & \multicolumn{1}{c|}{4\textsuperscript{th}} & \multicolumn{1}{c|}{5\textsuperscript{th}} & 6\textsuperscript{th} \\ \hline
                    \multicolumn{1}{|c|}{Identified}  & \cmark                 & \cmark                 & \cmark                 & \cmark                 & \cmark                 & \cmark  \\ \cline{1-1}
                    \multicolumn{1}{|c|}{$IoU$}  & 0.8956                 & 0.8650                 & 0.8239                 & 0.8124                 & 0.7121                 & 0.6841  \\ \cline{1-1}
                    \multicolumn{1}{|c|}{Median} & 21.0                   & 25.0                   & 25.0                   & 25.0                   & 28.0                   & 29.0  \\ \hline
                \end{tabular}
                \caption{Our algorithm}
                \label{tab:table_1b}
            \end{subtable}
            \caption{Tables of points}
            \label{tab:table_1}
        \end{table}
        Table \cref{tab:table_1} shows evaluation of the two different approaches. Let us comment the results.
        \begin{itemize}
            \item We can observe that both methods have found all object occurrences, $6/6$.
            \item The $IoU$ measure is higher in the RANSAC driven approach. This happens because in our method we consider as object region the one contained in the convex hull of seed key points, while in the RANSAC driven method the entire projected rectangle is considered.
            \item The median is lower in our approach. Though the difference is not very significant (due to the object planarity), this shows that our approach has the capacity of adaptation to different object parts deformation.
        \end{itemize}
        We can say that in this scenario the first approach works better. Obviously this is the optimal case in which an homography based approach work, since it is meant for this. However, we have shown these results to underline the fact that our method is also able to manage this type of scenario well.
    \subsection{One distorted object occurrence}
        \label{subsec:subsection_6.4.2}
        \subsubsection{Case 1}
            \begin{figure}[!h]
                \centering
                \begin{minipage}[c]{0.3\textwidth}
                \centering
                    \includegraphics[width=\linewidth]{pictures/experimental_results/figure_26/figure_26a}
                    \caption{One distorted object occurrence.}
                \end{minipage}
                \begin{minipage}[c]{0.3\textwidth}
                \centering
                    \includegraphics[width=\linewidth]{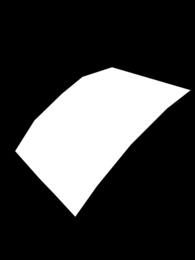}
                    \caption{Ground truth mask.}
                \end{minipage}
                \caption{Scene and mask.}
            \end{figure}
        \subsubsection{Case 2}
            \begin{figure}[!h]
                \centering
                \begin{minipage}[c]{0.3\textwidth}
                \centering
                    \includegraphics[width=\linewidth]{pictures/experimental_results/figure_26/figure_26b}
                    \caption{One distorted object occurrence.}
                \end{minipage}
                \begin{minipage}[c]{0.3\textwidth}
                \centering
                    \includegraphics[width=\linewidth]{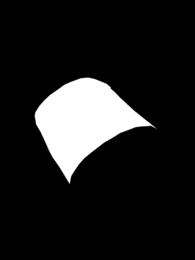}
                    \caption{Ground truth mask.}
                \end{minipage}
                \caption{Scene and mask.}
            \end{figure}
        \subsubsection{Case 3}
            \begin{figure}[!h]
                \centering
                \begin{minipage}[c]{0.3\textwidth}
                \centering
                    \includegraphics[width=\linewidth]{pictures/experimental_results/figure_26/figure_26c}
                    \caption{One distorted object occurrence.}
                \end{minipage}
                \begin{minipage}[c]{0.3\textwidth}
                \centering
                    \includegraphics[width=\linewidth]{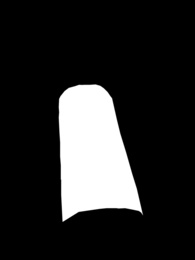}
                    \caption{Ground truth mask.}
                \end{minipage}
                \caption{Scene and mask.}
            \end{figure}
        \subsubsection{Case 4}
            \begin{figure}[!h]
                \centering
                \begin{minipage}[c]{0.3\textwidth}
                \centering
                    \includegraphics[width=\linewidth]{pictures/experimental_results/figure_26/figure_26d}
                    \caption{One distorted object occurrence.}
                \end{minipage}
                \begin{minipage}[c]{0.3\textwidth}
                \centering
                    \includegraphics[width=\linewidth]{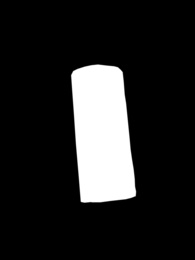}
                    \caption{Ground truth mask.}
                \end{minipage}
                \caption{Scene and mask.}
            \end{figure}
        \subsubsection{Case 5}
            \begin{figure}[H]
                \centering
                \begin{minipage}[c]{0.45\textwidth}
                \centering
                    \includegraphics[width=\linewidth]{pictures/experimental_results/figure_26/figure_26e}
                    \caption{One distorted object occurrence.}
                \end{minipage}
                \begin{minipage}[c]{0.45\textwidth}
                \centering
                    \includegraphics[width=\linewidth]{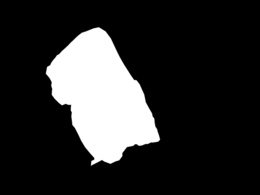}
                    \caption{Ground truth mask.}
                \end{minipage}
                \caption{Scene and mask.}
            \end{figure}
        \subsubsection{Case 6}
            \begin{figure}[!h]
                \centering
                \begin{minipage}[c]{0.45\textwidth}
                \centering
                    \includegraphics[width=\linewidth]{pictures/experimental_results/figure_26/figure_26f}
                    \caption{One distorted object occurrence.}
                \end{minipage}
                \begin{minipage}[c]{0.45\textwidth}
                \centering
                    \includegraphics[width=\linewidth]{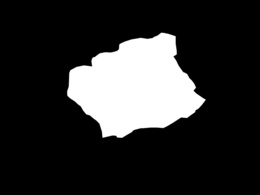}
                    \caption{Ground truth mask.}
                \end{minipage}
                \caption{Scene and mask.}
            \end{figure}
        Extensive figures of results of the different cases in \cref{appendix_A.2.1}, \ref{appendix_A.2.2}, \ref{appendix_A.2.3}, \ref{appendix_A.2.4}, \ref{appendix_A.2.5}, \ref{appendix_A.2.6} respectively.
        \begin{figure}[H]
            \centering
            \begin{subfigure}[!h]{0.4\linewidth}
                \includegraphics[width=\linewidth]{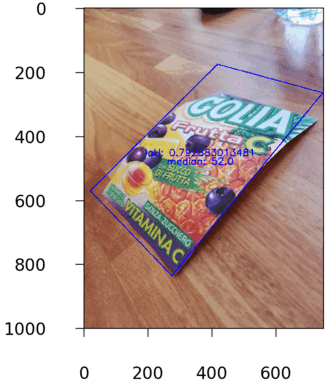}
            \end{subfigure}
            \begin{subfigure}[!h]{0.4\linewidth}
                \includegraphics[width=\linewidth]{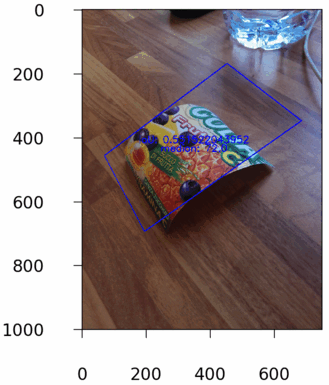}
            \end{subfigure}
            \begin{subfigure}[!h]{0.4\linewidth}
                \includegraphics[width=\linewidth]{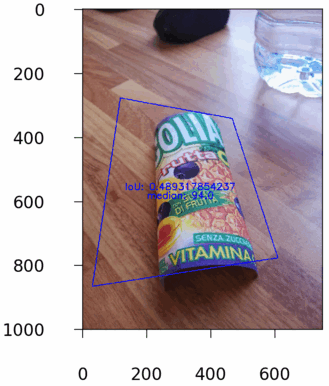}
            \end{subfigure}
            \begin{subfigure}[!h]{0.4\linewidth}
                \includegraphics[width=\linewidth]{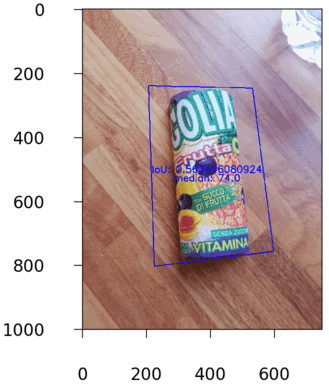}
            \end{subfigure}
            \begin{subfigure}[!h]{0.48\linewidth}
                \includegraphics[width=\linewidth]{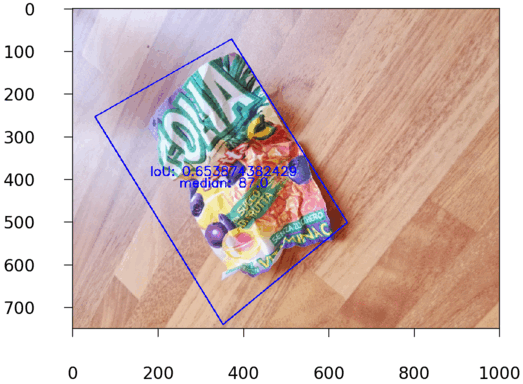}
            \end{subfigure}
            \begin{subfigure}[!h]{0.48\linewidth}
                \includegraphics[width=\linewidth]{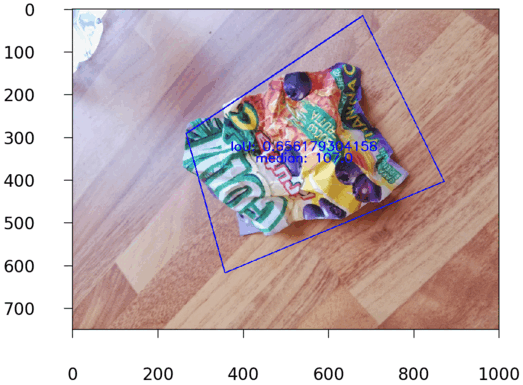}
            \end{subfigure}
            \caption{Objects found by homography-based RANSAC algorithm.}
            \label{fig:figure_77}
        \end{figure}
        \begin{figure}[H]
            \centering
            \begin{subfigure}[!h]{0.4\linewidth}
                \includegraphics[width=\linewidth]{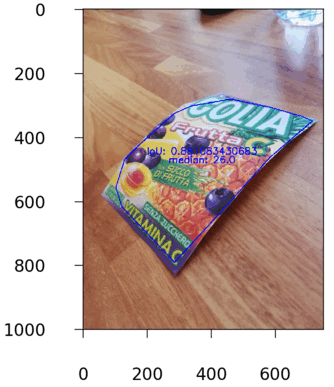}
            \end{subfigure}
            \begin{subfigure}[!h]{0.4\linewidth}
                \includegraphics[width=\linewidth]{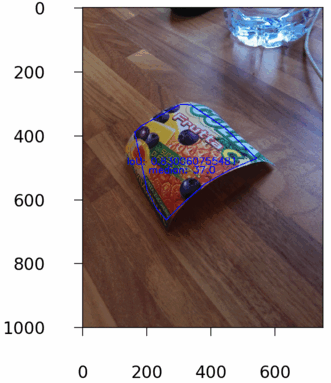}
            \end{subfigure}
            \begin{subfigure}[!h]{0.4\linewidth}
                \includegraphics[width=\linewidth]{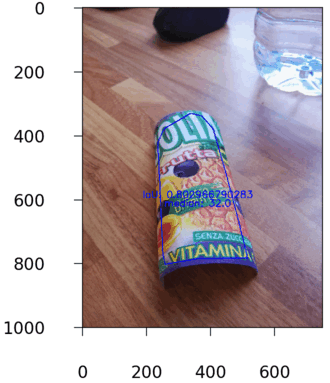}
            \end{subfigure}
            \begin{subfigure}[!h]{0.4\linewidth}
                \includegraphics[width=\linewidth]{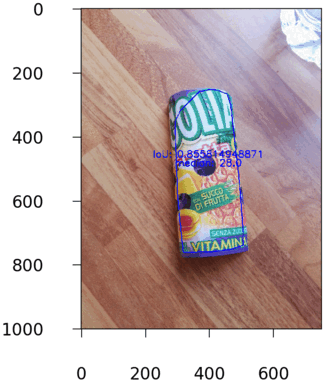}
            \end{subfigure}
            \begin{subfigure}[!h]{0.48\linewidth}
                \includegraphics[width=\linewidth]{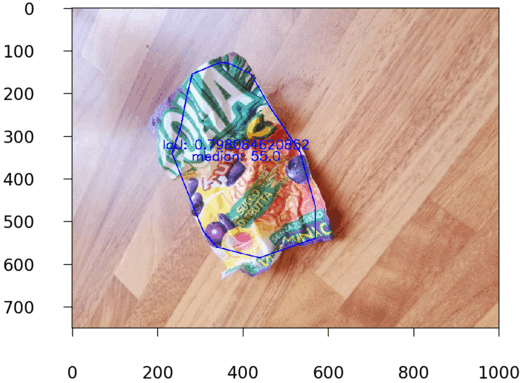}
            \end{subfigure}
            \begin{subfigure}[!h]{0.48\linewidth}
                \includegraphics[width=\linewidth]{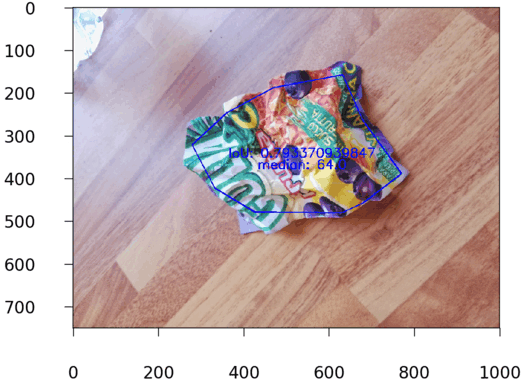}
            \end{subfigure}
            \caption{Objects found by our algorithm.}
            \label{fig:figure_78}
        \end{figure}
        \begin{table}[!ht]
            \centering
            \begin{subtable}{\textwidth}
                \centering
                \begin{tabular}{c|cccccc|}
                    \cline{2-7}
                                                 & \multicolumn{6}{c|}{Cases}                                                                                        \\ \cline{2-7} 
                                                 & \multicolumn{1}{c|}{1\textsuperscript{st}} & \multicolumn{1}{c|}{2\textsuperscript{nd}} & \multicolumn{1}{c|}{3\textsuperscript{rd}} & \multicolumn{1}{c|}{4\textsuperscript{th}} & \multicolumn{1}{c|}{5\textsuperscript{th}} & 6\textsuperscript{th} \\ \hline
                    \multicolumn{1}{|c|}{Identified}  & \cmark                 & \cmark                 & \cmark                 & \cmark                 & \cmark                 & \cmark  \\ \cline{1-1}
                    \multicolumn{1}{|c|}{$IoU$}  & 0.7929                 & 0.5515                 & 0.4893                 & 0.5634                 & 0.6539                 & 0.6562  \\ \cline{1-1}
                    \multicolumn{1}{|c|}{Median} & 52.0                   & 72.0                   & 94.0                   & 74.0                   & 87.0                   & 107.0  \\ \hline
                \end{tabular}
                \caption{Homography-based RANSAC}
                \label{tab:table_2a}
            \end{subtable}
            \\\vspace{5mm}
            \begin{subtable}{\textwidth}
                \centering
                \begin{tabular}{c|cccccc|}
                    \cline{2-7}
                                                 & \multicolumn{6}{c|}{Cases}                                                                                        \\ \cline{2-7} 
                                                 & \multicolumn{1}{c|}{1\textsuperscript{st}} & \multicolumn{1}{c|}{2\textsuperscript{nd}} & \multicolumn{1}{c|}{3\textsuperscript{rd}} & \multicolumn{1}{c|}{4\textsuperscript{th}} & \multicolumn{1}{c|}{5\textsuperscript{th}} & 6\textsuperscript{th} \\ \hline
                    \multicolumn{1}{|c|}{Identified}  & \cmark                 & \cmark                 & \cmark                 & \cmark                 & \cmark                 & \cmark  \\ \cline{1-1}
                    \multicolumn{1}{|c|}{$IoU$}  & 0.8811                 & 0.8301                 & 0.8030                 & 0.8558                 & 0.7981                 & 0.7934  \\ \cline{1-1}
                    \multicolumn{1}{|c|}{Median} & 26.0                   & 37.0                   & 32.0                   & 28.0                   & 55.0                   & 64.0  \\ \hline
                \end{tabular}
                \caption{Our algorithm}
                \label{tab:table_2b}
            \end{subtable}
            \caption{Tables of points}
            \label{tab:table_2}
        \end{table}
        \begin{itemize}
            \item Both methods have found all object occurrences, $6/6$.
            \item The $IoU$ measure is higher in our method, for any degree of distortion. We can observe a better performance when smooth distortions are present.
            \item The median is lower in our approach. This shows our ability to describe deformations.
        \end{itemize}
    \subsection{Many distorted object occurrences}
        \label{subsec:subsection_6.4.3}
        \begin{figure}[!h]
            \centering
            \includegraphics[width=0.5\linewidth]{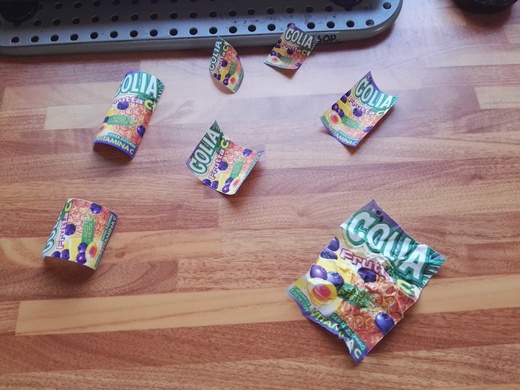}
            \caption{Many distorted object occurrences.}
            \label{fig:figure_91}
        \end{figure}
        \begin{figure}[!h]
            \centering
            \begin{subfigure}[!h]{0.32\linewidth}
                \includegraphics[width=\linewidth]{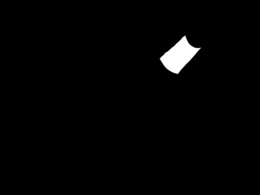}
            \end{subfigure}
            \begin{subfigure}[!h]{0.32\linewidth}
                \includegraphics[width=\linewidth]{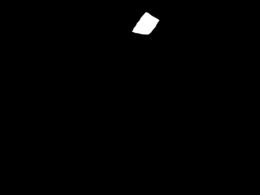}
            \end{subfigure}
            \begin{subfigure}[!h]{0.32\linewidth}
                \includegraphics[width=\linewidth]{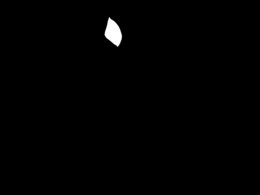}
            \end{subfigure}
            \begin{subfigure}[!h]{0.32\linewidth}
                \includegraphics[width=\linewidth]{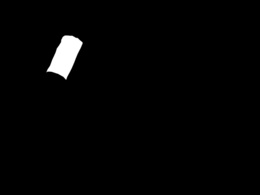}
            \end{subfigure}
            \begin{subfigure}[!h]{0.32\linewidth}
                \includegraphics[width=\linewidth]{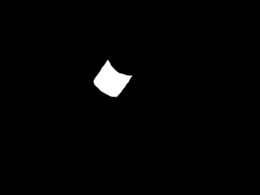}
            \end{subfigure}
            \begin{subfigure}[!h]{0.32\linewidth}
                \includegraphics[width=\linewidth]{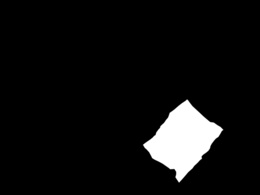}
            \end{subfigure}
            \caption{Ground truth mask for each object occurrence in \cref{fig:figure_91}.}
        \end{figure}
        Extensive figures of results of RANSAC driven and our algorithm in \cref{appendix_A.3.1}, \ref{appendix_A.3.2} respectively.
        \begin{figure}[!h]
            \centering
            \begin{subfigure}[!h]{0.48\linewidth}
                \includegraphics[width=\linewidth]{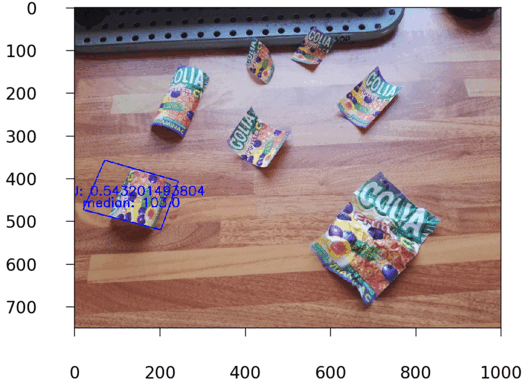}
            \end{subfigure}
            \begin{subfigure}[!h]{0.48\linewidth}
                \includegraphics[width=\linewidth]{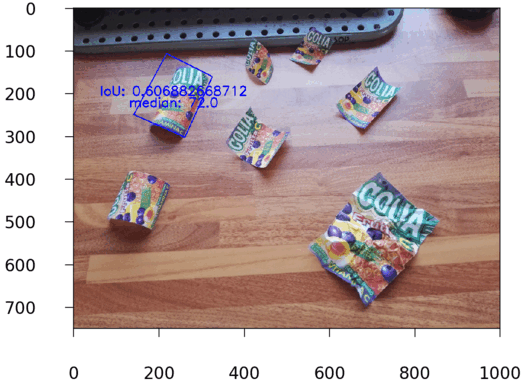}
            \end{subfigure}
            \begin{subfigure}[!h]{0.48\linewidth}
                \includegraphics[width=\linewidth]{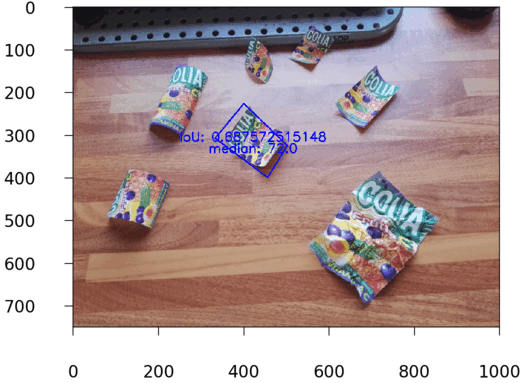}
            \end{subfigure}
            \begin{subfigure}[!h]{0.48\linewidth}
                \includegraphics[width=\linewidth]{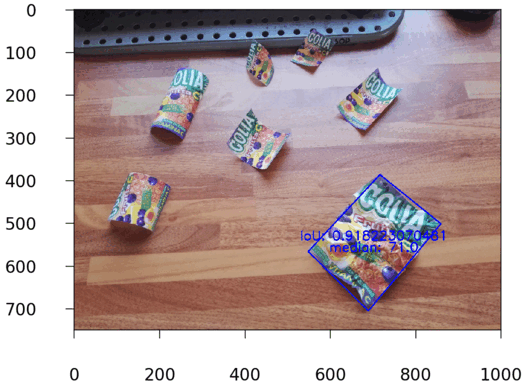}
            \end{subfigure}
            \begin{subfigure}[!h]{0.48\linewidth}
                \includegraphics[width=\linewidth]{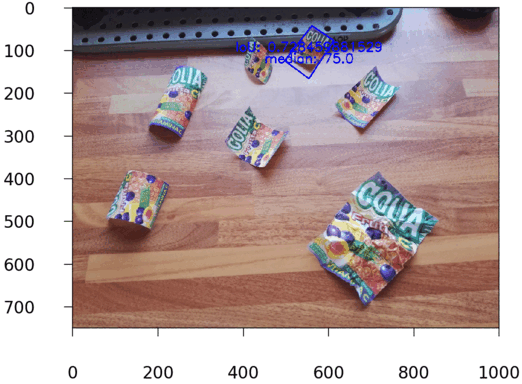}
            \end{subfigure}
            \begin{subfigure}[!h]{0.48\linewidth}
                \includegraphics[width=\linewidth]{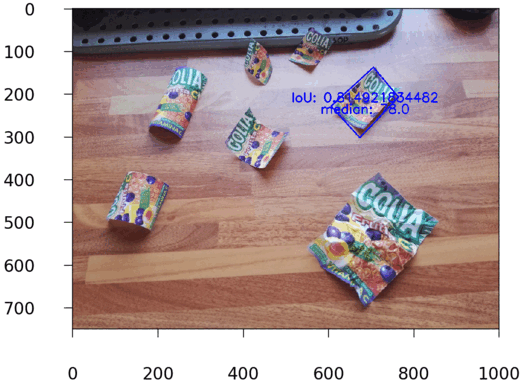}
            \end{subfigure}
            \begin{subfigure}[!h]{0.48\linewidth}
                \includegraphics[width=\linewidth]{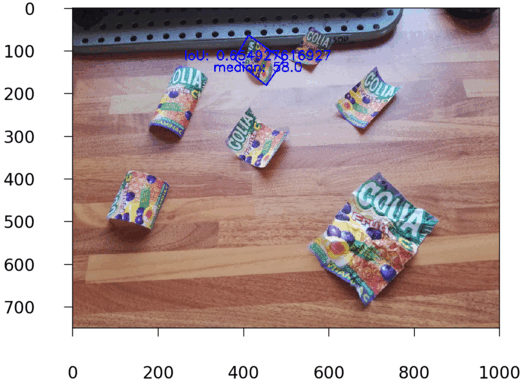}
            \end{subfigure}
            \caption{Object occurrences found by homography-based RANSAC algorithm.}
            \label{fig:figure_79}
        \end{figure}
        \begin{figure}[!h]
            \centering
            \begin{subfigure}[!h]{0.48\linewidth}
                \includegraphics[width=\linewidth]{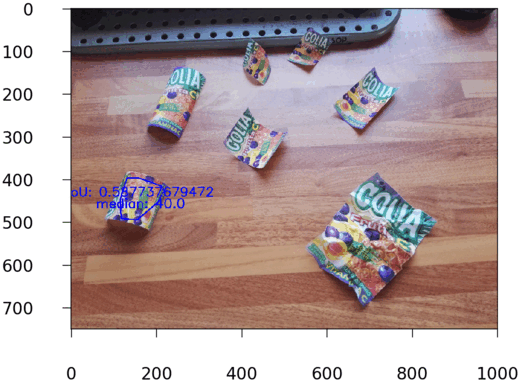}
            \end{subfigure}
            \begin{subfigure}[!h]{0.48\linewidth}
                \includegraphics[width=\linewidth]{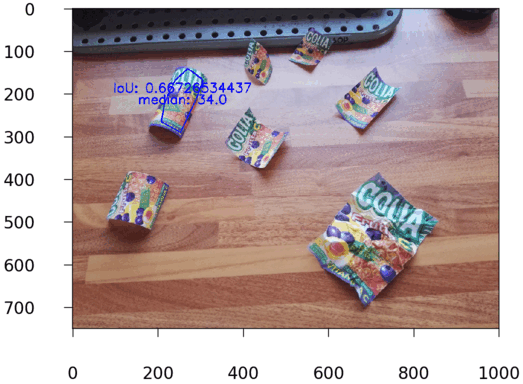}
            \end{subfigure}
            \begin{subfigure}[!h]{0.48\linewidth}
                \includegraphics[width=\linewidth]{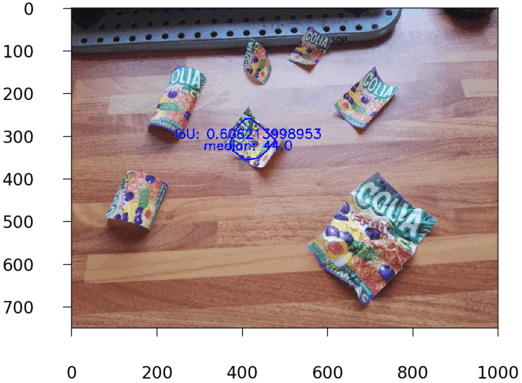}
            \end{subfigure}
            \begin{subfigure}[!h]{0.48\linewidth}
                \includegraphics[width=\linewidth]{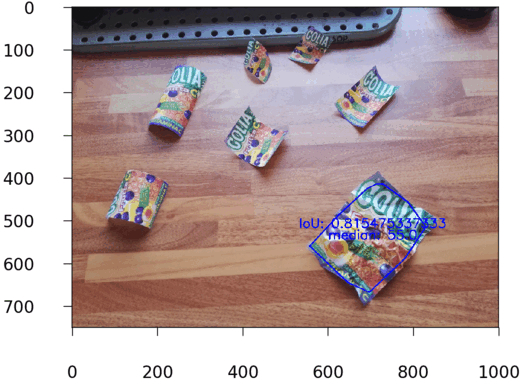}
            \end{subfigure}
            \begin{subfigure}[!h]{0.48\linewidth}
                \includegraphics[width=\linewidth]{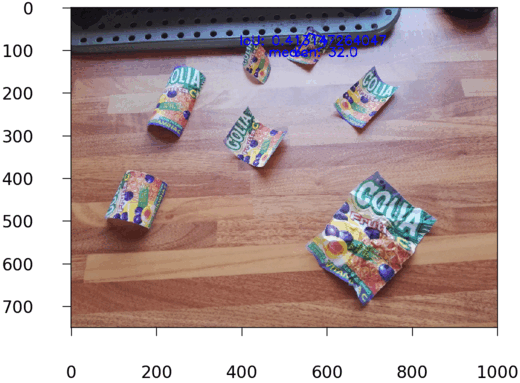}
            \end{subfigure}
            \begin{subfigure}[!h]{0.48\linewidth}
                \includegraphics[width=\linewidth]{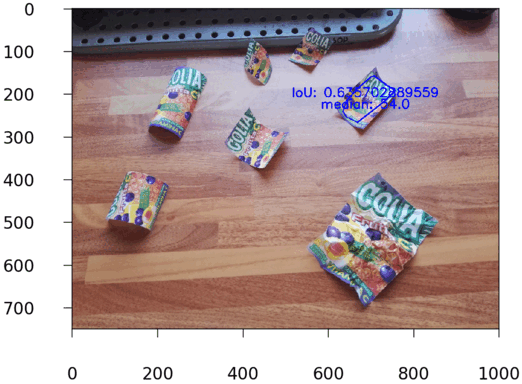}
            \end{subfigure}
            \begin{subfigure}[!h]{0.48\linewidth}
                \includegraphics[width=\linewidth]{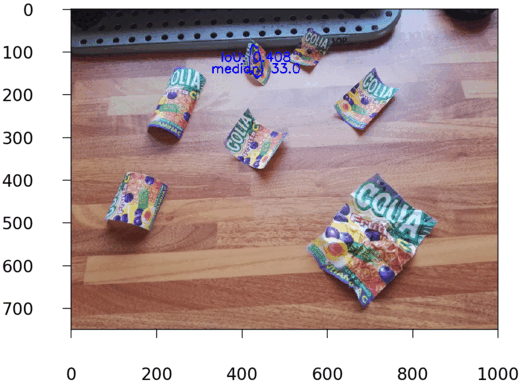}
            \end{subfigure}
            \caption{Object occurrences found by our algorithm.}
            \label{fig:figure_80}
        \end{figure}
        \begin{table}[!ht]
            \centering
            \begin{subtable}{\textwidth}
                \centering
                \begin{tabular}{c|ccccccc|}
                    \cline{2-8}
                                                 & \multicolumn{7}{c|}{Object occurrences}                                                                                        \\ \cline{2-8} 
                                                 & \multicolumn{1}{c|}{1\textsuperscript{st}} & \multicolumn{1}{c|}{2\textsuperscript{nd}} & \multicolumn{1}{c|}{3\textsuperscript{rd}} & \multicolumn{1}{c|}{4\textsuperscript{th}} & \multicolumn{1}{c|}{5\textsuperscript{th}} & \multicolumn{1}{c|}{6\textsuperscript{th}} & 7\textsuperscript{th} \\ \hline
                    \multicolumn{1}{|c|}{Identified}  & \cmark                 & \cmark                 & \cmark                 & \cmark                 & \cmark                 & \cmark  & \cmark \\ \cline{1-1}
                    \multicolumn{1}{|c|}{$IoU$}  & 0.5432                 & 0.6069                 & 0.6876                 & 0.9182                 & 0.7285                 & 0.8149  & 0.6549 \\ \cline{1-1}
                    \multicolumn{1}{|c|}{Median} & 103.0                   & 72.0                   & 72.0                   & 71.0                   & 75.0                   & 78.0  & 58.0 \\ \hline
                \end{tabular}
                \caption{Homography-based RANSAC}
                \label{tab:table_3a}
            \end{subtable}
            \\\vspace{5mm}
            \begin{subtable}{\textwidth}
                \centering
                \begin{tabular}{c|ccccccc|}
                    \cline{2-8}
                                                 & \multicolumn{7}{c|}{Object occurrences}                                                                                        \\ \cline{2-8} 
                                                 & \multicolumn{1}{c|}{1\textsuperscript{st}} & \multicolumn{1}{c|}{2\textsuperscript{nd}} & \multicolumn{1}{c|}{3\textsuperscript{rd}} & \multicolumn{1}{c|}{4\textsuperscript{th}} & \multicolumn{1}{c|}{5\textsuperscript{th}} & \multicolumn{1}{c|}{6\textsuperscript{th}} & 7\textsuperscript{th} \\ \hline
                    \multicolumn{1}{|c|}{Identified}  & \cmark                 & \cmark                 & \cmark                 & \cmark                 & \cmark                 & \cmark  & \cmark \\ \cline{1-1}
                    \multicolumn{1}{|c|}{$IoU$}  & 0.5877                 & 0.6673                 & 0.6062                 & 0.8155                 &    0.4131              & 0.6357  & 0.408 \\ \cline{1-1}
                    \multicolumn{1}{|c|}{Median} & 40.0                   & 34.0                   & 44.0                   & 55.0                   & 32.0                   & 54.0  & 33.0 \\ \hline
                \end{tabular}
                \caption{Our algorithm}
                \label{tab:table_3b}
            \end{subtable}
            \caption{Tables of points}
            \label{tab:table_3}
        \end{table}
        Evaluations in Table \cref{tab:table_2}.
        \begin{itemize}
            \item Both methods have found all object occurrences, $6/6$.
            \item The $IoU$ measure is generally higher in the RANSAC driven approach. Compared to the single object cases (\cref{subsec:subsection_6.4.3}) here we have a less detailed image (maximum dimension 1000 pixels for a larger landscape). This could have influenced the amount of good key points extracted, and consequently the number of available matches exploitable in seeds expansion.
            \item The median is lower in our approach. Showing also in that case that, inside the identified region, our approach has a greater capacity to adapt to deformations.
        \end{itemize}
    \thispagestyle{empty}
    \chapter{Future research directions and conclusions}
\label{ch:chapter_6}

\noindent
In this work we tackled the problem of object detection following a novel approach. In particular we shown the possibility to improve information coming from features and matches in order to deal with the non-trivial situation of distorted object occurrences.
\\
The main contributions are:
\begin{itemize}
    \item Extract a greater amount of information from the single template provided, exploiting the relative position of its features. This is done building triangulation of key points and using the produced mesh to iteratively grow the set of validated correspondences in a guided and object-dependent way (explained in \cref{sec:section_4.2} and \ref{sec:section_4.3}).
    \item Exploit all the information contained in key points (position, orientation and scale) and in their descriptors in order to validate the goodness of additional correspondences. We were inspired by the work of Oleg O. Sushkov and Claude Sammut\cite{ref:reference_1} to accomplish this task. In fact we utilized some concept introduced by them (described in \cref{subsubsec:subsubsection_4.3.5.1}, \ref{subsubsec:subsubsection_4.3.5.2}, \ref{subsubsec:subsubsection_4.3.5.3}), presented a new one (introduced in \cref{subsubsec:subsubsection_4.3.5.4}) and adapted their usage to follow our iterative approach.
\end{itemize}
We tested our work in scenarios with incremental difficulty, in which:
\begin{enumerate}
    \item Non-distorted object occurrences are present.
    \item Only one distorted object occurrence is present.
    \item Many distorted object occurrences are present.
\end{enumerate}
In the first case our method has performance similar to those of homography based method (\cref{subsec:subsection_6.4.1}); this scenario is the one for the which homography based method is intended for. In the second case our approach shows superior performance, proving its ability to deal with template object distortions (\cref{subsec:subsection_6.4.2}). In the last case the homography based approach has greater performance due to the presence of less key points in the scene and, as consequence, less correspondences between template and object instances (\cref{subsec:subsection_6.4.3}). In fact an homography based method needs only 4 support correspondences to estimate a full homography, while our method, in order to cover the whole object area, needs correspondences that refers to the entire object surface.
\\\\
Given these results, the first improvement to the method is the usage of a different feature detector and descriptor in order to have correspondences on the whole object area. In fact we need a detector able to find the same feature despite high perspective effects arising from object distortions. Although ASIFT allows to find more correspondences with respect to plain SIFT, this is not enough. Maybe a deep detector and descriptor could be able to recognize and describe more object parts under many deformations, and at the same time to reduce false positive correspondences.
\\
Another possible advancement is to delegate to a neural network, trained with synthetic distortions of the object, the decision about the relative weight of each consistency score and the subsequent threshold value (\cref{subsubsec:subsubsection_4.3.5.5}) for admission or rejection of a new match in the seed correspondences set.
\\
The last improvement is about execution time. In fact the parallelization of the execution is a natural way to deal with our multi-seed approach in a faster way, as explained in \cref{sec:section_5.2}.
    \thispagestyle{empty}
    
    \cleardoublepage
    \printbibliography
    
    \appendix
        \pagenumbering{Roman}
        \chapter{Experiments images}
\label{appendix_A}

\section{Many undistorted object occurrences}
    \paragraph{Feature-based object detection algorithm driven by RANSAC}
        \label{appendix_A.1.1}
        \begin{figure}[H]
    \centering
    \begin{subfigure}[!h]{0.7\linewidth}
        \includegraphics[width=\linewidth]{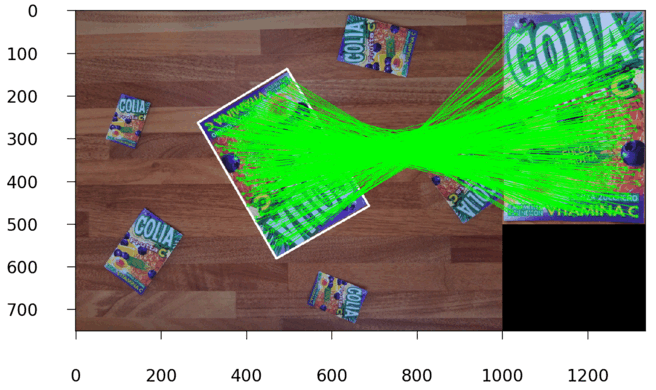}
        \caption{Clustered matches.}
        \label{fig:figure_30a}
    \end{subfigure}
    \begin{subfigure}[!h]{0.35\linewidth}
        \includegraphics[width=\linewidth]{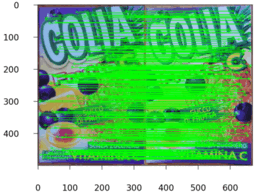}
        \caption{Rectified object occurrence.}
        \label{fig:figure_30b}
    \end{subfigure}
    \begin{subfigure}[!h]{0.35\linewidth}
        \includegraphics[width=\linewidth]{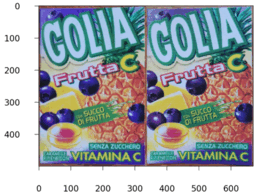}
        \caption{Template and histogram matched object.}
        \label{fig:figure_30c}
    \end{subfigure}
\end{figure}
\begin{figure}[H]
    \ContinuedFloat
    \centering
    \begin{subfigure}[!h]{0.3\linewidth}
        \includegraphics[width=\linewidth]{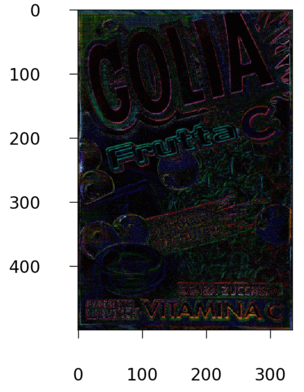}
        \caption{BGR absolute difference.}
        \label{fig:figure_30d}
    \end{subfigure}
    \begin{subfigure}[!h]{0.4\linewidth}
        \includegraphics[width=\linewidth]{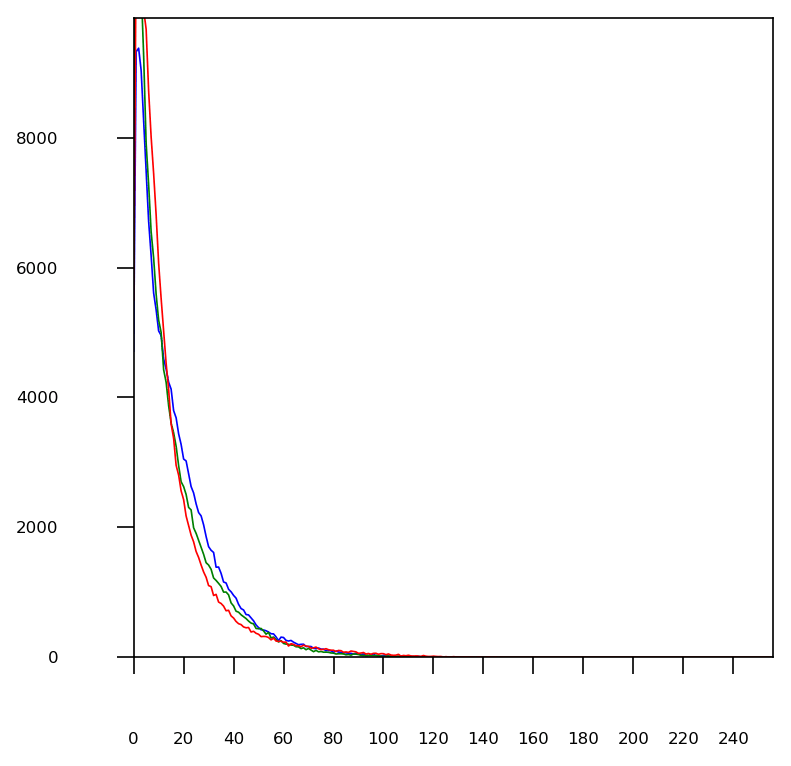}
        \caption{BGR difference histogram.}
        \label{fig:figure_30h}
    \end{subfigure}
    \begin{subfigure}[!h]{0.3\linewidth}
        \includegraphics[width=\linewidth]{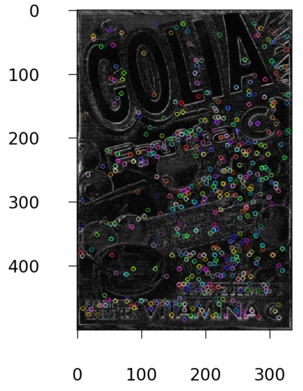}
        \caption{Pixelwise difference norm.}
        \label{fig:figure_30i}
    \end{subfigure}
    \begin{subfigure}[!h]{0.4\linewidth}
        \includegraphics[width=\linewidth]{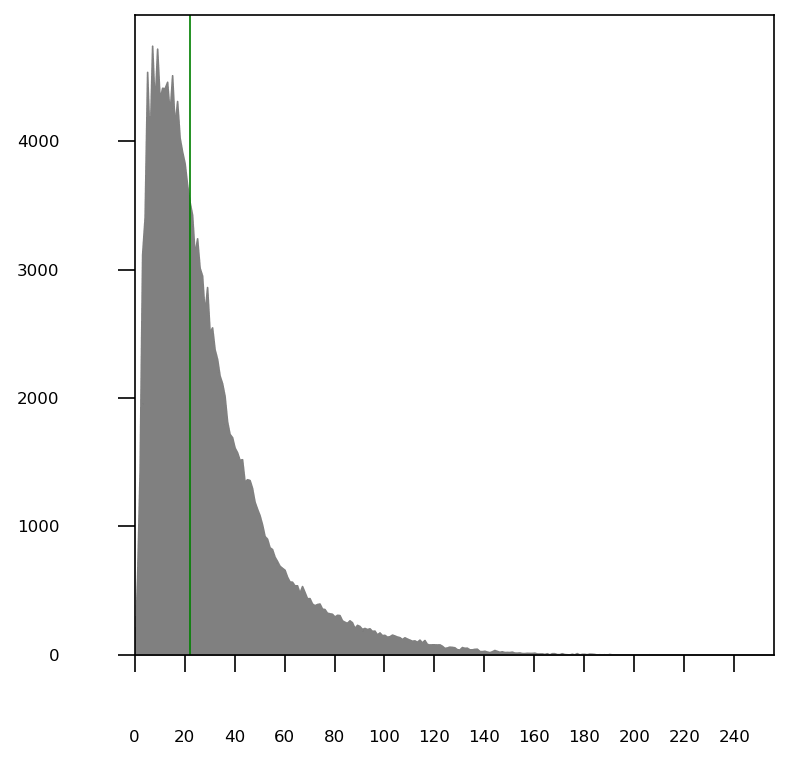}
        \caption{Pixelwise difference norm histogram.}
        \label{fig:figure_30j}
    \end{subfigure}
    \caption{1\textsuperscript{st} object occurrence.}
    \label{fig:figure_30}
\end{figure}
\begin{figure}[H]
    \centering
    \begin{subfigure}[!h]{0.7\linewidth}
        \includegraphics[width=\linewidth]{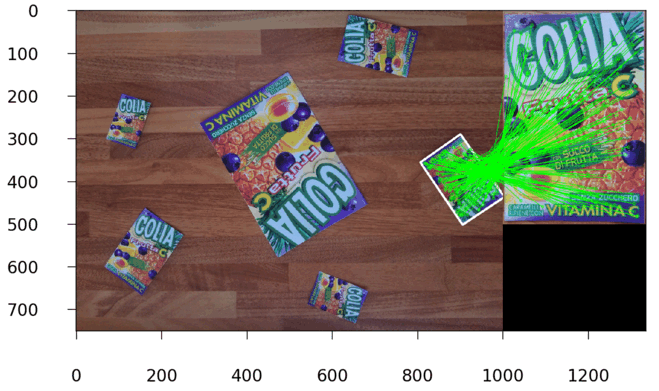}
        \caption{Clustered matches.}
    \end{subfigure}
    \begin{subfigure}[!h]{0.35\linewidth}
        \includegraphics[width=\linewidth]{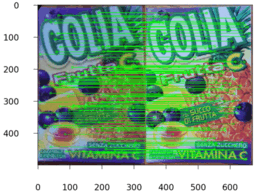}
        \caption{Rectified object occurrence.}
    \end{subfigure}
    \begin{subfigure}[!h]{0.35\linewidth}
        \includegraphics[width=\linewidth]{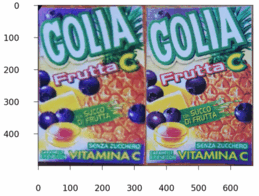}
        \caption{Template and histogram matched object.}
    \end{subfigure}
    \begin{subfigure}[!h]{0.3\linewidth}
        \includegraphics[width=\linewidth]{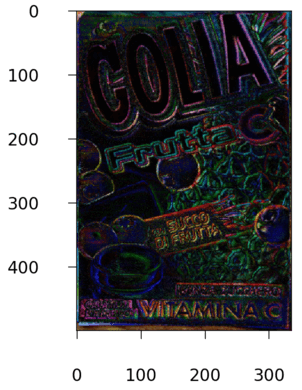}
        \caption{BGR absolute difference.}
    \end{subfigure}
    \begin{subfigure}[!h]{0.4\linewidth}
        \includegraphics[width=\linewidth]{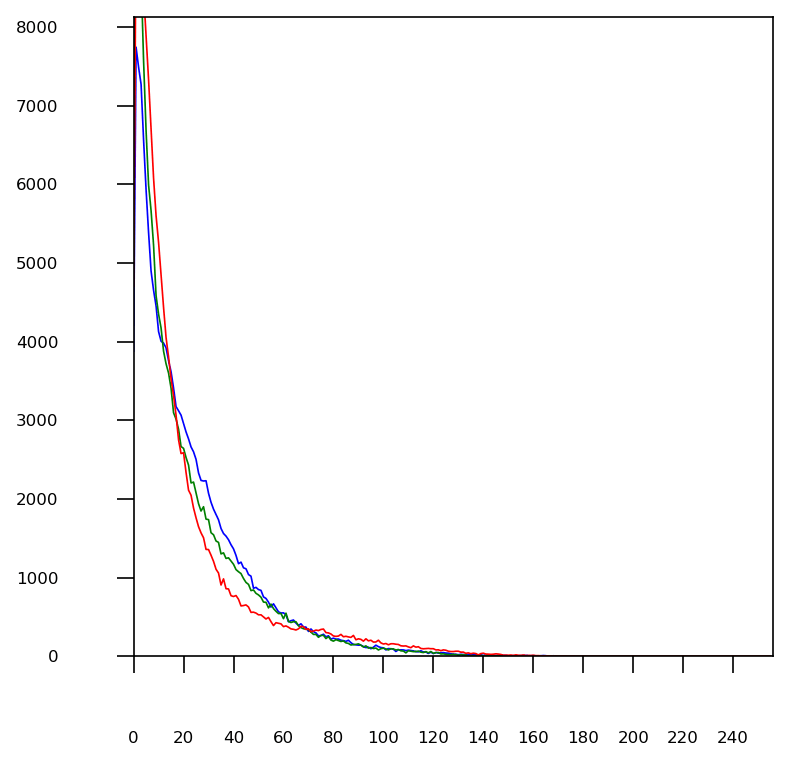}
        \caption{BGR difference histogram.}
    \end{subfigure}
\end{figure}
\begin{figure}[H]
    \ContinuedFloat
    \centering
    \begin{subfigure}[!h]{0.3\linewidth}
        \includegraphics[width=\linewidth]{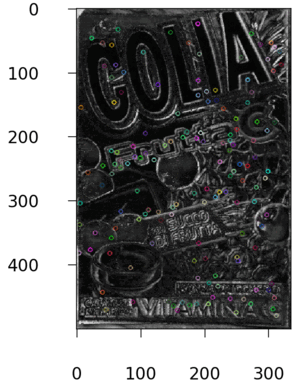}
        \caption{Pixelwise difference norm.}
    \end{subfigure}
    \begin{subfigure}[!h]{0.4\linewidth}
        \includegraphics[width=\linewidth]{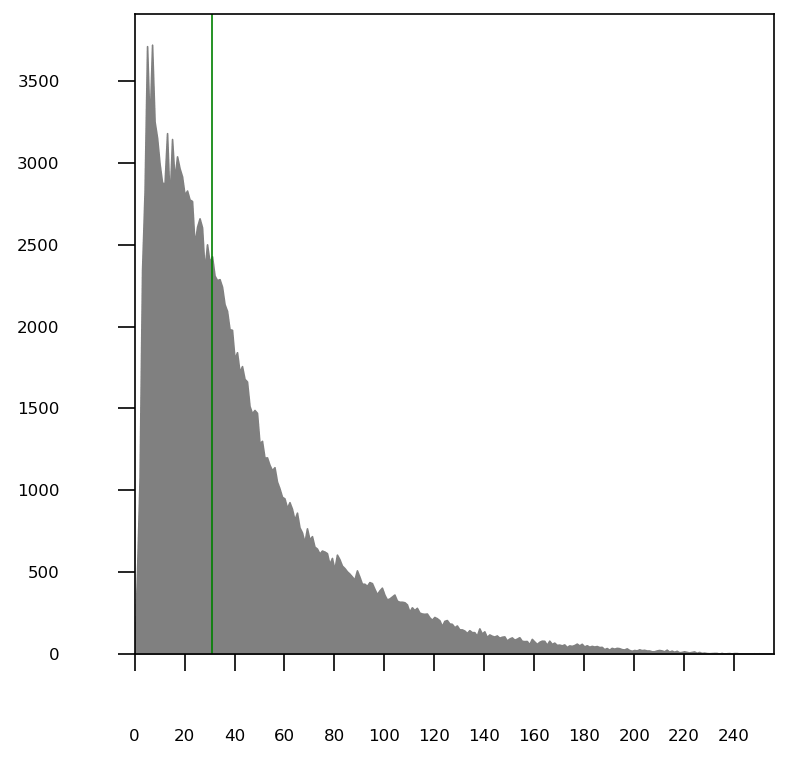}
        \caption{Pixelwise difference norm histogram.}
    \end{subfigure}
    \caption{2\textsuperscript{nd} object occurrence.}
    \label{fig:figure_31}
\end{figure}
\begin{figure}[H]
    \centering
    \begin{subfigure}[!h]{0.7\linewidth}
        \includegraphics[width=\linewidth]{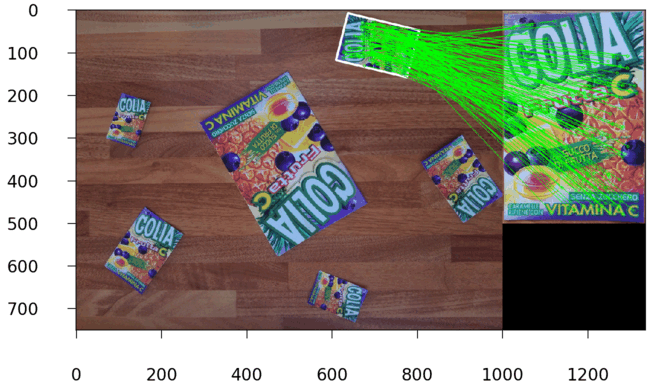}
        \caption{Clustered matches.}
    \end{subfigure}
    \begin{subfigure}[!h]{0.35\linewidth}
        \includegraphics[width=\linewidth]{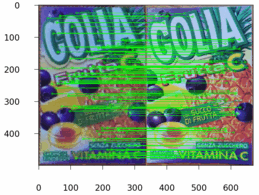}
        \caption{Rectified object occurrence.}
    \end{subfigure}
    \begin{subfigure}[!h]{0.35\linewidth}
        \includegraphics[width=\linewidth]{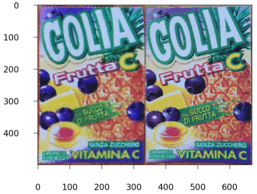}
        \caption{Template and histogram matched object.}
    \end{subfigure}
    \begin{subfigure}[!h]{0.3\linewidth}
        \includegraphics[width=\linewidth]{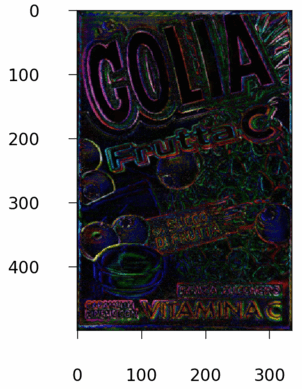}
        \caption{BGR absolute difference.}
    \end{subfigure}
    \begin{subfigure}[!h]{0.4\linewidth}
        \includegraphics[width=\linewidth]{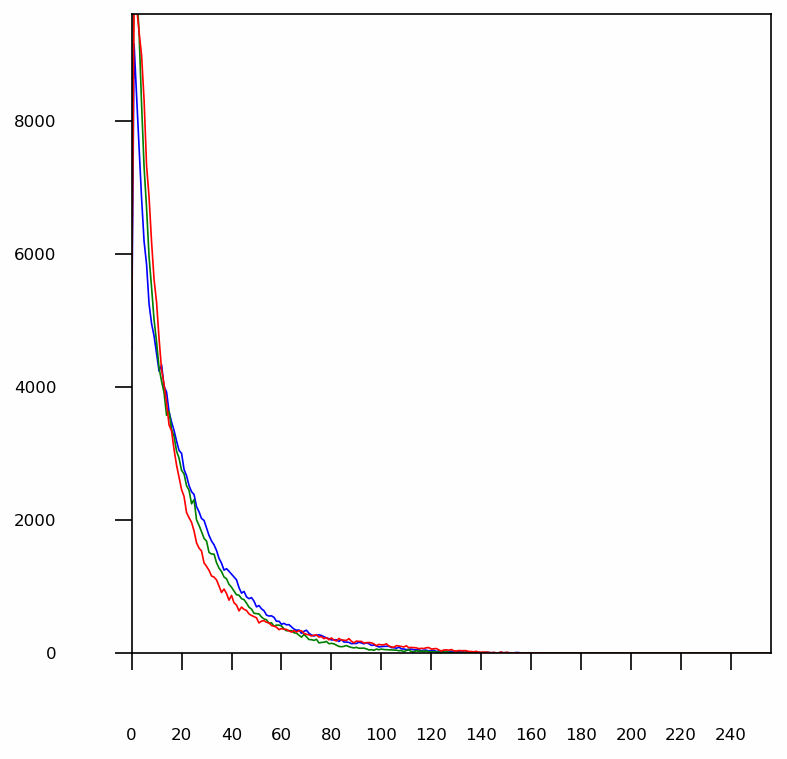}
        \caption{BGR difference histogram.}
    \end{subfigure}
\end{figure}
\begin{figure}[H]
    \ContinuedFloat
    \centering
    \begin{subfigure}[!h]{0.3\linewidth}
        \includegraphics[width=\linewidth]{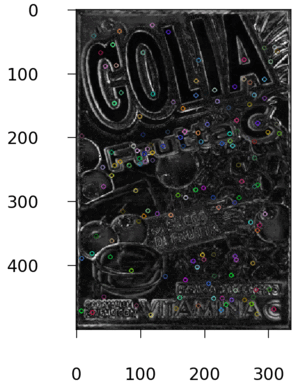}
        \caption{Pixelwise difference norm.}
    \end{subfigure}
    \begin{subfigure}[!h]{0.4\linewidth}
        \includegraphics[width=\linewidth]{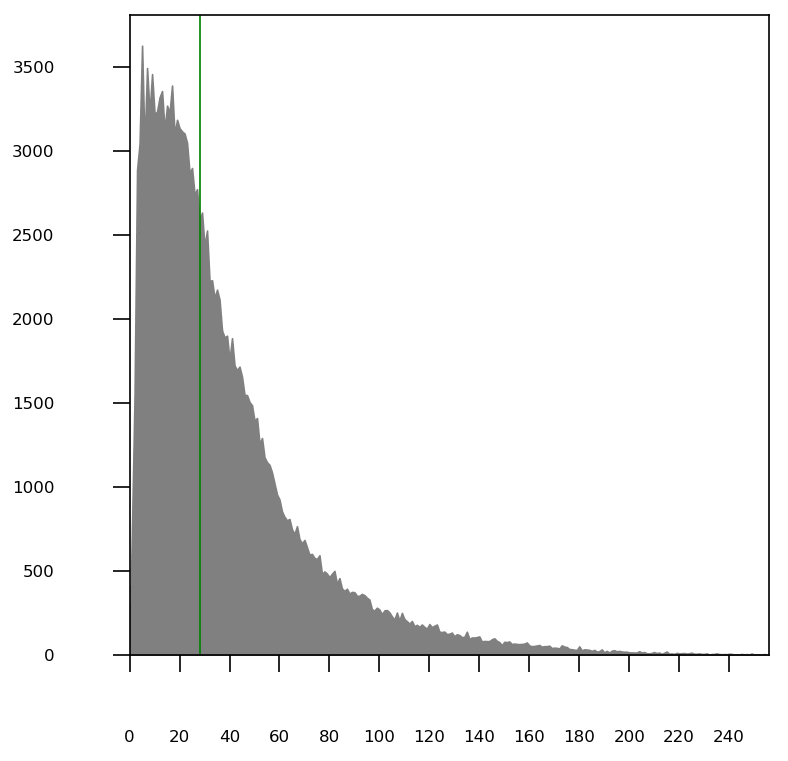}
        \caption{Pixelwise difference norm histogram.}
    \end{subfigure}
    \caption{3\textsuperscript{rd} object occurrence.}
    \label{fig:figure_32}
\end{figure}
\begin{figure}[H]
    \centering
    \begin{subfigure}[!h]{0.7\linewidth}
        \includegraphics[width=\linewidth]{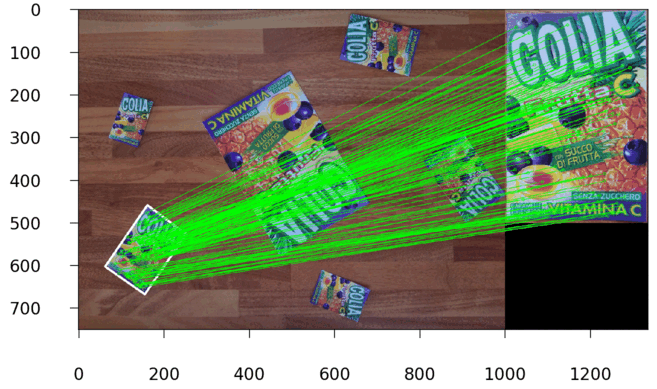}
        \caption{Clustered matches.}
    \end{subfigure}
    \begin{subfigure}[!h]{0.35\linewidth}
        \includegraphics[width=\linewidth]{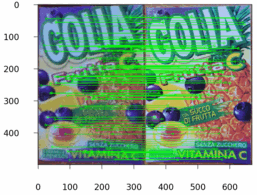}
        \caption{Rectified object occurrence.}
    \end{subfigure}
    \begin{subfigure}[!h]{0.35\linewidth}
        \includegraphics[width=\linewidth]{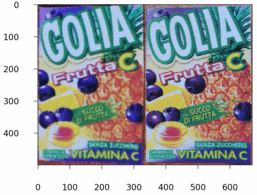}
        \caption{Template and histogram matched object.}
    \end{subfigure}
    \begin{subfigure}[!h]{0.3\linewidth}
        \includegraphics[width=\linewidth]{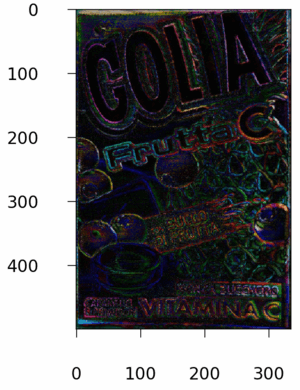}
        \caption{BGR absolute difference.}
    \end{subfigure}
    \begin{subfigure}[!h]{0.4\linewidth}
        \includegraphics[width=\linewidth]{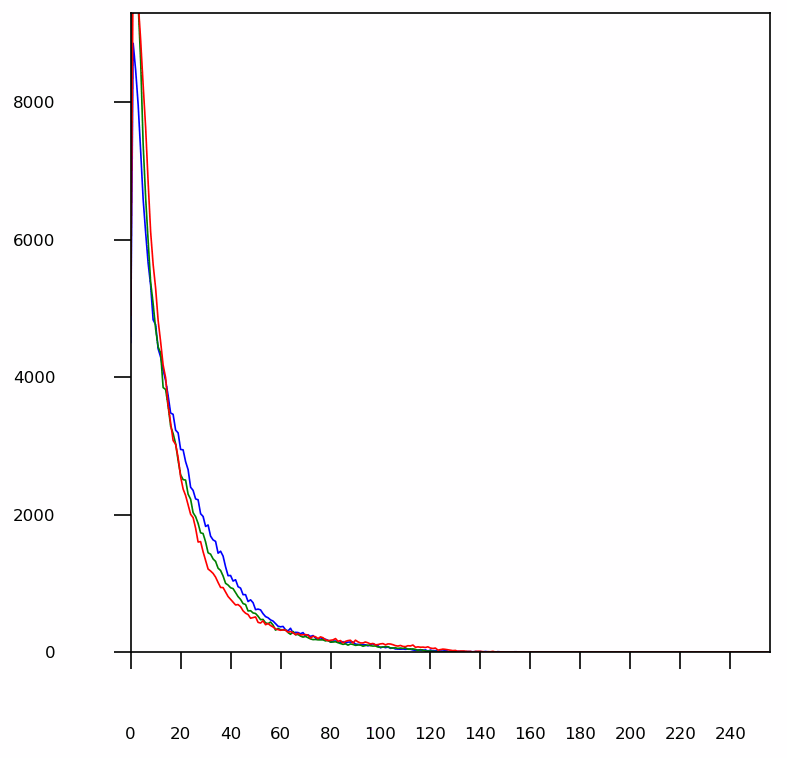}
        \caption{BGR difference histogram.}
    \end{subfigure}
\end{figure}
\begin{figure}[H]
    \ContinuedFloat
    \centering
    \begin{subfigure}[!h]{0.3\linewidth}
        \includegraphics[width=\linewidth]{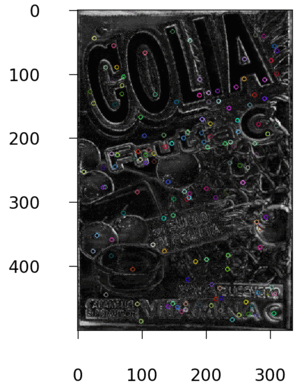}
        \caption{Pixelwise difference norm.}
    \end{subfigure}
    \begin{subfigure}[!h]{0.4\linewidth}
        \includegraphics[width=\linewidth]{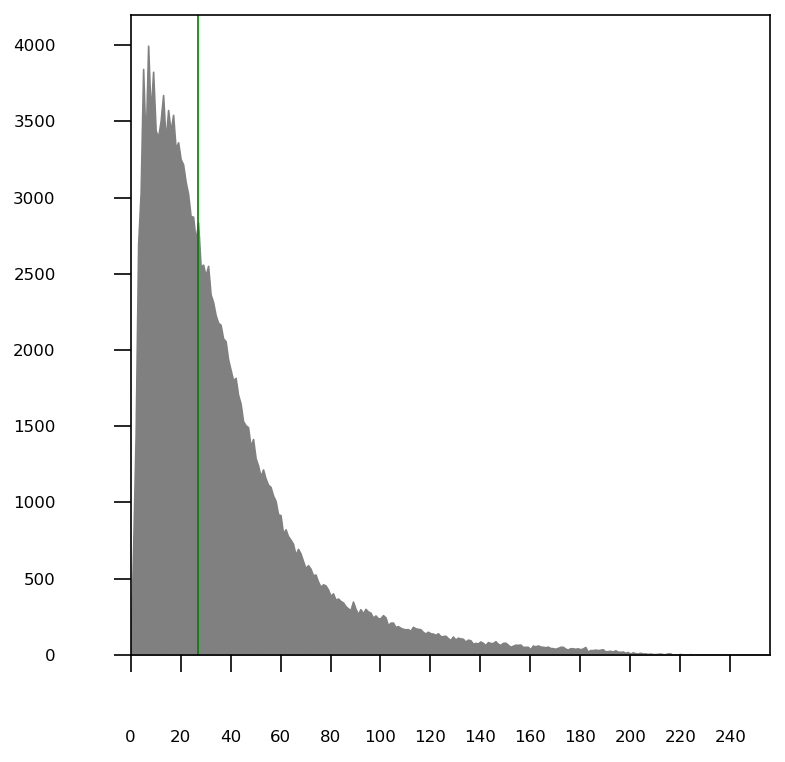}
        \caption{Pixelwise difference norm histogram.}
    \end{subfigure}
    \caption{4\textsuperscript{th} object occurrence.}
    \label{fig:figure_33}
\end{figure}
\begin{figure}[H]
    \centering
    \begin{subfigure}[!h]{0.7\linewidth}
        \includegraphics[width=\linewidth]{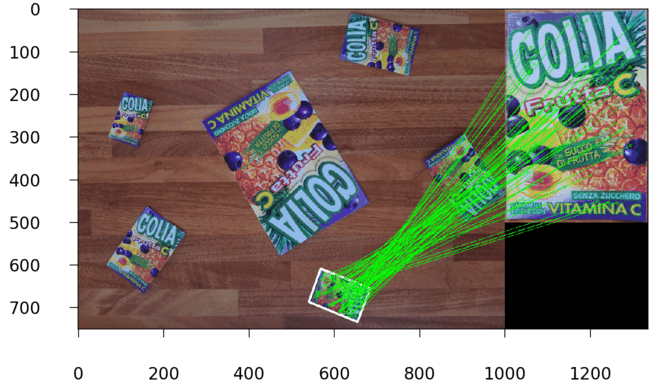}
        \caption{Clustered matches.}
    \end{subfigure}
    \begin{subfigure}[!h]{0.35\linewidth}
        \includegraphics[width=\linewidth]{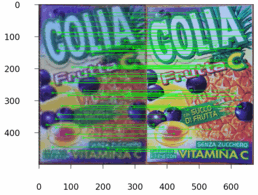}
        \caption{Rectified object occurrence.}
    \end{subfigure}
    \begin{subfigure}[!h]{0.35\linewidth}
        \includegraphics[width=\linewidth]{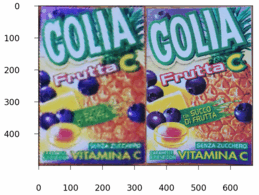}
        \caption{Template and histogram matched object.}
    \end{subfigure}
    \begin{subfigure}[!h]{0.3\linewidth}
        \includegraphics[width=\linewidth]{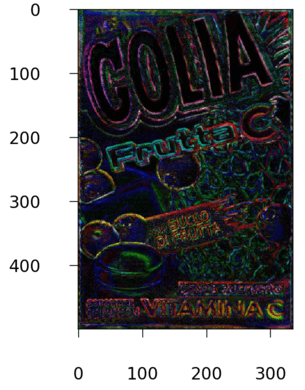}
        \caption{BGR absolute difference.}
    \end{subfigure}
    \begin{subfigure}[!h]{0.4\linewidth}
        \includegraphics[width=\linewidth]{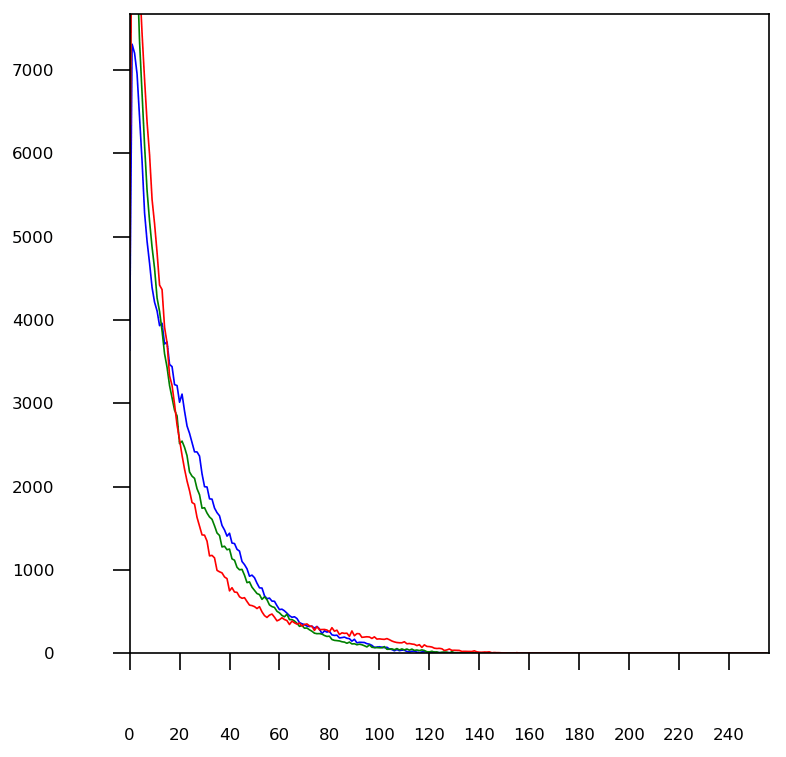}
        \caption{BGR difference histogram.}
    \end{subfigure}
\end{figure}
\begin{figure}[H]
    \ContinuedFloat
    \centering
    \begin{subfigure}[!h]{0.3\linewidth}
        \includegraphics[width=\linewidth]{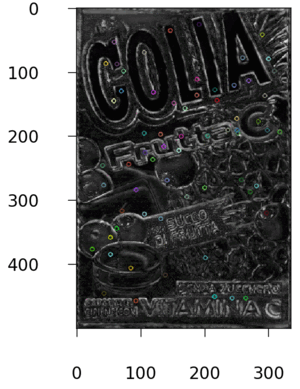}
        \caption{Pixelwise difference norm.}
    \end{subfigure}
    \begin{subfigure}[!h]{0.4\linewidth}
        \includegraphics[width=\linewidth]{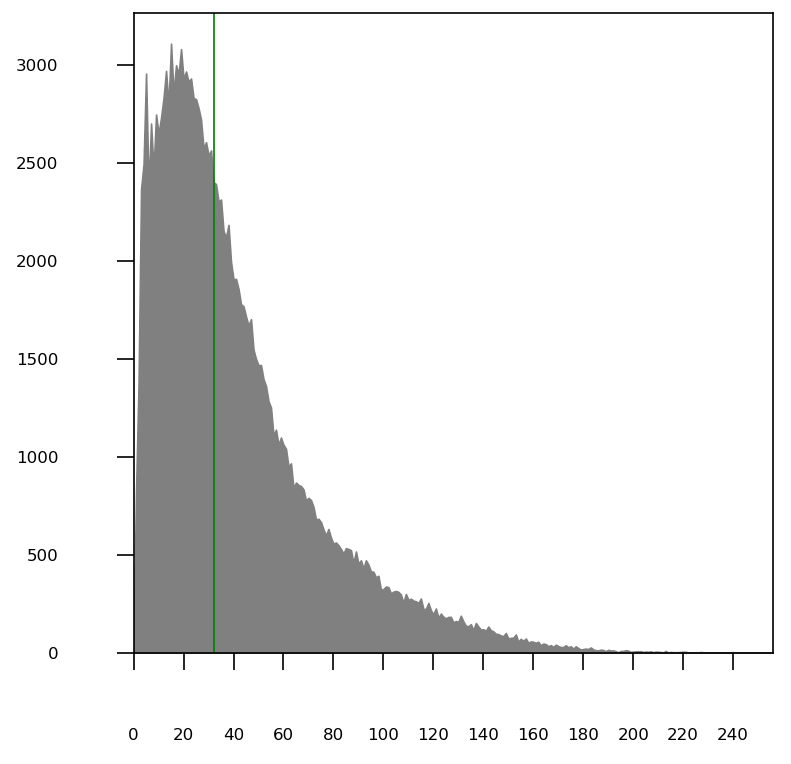}
        \caption{Pixelwise difference norm histogram.}
    \end{subfigure}
    \caption{5\textsuperscript{th} object occurrence.}
    \label{fig:figure_34}
\end{figure}
\begin{figure}[H]
    \centering
    \begin{subfigure}[!h]{0.7\linewidth}
        \includegraphics[width=\linewidth]{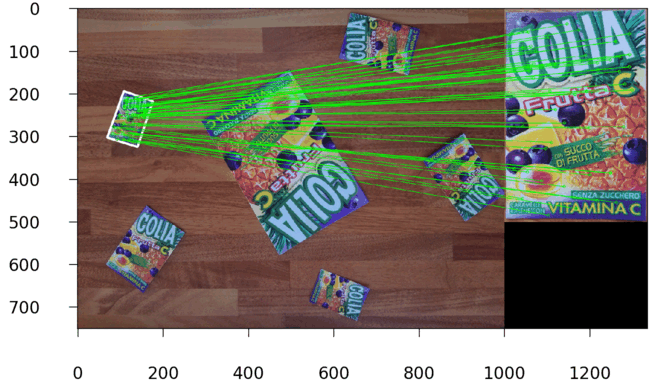}
        \caption{Clustered matches.}
    \end{subfigure}
    \begin{subfigure}[!h]{0.35\linewidth}
        \includegraphics[width=\linewidth]{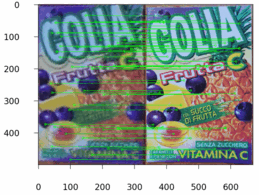}
        \caption{Rectified object occurrence.}
    \end{subfigure}
    \begin{subfigure}[!h]{0.35\linewidth}
        \includegraphics[width=\linewidth]{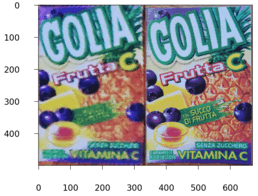}
        \caption{Template and histogram matched object.}
    \end{subfigure}
    \begin{subfigure}[!h]{0.3\linewidth}
        \includegraphics[width=\linewidth]{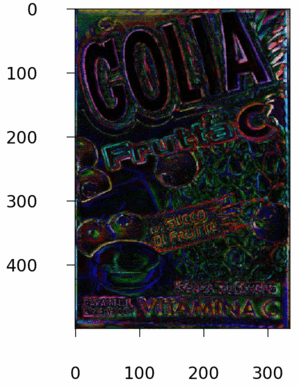}
        \caption{BGR absolute difference.}
    \end{subfigure}
    \begin{subfigure}[!h]{0.4\linewidth}
        \includegraphics[width=\linewidth]{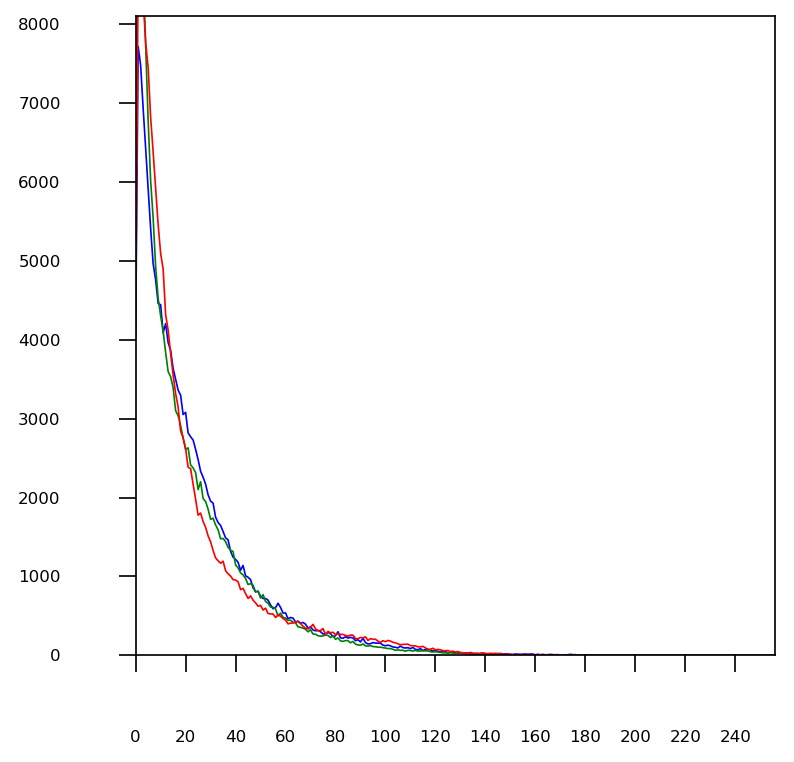}
        \caption{BGR difference histogram.}
    \end{subfigure}
\end{figure}
\begin{figure}[H]
    \ContinuedFloat
    \centering
    \begin{subfigure}[!h]{0.3\linewidth}
        \includegraphics[width=\linewidth]{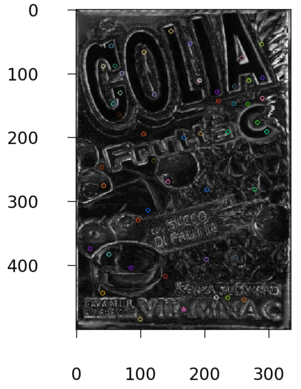}
        \caption{Pixelwise difference norm.}
    \end{subfigure}
    \begin{subfigure}[!h]{0.4\linewidth}
        \includegraphics[width=\linewidth]{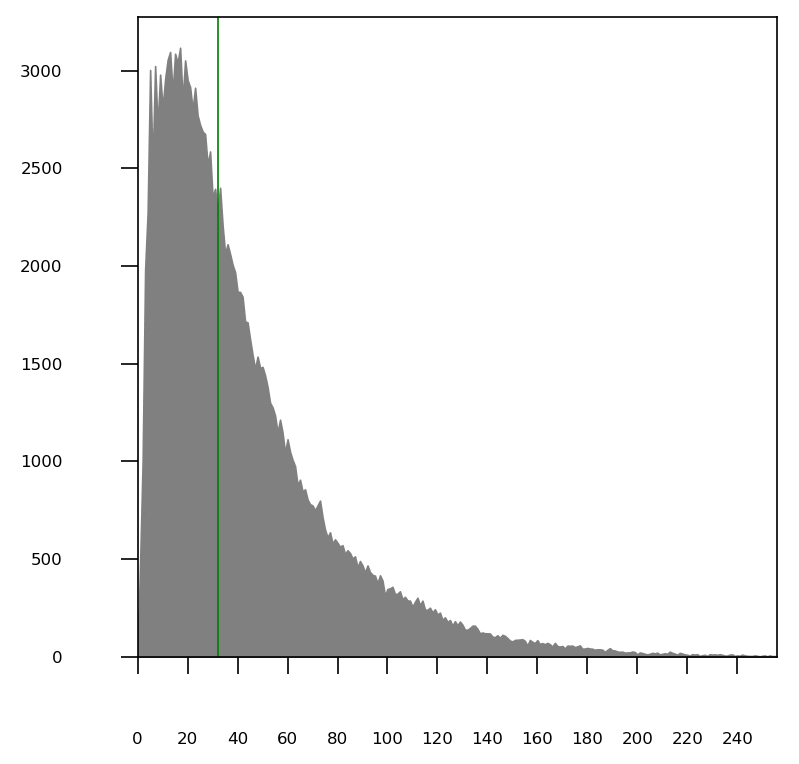}
        \caption{Pixelwise difference norm histogram.}
    \end{subfigure}
    \caption{6\textsuperscript{th} object occurrence.}
    \label{fig:figure_35}
\end{figure}
    \paragraph{Our algorithm}
        \label{appendix_A.1.2}
        \begin{figure}[H]
    \centering
    \begin{subfigure}[!h]{0.7\linewidth}
        \includegraphics[width=\linewidth]{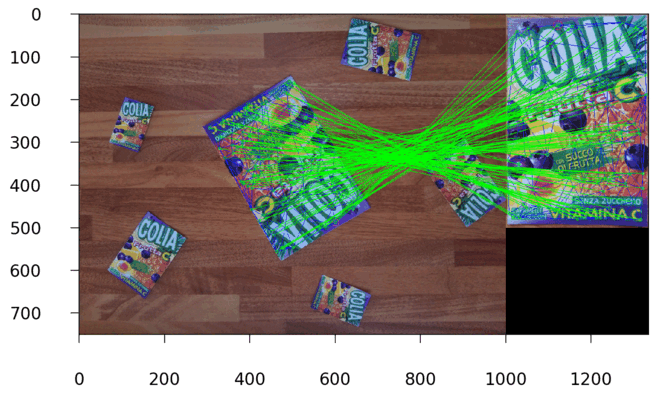}
        \caption{Expanded seed.}
        \label{fig:figure_36a}
    \end{subfigure}
    \begin{subfigure}[!h]{0.35\linewidth}
        \includegraphics[width=\linewidth]{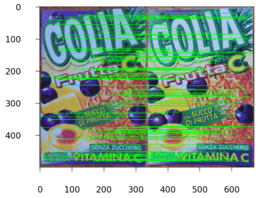}
        \caption{Rectified object occurrence.}
        \label{fig:figure_36b}
    \end{subfigure}
    \begin{subfigure}[!h]{0.35\linewidth}
        \includegraphics[width=\linewidth]{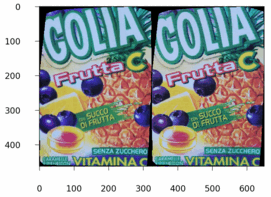}
        \caption{Template and histogram matched object.}
        \label{fig:figure_36c}
    \end{subfigure}
    \begin{subfigure}[!h]{0.3\linewidth}
        \includegraphics[width=\linewidth]{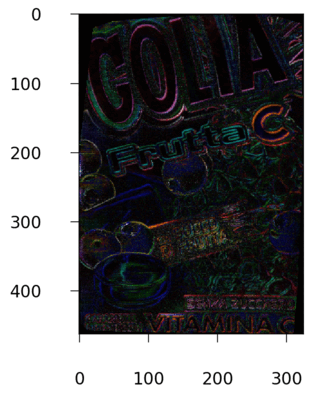}
        \caption{BGR absolute difference.}
        \label{fig:figure_36d}
    \end{subfigure}
    \begin{subfigure}[!h]{0.4\linewidth}
        \includegraphics[width=\linewidth]{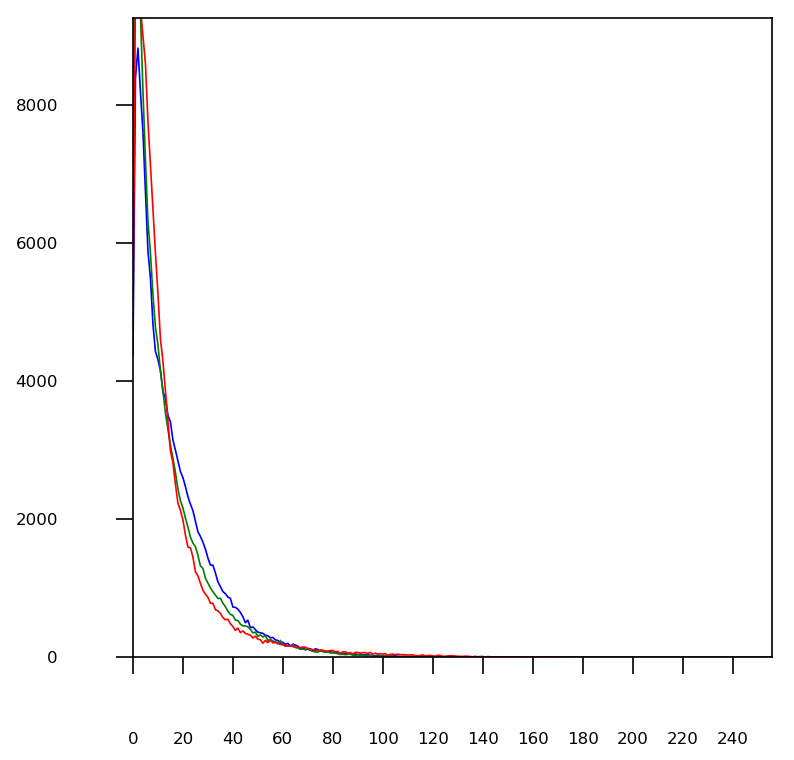}
        \caption{BGR difference histogram.}
        \label{fig:figure_36h}
    \end{subfigure}
\end{figure}
\begin{figure}[H]
    \ContinuedFloat
    \centering
    \begin{subfigure}[!h]{0.3\linewidth}
        \includegraphics[width=\linewidth]{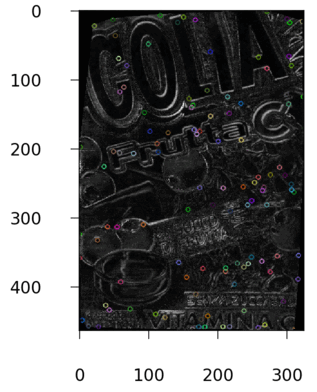}
        \caption{Pixelwise difference norm.}
        \label{fig:figure_36i}
    \end{subfigure}
    \begin{subfigure}[!h]{0.4\linewidth}
        \includegraphics[width=\linewidth]{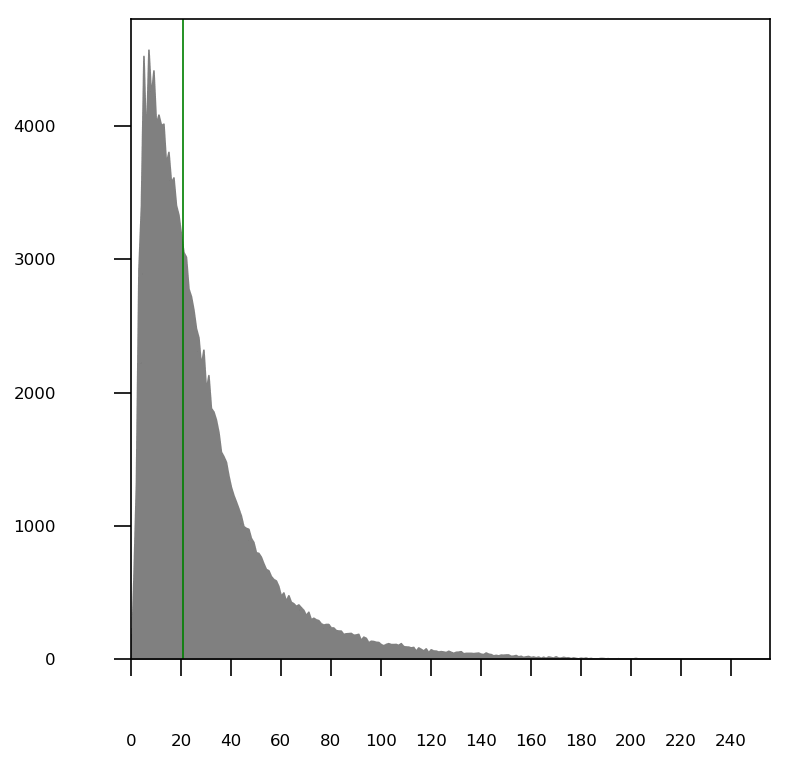}
        \caption{Pixelwise difference norm histogram.}
        \label{fig:figure_36j}
    \end{subfigure}
    \caption{1\textsuperscript{st} object occurrence (union of seeds \#0-1-2-4-6-8).}
    \label{fig:figure_36}
\end{figure}
\begin{figure}[H]
    \centering
    \begin{subfigure}[!h]{0.7\linewidth}
        \includegraphics[width=\linewidth]{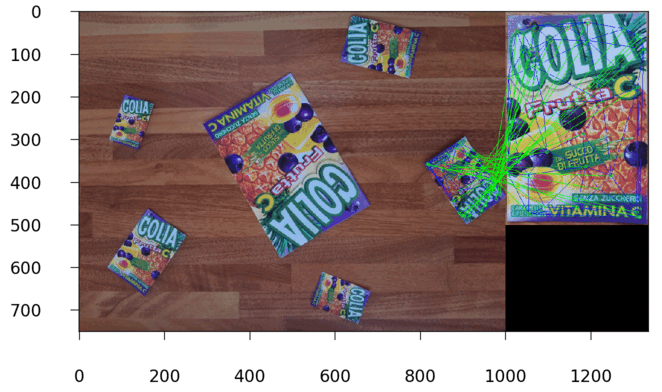}
        \caption{Expanded seed.}
    \end{subfigure}
    \begin{subfigure}[!h]{0.35\linewidth}
        \includegraphics[width=\linewidth]{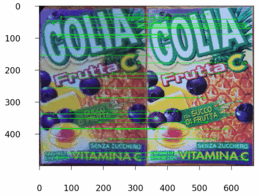}
        \caption{Rectified object occurrence.}
    \end{subfigure}
    \begin{subfigure}[!h]{0.35\linewidth}
        \includegraphics[width=\linewidth]{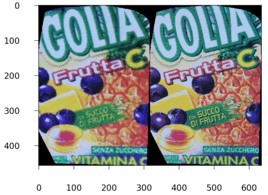}
        \caption{Template and histogram matched object.}
    \end{subfigure}
    \begin{subfigure}[!h]{0.3\linewidth}
        \includegraphics[width=\linewidth]{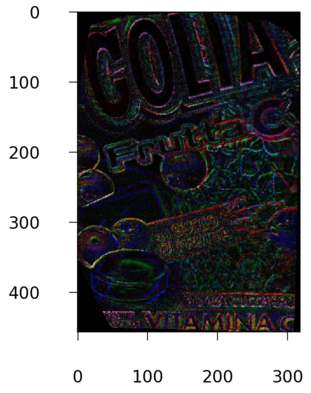}
        \caption{BGR absolute difference.}
    \end{subfigure}
    \begin{subfigure}[!h]{0.4\linewidth}
        \includegraphics[width=\linewidth]{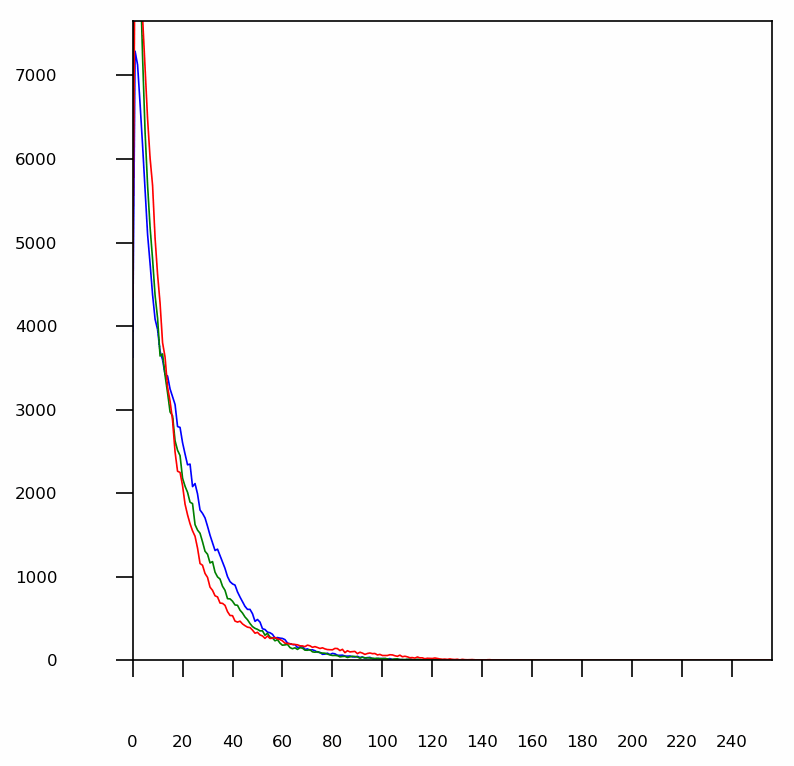}
        \caption{BGR difference histogram.}
    \end{subfigure}
\end{figure}
\begin{figure}[H]
    \ContinuedFloat
    \centering
    \begin{subfigure}[!h]{0.3\linewidth}
        \includegraphics[width=\linewidth]{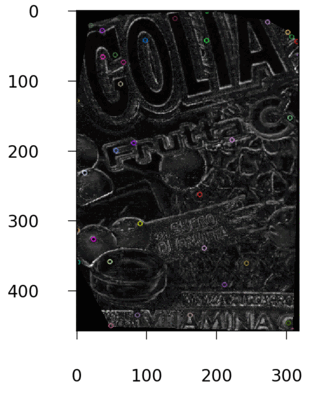}
        \caption{Pixelwise difference norm.}
    \end{subfigure}
    \begin{subfigure}[!h]{0.4\linewidth}
        \includegraphics[width=\linewidth]{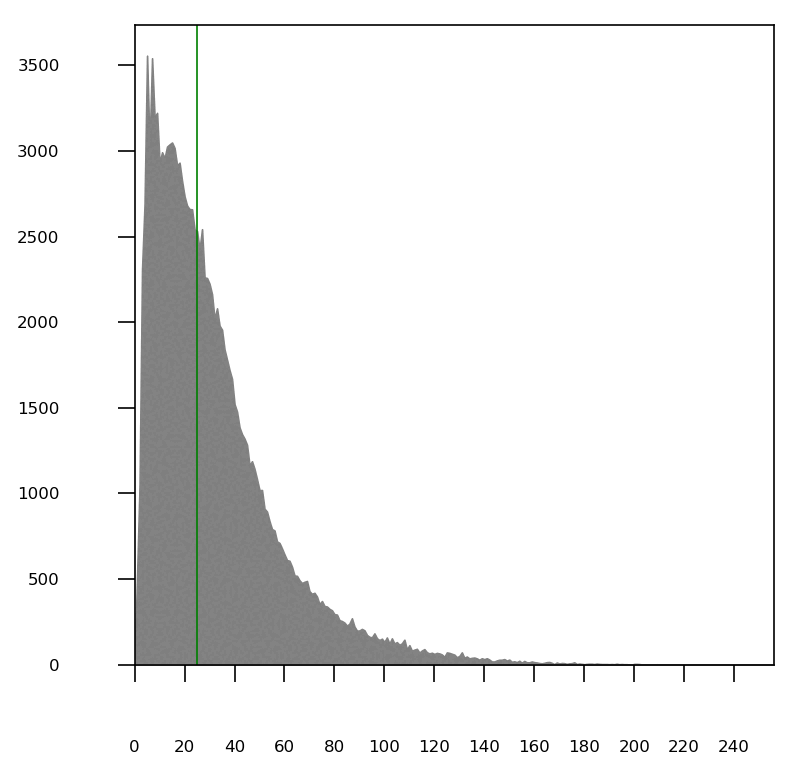}
        \caption{Pixelwise difference norm histogram.}
    \end{subfigure}
    \caption{2\textsuperscript{th} object occurrence (seed \#9).}
    \label{fig:figure_37}
\end{figure}
\begin{figure}[H]
    \centering
    \begin{subfigure}[!h]{0.7\linewidth}
        \includegraphics[width=\linewidth]{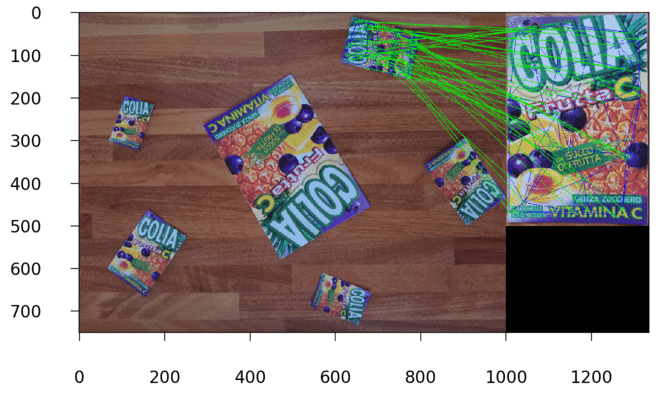}
        \caption{Expanded seed.}
    \end{subfigure}
    \begin{subfigure}[!h]{0.35\linewidth}
        \includegraphics[width=\linewidth]{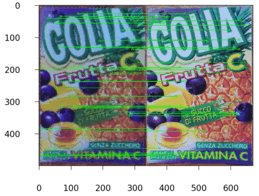}
        \caption{Rectified object occurrence.}
    \end{subfigure}
    \begin{subfigure}[!h]{0.35\linewidth}
        \includegraphics[width=\linewidth]{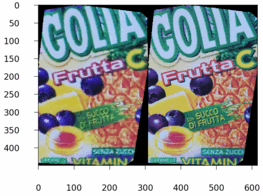}
        \caption{Template and histogram matched object.}
    \end{subfigure}
    \begin{subfigure}[!h]{0.3\linewidth}
        \includegraphics[width=\linewidth]{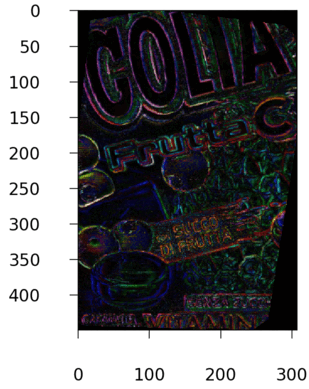}
        \caption{BGR absolute difference.}
    \end{subfigure}
    \begin{subfigure}[!h]{0.4\linewidth}
        \includegraphics[width=\linewidth]{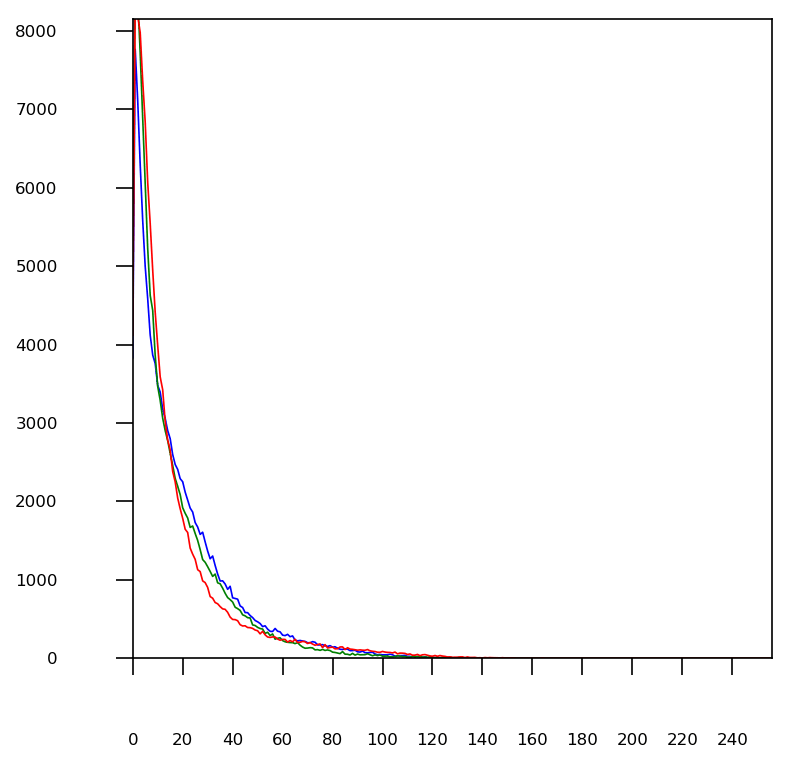}
        \caption{BGR difference histogram.}
    \end{subfigure}
\end{figure}
\begin{figure}[H]
    \ContinuedFloat
    \centering
    \begin{subfigure}[!h]{0.3\linewidth}
        \includegraphics[width=\linewidth]{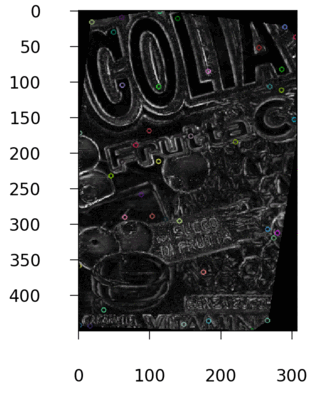}
        \caption{Pixelwise difference norm.}
    \end{subfigure}
    \begin{subfigure}[!h]{0.4\linewidth}
        \includegraphics[width=\linewidth]{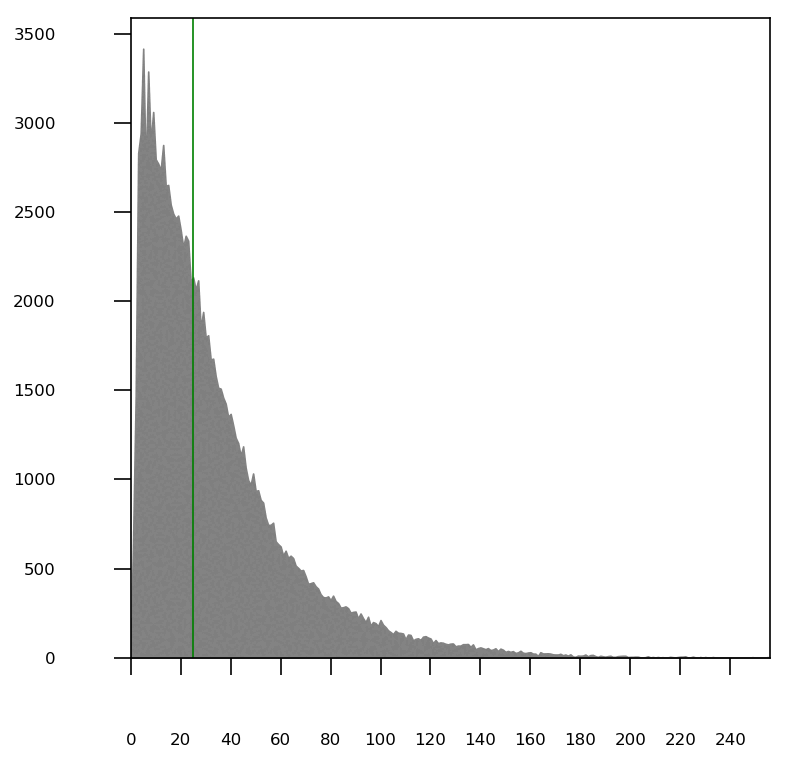}
        \caption{Pixelwise difference norm histogram.}
    \end{subfigure}
    \caption{3\textsuperscript{rd} object occurrence (union of seeds \#5-7).}
    \label{fig:figure_38}
\end{figure}
\begin{figure}[H]
    \centering
    \begin{subfigure}[!h]{0.7\linewidth}
        \includegraphics[width=\linewidth]{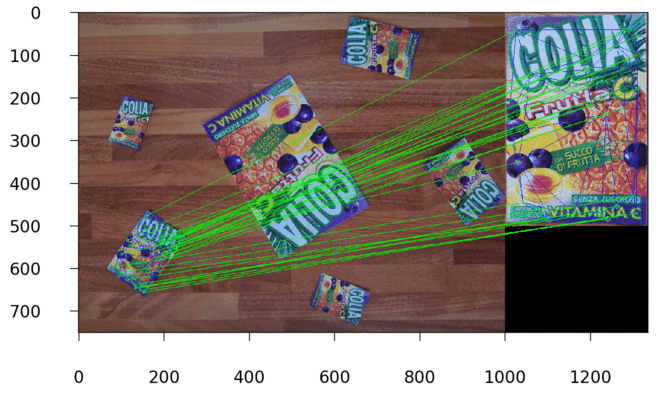}
        \caption{Expanded seed.}
    \end{subfigure}
    \begin{subfigure}[!h]{0.35\linewidth}
        \includegraphics[width=\linewidth]{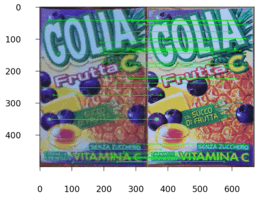}
        \caption{Rectified object occurrence.}
    \end{subfigure}
    \begin{subfigure}[!h]{0.35\linewidth}
        \includegraphics[width=\linewidth]{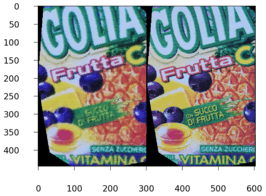}
        \caption{Template and histogram matched object.}
    \end{subfigure}
    \begin{subfigure}[!h]{0.3\linewidth}
        \includegraphics[width=\linewidth]{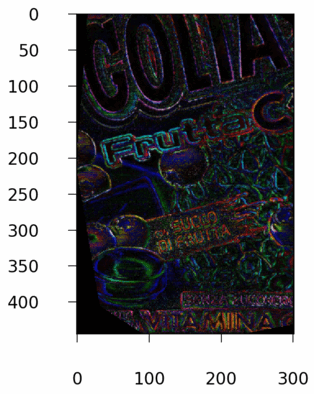}
        \caption{BGR absolute difference.}
    \end{subfigure}
    \begin{subfigure}[!h]{0.4\linewidth}
        \includegraphics[width=\linewidth]{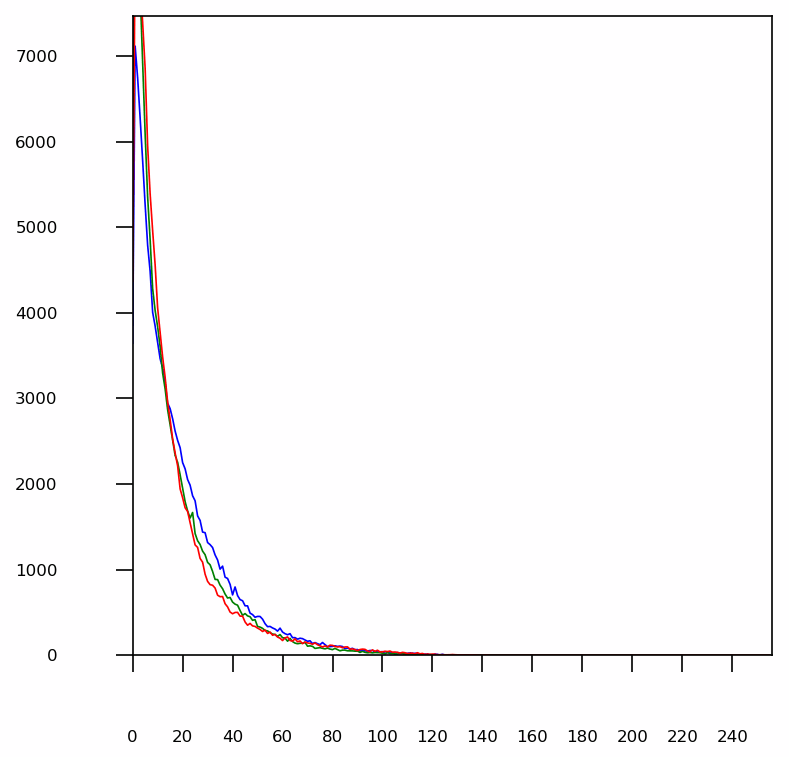}
        \caption{BGR difference histogram.}
    \end{subfigure}
\end{figure}
\begin{figure}[H]
    \ContinuedFloat
    \centering
    \begin{subfigure}[!h]{0.3\linewidth}
        \includegraphics[width=\linewidth]{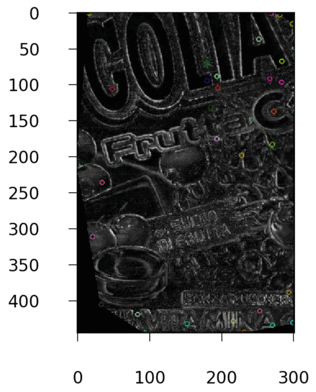}
        \caption{Pixelwise difference norm.}
    \end{subfigure}
    \begin{subfigure}[!h]{0.4\linewidth}
        \includegraphics[width=\linewidth]{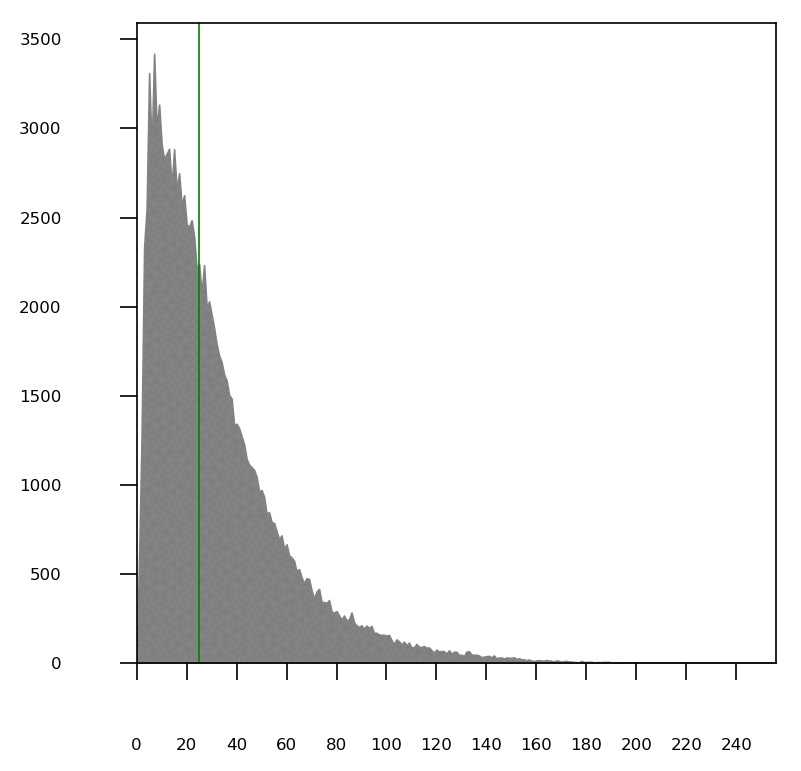}
        \caption{Pixelwise difference norm histogram.}
    \end{subfigure}
    \caption{4\textsuperscript{nd} object occurrence (seed \#3).}
    \label{fig:figure_39}
\end{figure}
\begin{figure}[H]
    \centering
    \begin{subfigure}[!h]{0.7\linewidth}
        \includegraphics[width=\linewidth]{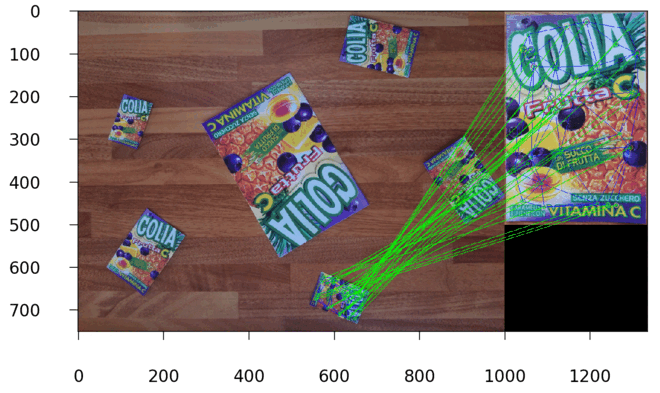}
        \caption{Expanded seed.}
    \end{subfigure}
    \begin{subfigure}[!h]{0.35\linewidth}
        \includegraphics[width=\linewidth]{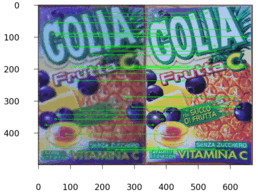}
        \caption{Rectified object occurrence.}
    \end{subfigure}
    \begin{subfigure}[!h]{0.35\linewidth}
        \includegraphics[width=\linewidth]{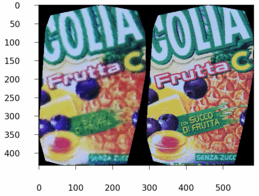}
        \caption{Template and histogram matched object.}
    \end{subfigure}
    \begin{subfigure}[!h]{0.3\linewidth}
        \includegraphics[width=\linewidth]{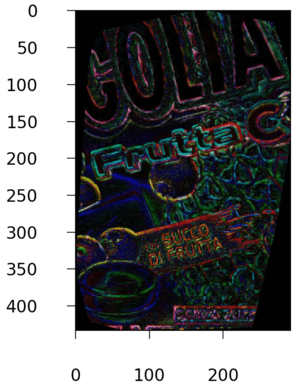}
        \caption{BGR absolute difference.}
    \end{subfigure}
    \begin{subfigure}[!h]{0.4\linewidth}
        \includegraphics[width=\linewidth]{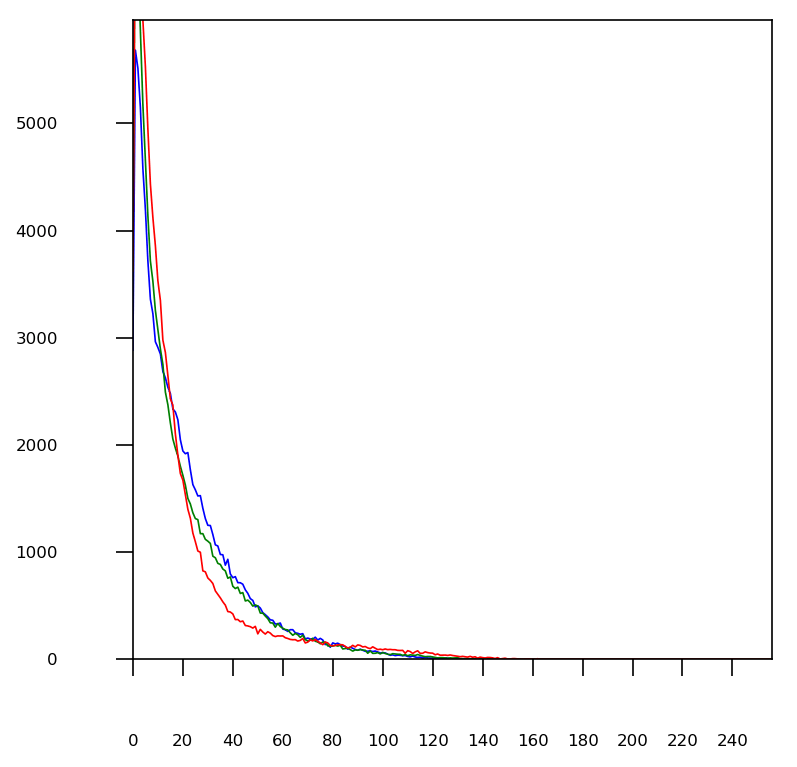}
        \caption{BGR difference histogram.}
    \end{subfigure}
\end{figure}
\begin{figure}[H]
    \ContinuedFloat
    \centering
    \begin{subfigure}[!h]{0.3\linewidth}
        \includegraphics[width=\linewidth]{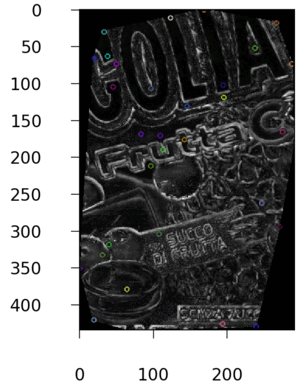}
        \caption{Pixelwise difference norm.}
    \end{subfigure}
    \begin{subfigure}[!h]{0.4\linewidth}
        \includegraphics[width=\linewidth]{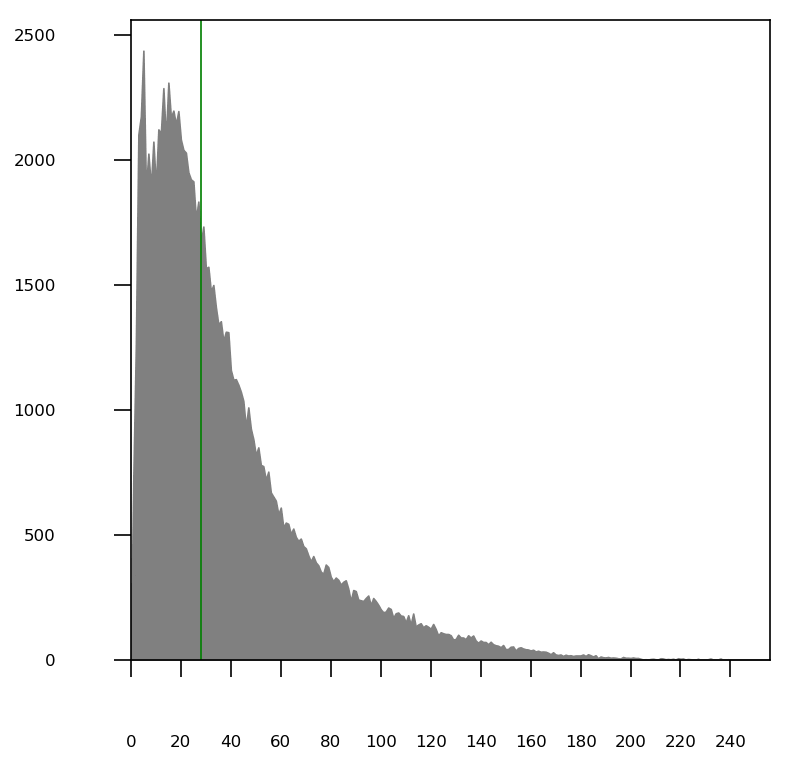}
        \caption{Pixelwise difference norm histogram.}
    \end{subfigure}
    \caption{5\textsuperscript{th} object occurrence (union of seeds \#15-16-17-18-19).}
    \label{fig:figure_40}
\end{figure}
\begin{figure}[H]
    \centering
    \begin{subfigure}[!h]{0.7\linewidth}
        \includegraphics[width=\linewidth]{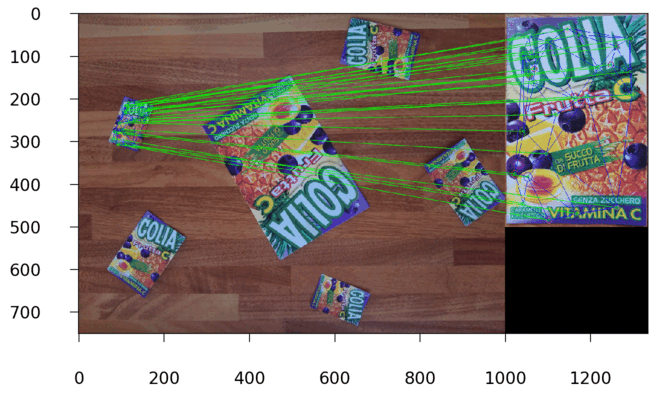}
        \caption{Expanded seed.}
    \end{subfigure}
    \begin{subfigure}[!h]{0.35\linewidth}
        \includegraphics[width=\linewidth]{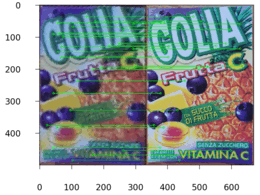}
        \caption{Rectified object occurrence.}
    \end{subfigure}
    \begin{subfigure}[!h]{0.35\linewidth}
        \includegraphics[width=\linewidth]{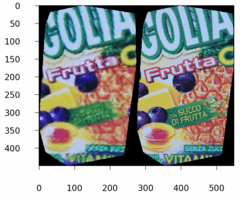}
        \caption{Template and histogram matched object.}
    \end{subfigure}
    \begin{subfigure}[!h]{0.3\linewidth}
        \includegraphics[width=\linewidth]{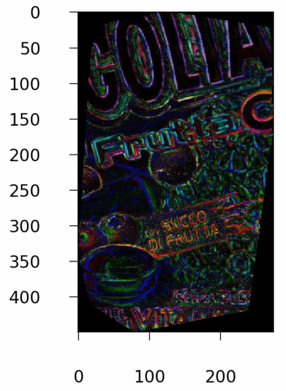}
        \caption{BGR absolute difference.}
    \end{subfigure}
    \begin{subfigure}[!h]{0.4\linewidth}
        \includegraphics[width=\linewidth]{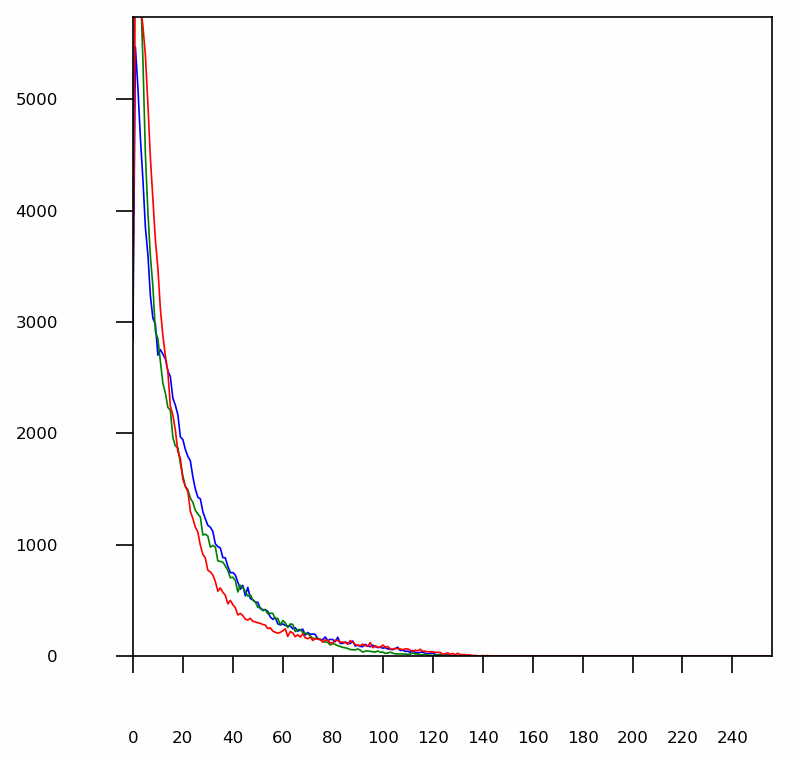}
        \caption{BGR difference histogram.}
    \end{subfigure}
\end{figure}
\begin{figure}[H]
    \ContinuedFloat
    \centering
    \begin{subfigure}[!h]{0.3\linewidth}
        \includegraphics[width=\linewidth]{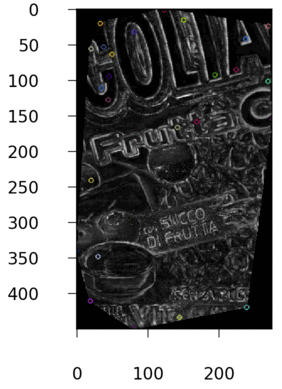}
        \caption{Pixelwise difference norm.}
    \end{subfigure}
    \begin{subfigure}[!h]{0.4\linewidth}
        \includegraphics[width=\linewidth]{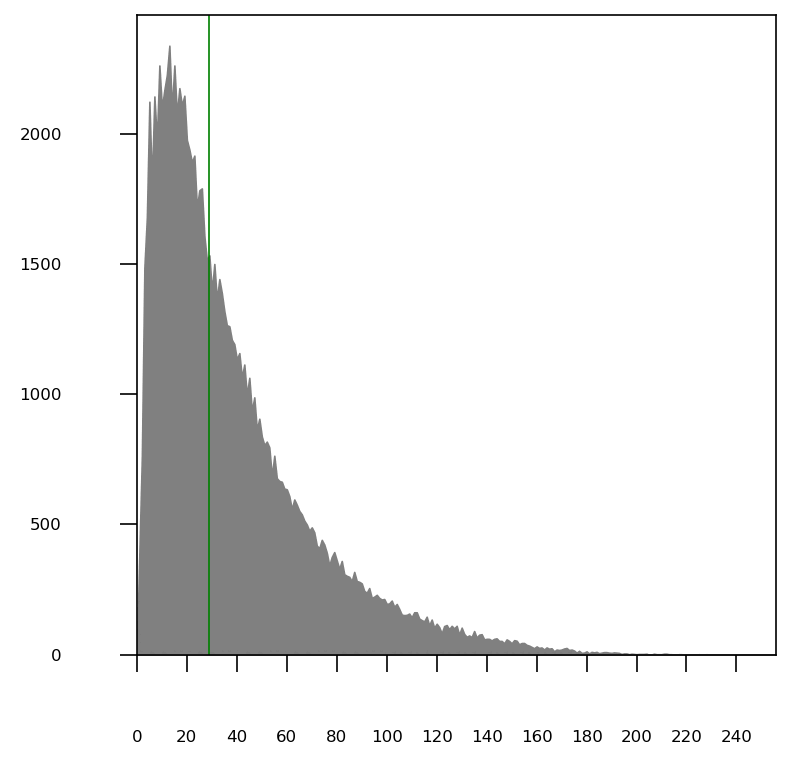}
        \caption{Pixelwise difference norm histogram.}
    \end{subfigure}
    \caption{6\textsuperscript{th} object occurrence (union of seeds \#10-11-12-13).}
    \label{fig:figure_41}
\end{figure}
\section{One distorted object occurrence}
        \subsection{Case 1}
    \label{appendix_A.2.1}
    \paragraph{Feature-based object detection algorithm driven by RANSAC}
        \begin{figure}[H]
            \centering
            \begin{subfigure}[!h]{0.4\linewidth}
                \includegraphics[width=\linewidth]{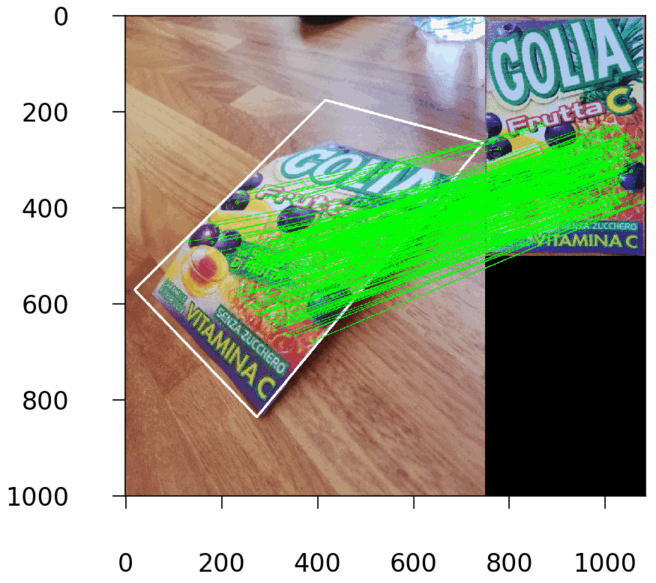}
                \caption{Clustered matches.}
                \label{fig:figure_42a}
            \end{subfigure}
            \begin{subfigure}[!h]{0.35\linewidth}
                \includegraphics[width=\linewidth]{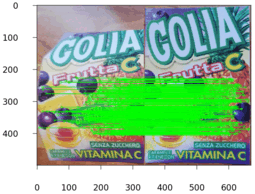}
                \caption{Rectified object occurrence.}
                \label{fig:figure_42b}
            \end{subfigure}
            \begin{subfigure}[!h]{0.35\linewidth}
                \includegraphics[width=\linewidth]{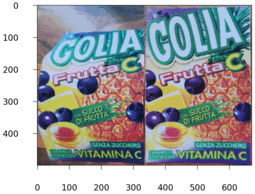}
                \caption{Template and histogram matched object.}
                \label{fig:figure_42c}
            \end{subfigure}
            \begin{subfigure}[!h]{0.3\linewidth}
                \includegraphics[width=\linewidth]{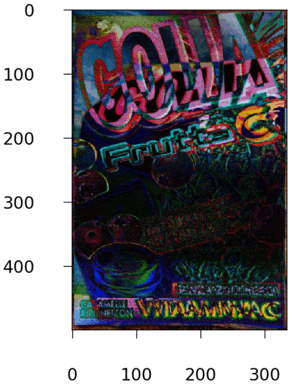}
                \caption{BGR absolute difference.}
                \label{fig:figure_42d}
            \end{subfigure}
            \begin{subfigure}[!h]{0.4\linewidth}
                \includegraphics[width=\linewidth]{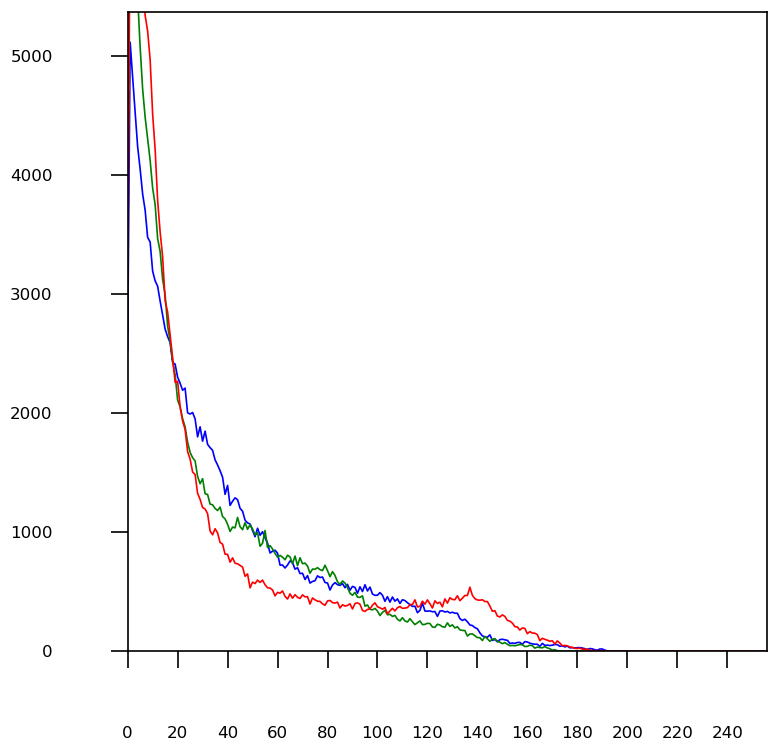}
                \caption{BGR difference histogram.}
                \label{fig:figure_42h}
            \end{subfigure}
            \begin{subfigure}[!h]{0.3\linewidth}
                \includegraphics[width=\linewidth]{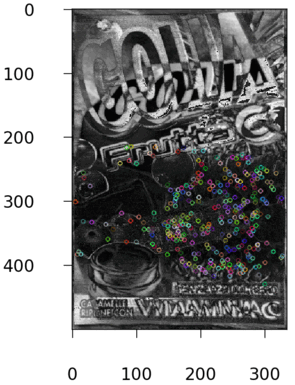}
                \caption{Pixelwise difference norm.}
                \label{fig:figure_42i}
            \end{subfigure}
        \end{figure}
        \begin{figure}[H]
            \ContinuedFloat
            \centering
            \begin{subfigure}[!h]{0.4\linewidth}
                \includegraphics[width=\linewidth]{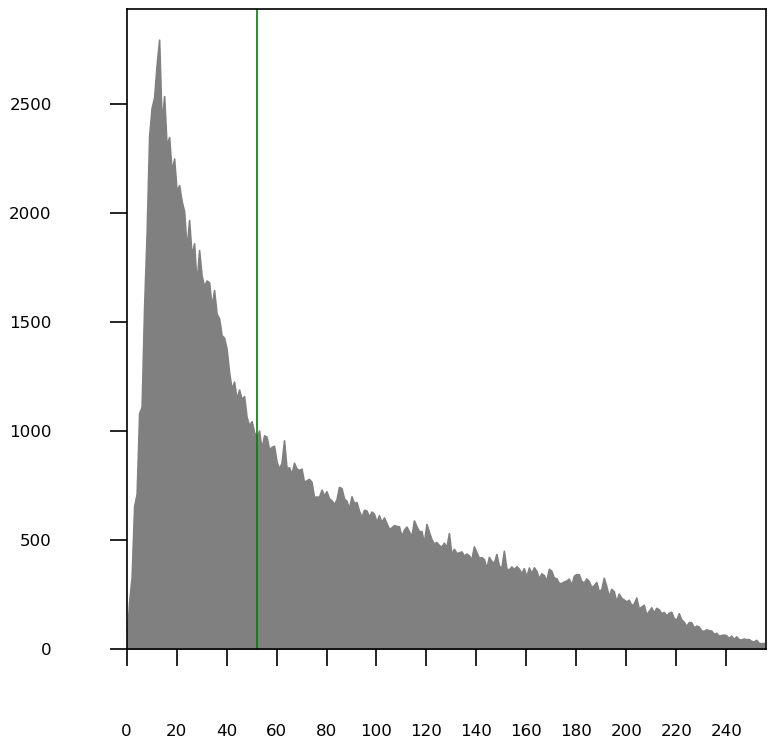}
                \caption{Pixelwise difference norm histogram.}
                \label{fig:figure_42j}
            \end{subfigure}
            \caption{Object occurrence.}
            \label{fig:figure_42}
        \end{figure}
    \paragraph{Our algorithm}
        \begin{figure}[H]
            \centering
            \begin{subfigure}[!h]{0.4\linewidth}
                \includegraphics[width=\linewidth]{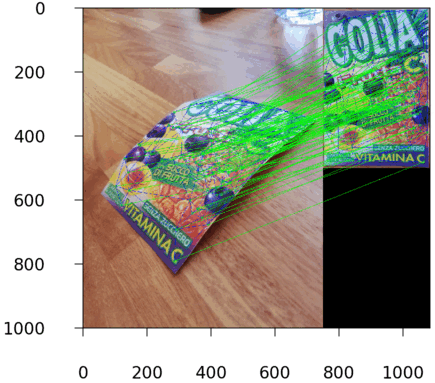}
                \caption{Expanded seed.}
                \label{fig:figure_48a}
            \end{subfigure}
            \begin{subfigure}[!h]{0.35\linewidth}
                \includegraphics[width=\linewidth]{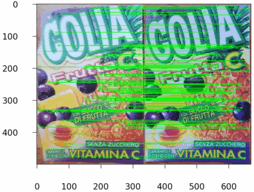}
                \caption{Rectified object occurrence.}
                \label{fig:figure_48b}
            \end{subfigure}
            \begin{subfigure}[!h]{0.35\linewidth}
                \includegraphics[width=\linewidth]{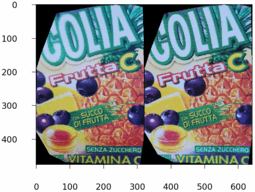}
                \caption{Template and histogram matched object.}
                \label{fig:figure_48c}
            \end{subfigure}
            \begin{subfigure}[!h]{0.3\linewidth}
                \includegraphics[width=\linewidth]{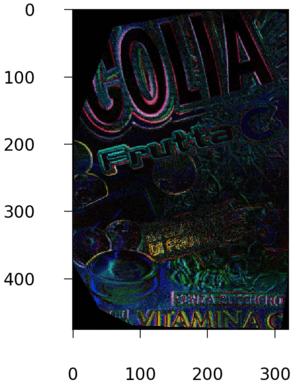}
                \caption{BGR absolute difference.}
                \label{fig:figure_48d}
            \end{subfigure}
            \begin{subfigure}[!h]{0.4\linewidth}
                \includegraphics[width=\linewidth]{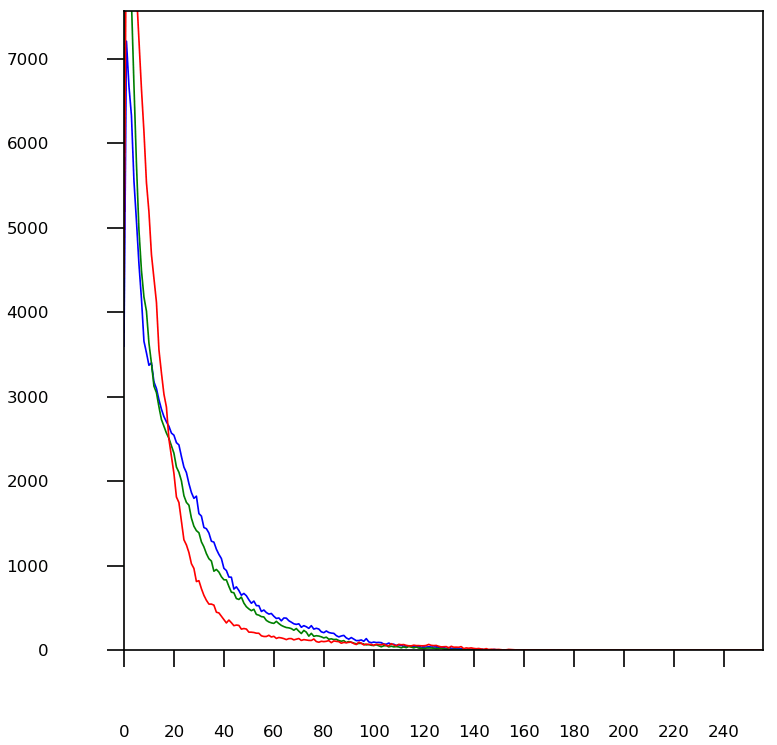}
                \caption{BGR difference histogram.}
                \label{fig:figure_48h}
            \end{subfigure}
            \begin{subfigure}[!h]{0.3\linewidth}
                \includegraphics[width=\linewidth]{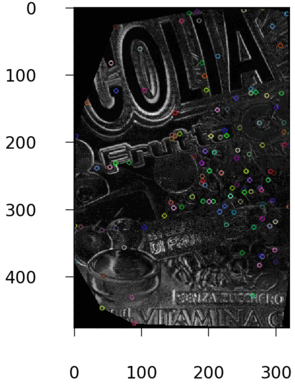}
                \caption{Pixelwise difference norm.}
                \label{fig:figure_48i}
            \end{subfigure}
        \end{figure}
        \begin{figure}[H]
            \ContinuedFloat
            \centering
            \begin{subfigure}[!h]{0.4\linewidth}
                \includegraphics[width=\linewidth]{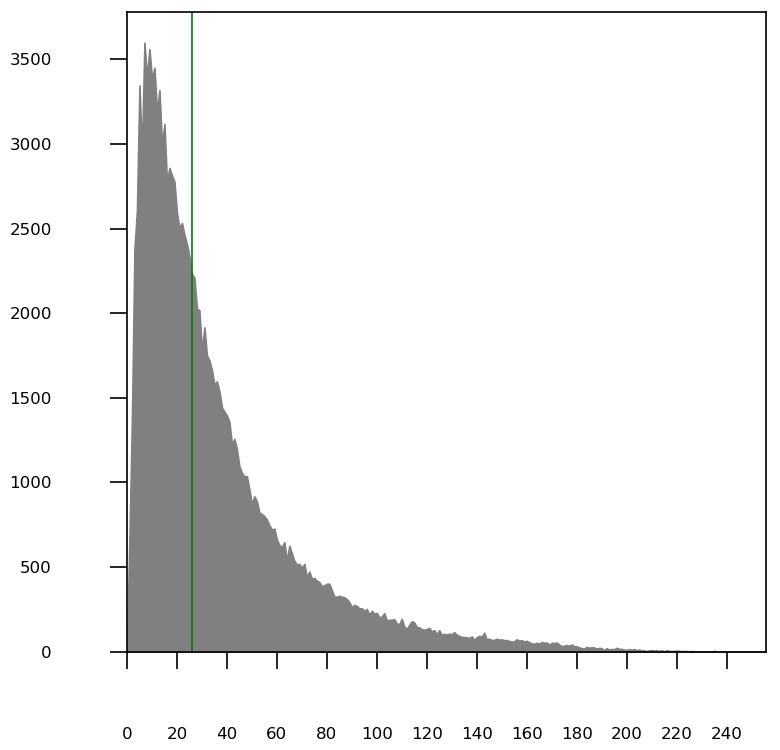}
                \caption{Pixelwise difference norm histogram.}
                \label{fig:figure_48j}
            \end{subfigure}
            \caption{Object occurrence (union of seeds \#16-18-19-25-27-29).}
            \label{fig:figure_48}
        \end{figure}
\subsection{Case 2}
    \label{appendix_A.2.2}
    \paragraph{Feature-based object detection algorithm driven by RANSAC}
        \begin{figure}[H]
            \centering
            \begin{subfigure}[!h]{0.4\linewidth}
                \includegraphics[width=\linewidth]{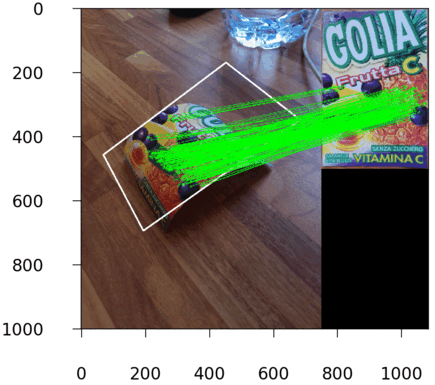}
                \caption{Clustered matches.}
                \label{fig:figure_43a}
            \end{subfigure}
            \begin{subfigure}[!h]{0.35\linewidth}
                \includegraphics[width=\linewidth]{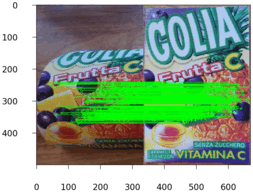}
                \caption{Rectified object occurrence.}
                \label{fig:figure_43b}
            \end{subfigure}
            \begin{subfigure}[!h]{0.35\linewidth}
                \includegraphics[width=\linewidth]{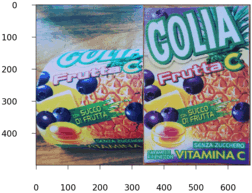}
                \caption{Template and histogram matched object.}
                \label{fig:figure_43c}
            \end{subfigure}
            \begin{subfigure}[!h]{0.3\linewidth}
                \includegraphics[width=\linewidth]{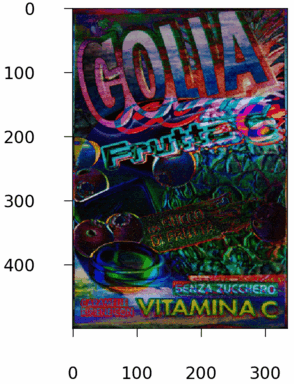}
                \caption{BGR absolute difference.}
                \label{fig:figure_43d}
            \end{subfigure}
            \begin{subfigure}[!h]{0.4\linewidth}
                \includegraphics[width=\linewidth]{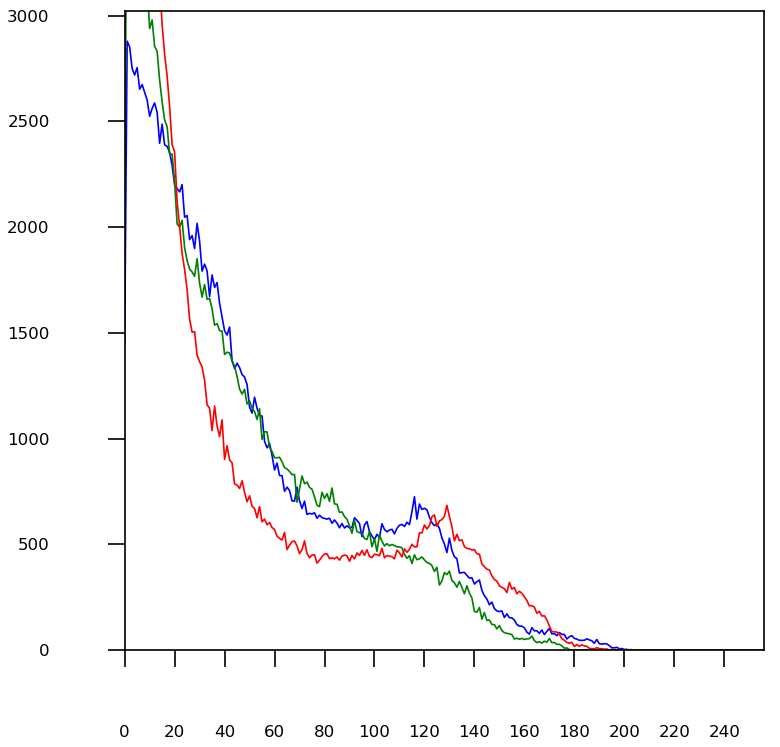}
                \caption{BGR difference histogram.}
                \label{fig:figure_43h}
            \end{subfigure}
            \begin{subfigure}[!h]{0.3\linewidth}
                \includegraphics[width=\linewidth]{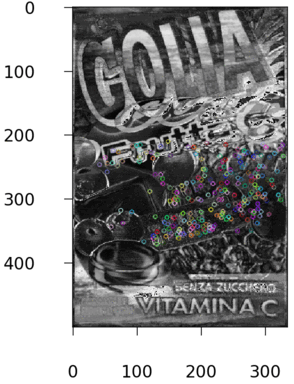}
                \caption{Pixelwise difference norm.}
                \label{fig:figure_43i}
            \end{subfigure}
        \end{figure}
        \begin{figure}[H]
            \ContinuedFloat
            \centering
            \begin{subfigure}[!h]{0.4\linewidth}
                \includegraphics[width=\linewidth]{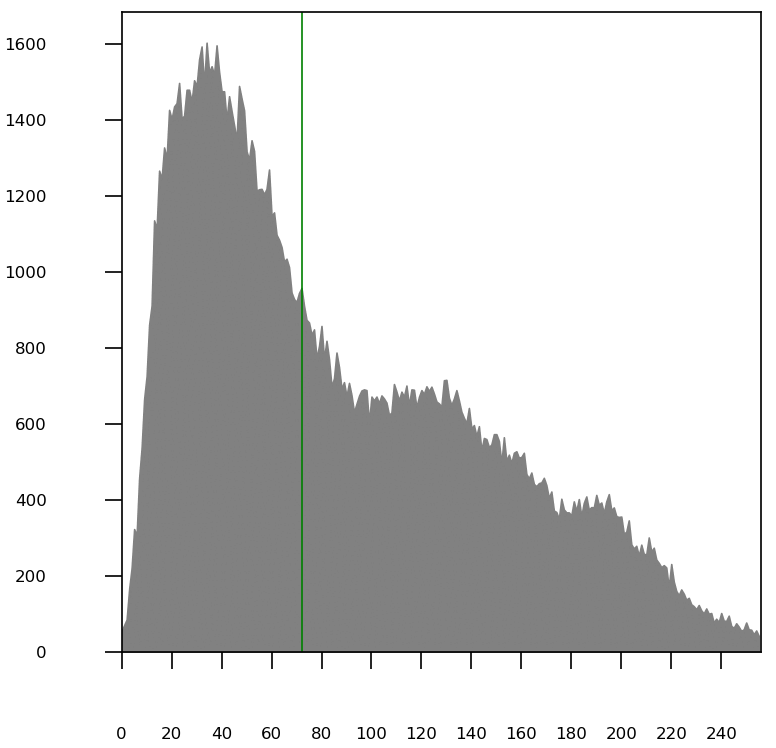}
                \caption{Pixelwise difference norm histogram.}
                \label{fig:figure_43j}
            \end{subfigure}
            \caption{Object occurrence.}
            \label{fig:figure_43}
        \end{figure}
    \paragraph{Our algorithm}
        \begin{figure}[H]
            \centering
            \begin{subfigure}[!h]{0.4\linewidth}
                \includegraphics[width=\linewidth]{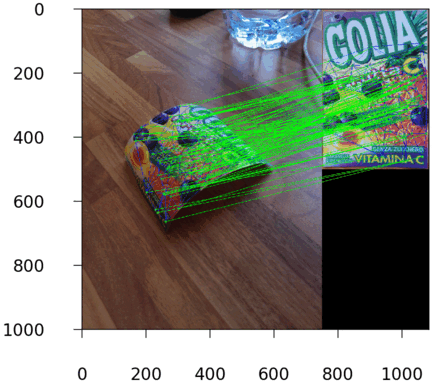}
                \caption{Expanded seed.}
                \label{fig:figure_49a}
            \end{subfigure}
            \begin{subfigure}[!h]{0.35\linewidth}
                \includegraphics[width=\linewidth]{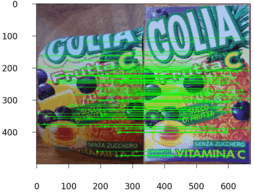}
                \caption{Rectified object occurrence.}
                \label{fig:figure_49b}
            \end{subfigure}
            \begin{subfigure}[!h]{0.35\linewidth}
                \includegraphics[width=\linewidth]{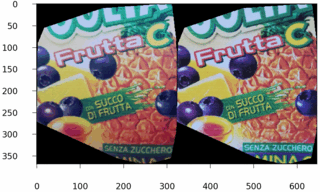}
                \caption{Template and histogram matched object.}
                \label{fig:figure_49c}
            \end{subfigure}
            \begin{subfigure}[!h]{0.3\linewidth}
                \includegraphics[width=\linewidth]{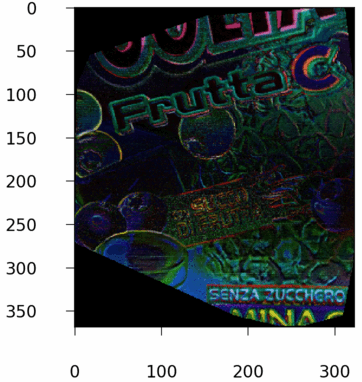}
                \caption{BGR absolute difference.}
                \label{fig:figure_49d}
            \end{subfigure}
            \begin{subfigure}[!h]{0.4\linewidth}
                \includegraphics[width=\linewidth]{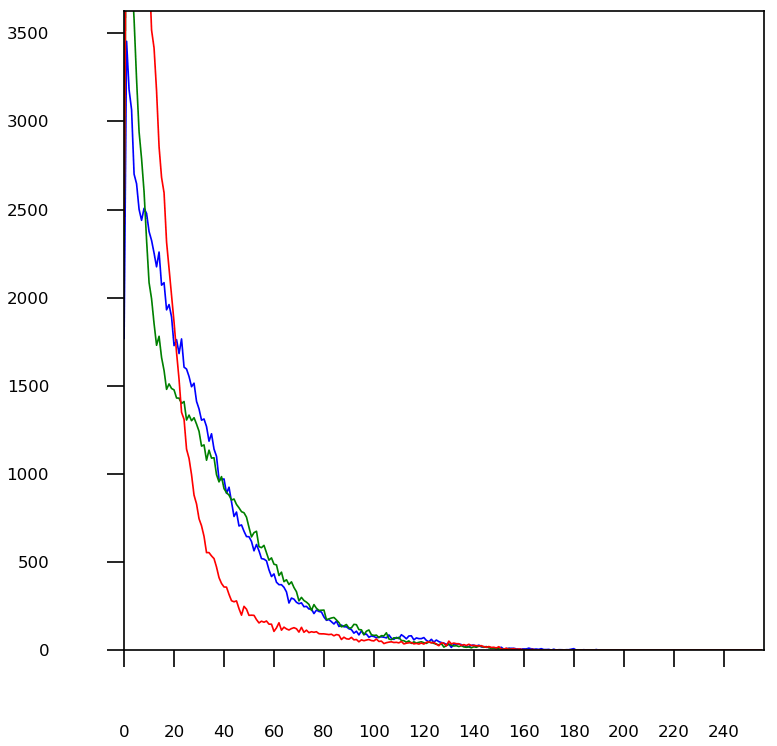}
                \caption{BGR difference histogram.}
                \label{fig:figure_49h}
            \end{subfigure}
            \begin{subfigure}[!h]{0.3\linewidth}
                \includegraphics[width=\linewidth]{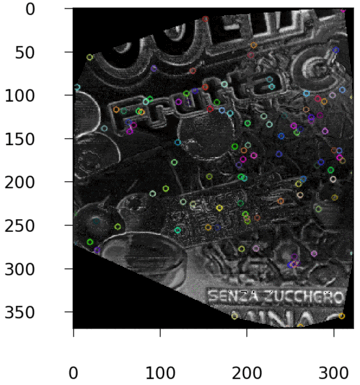}
                \caption{Pixelwise difference norm.}
                \label{fig:figure_49i}
            \end{subfigure}
        \end{figure}
        \begin{figure}[H]
            \ContinuedFloat
            \centering
            \begin{subfigure}[!h]{0.4\linewidth}
                \includegraphics[width=\linewidth]{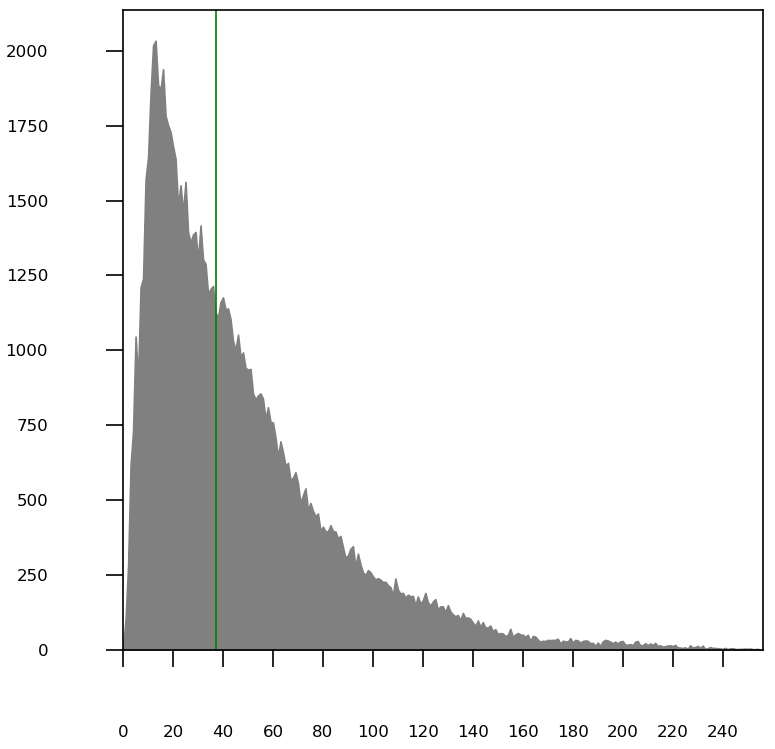}
                \caption{Pixelwise difference norm histogram.}
                \label{fig:figure_49j}
            \end{subfigure}
            \caption{Object occurrence (union of seeds \#4-5-7-11-16-25-26).}
            \label{fig:figure_49}
        \end{figure}
\subsection{Case 3}
    \label{appendix_A.2.3}
    \paragraph{Feature-based object detection algorithm driven by RANSAC}
        \begin{figure}[H]
            \centering
            \begin{subfigure}[!h]{0.4\linewidth}
                \includegraphics[width=\linewidth]{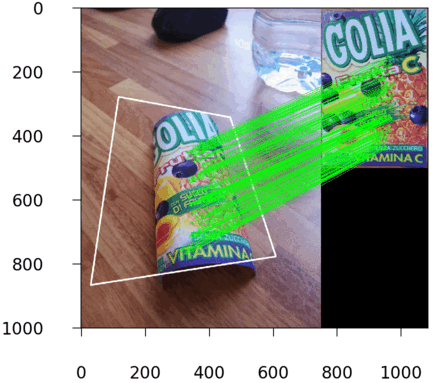}
                \caption{Clustered matches.}
                \label{fig:figure_44a}
            \end{subfigure}
            \begin{subfigure}[!h]{0.35\linewidth}
                \includegraphics[width=\linewidth]{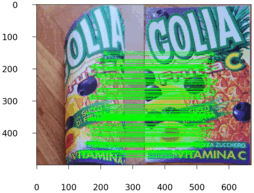}
                \caption{Rectified object occurrence.}
                \label{fig:figure_44b}
            \end{subfigure}
            \begin{subfigure}[!h]{0.35\linewidth}
                \includegraphics[width=\linewidth]{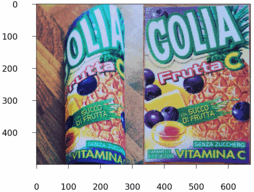}
                \caption{Template and histogram matched object.}
                \label{fig:figure_44c}
            \end{subfigure}
            \begin{subfigure}[!h]{0.3\linewidth}
                \includegraphics[width=\linewidth]{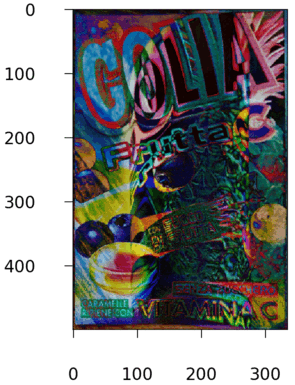}
                \caption{BGR absolute difference.}
                \label{fig:figure_44d}
            \end{subfigure}
            \begin{subfigure}[!h]{0.4\linewidth}
                \includegraphics[width=\linewidth]{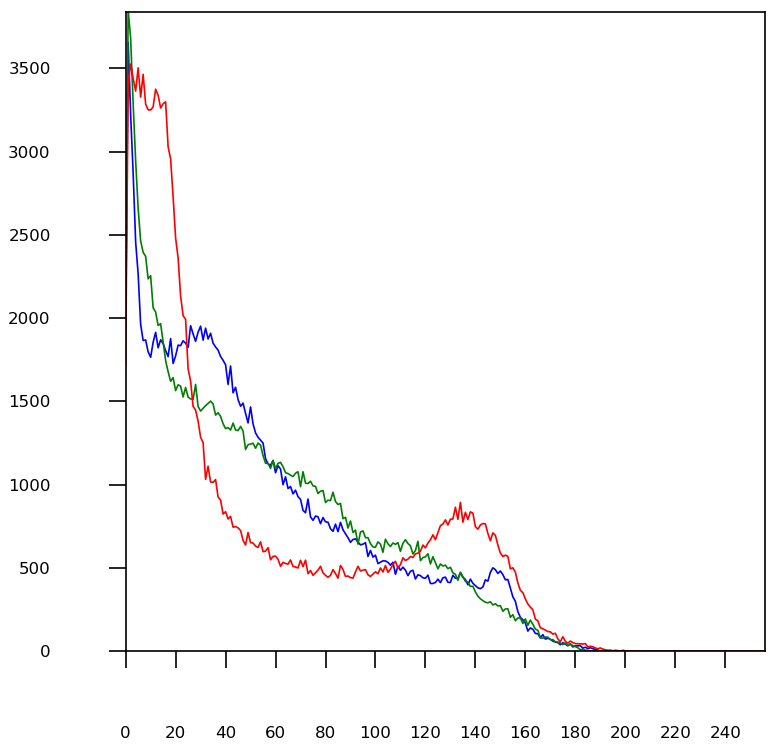}
                \caption{BGR difference histogram.}
                \label{fig:figure_44h}
            \end{subfigure}
            \begin{subfigure}[!h]{0.3\linewidth}
                \includegraphics[width=\linewidth]{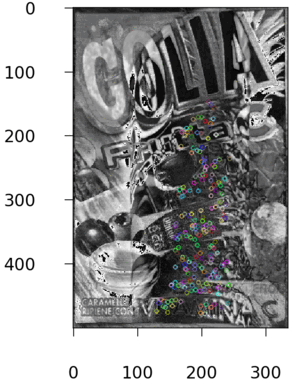}
                \caption{Pixelwise difference norm.}
                \label{fig:figure_44i}
            \end{subfigure}
        \end{figure}
        \begin{figure}[H]
            \ContinuedFloat
            \centering
            \begin{subfigure}[!h]{0.4\linewidth}
                \includegraphics[width=\linewidth]{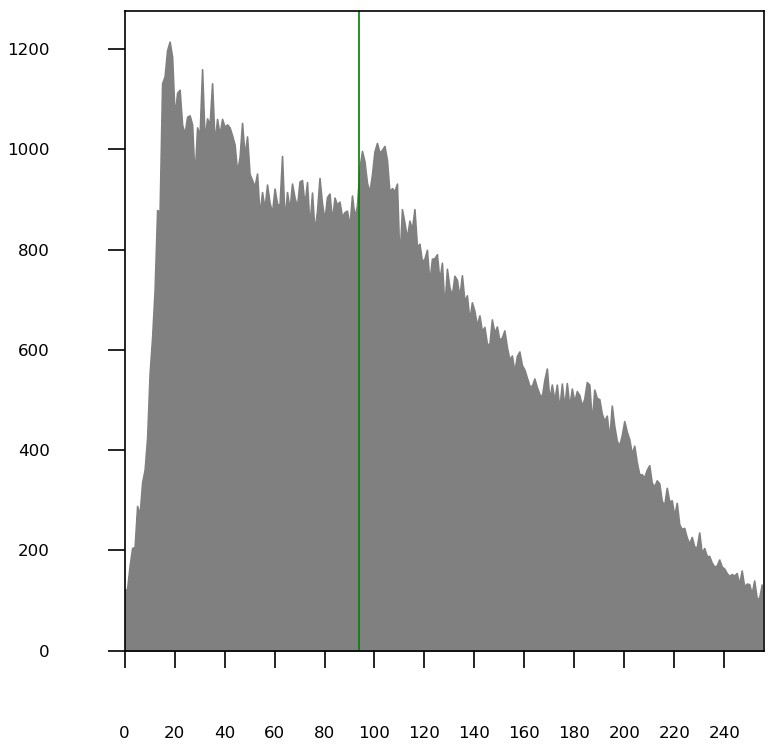}
                \caption{Pixelwise difference norm histogram.}
                \label{fig:figure_44j}
            \end{subfigure}
            \caption{Object occurrence.}
            \label{fig:figure_44}
        \end{figure}
    \paragraph{Our algorithm}
        \begin{figure}[H]
            \centering
            \begin{subfigure}[!h]{0.4\linewidth}
                \includegraphics[width=\linewidth]{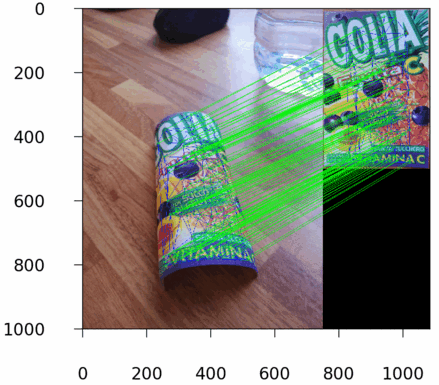}
                \caption{Expanded seed.}
                \label{fig:figure_50a}
            \end{subfigure}
            \begin{subfigure}[!h]{0.35\linewidth}
                \includegraphics[width=\linewidth]{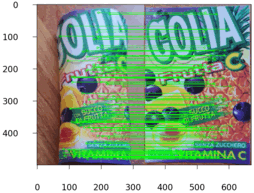}
                \caption{Rectified object occurrence.}
                \label{fig:figure_50b}
            \end{subfigure}
            \begin{subfigure}[!h]{0.35\linewidth}
                \includegraphics[width=\linewidth]{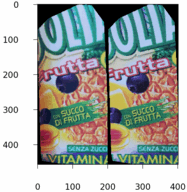}
                \caption{Template and histogram matched object.}
                \label{fig:figure_50c}
            \end{subfigure}
            \begin{subfigure}[!h]{0.3\linewidth}
                \includegraphics[width=\linewidth]{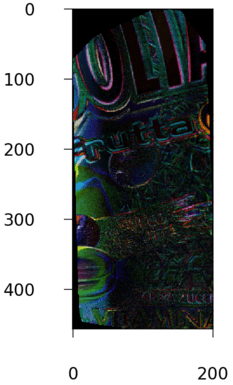}
                \caption{BGR absolute difference.}
                \label{fig:figure_50d}
            \end{subfigure}
            \begin{subfigure}[!h]{0.4\linewidth}
                \includegraphics[width=\linewidth]{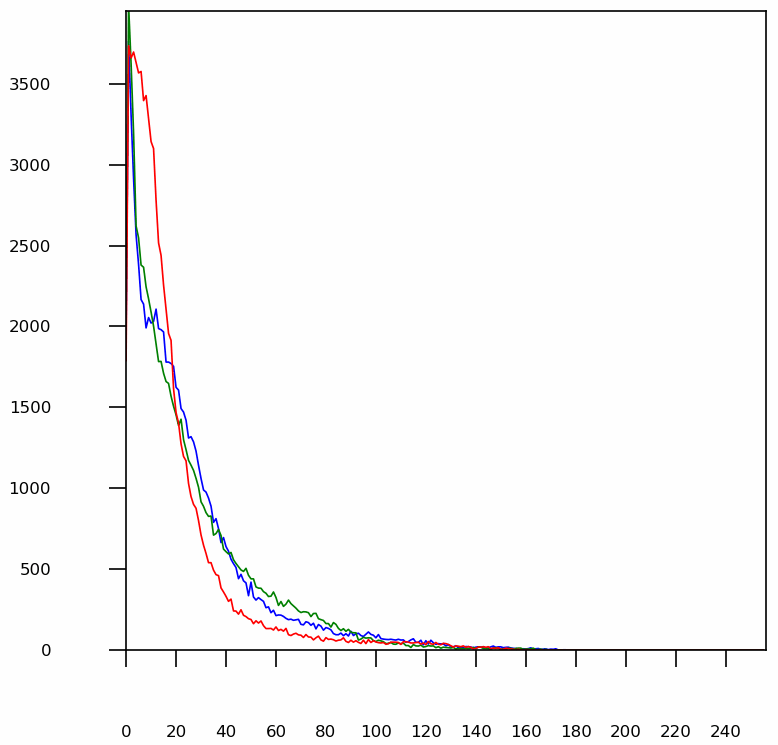}
                \caption{BGR difference histogram.}
                \label{fig:figure_50h}
            \end{subfigure}
            \begin{subfigure}[!h]{0.3\linewidth}
                \includegraphics[width=\linewidth]{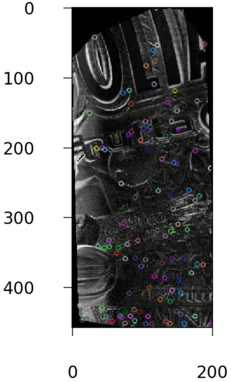}
                \caption{Pixelwise difference norm.}
                \label{fig:figure_50i}
            \end{subfigure}
        \end{figure}
        \begin{figure}[H]
            \ContinuedFloat
            \centering
            \begin{subfigure}[!h]{0.4\linewidth}
                \includegraphics[width=\linewidth]{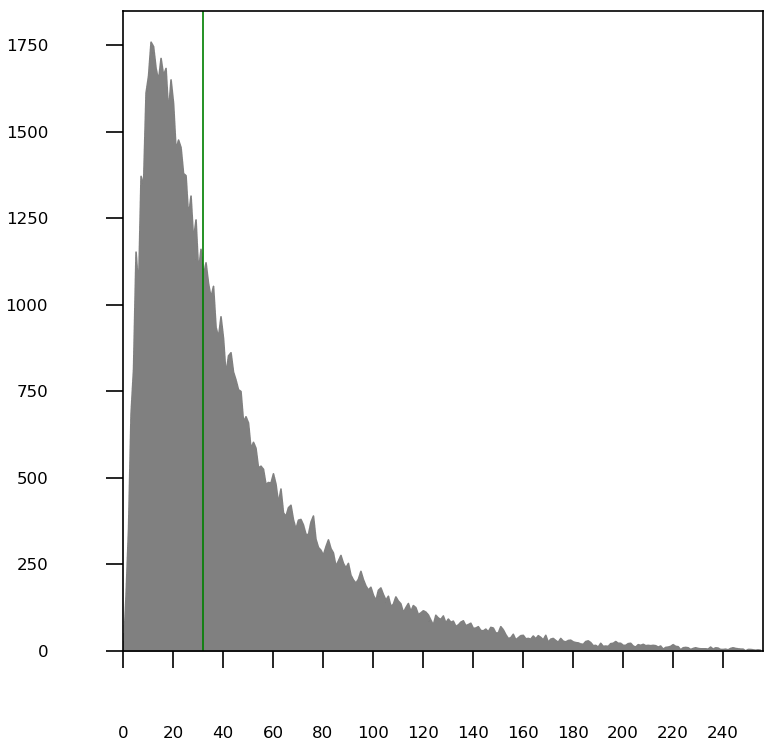}
                \caption{Pixelwise difference norm histogram.}
                \label{fig:figure_50j}
            \end{subfigure}
            \caption{Object occurrence (union of seeds \#0-1-4-10-15-26).}
            \label{fig:figure_50}
        \end{figure}
\subsection{Case 4}
    \label{appendix_A.2.4}
    \paragraph{Feature-based object detection algorithm driven by RANSAC}
        \begin{figure}[H]
            \centering
            \begin{subfigure}[!h]{0.4\linewidth}
                \includegraphics[width=\linewidth]{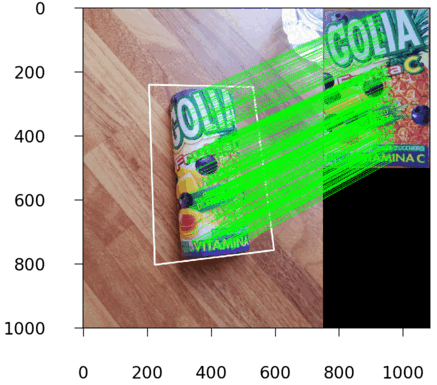}
                \caption{Clustered matches.}
                \label{fig:figure_45a}
            \end{subfigure}
            \begin{subfigure}[!h]{0.35\linewidth}
                \includegraphics[width=\linewidth]{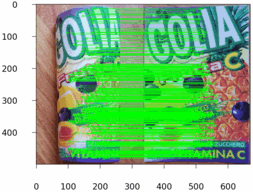}
                \caption{Rectified object occurrence.}
                \label{fig:figure_45b}
            \end{subfigure}
            \begin{subfigure}[!h]{0.35\linewidth}
                \includegraphics[width=\linewidth]{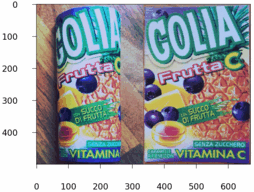}
                \caption{Template and histogram matched object.}
                \label{fig:figure_45c}
            \end{subfigure}
            \begin{subfigure}[!h]{0.3\linewidth}
                \includegraphics[width=\linewidth]{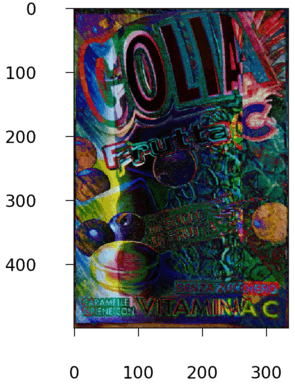}
                \caption{BGR absolute difference.}
                \label{fig:figure_45d}
            \end{subfigure}
            \begin{subfigure}[!h]{0.4\linewidth}
                \includegraphics[width=\linewidth]{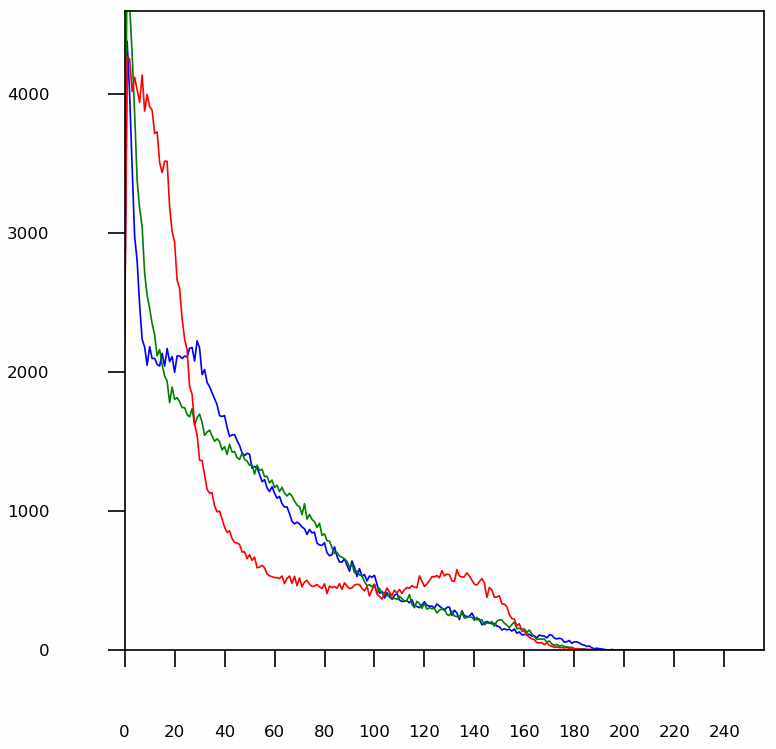}
                \caption{BGR difference histogram.}
                \label{fig:figure_45h}
            \end{subfigure}
            \begin{subfigure}[!h]{0.3\linewidth}
                \includegraphics[width=\linewidth]{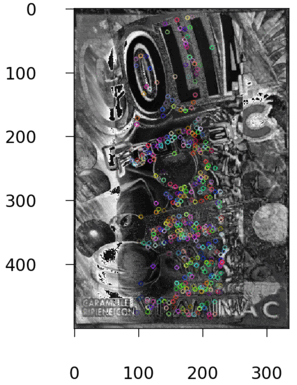}
                \caption{Pixelwise difference norm.}
                \label{fig:figure_45i}
            \end{subfigure}
        \end{figure}
        \begin{figure}[H]
            \ContinuedFloat
            \centering
            \begin{subfigure}[!h]{0.4\linewidth}
                \includegraphics[width=\linewidth]{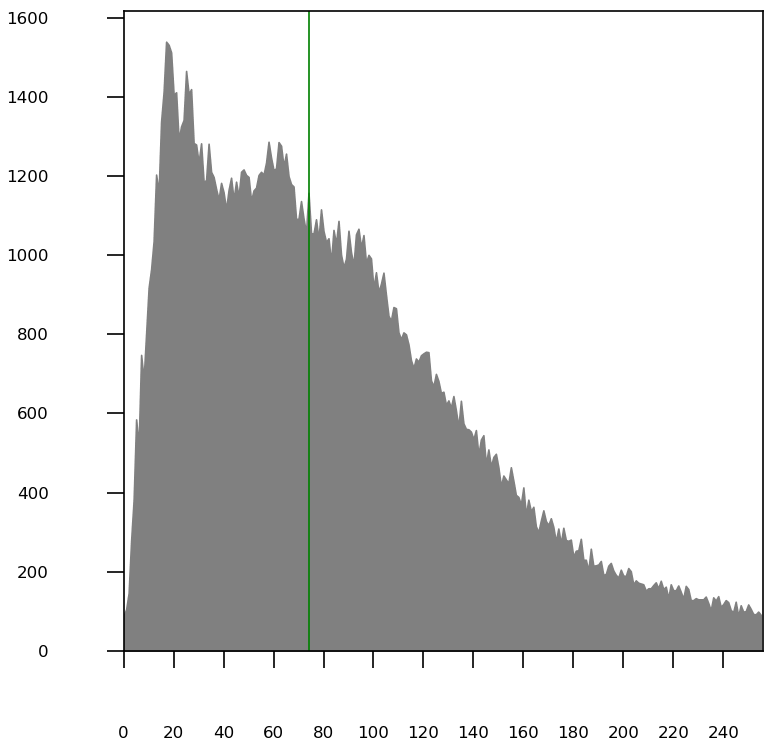}
                \caption{Pixelwise difference norm histogram.}
                \label{fig:figure_45j}
            \end{subfigure}
            \caption{Object occurrence.}
            \label{fig:figure_45}
        \end{figure}
    \paragraph{Our algorithm}
        \begin{figure}[H]
            \centering
            \begin{subfigure}[!h]{0.4\linewidth}
                \includegraphics[width=\linewidth]{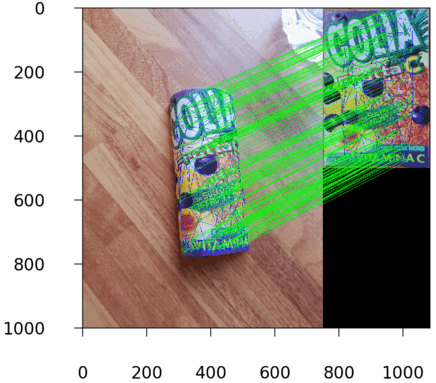}
                \caption{Expanded seed.}
                \label{fig:figure_51a}
            \end{subfigure}
            \begin{subfigure}[!h]{0.35\linewidth}
                \includegraphics[width=\linewidth]{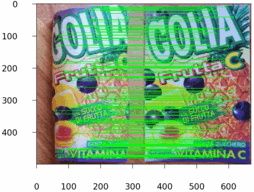}
                \caption{Rectified object occurrence.}
                \label{fig:figure_51b}
            \end{subfigure}
            \begin{subfigure}[!h]{0.35\linewidth}
                \includegraphics[width=\linewidth]{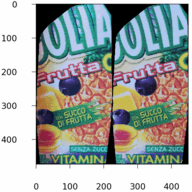}
                \caption{Template and histogram matched object.}
                \label{fig:figure_51c}
            \end{subfigure}
            \begin{subfigure}[!h]{0.3\linewidth}
                \includegraphics[width=\linewidth]{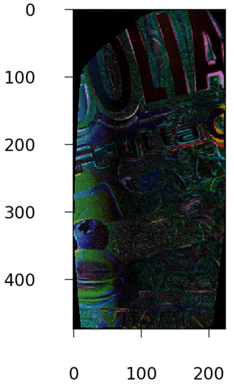}
                \caption{BGR absolute difference.}
                \label{fig:figure_51d}
            \end{subfigure}
            \begin{subfigure}[!h]{0.4\linewidth}
                \includegraphics[width=\linewidth]{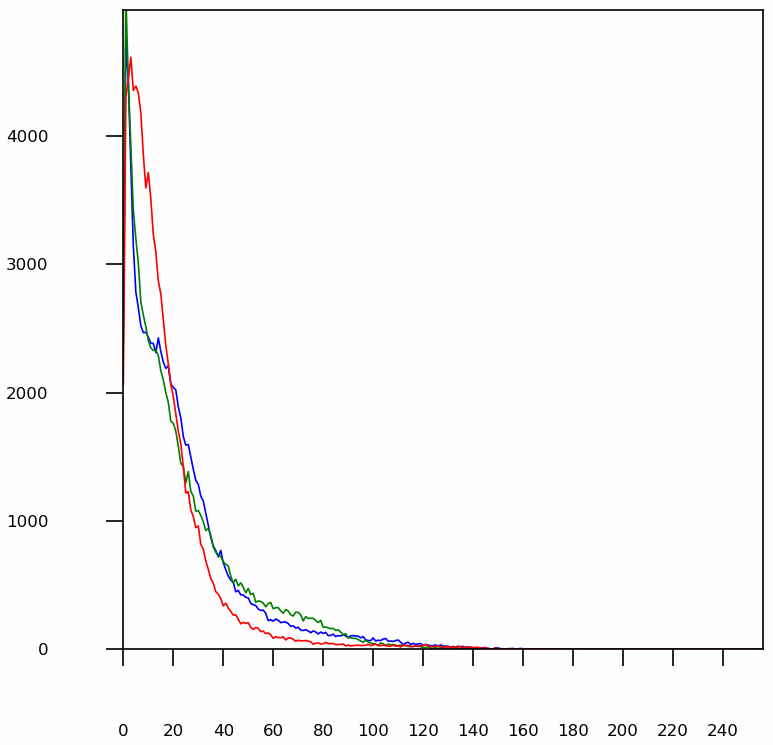}
                \caption{BGR difference histogram.}
                \label{fig:figure_51h}
            \end{subfigure}
            \begin{subfigure}[!h]{0.3\linewidth}
                \includegraphics[width=\linewidth]{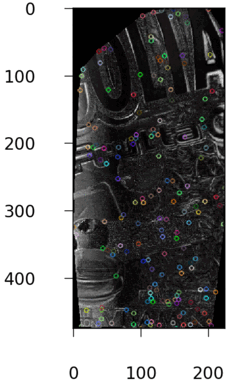}
                \caption{Pixelwise difference norm.}
                \label{fig:figure_51i}
            \end{subfigure}
        \end{figure}
        \begin{figure}[H]
            \ContinuedFloat
            \centering
            \begin{subfigure}[!h]{0.4\linewidth}
                \includegraphics[width=\linewidth]{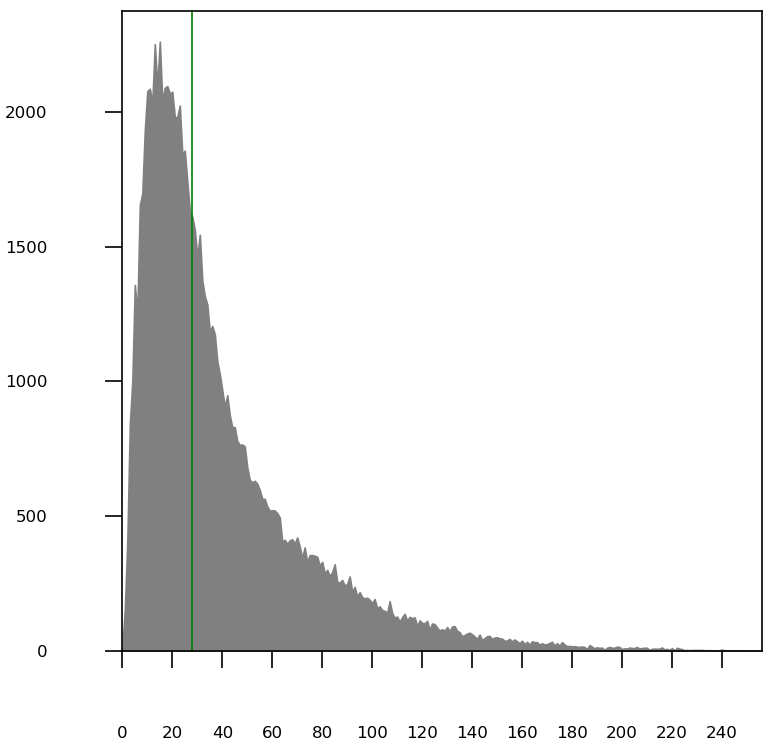}
                \caption{Pixelwise difference norm histogram.}
                \label{fig:figure_51j}
            \end{subfigure}
            \caption{Object occurrence (union of seeds \#4-5-7-9-10-23-25-27-29).}
            \label{fig:figure_51}
        \end{figure}
\subsection{Case 5}
    \label{appendix_A.2.5}
    \paragraph{Feature-based object detection algorithm driven by RANSAC}
        \begin{figure}[H]
            \centering
            \begin{subfigure}[!h]{0.4\linewidth}
                \includegraphics[width=\linewidth]{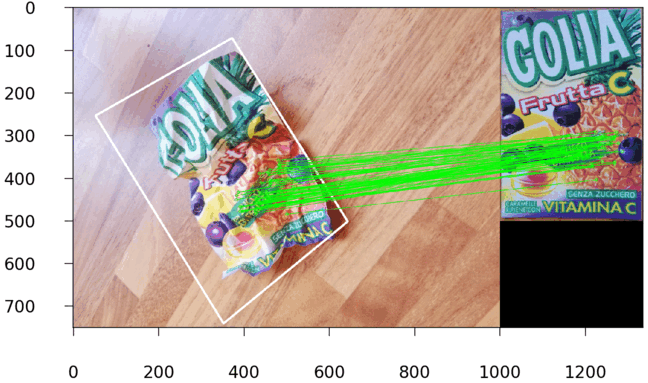}
                \caption{Clustered matches.}
                \label{fig:figure_46a}
            \end{subfigure}
            \begin{subfigure}[!h]{0.35\linewidth}
                \includegraphics[width=\linewidth]{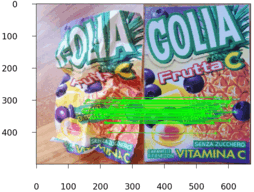}
                \caption{Rectified object occurrence.}
                \label{fig:figure_46b}
            \end{subfigure}
            \begin{subfigure}[!h]{0.35\linewidth}
                \includegraphics[width=\linewidth]{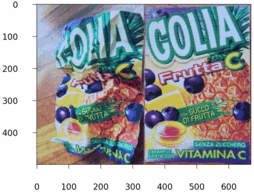}
                \caption{Template and histogram matched object.}
                \label{fig:figure_46c}
            \end{subfigure}
            \begin{subfigure}[!h]{0.3\linewidth}
                \includegraphics[width=\linewidth]{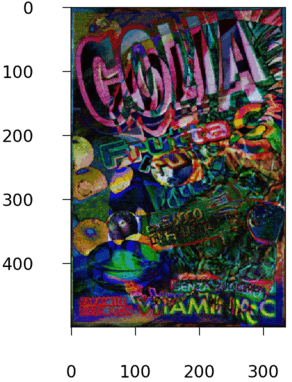}
                \caption{BGR absolute difference.}
                \label{fig:figure_46d}
            \end{subfigure}
            \begin{subfigure}[!h]{0.4\linewidth}
                \includegraphics[width=\linewidth]{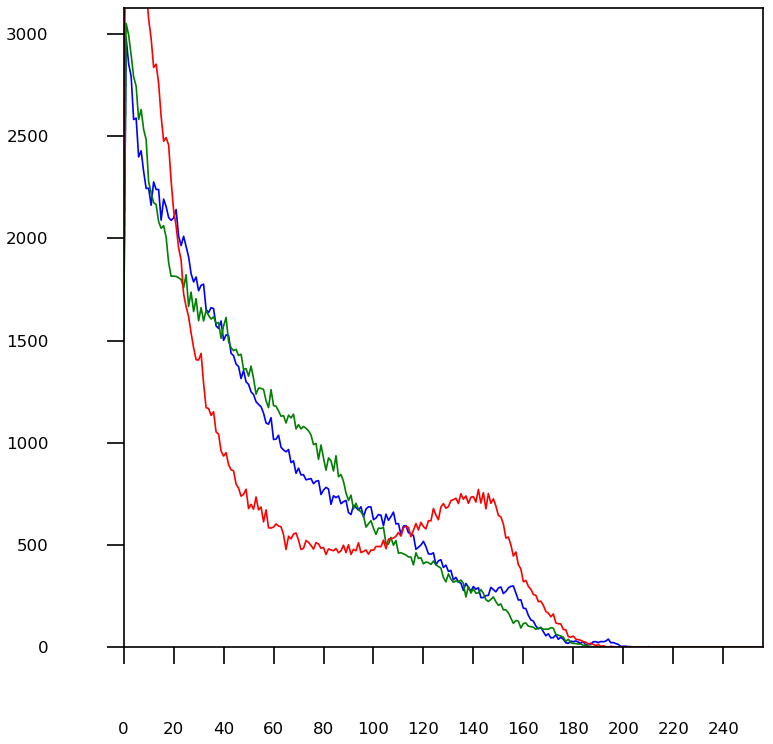}
                \caption{BGR difference histogram.}
                \label{fig:figure_46h}
            \end{subfigure}
            \begin{subfigure}[!h]{0.3\linewidth}
                \includegraphics[width=\linewidth]{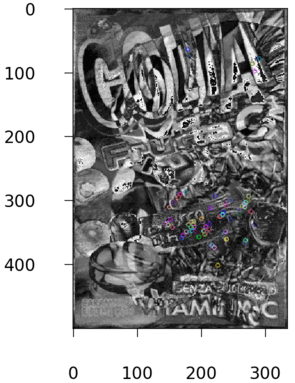}
                \caption{Pixelwise difference norm.}
                \label{fig:figure_46i}
            \end{subfigure}
        \end{figure}
        \begin{figure}[H]
            \ContinuedFloat
            \centering
            \begin{subfigure}[!h]{0.4\linewidth}
                \includegraphics[width=\linewidth]{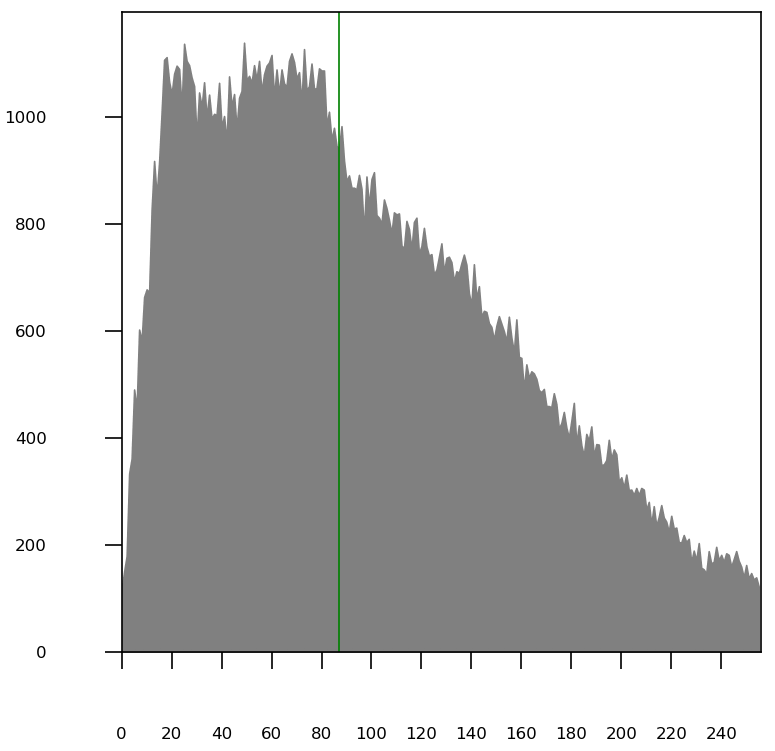}
                \caption{Pixelwise difference norm histogram.}
                \label{fig:figure_46j}
            \end{subfigure}
            \caption{Object occurrence.}
            \label{fig:figure_46}
        \end{figure}
    \paragraph{Our algorithm}
        \begin{figure}[H]
            \centering
            \begin{subfigure}[!h]{0.4\linewidth}
                \includegraphics[width=\linewidth]{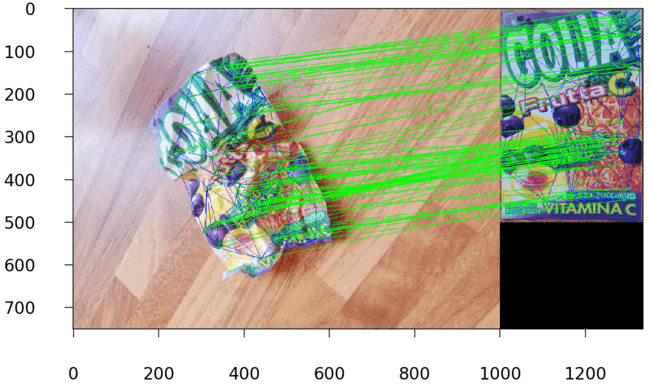}
                \caption{Expanded seed.}
                \label{fig:figure_52a}
            \end{subfigure}
            \begin{subfigure}[!h]{0.35\linewidth}
                \includegraphics[width=\linewidth]{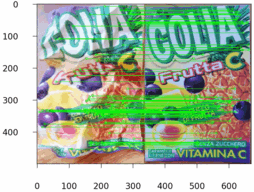}
                \caption{Rectified object occurrence.}
                \label{fig:figure_52b}
            \end{subfigure}
            \begin{subfigure}[!h]{0.35\linewidth}
                \includegraphics[width=\linewidth]{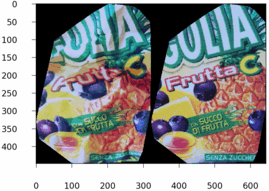}
                \caption{Template and histogram matched object.}
                \label{fig:figure_52c}
            \end{subfigure}
            \begin{subfigure}[!h]{0.3\linewidth}
                \includegraphics[width=\linewidth]{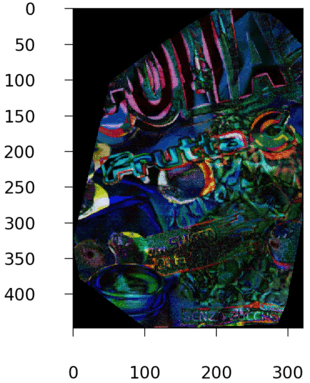}
                \caption{BGR absolute difference.}
                \label{fig:figure_52d}
            \end{subfigure}
            \begin{subfigure}[!h]{0.4\linewidth}
                \includegraphics[width=\linewidth]{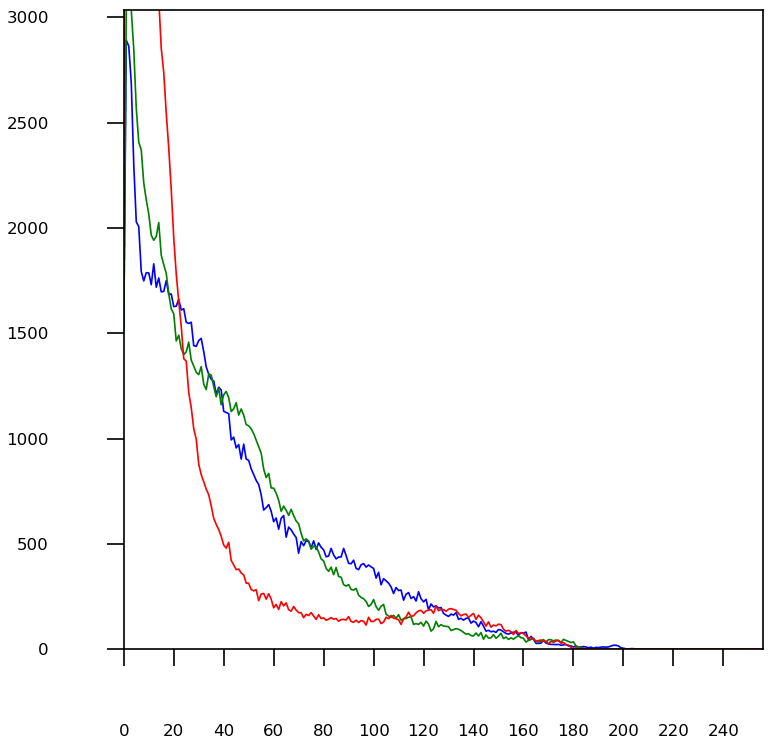}
                \caption{BGR difference histogram.}
                \label{fig:figure_52h}
            \end{subfigure}
            \begin{subfigure}[!h]{0.3\linewidth}
                \includegraphics[width=\linewidth]{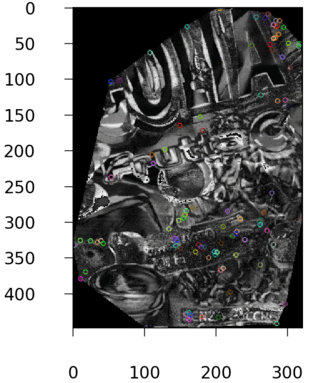}
                \caption{Pixelwise difference norm.}
                \label{fig:figure_52i}
            \end{subfigure}
        \end{figure}
        \begin{figure}[H]
            \ContinuedFloat
            \centering
            \begin{subfigure}[!h]{0.4\linewidth}
                \includegraphics[width=\linewidth]{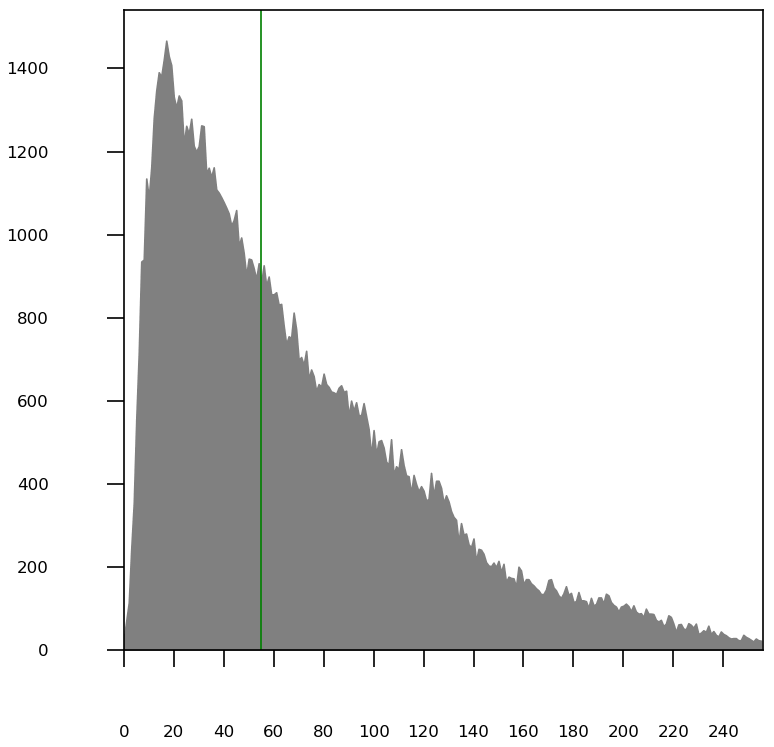}
                \caption{Pixelwise difference norm histogram.}
                \label{fig:figure_52j}
            \end{subfigure}
            \caption{Object occurrence (union of seeds \#0-1-3-4-5-6-7-8-9-10-11-12-13-15-16-17-18-19-20-21-22-23-24-25-26-27-28).}
            \label{fig:figure_52}
        \end{figure}
\subsection{Case 6}
    \label{appendix_A.2.6}
    \paragraph{Feature-based object detection algorithm driven by RANSAC}
        \begin{figure}[H]
            \centering
            \begin{subfigure}[!h]{0.4\linewidth}
                \includegraphics[width=\linewidth]{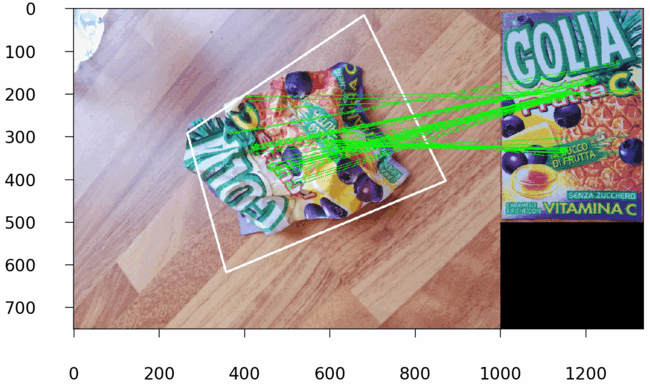}
                \caption{Clustered matches.}
                \label{fig:figure_47a}
            \end{subfigure}
            \begin{subfigure}[!h]{0.35\linewidth}
                \includegraphics[width=\linewidth]{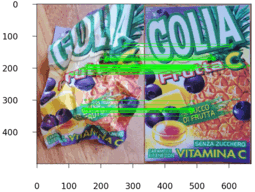}
                \caption{Rectified object occurrence.}
                \label{fig:figure_47b}
            \end{subfigure}
            \begin{subfigure}[!h]{0.35\linewidth}
                \includegraphics[width=\linewidth]{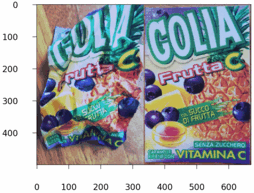}
                \caption{Template and histogram matched object.}
                \label{fig:figure_47c}
            \end{subfigure}
            \begin{subfigure}[!h]{0.3\linewidth}
                \includegraphics[width=\linewidth]{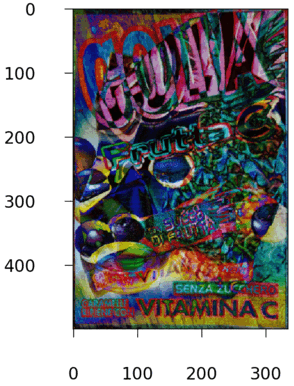}
                \caption{BGR absolute difference.}
                \label{fig:figure_47d}
            \end{subfigure}
            \begin{subfigure}[!h]{0.4\linewidth}
                \includegraphics[width=\linewidth]{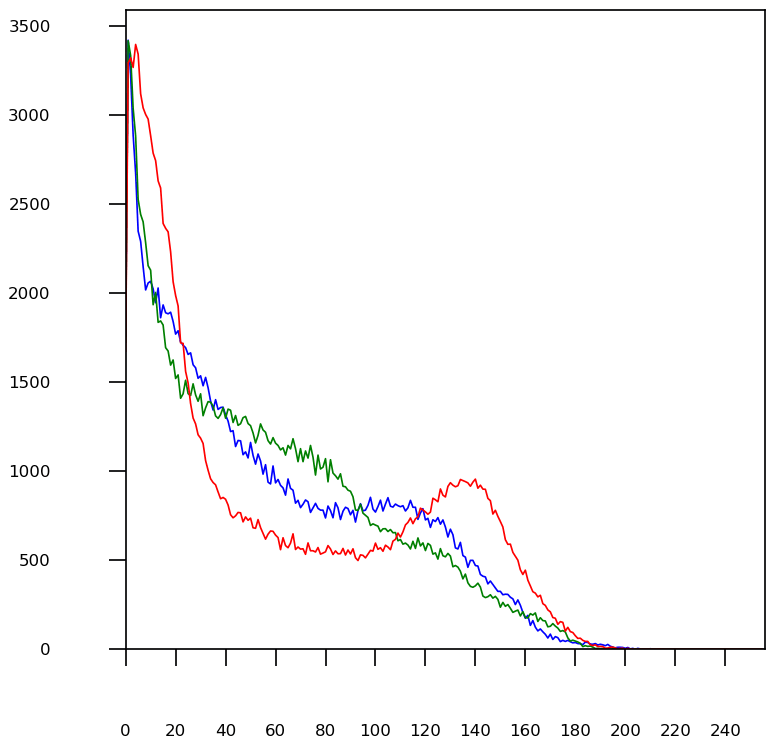}
                \caption{BGR difference histogram.}
                \label{fig:figure_47h}
            \end{subfigure}
            \begin{subfigure}[!h]{0.3\linewidth}
                \includegraphics[width=\linewidth]{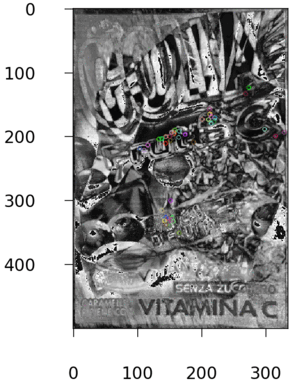}
                \caption{Pixelwise difference norm.}
                \label{fig:figure_47i}
            \end{subfigure}
        \end{figure}
        \begin{figure}[H]
            \ContinuedFloat
            \centering
            \begin{subfigure}[!h]{0.4\linewidth}
                \includegraphics[width=\linewidth]{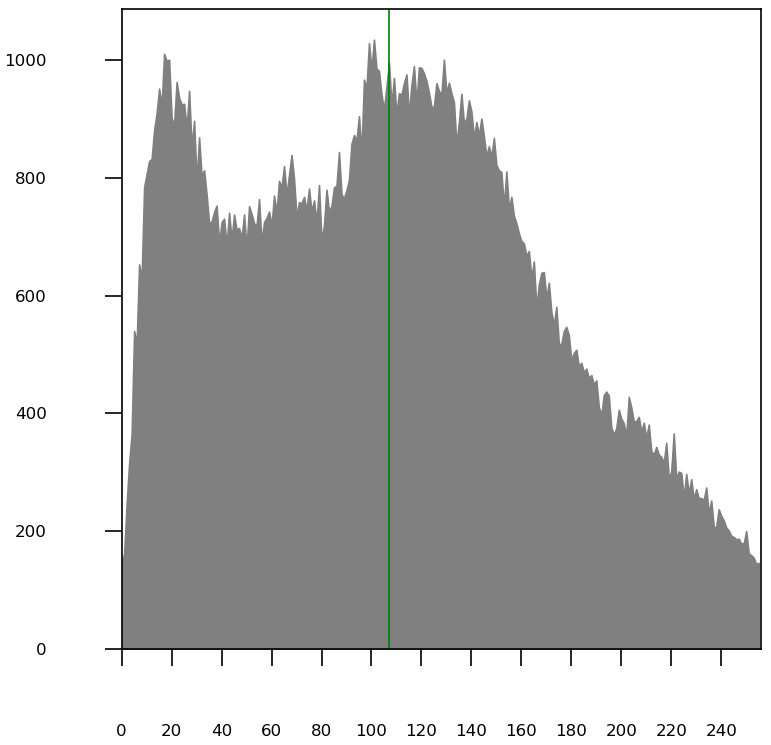}
                \caption{Pixelwise difference norm histogram.}
                \label{fig:figure_47j}
            \end{subfigure}
            \caption{Object occurrence.}
            \label{fig:figure_47}
        \end{figure}
    \paragraph{Our algorithm}
        \begin{figure}[H]
            \centering
            \begin{subfigure}[!h]{0.4\linewidth}
                \includegraphics[width=\linewidth]{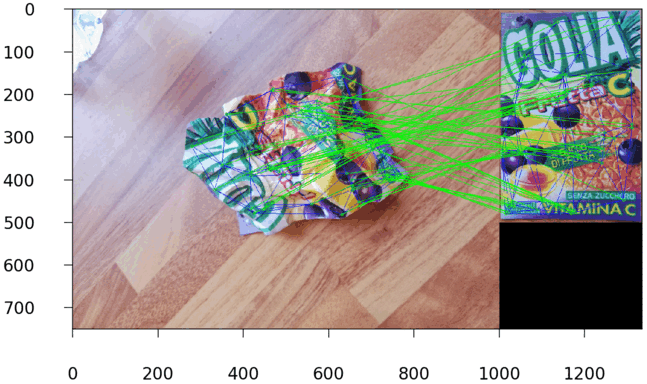}
                \caption{Expanded seed.}
                \label{fig:figure_53a}
            \end{subfigure}
            \begin{subfigure}[!h]{0.35\linewidth}
                \includegraphics[width=\linewidth]{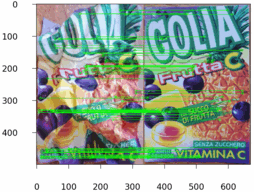}
                \caption{Rectified object occurrence.}
                \label{fig:figure_53b}
            \end{subfigure}
            \begin{subfigure}[!h]{0.35\linewidth}
                \includegraphics[width=\linewidth]{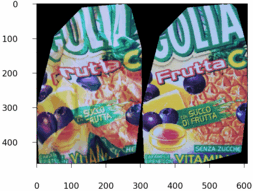}
                \caption{Template and histogram matched object.}
                \label{fig:figure_53c}
            \end{subfigure}
            \begin{subfigure}[!h]{0.3\linewidth}
                \includegraphics[width=\linewidth]{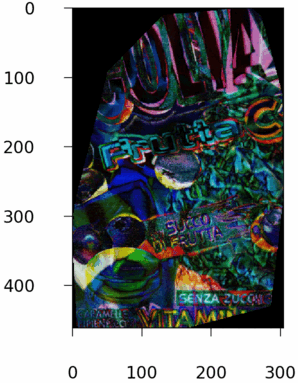}
                \caption{BGR absolute difference.}
                \label{fig:figure_53d}
            \end{subfigure}
            \begin{subfigure}[!h]{0.4\linewidth}
                \includegraphics[width=\linewidth]{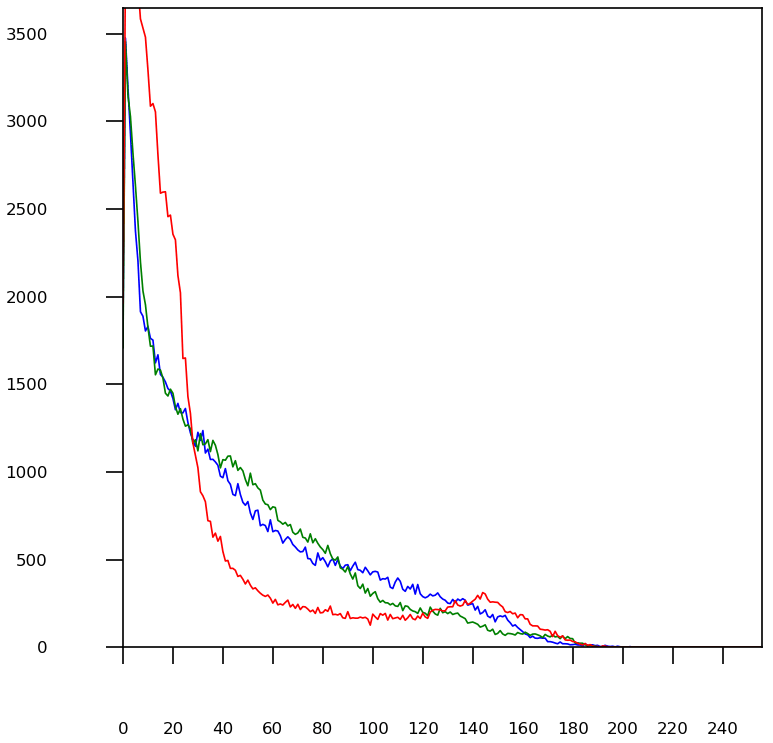}
                \caption{BGR difference histogram.}
                \label{fig:figure_53h}
            \end{subfigure}
            \begin{subfigure}[!h]{0.3\linewidth}
                \includegraphics[width=\linewidth]{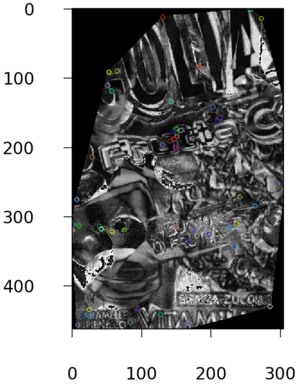}
                \caption{Pixelwise difference norm.}
                \label{fig:figure_53i}
            \end{subfigure}
        \end{figure}
        \begin{figure}[H]
            \ContinuedFloat
            \centering
            \begin{subfigure}[!h]{0.4\linewidth}
                \includegraphics[width=\linewidth]{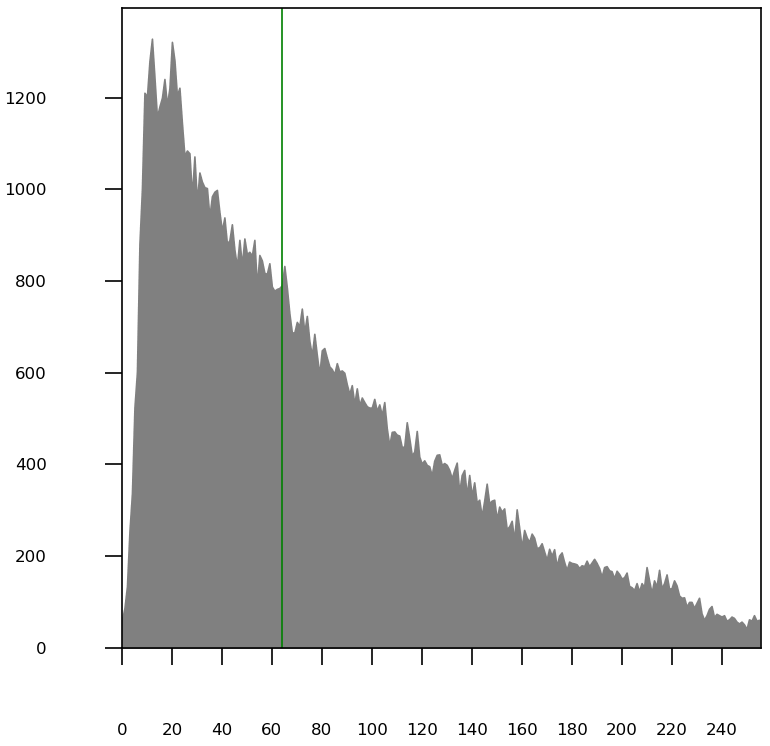}
                \caption{Pixelwise difference norm histogram.}
                \label{fig:figure_53j}
            \end{subfigure}
            \caption{Object occurrence (union of seeds \#5-8-11-12-13-14-15-16-17-18-20-21-24-25-26-27-28-29).}
            \label{fig:figure_53}
        \end{figure}
\section{Many distorted object occurrences}
    \paragraph{Feature-based object detection algorithm driven by RANSAC}
        \label{appendix_A.3.1}
        \begin{figure}[H]
    \centering
    \begin{subfigure}[!h]{0.7\linewidth}
        \includegraphics[width=\linewidth]{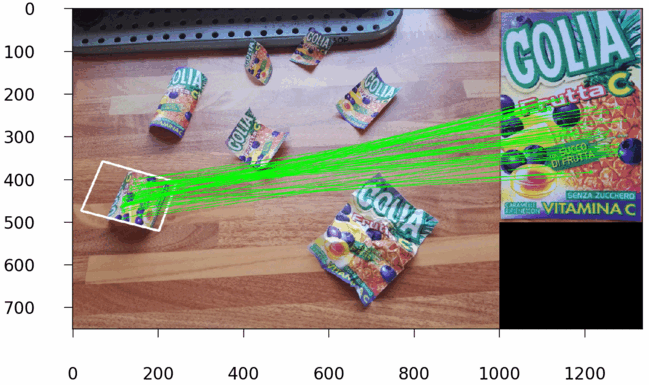}
        \caption{Clustered matches.}
        \label{fig:figure_61a}
    \end{subfigure}
    \begin{subfigure}[!h]{0.35\linewidth}
        \includegraphics[width=\linewidth]{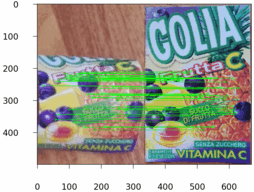}
        \caption{Rectified object occurrence.}
        \label{fig:figure_61b}
    \end{subfigure}
    \begin{subfigure}[!h]{0.35\linewidth}
        \includegraphics[width=\linewidth]{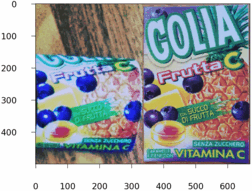}
        \caption{Template and histogram matched object.}
        \label{fig:figure_61c}
    \end{subfigure}
    \begin{subfigure}[!h]{0.3\linewidth}
        \includegraphics[width=\linewidth]{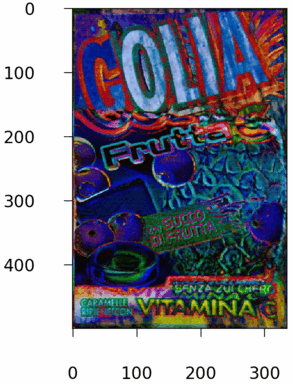}
        \caption{BGR absolute difference.}
        \label{fig:figure_61d}
    \end{subfigure}
    \begin{subfigure}[!h]{0.4\linewidth}
        \includegraphics[width=\linewidth]{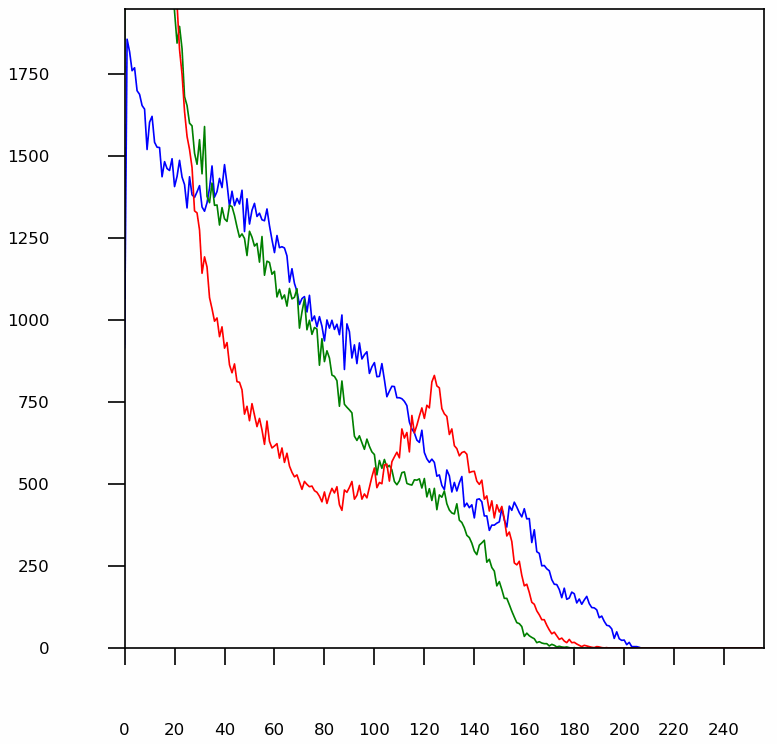}
        \caption{BGR difference histogram.}
        \label{fig:figure_61h}
    \end{subfigure}
\end{figure}
\begin{figure}[H]
    \ContinuedFloat
    \centering
    \begin{subfigure}[!h]{0.3\linewidth}
        \includegraphics[width=\linewidth]{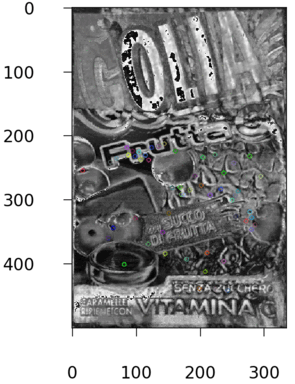}
        \caption{Pixelwise difference norm.}
        \label{fig:figure_61i}
    \end{subfigure}
    \begin{subfigure}[!h]{0.4\linewidth}
        \includegraphics[width=\linewidth]{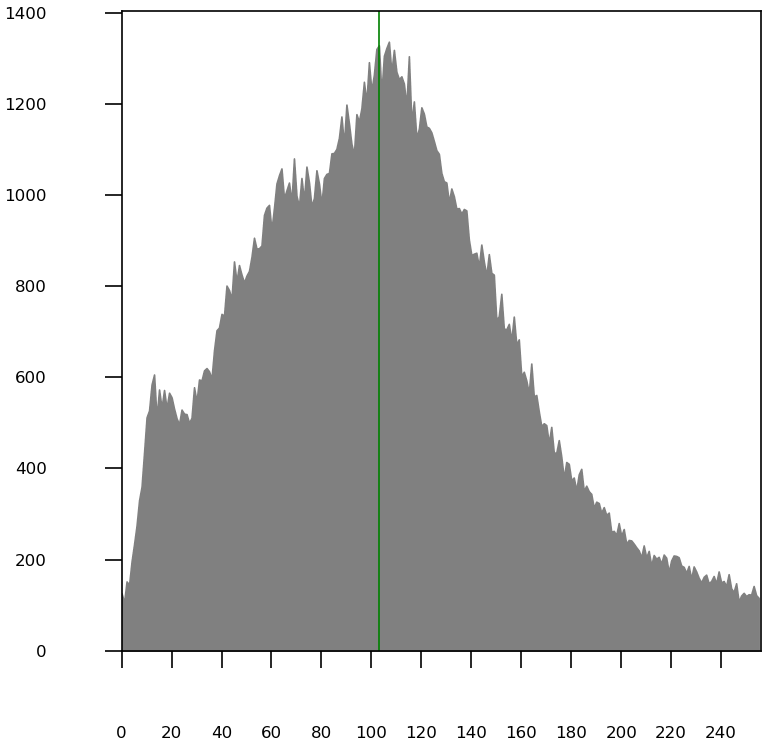}
        \caption{Pixelwise difference norm histogram.}
        \label{fig:figure_61j}
    \end{subfigure}
    \caption{1\textsuperscript{st} object occurrence.}
    \label{fig:figure_61}
\end{figure}
\begin{figure}[H]
    \centering
    \begin{subfigure}[!h]{0.7\linewidth}
        \includegraphics[width=\linewidth]{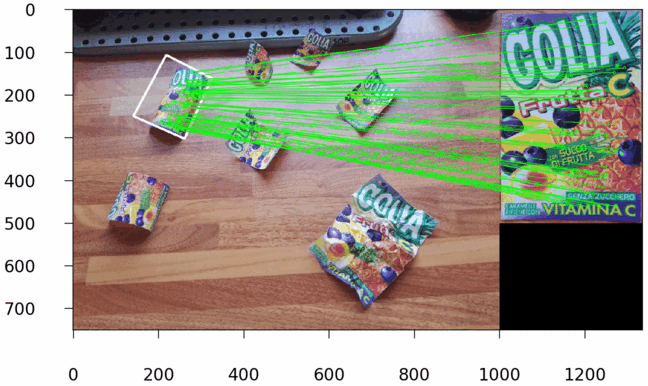}
        \caption{Clustered matches.}
    \end{subfigure}
    \begin{subfigure}[!h]{0.35\linewidth}
        \includegraphics[width=\linewidth]{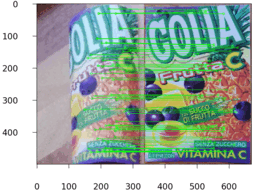}
        \caption{Rectified object occurrence.}
    \end{subfigure}
    \begin{subfigure}[!h]{0.35\linewidth}
        \includegraphics[width=\linewidth]{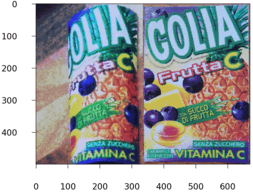}
        \caption{Template and histogram matched object.}
    \end{subfigure}
    \begin{subfigure}[!h]{0.3\linewidth}
        \includegraphics[width=\linewidth]{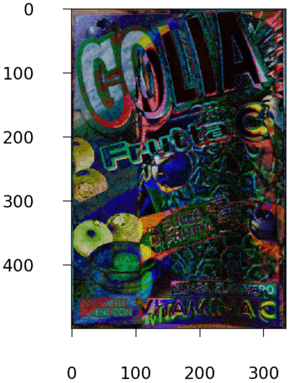}
        \caption{BGR absolute difference.}
    \end{subfigure}
    \begin{subfigure}[!h]{0.4\linewidth}
        \includegraphics[width=\linewidth]{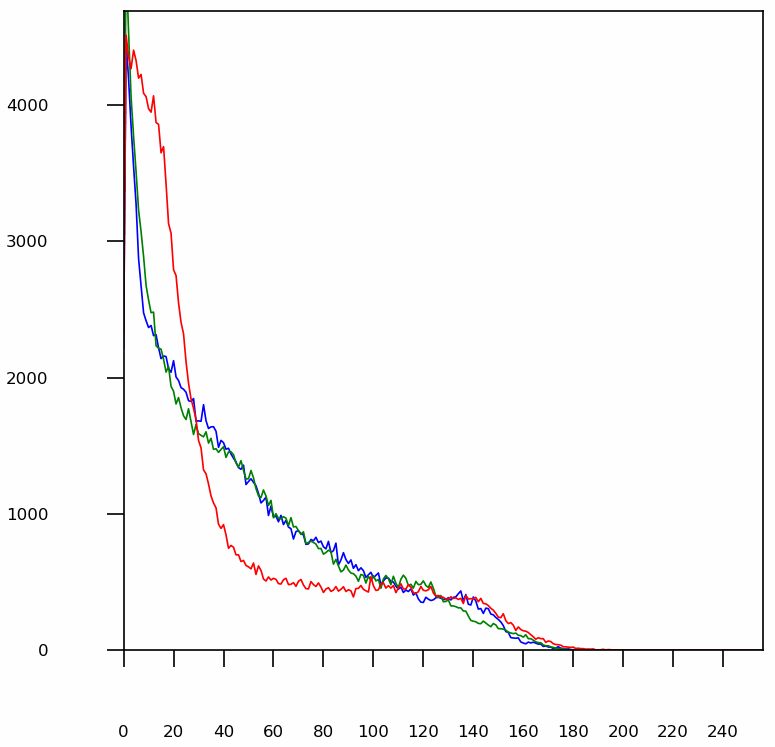}
        \caption{BGR difference histogram.}
    \end{subfigure}
\end{figure}
\begin{figure}[H]
    \centering
    \begin{subfigure}[!h]{0.3\linewidth}
        \includegraphics[width=\linewidth]{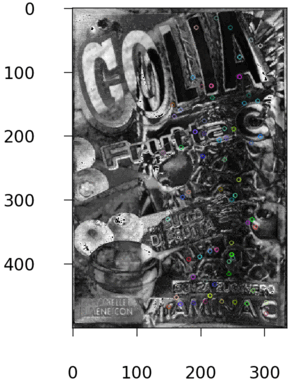}
        \caption{Pixelwise difference norm.}
    \end{subfigure}
    \begin{subfigure}[!h]{0.4\linewidth}
        \includegraphics[width=\linewidth]{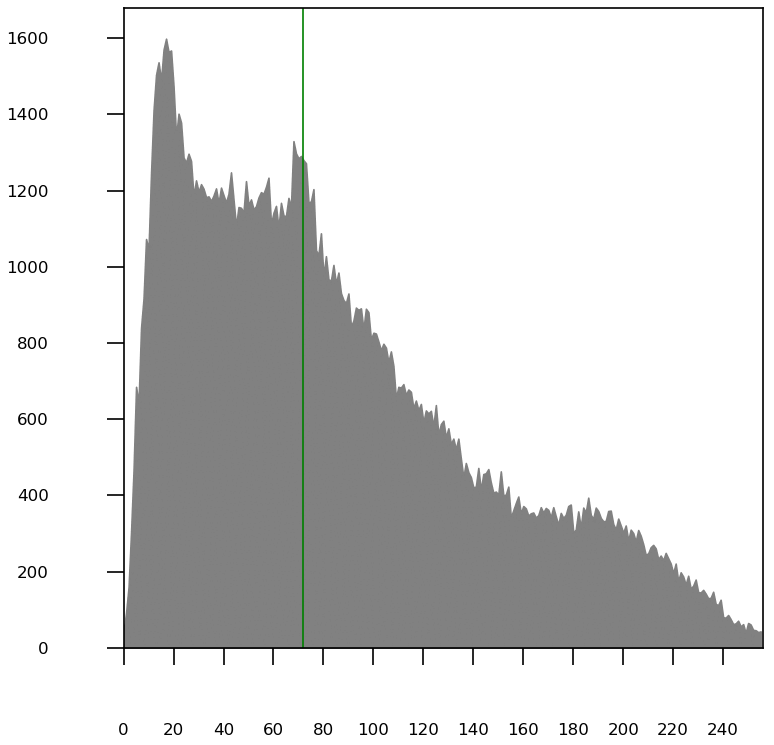}
        \caption{Pixelwise difference norm histogram.}
    \end{subfigure}
    \caption{2\textsuperscript{nd} object occurrence.}
    \label{fig:figure_62}
\end{figure}
\begin{figure}[H]
    \centering
    \begin{subfigure}[!h]{0.7\linewidth}
        \includegraphics[width=\linewidth]{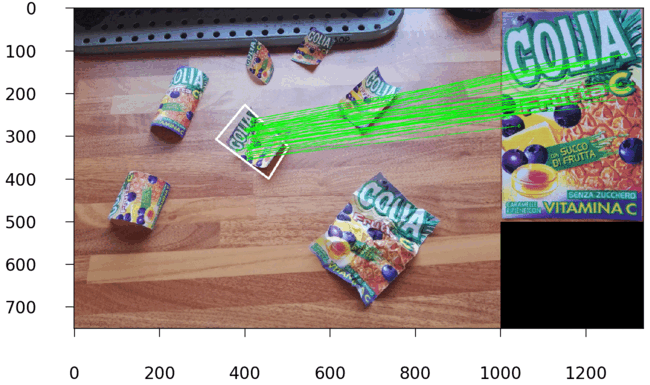}
        \caption{Clustered matches.}
    \end{subfigure}
    \begin{subfigure}[!h]{0.35\linewidth}
        \includegraphics[width=\linewidth]{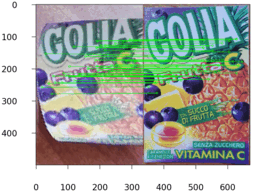}
        \caption{Rectified object occurrence.}
    \end{subfigure}
    \begin{subfigure}[!h]{0.35\linewidth}
        \includegraphics[width=\linewidth]{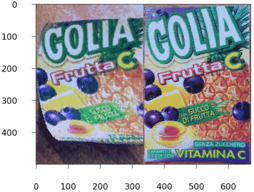}
        \caption{Template and histogram matched object.}
    \end{subfigure}
    \begin{subfigure}[!h]{0.3\linewidth}
        \includegraphics[width=\linewidth]{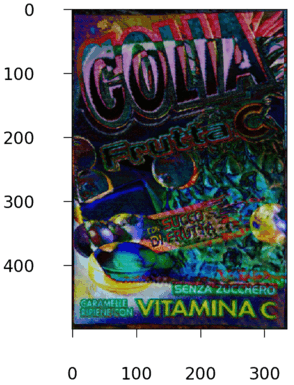}
        \caption{BGR absolute difference.}
    \end{subfigure}
    \begin{subfigure}[!h]{0.4\linewidth}
        \includegraphics[width=\linewidth]{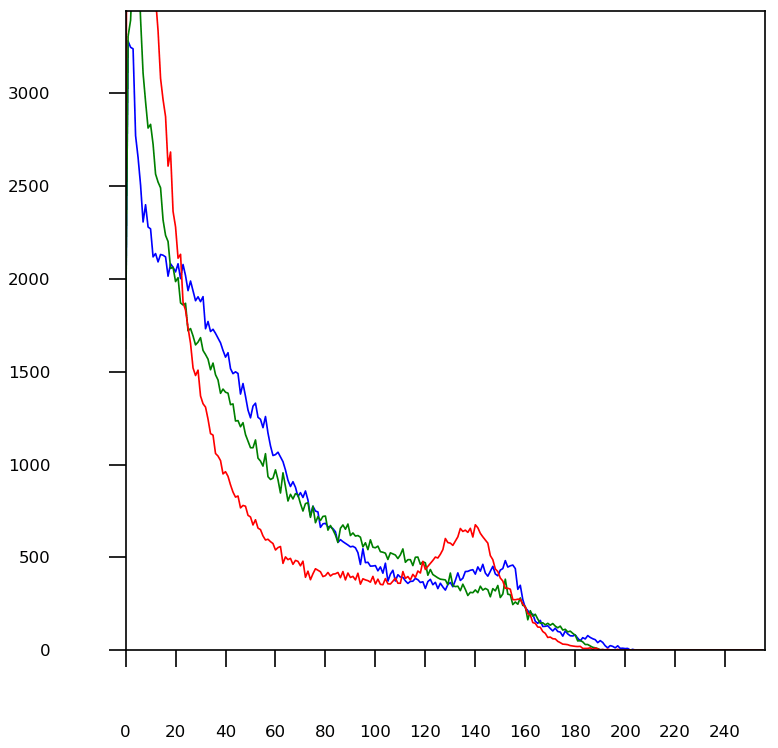}
        \caption{BGR difference histogram.}
    \end{subfigure}
\end{figure}
\begin{figure}[H]
    \centering
    \begin{subfigure}[!h]{0.3\linewidth}
        \includegraphics[width=\linewidth]{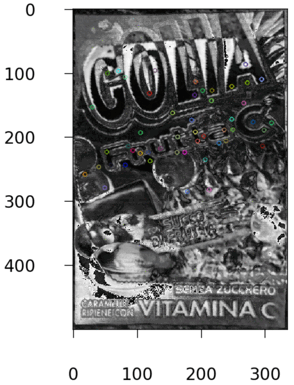}
        \caption{Pixelwise difference norm.}
    \end{subfigure}
    \begin{subfigure}[!h]{0.4\linewidth}
        \includegraphics[width=\linewidth]{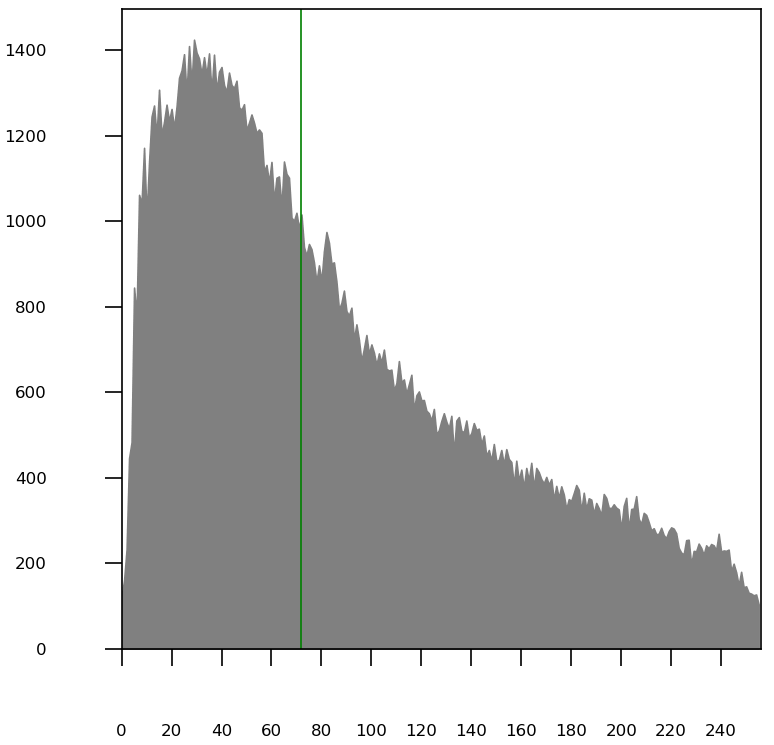}
        \caption{Pixelwise difference norm histogram.}
    \end{subfigure}
    \caption{3\textsuperscript{rd} object occurrence.}
    \label{fig:figure_63}
\end{figure}
\begin{figure}[H]
    \centering
    \begin{subfigure}[!h]{0.7\linewidth}
        \includegraphics[width=\linewidth]{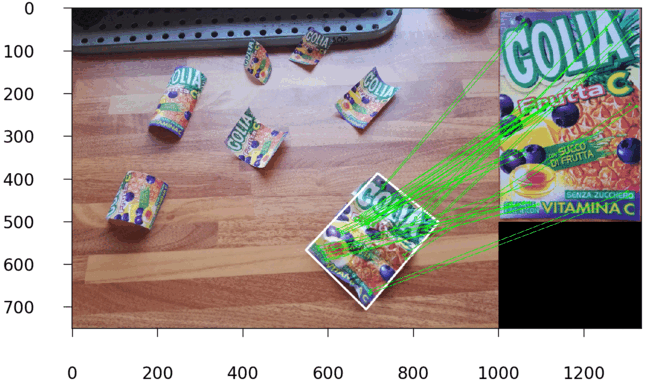}
        \caption{Clustered matches.}
    \end{subfigure}
    \begin{subfigure}[!h]{0.35\linewidth}
        \includegraphics[width=\linewidth]{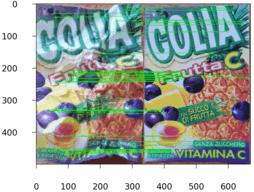}
        \caption{Rectified object occurrence.}
    \end{subfigure}
    \begin{subfigure}[!h]{0.35\linewidth}
        \includegraphics[width=\linewidth]{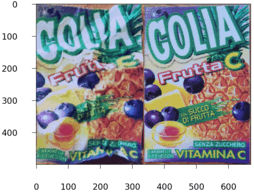}
        \caption{Template and histogram matched object.}
    \end{subfigure}
    \begin{subfigure}[!h]{0.3\linewidth}
        \includegraphics[width=\linewidth]{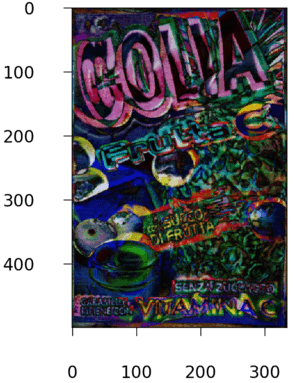}
        \caption{BGR absolute difference.}
    \end{subfigure}
    \begin{subfigure}[!h]{0.4\linewidth}
        \includegraphics[width=\linewidth]{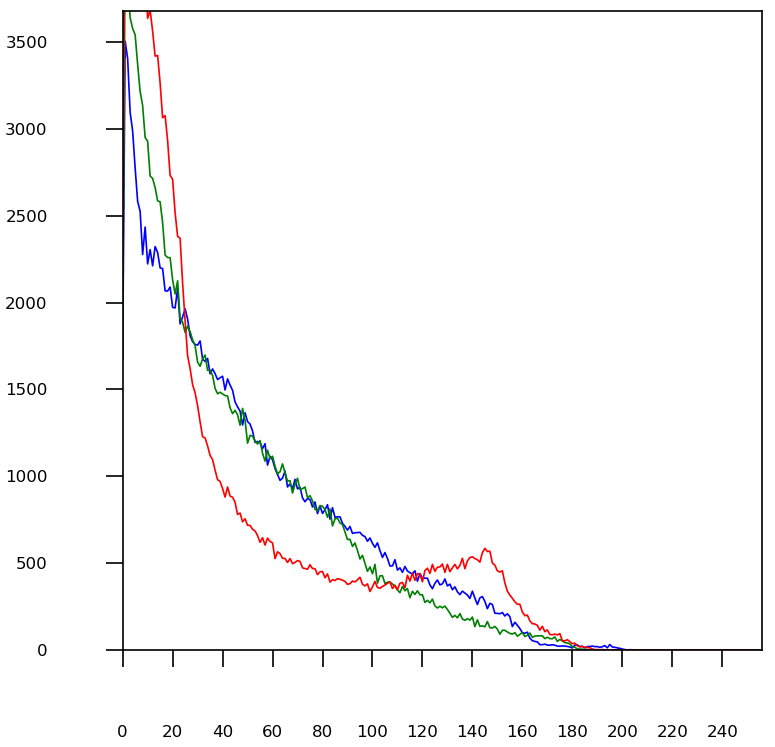}
        \caption{BGR difference histogram.}
    \end{subfigure}
\end{figure}
\begin{figure}[H]
    \centering
    \begin{subfigure}[!h]{0.3\linewidth}
        \includegraphics[width=\linewidth]{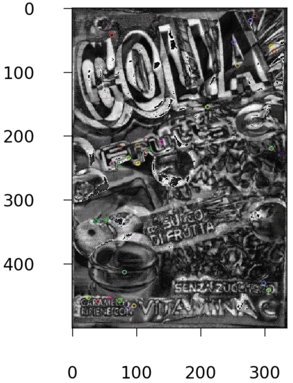}
        \caption{Pixelwise difference norm.}
    \end{subfigure}
    \begin{subfigure}[!h]{0.4\linewidth}
        \includegraphics[width=\linewidth]{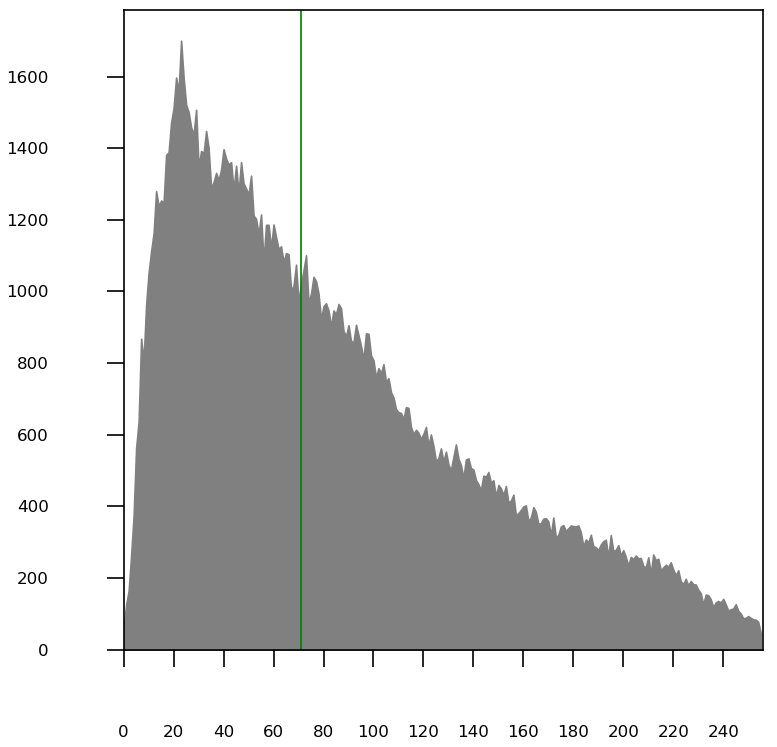}
        \caption{Pixelwise difference norm histogram.}
    \end{subfigure}
    \caption{4\textsuperscript{th} object occurrence.}
    \label{fig:figure_64}
\end{figure}
\begin{figure}[H]
    \centering
    \begin{subfigure}[!h]{0.7\linewidth}
        \includegraphics[width=\linewidth]{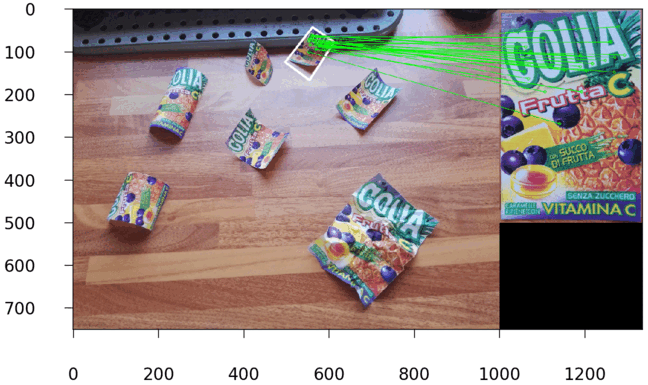}
        \caption{Clustered matches.}
    \end{subfigure}
    \begin{subfigure}[!h]{0.35\linewidth}
        \includegraphics[width=\linewidth]{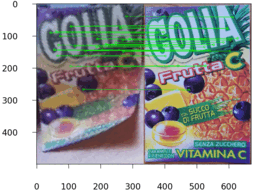}
        \caption{Rectified object occurrence.}
    \end{subfigure}
    \begin{subfigure}[!h]{0.35\linewidth}
        \includegraphics[width=\linewidth]{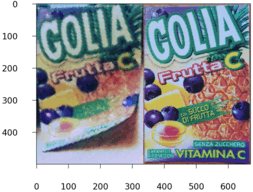}
        \caption{Template and histogram matched object.}
    \end{subfigure}
    \begin{subfigure}[!h]{0.3\linewidth}
        \includegraphics[width=\linewidth]{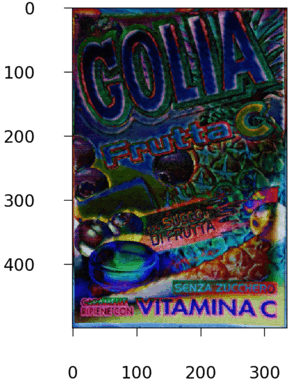}
        \caption{BGR absolute difference.}
    \end{subfigure}
    \begin{subfigure}[!h]{0.4\linewidth}
        \includegraphics[width=\linewidth]{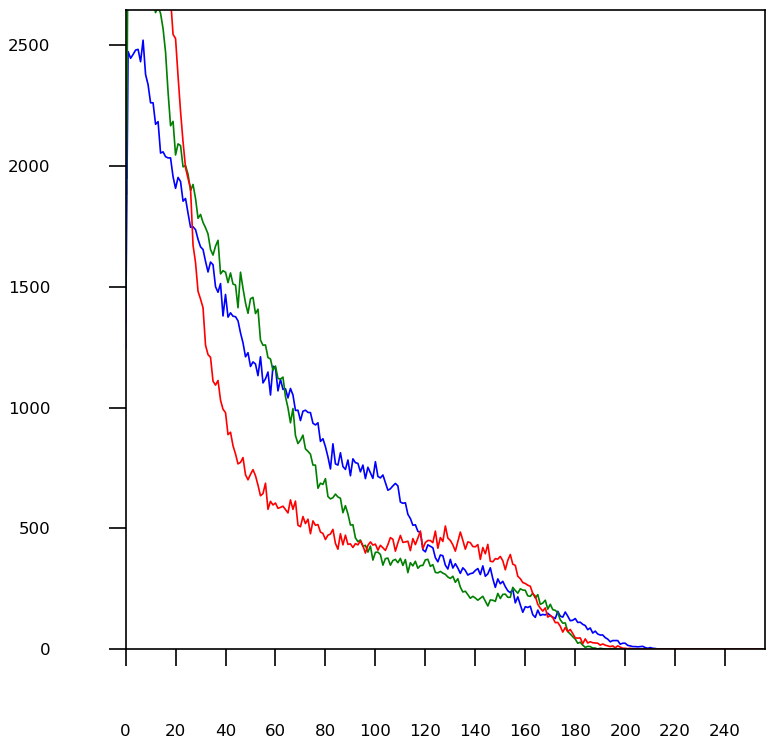}
        \caption{BGR difference histogram.}
    \end{subfigure}
\end{figure}
\begin{figure}[H]
    \centering
    \begin{subfigure}[!h]{0.3\linewidth}
        \includegraphics[width=\linewidth]{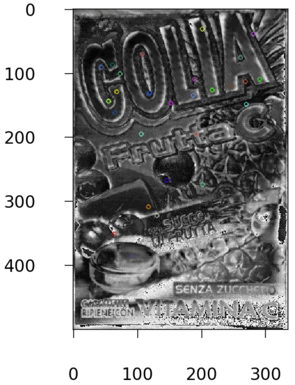}
        \caption{Pixelwise difference norm.}
    \end{subfigure}
    \begin{subfigure}[!h]{0.4\linewidth}
        \includegraphics[width=\linewidth]{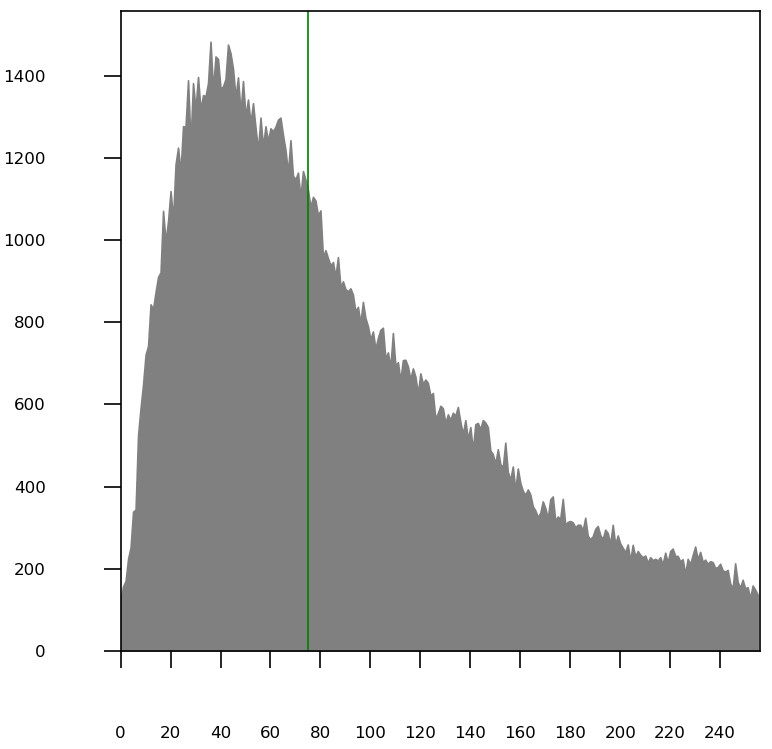}
        \caption{Pixelwise difference norm histogram.}
    \end{subfigure}
    \caption{5\textsuperscript{th} object occurrence.}
    \label{fig:figure_65}
\end{figure}
\begin{figure}[H]
    \centering
    \begin{subfigure}[!h]{0.7\linewidth}
        \includegraphics[width=\linewidth]{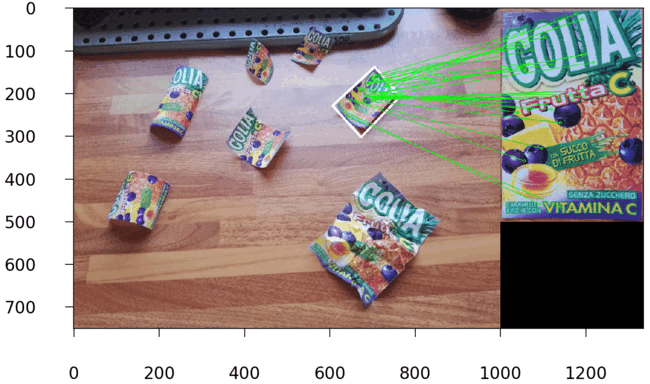}
        \caption{Clustered matches.}
    \end{subfigure}
    \begin{subfigure}[!h]{0.35\linewidth}
        \includegraphics[width=\linewidth]{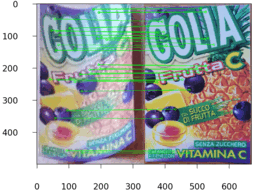}
        \caption{Rectified object occurrence.}
    \end{subfigure}
    \begin{subfigure}[!h]{0.35\linewidth}
        \includegraphics[width=\linewidth]{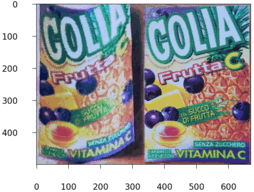}
        \caption{Template and histogram matched object.}
    \end{subfigure}
    \begin{subfigure}[!h]{0.3\linewidth}
        \includegraphics[width=\linewidth]{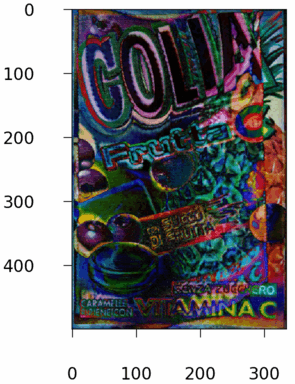}
        \caption{BGR absolute difference.}
    \end{subfigure}
    \begin{subfigure}[!h]{0.4\linewidth}
        \includegraphics[width=\linewidth]{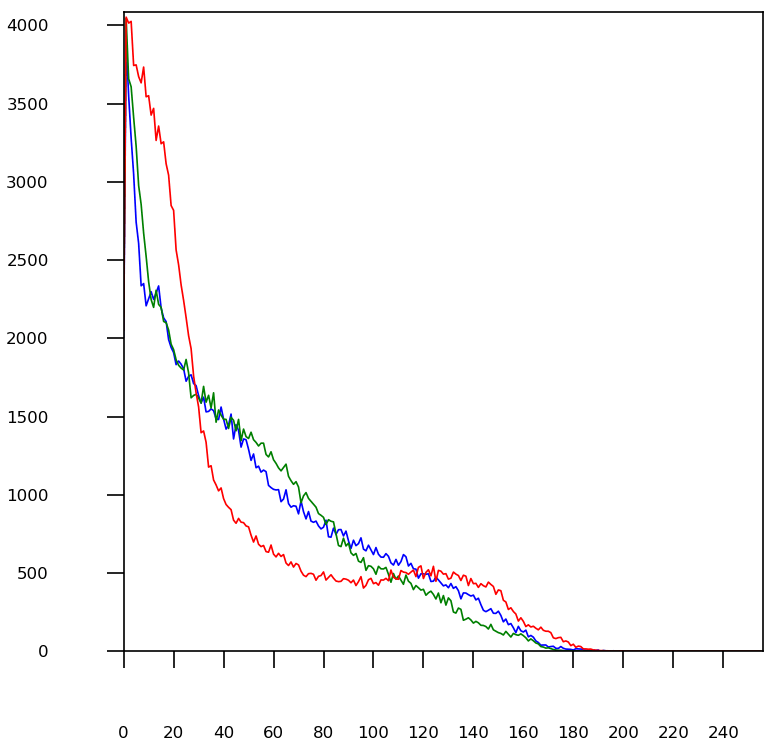}
        \caption{BGR difference histogram.}
    \end{subfigure}
\end{figure}
\begin{figure}[H]
    \centering
    \begin{subfigure}[!h]{0.3\linewidth}
        \includegraphics[width=\linewidth]{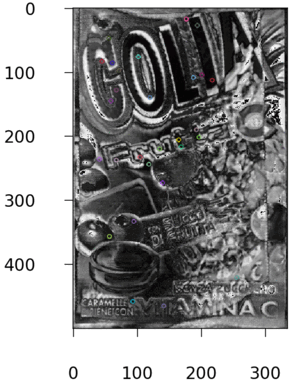}
        \caption{Pixelwise difference norm.}
    \end{subfigure}
    \begin{subfigure}[!h]{0.4\linewidth}
        \includegraphics[width=\linewidth]{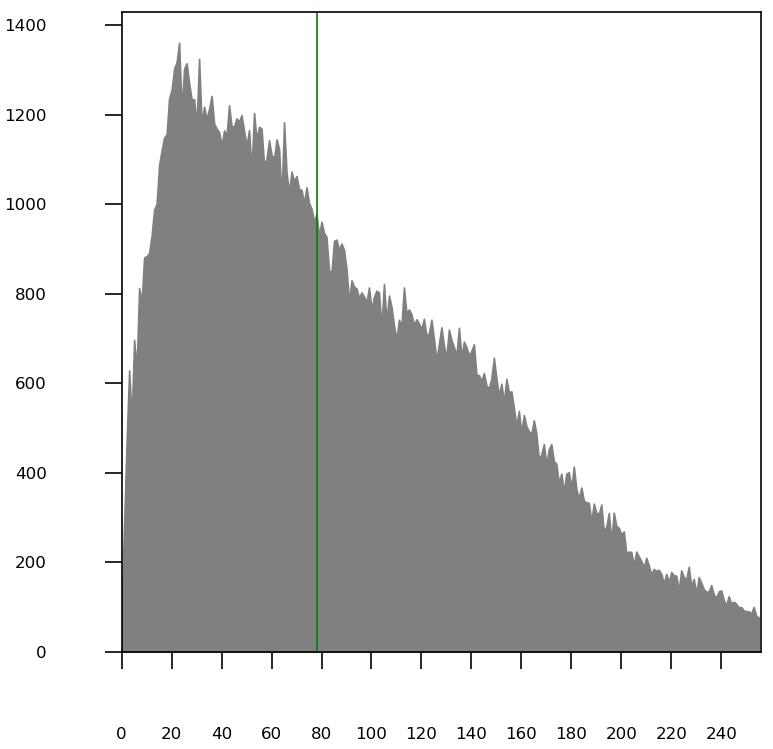}
        \caption{Pixelwise difference norm histogram.}
    \end{subfigure}
    \caption{6\textsuperscript{th} object occurrence.}
    \label{fig:figure_66}
\end{figure}
\begin{figure}[H]
    \centering
    \begin{subfigure}[!h]{0.7\linewidth}
        \includegraphics[width=\linewidth]{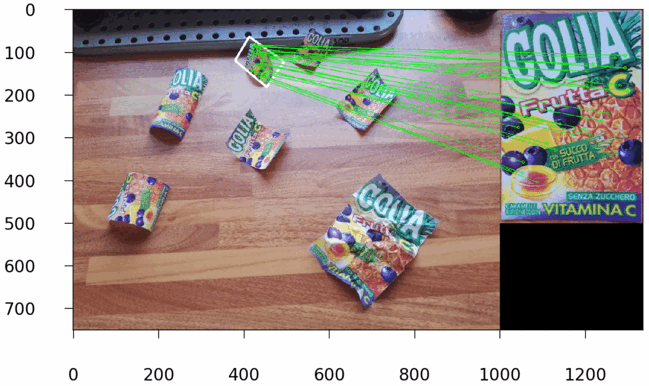}
        \caption{Clustered matches.}
    \end{subfigure}
    \begin{subfigure}[!h]{0.35\linewidth}
        \includegraphics[width=\linewidth]{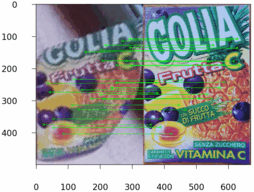}
        \caption{Rectified object occurrence.}
    \end{subfigure}
    \begin{subfigure}[!h]{0.35\linewidth}
        \includegraphics[width=\linewidth]{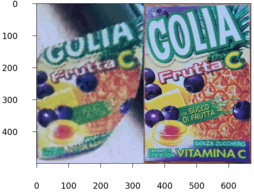}
        \caption{Template and histogram matched object.}
    \end{subfigure}
    \begin{subfigure}[!h]{0.3\linewidth}
        \includegraphics[width=\linewidth]{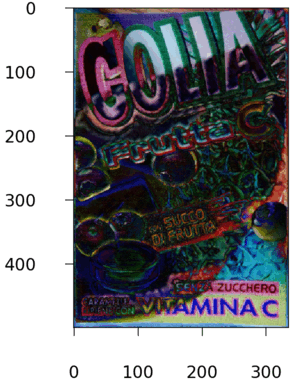}
        \caption{BGR absolute difference.}
    \end{subfigure}
    \begin{subfigure}[!h]{0.4\linewidth}
        \includegraphics[width=\linewidth]{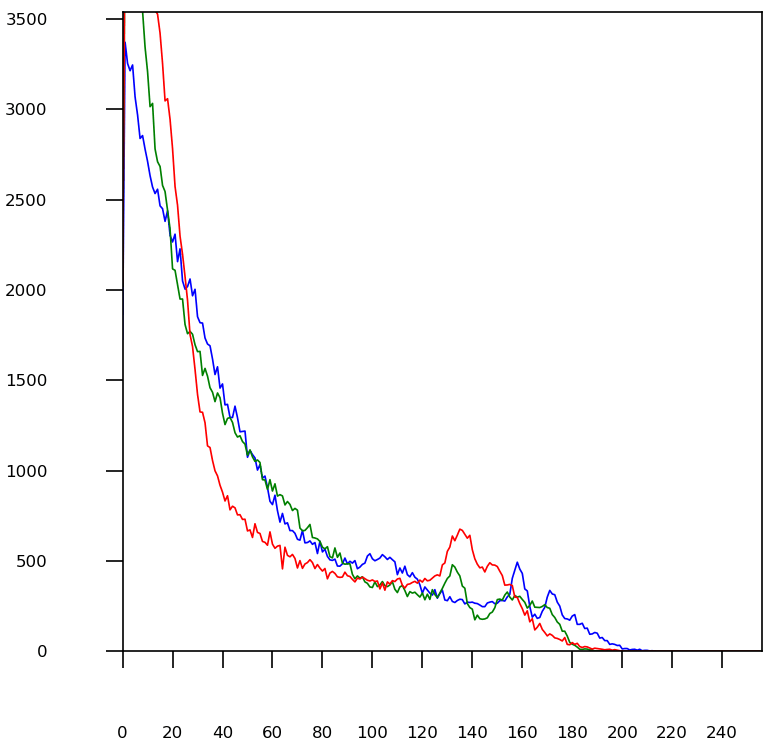}
        \caption{BGR difference histogram.}
    \end{subfigure}
\end{figure}
\begin{figure}[H]
    \centering
    \begin{subfigure}[!h]{0.3\linewidth}
        \includegraphics[width=\linewidth]{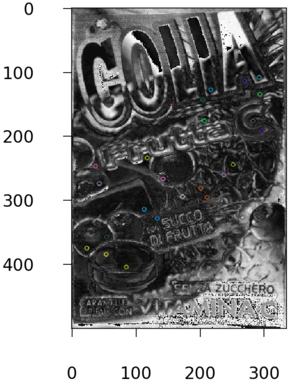}
        \caption{Pixelwise difference norm.}
    \end{subfigure}
    \begin{subfigure}[!h]{0.4\linewidth}
        \includegraphics[width=\linewidth]{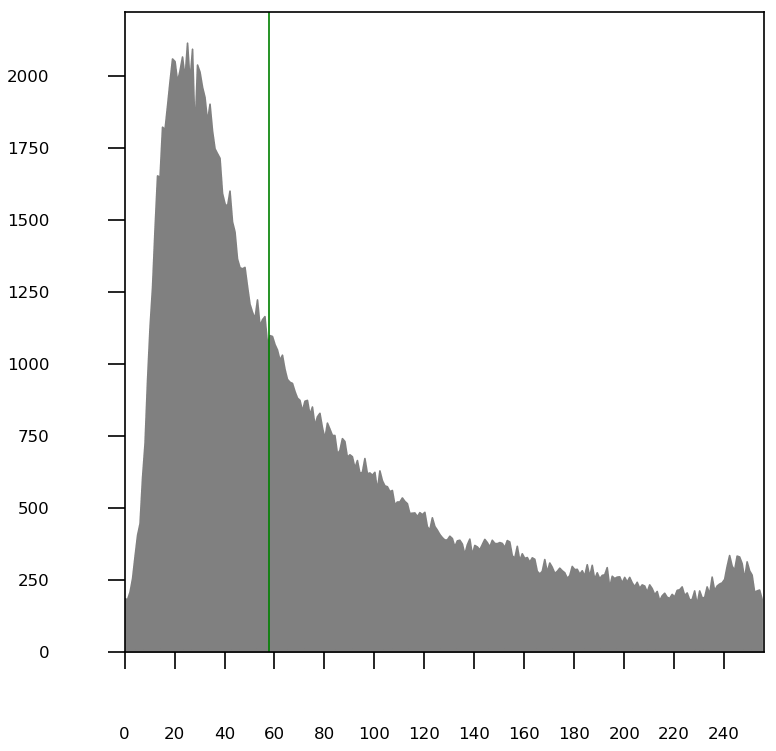}
        \caption{Pixelwise difference norm histogram.}
    \end{subfigure}
    \caption{7\textsuperscript{th} object occurrence.}
    \label{fig:figure_67}
\end{figure}
    \paragraph{Our algorithm}
        \label{appendix_A.3.2}
        \begin{figure}[H]
    \centering
    \begin{subfigure}[!h]{0.7\linewidth}
        \includegraphics[width=\linewidth]{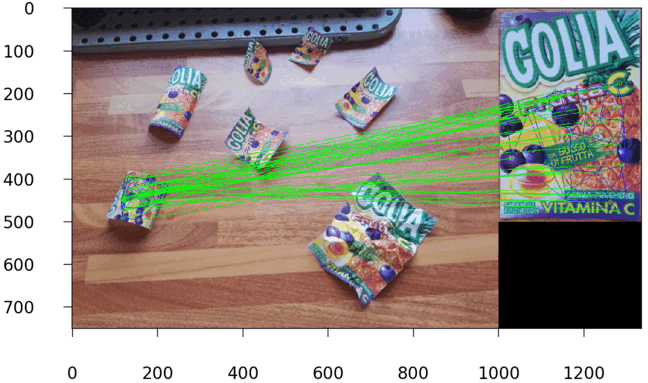}
        \caption{Expanded seed.}
        \label{fig:figure_68a}
    \end{subfigure}
    \begin{subfigure}[!h]{0.35\linewidth}
        \includegraphics[width=\linewidth]{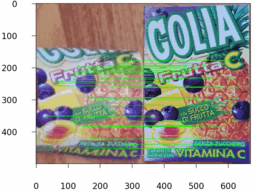}
        \caption{Rectified object occurrence.}
        \label{fig:figure_68b}
    \end{subfigure}
    \begin{subfigure}[!h]{0.35\linewidth}
        \includegraphics[width=\linewidth]{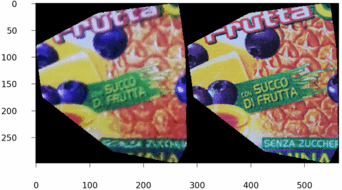}
        \caption{Template and histogram matched object.}
        \label{fig:figure_68c}
    \end{subfigure}
    \begin{subfigure}[!h]{0.3\linewidth}
        \includegraphics[width=\linewidth]{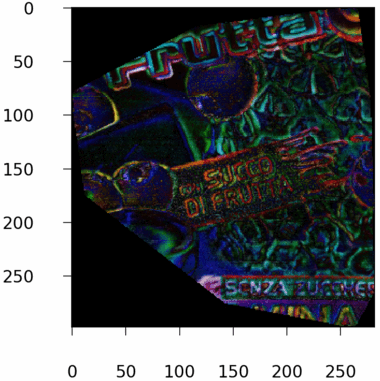}
        \caption{BGR absolute difference.}
        \label{fig:figure_68d}
    \end{subfigure}
    \begin{subfigure}[!h]{0.4\linewidth}
        \includegraphics[width=\linewidth]{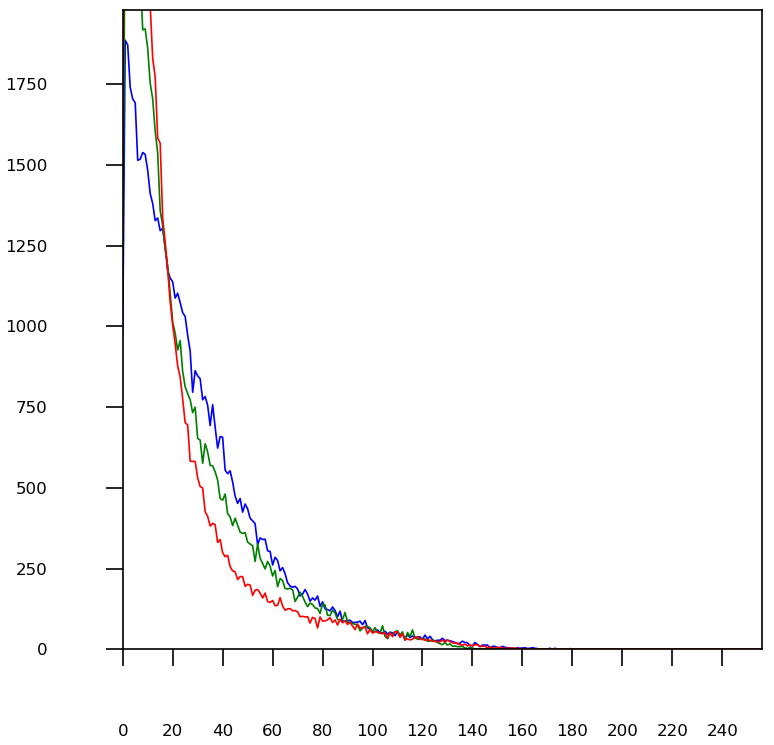}
        \caption{BGR difference histogram.}
        \label{fig:figure_68h}
    \end{subfigure}
\end{figure}
\begin{figure}[H]
    \ContinuedFloat
    \centering
    \begin{subfigure}[!h]{0.3\linewidth}
        \includegraphics[width=\linewidth]{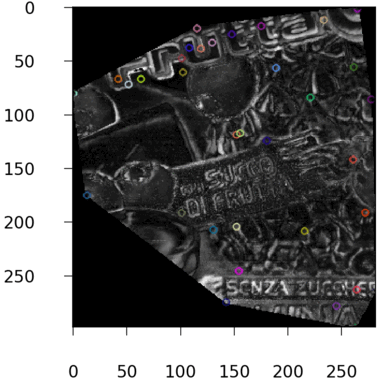}
        \caption{Pixelwise difference norm.}
        \label{fig:figure_68i}
    \end{subfigure}
    \begin{subfigure}[!h]{0.4\linewidth}
        \includegraphics[width=\linewidth]{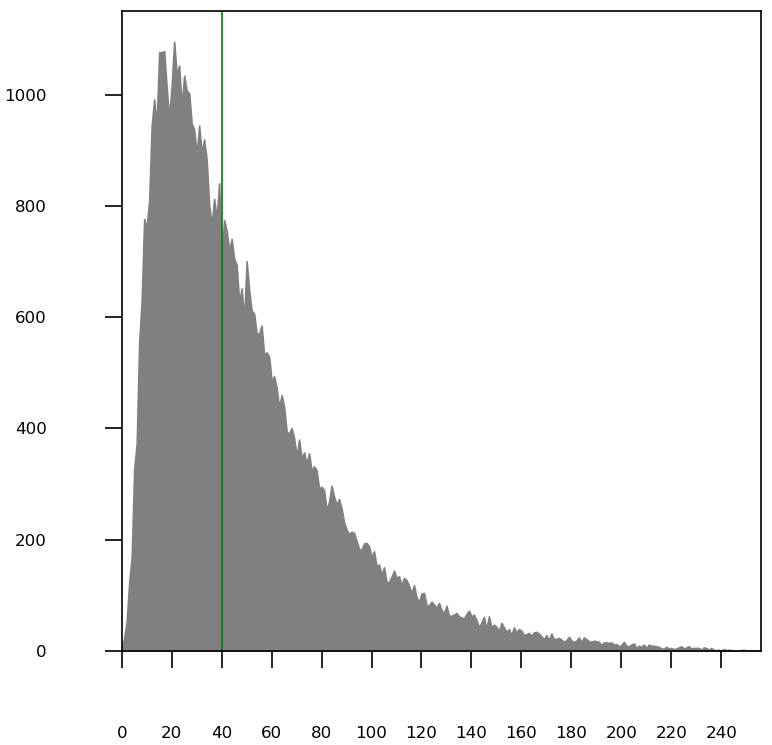}
        \caption{Pixelwise difference norm histogram.}
        \label{fig:figure_68j}
    \end{subfigure}
    \caption{1\textsuperscript{rd} object occurrence (union of seeds \#16-18-19).}
    \label{fig:figure_68}
\end{figure}
\begin{figure}[H]
    \centering
    \begin{subfigure}[!h]{0.7\linewidth}
        \includegraphics[width=\linewidth]{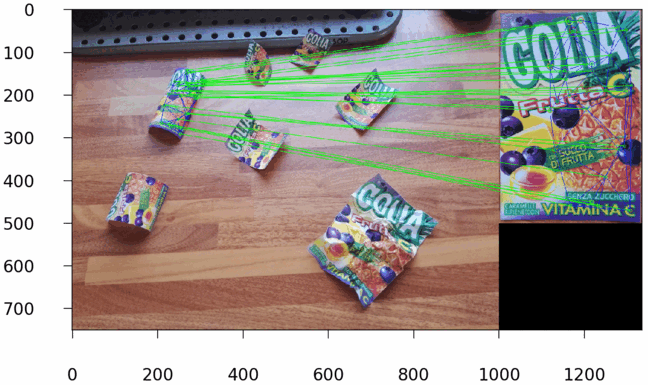}
        \caption{Expanded seed.}
    \end{subfigure}
    \begin{subfigure}[!h]{0.35\linewidth}
        \includegraphics[width=\linewidth]{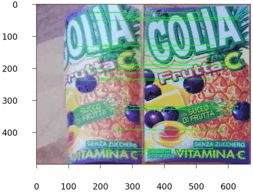}
        \caption{Rectified object occurrence.}
    \end{subfigure}
    \begin{subfigure}[!h]{0.35\linewidth}
        \includegraphics[width=\linewidth]{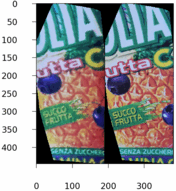}
        \caption{Template and histogram matched object.}
    \end{subfigure}
    \begin{subfigure}[!h]{0.3\linewidth}
        \includegraphics[width=\linewidth]{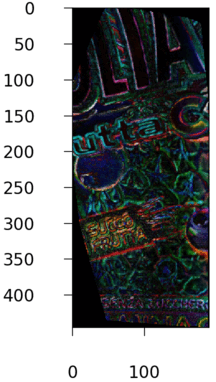}
        \caption{BGR absolute difference.}
    \end{subfigure}
    \begin{subfigure}[!h]{0.4\linewidth}
        \includegraphics[width=\linewidth]{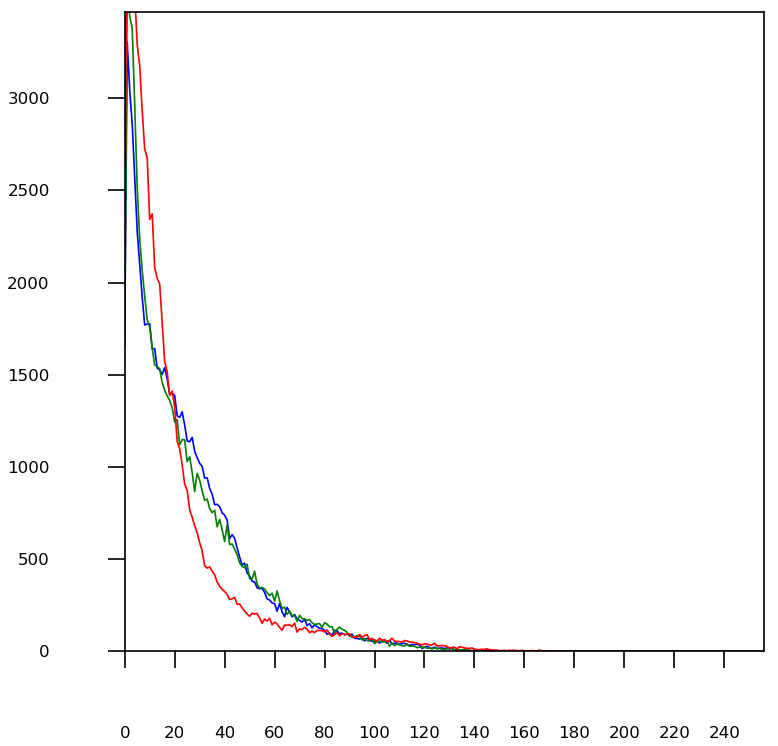}
        \caption{BGR difference histogram.}
    \end{subfigure}
\end{figure}
\begin{figure}[H]
    \ContinuedFloat
    \centering
    \begin{subfigure}[!h]{0.3\linewidth}
        \includegraphics[width=\linewidth]{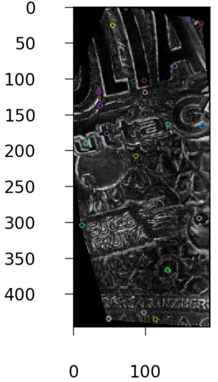}
        \caption{Pixelwise difference norm.}
    \end{subfigure}
    \begin{subfigure}[!h]{0.4\linewidth}
        \includegraphics[width=\linewidth]{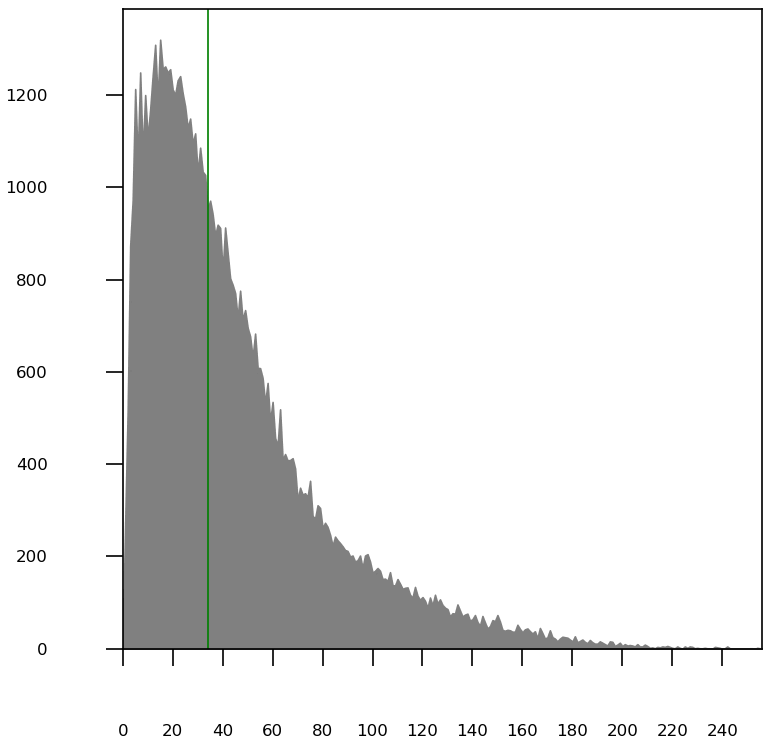}
        \caption{Pixelwise difference norm histogram.}
    \end{subfigure}
    \caption{2\textsuperscript{st} object occurrence (union of seeds \#2-3).}
    \label{fig:figure_69}
\end{figure}
\begin{figure}[H]
    \centering
    \begin{subfigure}[!h]{0.7\linewidth}
        \includegraphics[width=\linewidth]{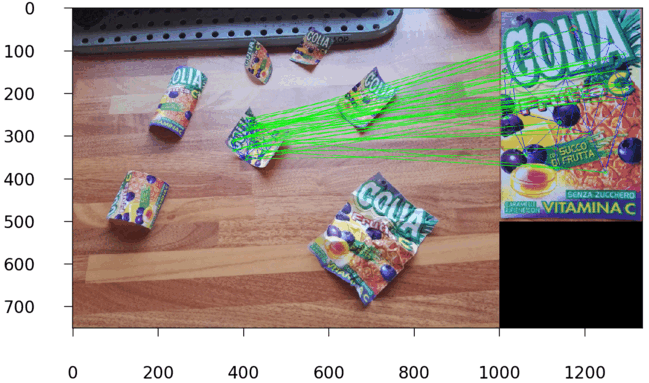}
        \caption{Expanded seed.}
    \end{subfigure}
    \begin{subfigure}[!h]{0.35\linewidth}
        \includegraphics[width=\linewidth]{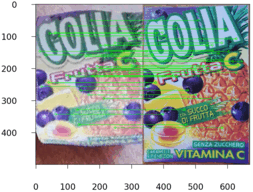}
        \caption{Rectified object occurrence.}
    \end{subfigure}
    \begin{subfigure}[!h]{0.35\linewidth}
        \includegraphics[width=\linewidth]{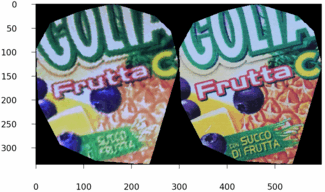}
        \caption{Template and histogram matched object.}
    \end{subfigure}
    \begin{subfigure}[!h]{0.3\linewidth}
        \includegraphics[width=\linewidth]{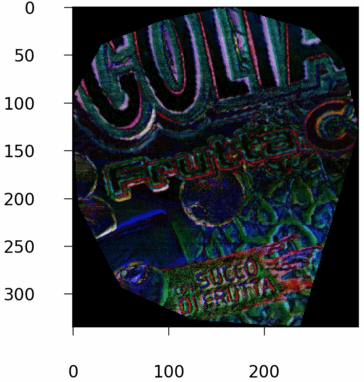}
        \caption{BGR absolute difference.}
    \end{subfigure}
    \begin{subfigure}[!h]{0.4\linewidth}
        \includegraphics[width=\linewidth]{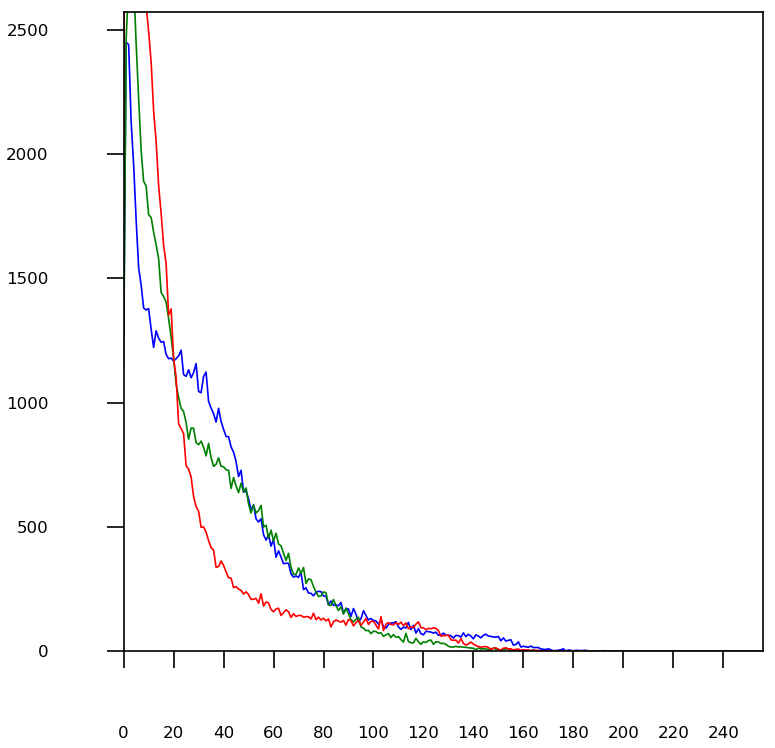}
        \caption{BGR difference histogram.}
    \end{subfigure}
\end{figure}
\begin{figure}[H]
    \centering
    \begin{subfigure}[!h]{0.3\linewidth}
        \includegraphics[width=\linewidth]{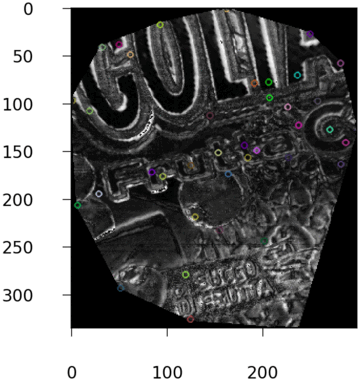}
        \caption{Pixelwise difference norm.}
    \end{subfigure}
    \begin{subfigure}[!h]{0.4\linewidth}
        \includegraphics[width=\linewidth]{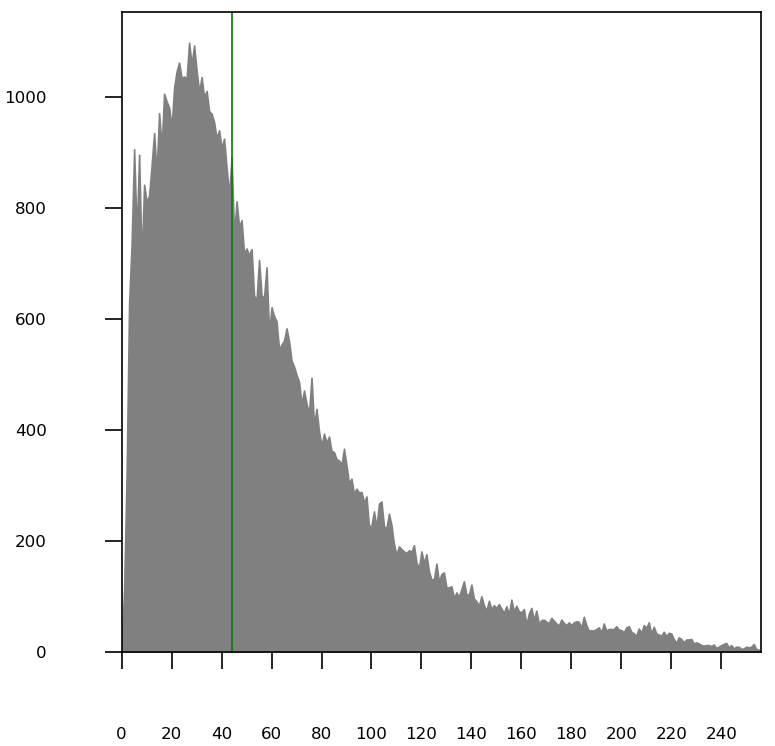}
        \caption{Pixelwise difference norm histogram.}
    \end{subfigure}
    \caption{3\textsuperscript{nd} object occurrence (union of seeds \#14-26).}
    \label{fig:figure_70}
\end{figure}
\begin{figure}[H]
    \centering
    \begin{subfigure}[!h]{0.7\linewidth}
        \includegraphics[width=\linewidth]{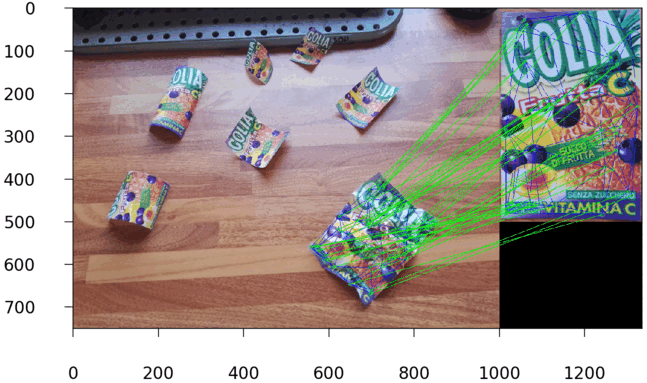}
        \caption{Expanded seed.}
    \end{subfigure}
    \begin{subfigure}[!h]{0.35\linewidth}
        \includegraphics[width=\linewidth]{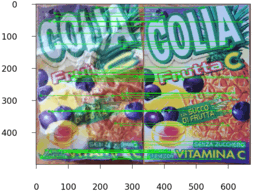}
        \caption{Rectified object occurrence.}
    \end{subfigure}
    \begin{subfigure}[!h]{0.35\linewidth}
        \includegraphics[width=\linewidth]{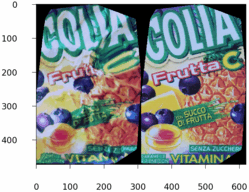}
        \caption{Template and histogram matched object.}
    \end{subfigure}
    \begin{subfigure}[!h]{0.3\linewidth}
        \includegraphics[width=\linewidth]{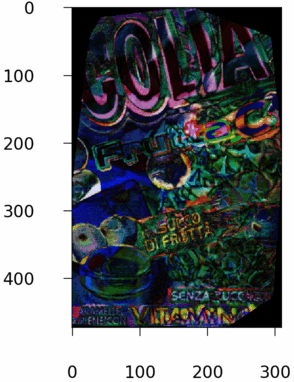}
        \caption{BGR absolute difference.}
    \end{subfigure}
    \begin{subfigure}[!h]{0.4\linewidth}
        \includegraphics[width=\linewidth]{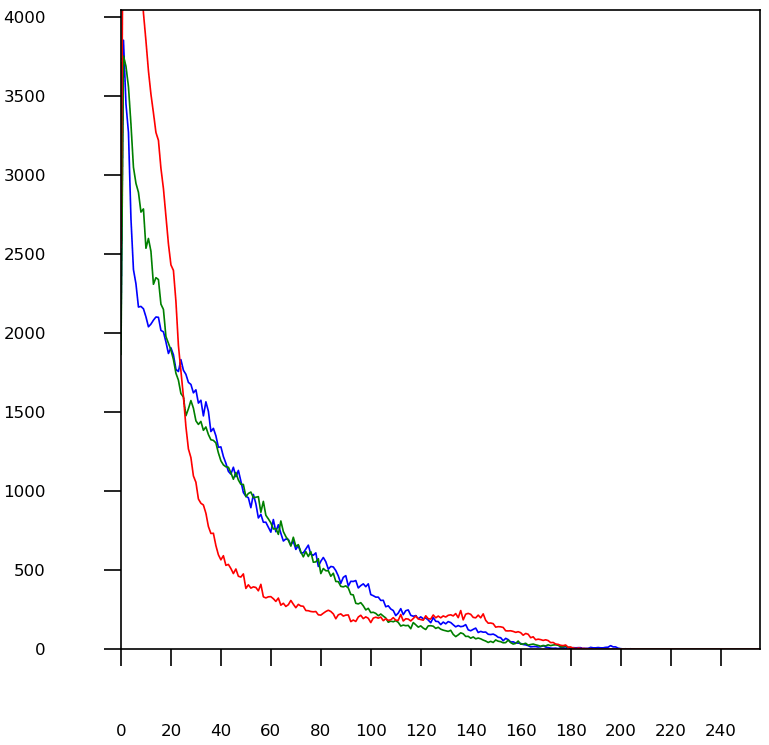}
        \caption{BGR difference histogram.}
    \end{subfigure}
\end{figure}
\begin{figure}[H]
    \centering
    \begin{subfigure}[!h]{0.3\linewidth}
        \includegraphics[width=\linewidth]{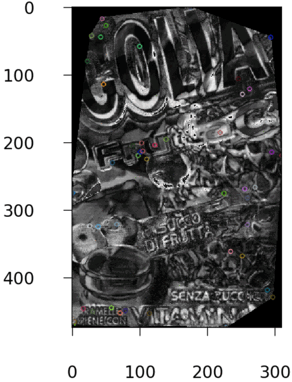}
        \caption{Pixelwise difference norm.}
    \end{subfigure}
    \begin{subfigure}[!h]{0.4\linewidth}
        \includegraphics[width=\linewidth]{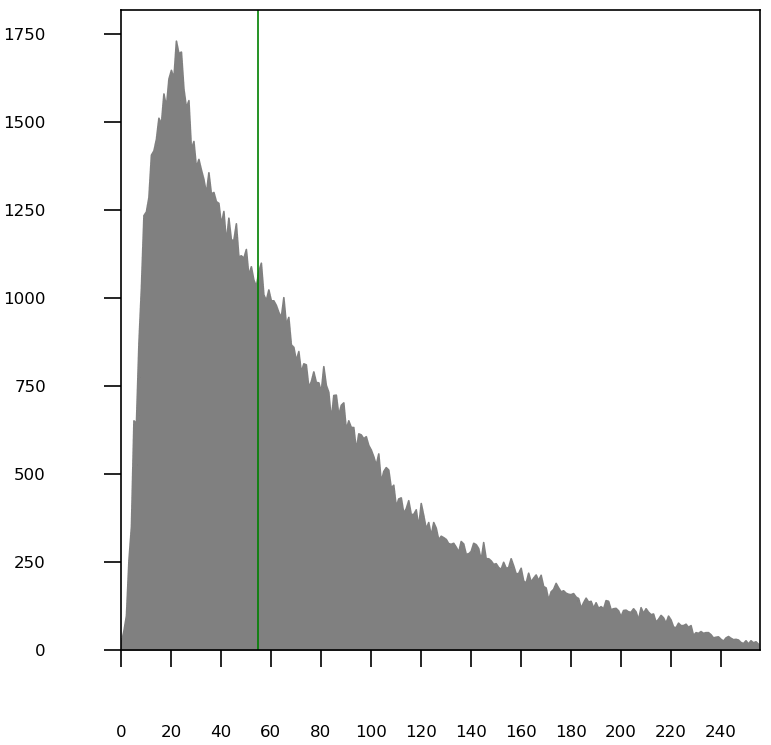}
        \caption{Pixelwise difference norm histogram.}
    \end{subfigure}
    \caption{4\textsuperscript{th} object occurrence (union of seeds \#39-41-42-43-44).}
    \label{fig:figure_71}
\end{figure}
\begin{figure}[H]
    \centering
    \begin{subfigure}[!h]{0.7\linewidth}
        \includegraphics[width=\linewidth]{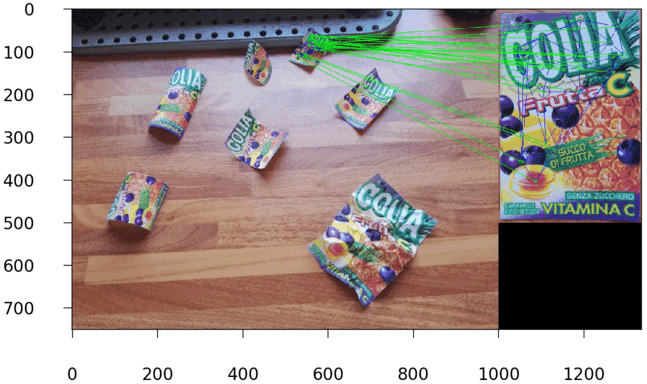}
        \caption{Expanded seed.}
    \end{subfigure}
    \begin{subfigure}[!h]{0.35\linewidth}
        \includegraphics[width=\linewidth]{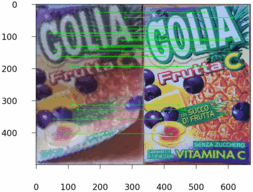}
        \caption{Rectified object occurrence.}
    \end{subfigure}
    \begin{subfigure}[!h]{0.35\linewidth}
        \includegraphics[width=\linewidth]{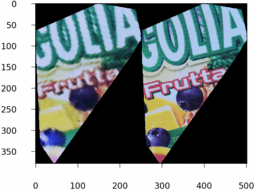}
        \caption{Template and histogram matched object.}
    \end{subfigure}
    \begin{subfigure}[!h]{0.3\linewidth}
        \includegraphics[width=\linewidth]{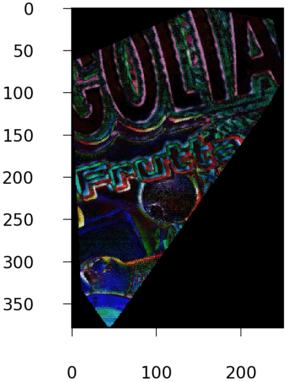}
        \caption{BGR absolute difference.}
    \end{subfigure}
    \begin{subfigure}[!h]{0.4\linewidth}
        \includegraphics[width=\linewidth]{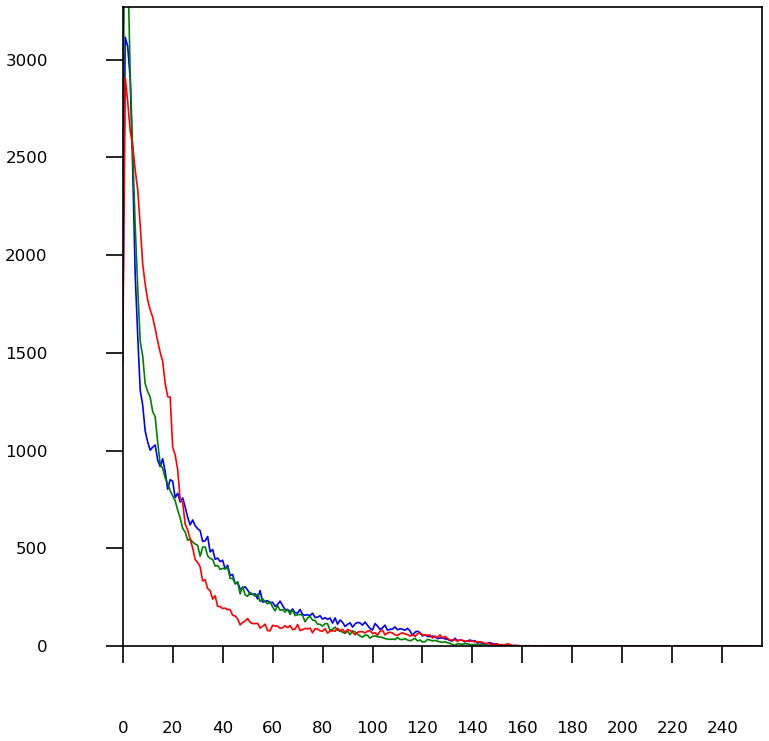}
        \caption{BGR difference histogram.}
    \end{subfigure}
\end{figure}
\begin{figure}[H]
    \centering
    \begin{subfigure}[!h]{0.3\linewidth}
        \includegraphics[width=\linewidth]{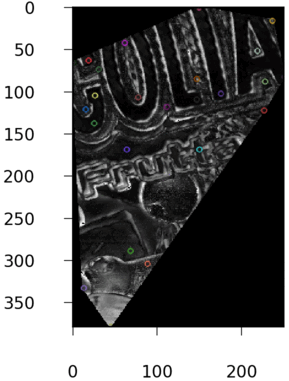}
        \caption{Pixelwise difference norm.}
    \end{subfigure}
    \begin{subfigure}[!h]{0.4\linewidth}
        \includegraphics[width=\linewidth]{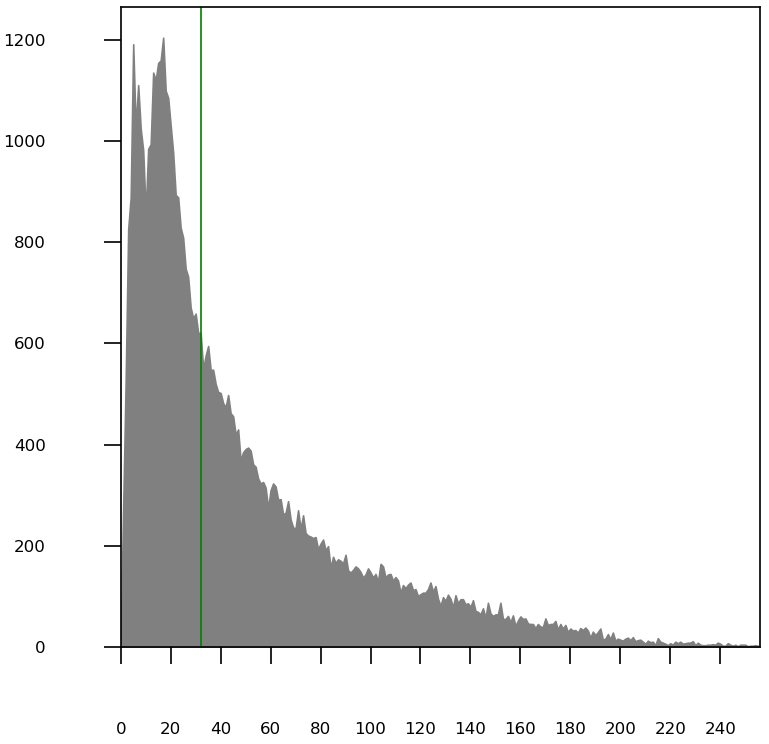}
        \caption{Pixelwise difference norm histogram.}
    \end{subfigure}
    \caption{5\textsuperscript{th} object occurrence (union of seeds \#26-28-29).}
    \label{fig:figure_72}
\end{figure}
\begin{figure}[H]
    \centering
    \begin{subfigure}[!h]{0.7\linewidth}
        \includegraphics[width=\linewidth]{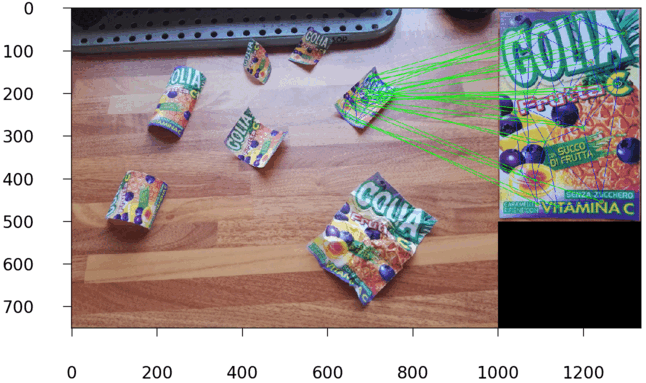}
        \caption{Expanded seed.}
    \end{subfigure}
    \begin{subfigure}[!h]{0.35\linewidth}
        \includegraphics[width=\linewidth]{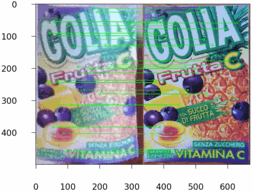}
        \caption{Rectified object occurrence.}
    \end{subfigure}
    \begin{subfigure}[!h]{0.35\linewidth}
        \includegraphics[width=\linewidth]{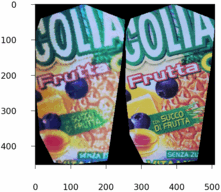}
        \caption{Template and histogram matched object.}
    \end{subfigure}
    \begin{subfigure}[!h]{0.3\linewidth}
        \includegraphics[width=\linewidth]{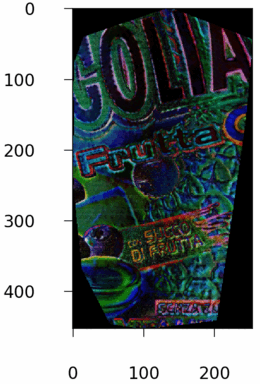}
        \caption{BGR absolute difference.}
    \end{subfigure}
    \begin{subfigure}[!h]{0.4\linewidth}
        \includegraphics[width=\linewidth]{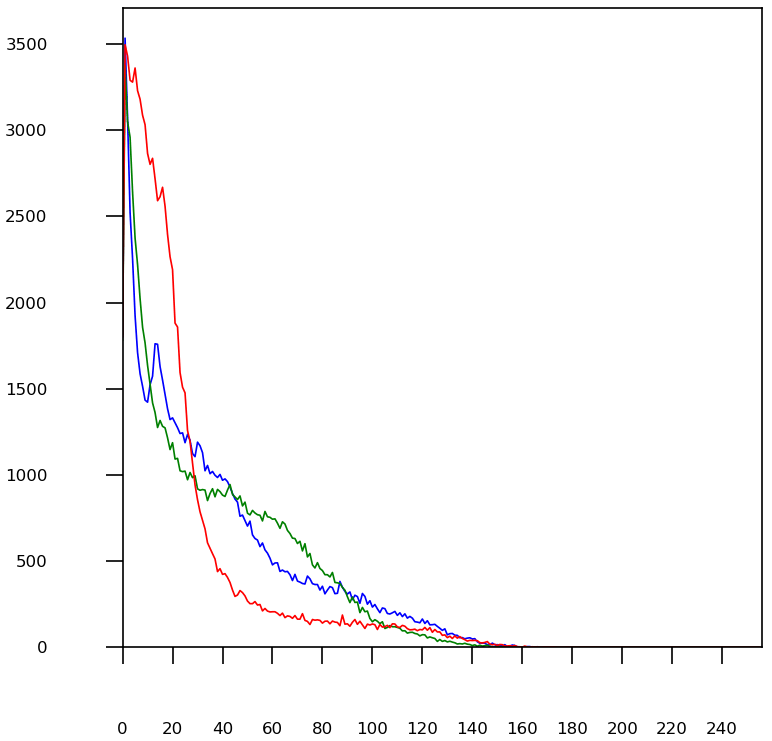}
        \caption{BGR difference histogram.}
    \end{subfigure}
\end{figure}
\begin{figure}[H]
    \ContinuedFloat
    \centering
    \begin{subfigure}[!h]{0.3\linewidth}
        \includegraphics[width=\linewidth]{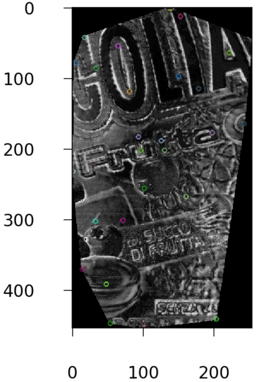}
        \caption{Pixelwise difference norm.}
    \end{subfigure}
    \begin{subfigure}[!h]{0.4\linewidth}
        \includegraphics[width=\linewidth]{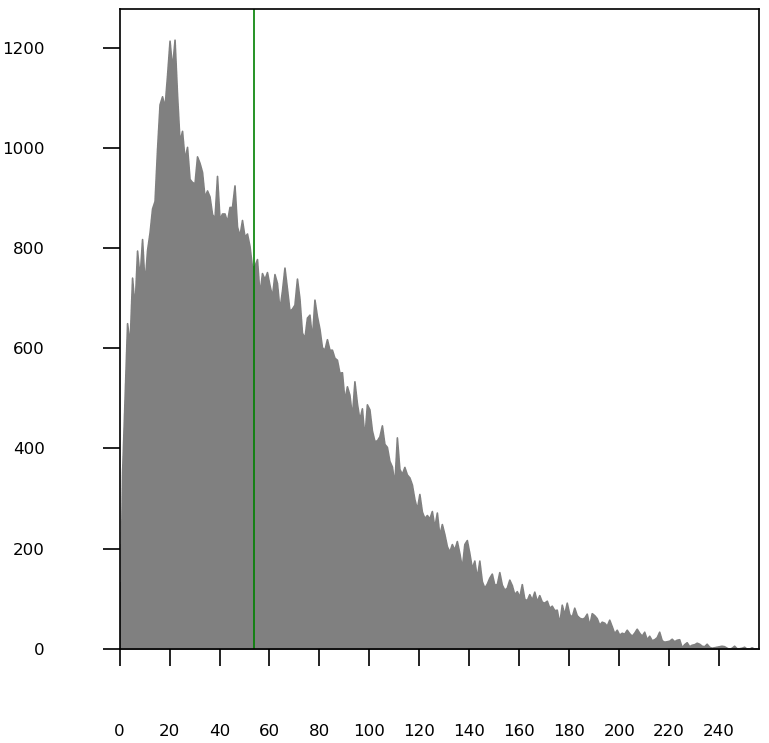}
        \caption{Pixelwise difference norm histogram.}
    \end{subfigure}
    \caption{6\textsuperscript{th} object occurrence (union of seeds \#74).}
    \label{fig:figure_73}
\end{figure}
\begin{figure}[H]
    \centering
    \begin{subfigure}[!h]{0.7\linewidth}
        \includegraphics[width=\linewidth]{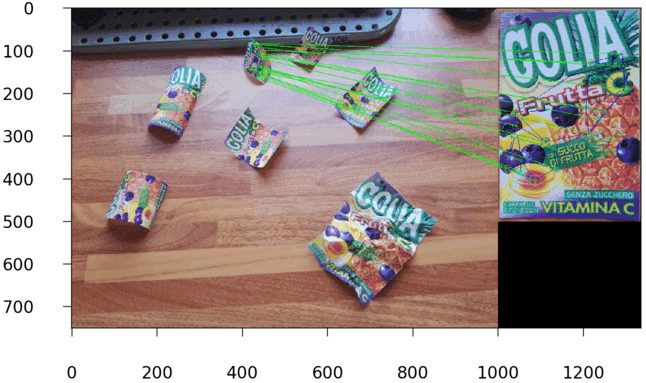}
        \caption{Expanded seed.}
    \end{subfigure}
    \begin{subfigure}[!h]{0.35\linewidth}
        \includegraphics[width=\linewidth]{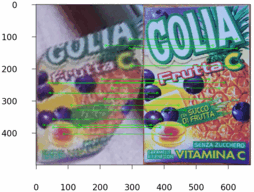}
        \caption{Rectified object occurrence.}
    \end{subfigure}
    \begin{subfigure}[!h]{0.35\linewidth}
        \includegraphics[width=\linewidth]{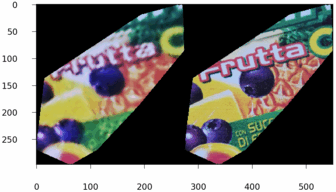}
        \caption{Template and histogram matched object.}
    \end{subfigure}
    \begin{subfigure}[!h]{0.3\linewidth}
        \includegraphics[width=\linewidth]{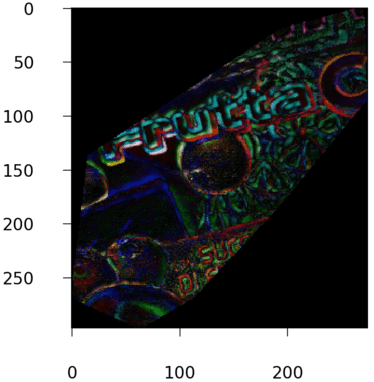}
        \caption{BGR absolute difference.}
    \end{subfigure}
    \begin{subfigure}[!h]{0.4\linewidth}
        \includegraphics[width=\linewidth]{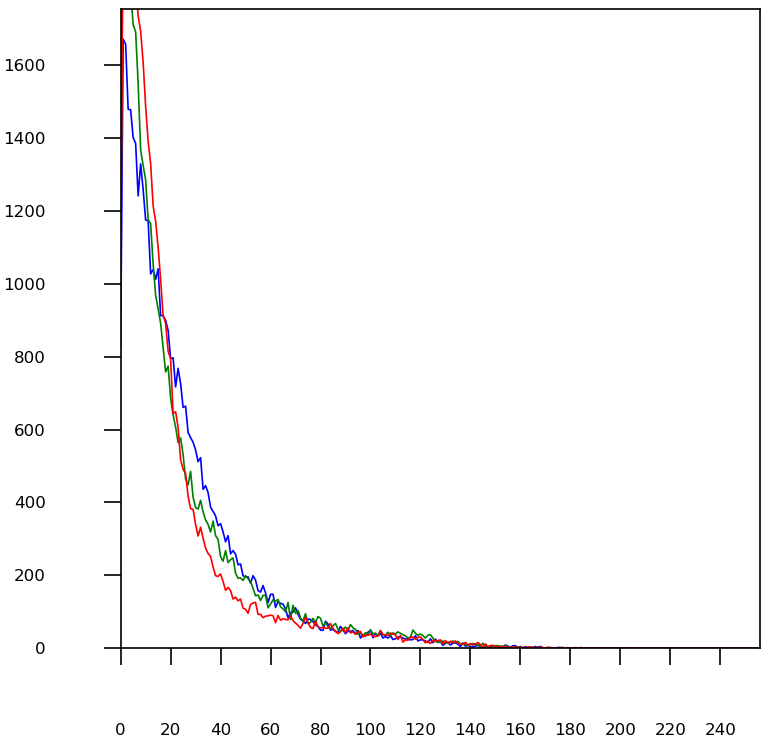}
        \caption{BGR difference histogram.}
    \end{subfigure}
\end{figure}
\begin{figure}[H]
    \centering
    \begin{subfigure}[!h]{0.3\linewidth}
        \includegraphics[width=\linewidth]{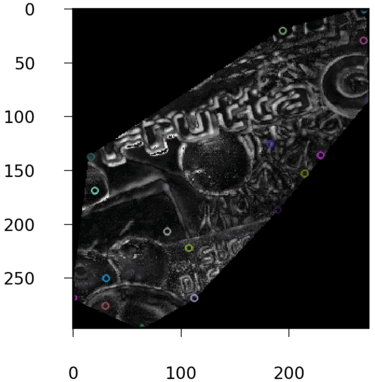}
        \caption{Pixelwise difference norm.}
    \end{subfigure}
    \begin{subfigure}[!h]{0.4\linewidth}
        \includegraphics[width=\linewidth]{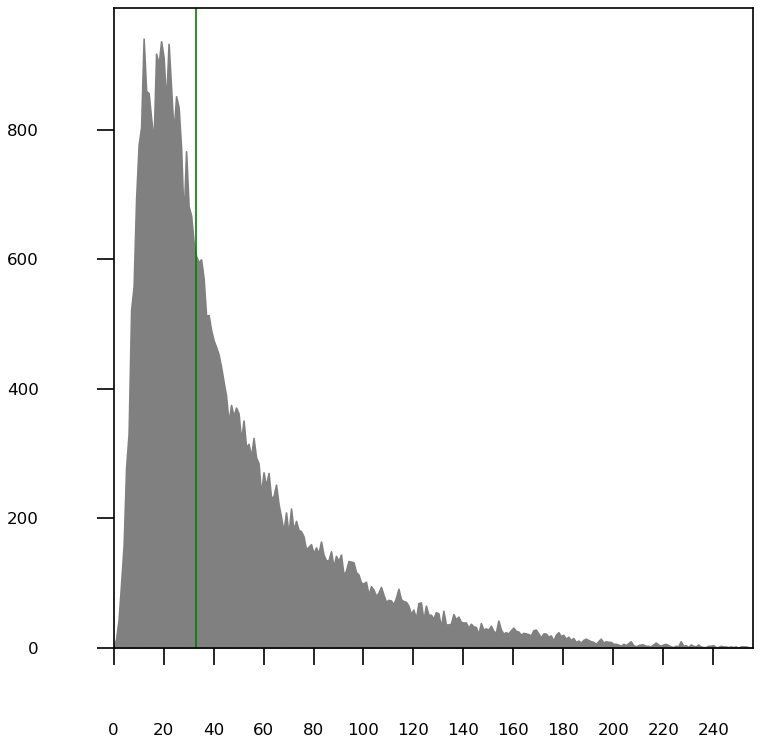}
        \caption{Pixelwise difference norm histogram.}
    \end{subfigure}
    \caption{7\textsuperscript{th} object occurrence (seed \#63).}
    \label{fig:figure_74}
\end{figure}
        \thispagestyle{empty}
        \include{sections/appendices/appendix_B}
        \thispagestyle{empty}
        \include{sections/appendices/appendix_C}
        \thispagestyle{empty}
        \include{sections/appendices/appendix_D}
        \thispagestyle{empty}
        \include{sections/appendices/appendix_E}
        \thispagestyle{empty}
        \include{sections/appendices/appendix_F}
        \thispagestyle{empty}
\end{document}